\documentclass[final]{cvpr}

\usepackage{times}
\usepackage{epsfig}
\usepackage{graphicx}
\usepackage{amsmath}
\usepackage{amssymb}
\usepackage{array}
\usepackage{comment} 
\usepackage{capt-of}
\usepackage{booktabs}
\usepackage{tabularx}
\usepackage{soul}
\usepackage{siunitx}

\makeatletter
\let\BaseCaption\@makecaption
\makeatother

\usepackage{subcaption}

\makeatletter
\let\@makecaption\BaseCaption
\makeatother

\graphicspath{ {./images/} }

\usepackage[pagebackref=true,breaklinks=true,colorlinks,bookmarks=false]{hyperref}

\def\cvprPaperID{7078} %
\def\confYear{CVPR 2021}
\begin{document}

\title{Encoding in Style: a StyleGAN Encoder for Image-to-Image Translation}

\author{
 Elad Richardson\textsuperscript{1} \qquad Yuval Alaluf\textsuperscript{1,2}
 \qquad Or Patashnik\textsuperscript{1,2} \qquad Yotam Nitzan\textsuperscript{2} \\ \qquad Yaniv Azar\textsuperscript{1} \qquad Stav Shapiro\textsuperscript{1} \qquad Daniel Cohen-Or\textsuperscript{2} \\ \\
 \textsuperscript{1}Penta-AI \qquad  \textsuperscript{2}Tel-Aviv University \\
}

\definecolor{darkg}{rgb}{0,0.6,0}
\definecolor{lgreen}{rgb}{0,0.5,0}
\definecolor{amethyst}{rgb}{0.6, 0.4, 0.8}
\definecolor{orange}{rgb}{0.93,0.48,0.03}
\definecolor{blueviolet}{rgb}{0.54,0.16,0.88}

\newcommand{\erc}[1]{{\color{cyan}\textbf{ER:} #1}}
\newcommand{\dcc}[1]{{\color{blue}\textbf{DC:} #1}}
\newcommand{\dc}[1]{{\color{blue}{#1}}}
\newcommand{\yac}[1]{{\color{lgreen}\textbf{YA:} #1}}
\newcommand{\ya}[1]{{\color{lgreen}#1}}
\newcommand{\ync}[1]{{\color{orange}\textbf{YN:} #1}}
\newcommand{\yn}[1]{{\color{orange}{#1}}}
\newcommand{\opc}[1]{{\color{amethyst}\textbf{OP:} #1}}
\newcommand{\op}[1]{{\color{amethyst}{#1}}}
\newcommand{\ssc}[1]{{\color{magenta}\textbf{SS:} #1}}
\newcommand{\ayc}[1]{{\color{blueviolet}\textbf{AY:} #1}}

\pagestyle{plain}

\twocolumn[{%
\renewcommand\twocolumn[1][]{#1}%
\vspace{-1em}
\maketitle
\vspace{-1em}
\begin{center}
    \centering
    \setlength{\tabcolsep}{1pt}
    \vspace{-0.2cm}
    \includegraphics[height=0.1625\linewidth]{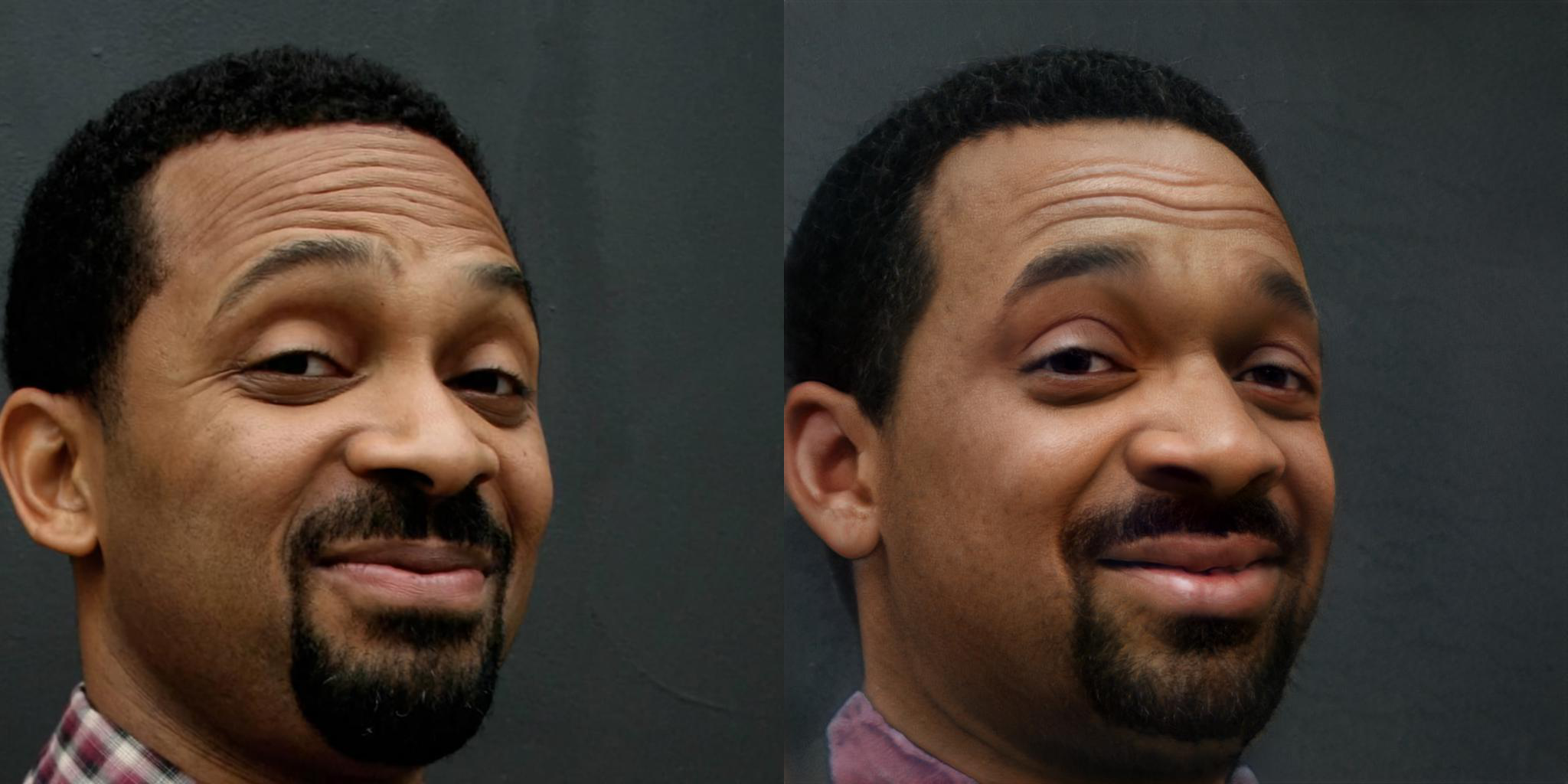}
    \includegraphics[height=0.1625\linewidth]{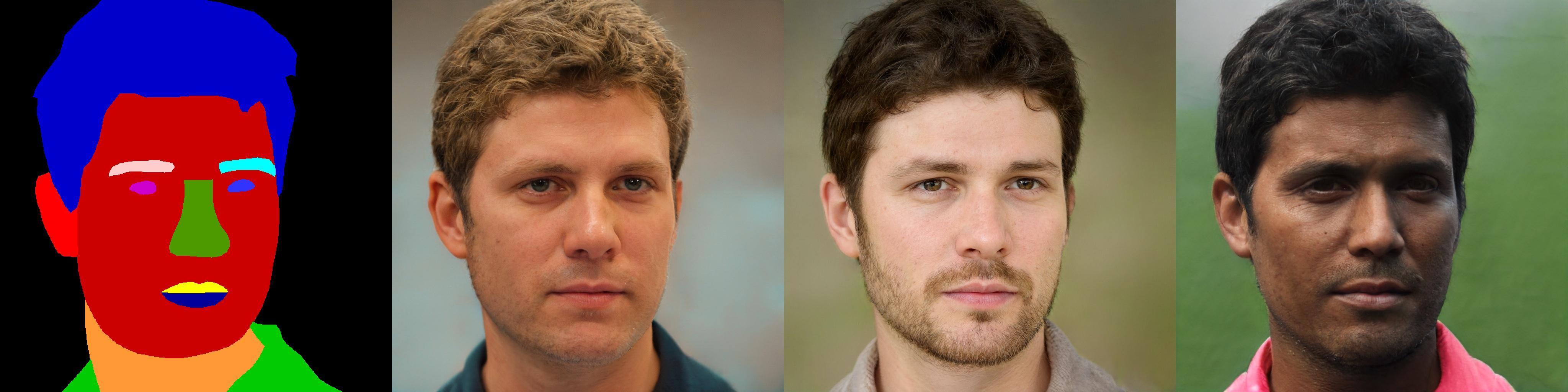}
    
    \vspace{0.05cm}
    \includegraphics[height=0.1625\linewidth]{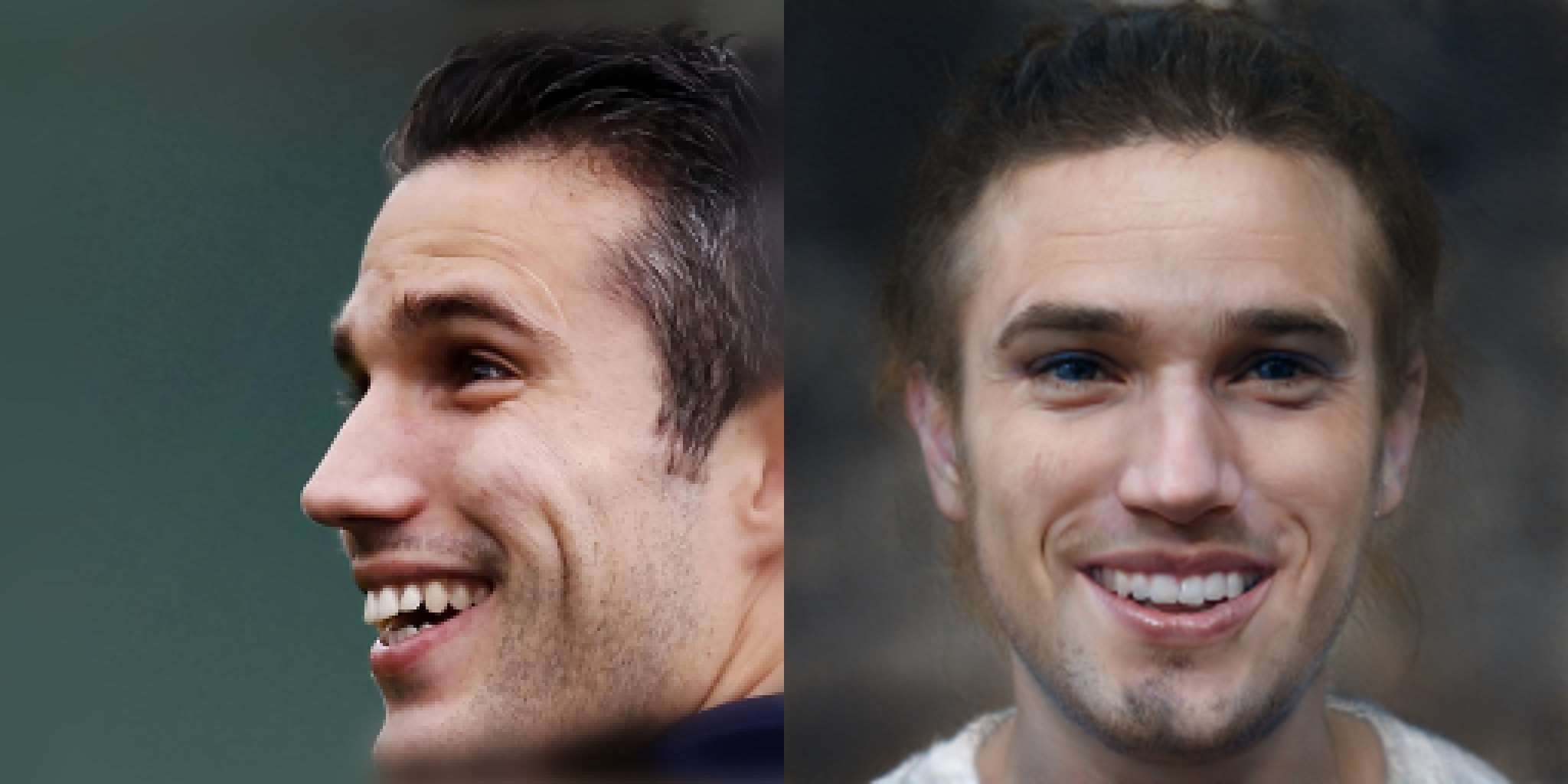}
    \includegraphics[height=0.1625\linewidth]{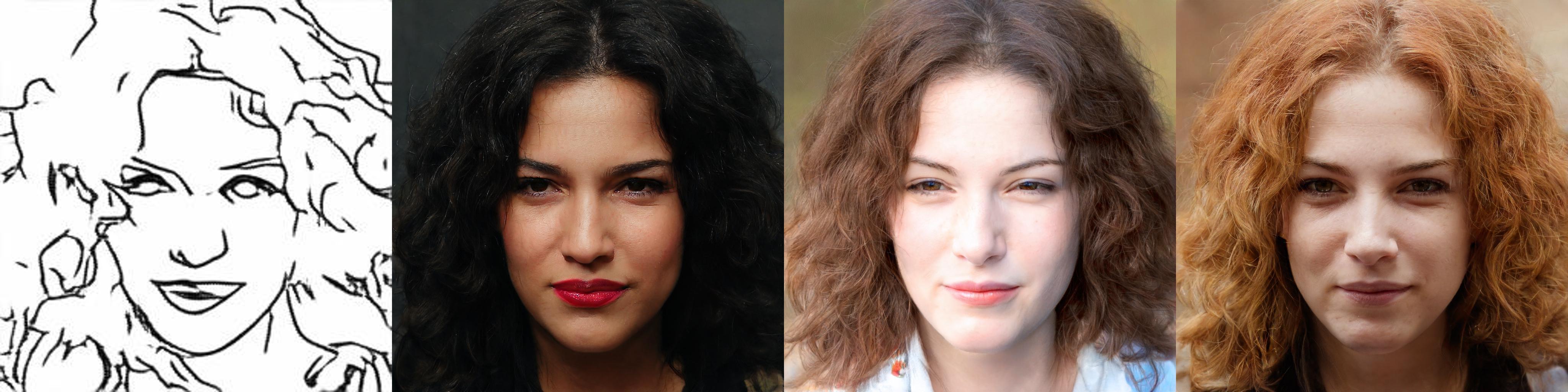}

    \vspace{0.05cm}
    \includegraphics[height=0.1625\linewidth]{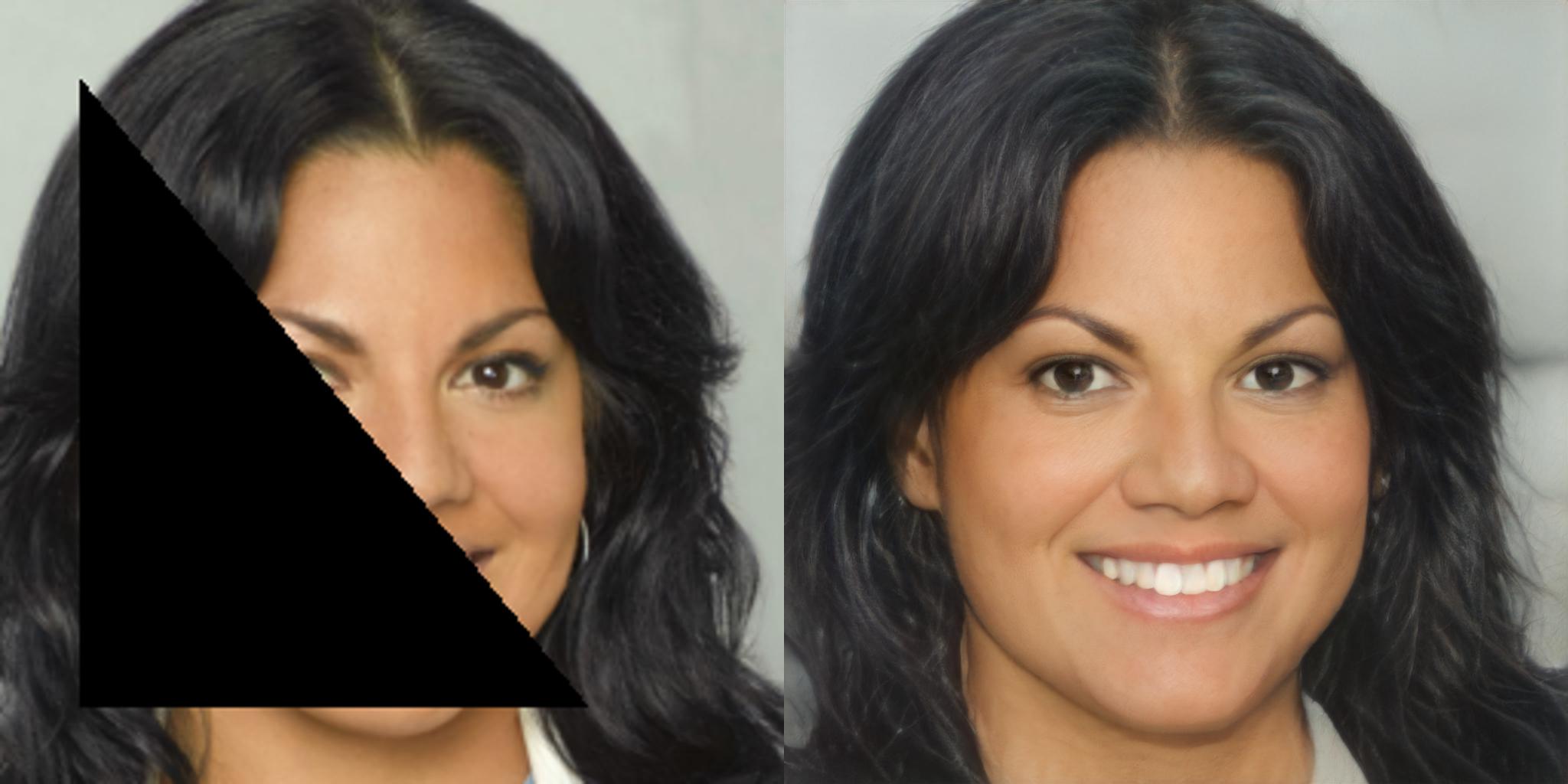}
    \includegraphics[height=0.1625\linewidth]{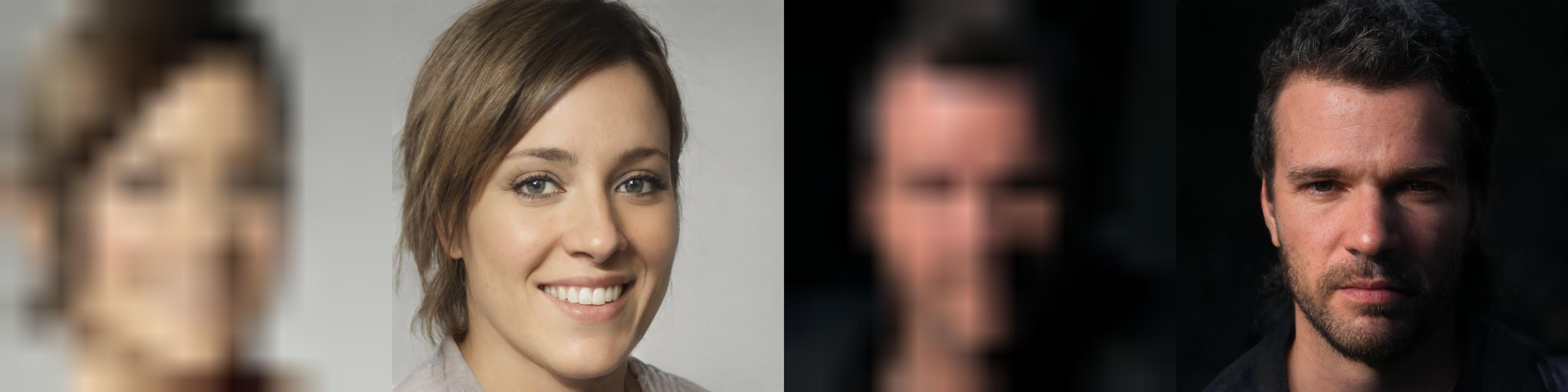}
    \captionof{figure}{The proposed pixel2style2pixel framework can be used to solve a wide variety of image-to-image translation tasks. Here we show results of pSp on StyleGAN inversion, multi-modal conditional image synthesis, facial frontalization, inpainting and super-resolution.}
    \label{teaser}
\end{center}%
}]

\begin{abstract}
We present a generic image-to-image translation framework, \textit{pixel2style2pixel (pSp)}. 
Our pSp framework is based on a novel encoder network that directly generates a series of style vectors which are fed into a pretrained StyleGAN generator, forming the extended $\mathcal{W+}$ latent space. 
We first show that our encoder can \textit{directly} embed real images into $\mathcal{W+}$, with no additional optimization.
Next, we propose utilizing our encoder to directly solve image-to-image translation tasks, defining them as encoding problems from some input domain into the latent domain. By deviating from the standard ``\textit{invert first, edit later}'' methodology used with previous StyleGAN encoders, our approach can handle a variety of tasks even when the input image is not represented in the StyleGAN domain.
We show that solving translation tasks through StyleGAN significantly simplifies the training process, as no adversary is required, has better support for solving tasks without pixel-to-pixel correspondence, and inherently supports multi-modal synthesis via the resampling of styles. 
Finally, we demonstrate the potential of our framework on a variety of facial image-to-image translation tasks, even when compared to state-of-the-art solutions designed specifically for a single task, and further show that it can be extended beyond the human facial domain. 
Code is available at \url{https://github.com/eladrich/pixel2style2pixel}.
\end{abstract}

\vspace{-0.75cm}
\section{Introduction}
In recent years, Generative Adversarial Networks (GANs) have significantly advanced image synthesis, particularly on face images. State-of-the-art image generation methods have achieved high visual quality and fidelity, and can now generate images with phenomenal realism. Most notably, StyleGAN \cite{karras2019style, karras2020analyzing} proposes a novel style-based generator architecture and attains state-of-the-art visual quality on high-resolution images. 
Moreover, it has been demonstrated that it has a disentangled latent space, $\mathcal{W}$ \cite{yang2019semantic,collins2020editing,shen2020interpreting}, 
which offers control and editing capabilities.

Recently, numerous methods have shown competence in controlling StyleGAN's latent space and performing meaningful manipulations in $\mathcal{W}$ ~\cite{jahanian2019steerability, shen2020interpreting, tewari2020stylerig, harkonen2020ganspace}. These methods follow an ``\textit{invert first, edit later}'' approach, where one first inverts an image into StyleGAN's latent space and then edits the latent code in a semantically meaningful manner to obtain a new code that is then used by StyleGAN to generate the output image.
However, it has been shown that inverting a real image into a $512$-dimensional vector $\textbf{w}\in \mathcal{W}$ does not lead to an accurate reconstruction. Motivated by this, it has become common practice \cite{abdal2019image2stylegan, abdal2020image2stylegan++, pbayliesstyleganencoder, zhu2020domain, Abdal:StyleFlow:Arxiv:2020} to encode real images into an extended latent space, $\mathcal{W+}$, defined by the concatenation of $18$ different $512$-dimensional $\textbf{w}$ vectors, one for each input layer of StyleGAN. These works usually resort to using per-image optimization over $\mathcal{W+}$, requiring several minutes for a single image. To accelerate this optimization process, some methods \cite{pbayliesstyleganencoder, zhu2020domain} have trained an encoder to infer an approximate vector in $\mathcal{W+}$ which serves as a good initial point from which additional optimization is required. However, a fast and accurate  inversion of real images into $\mathcal{W+}$ remains a challenge.

In this paper, we first introduce a novel encoder architecture tasked with encoding an arbitrary image directly into $\mathcal{W+}$. The encoder is based on a Feature Pyramid Network \cite{lin2017feature}, where style vectors are extracted from different pyramid scales and inserted directly into a \textit{fixed, pretrained StyleGAN generator} in correspondence to their spatial scales.
We show that our encoder can directly reconstruct real input images, allowing one to perform latent space manipulations without requiring time-consuming optimization.
While these manipulations allow for extensive editing of real images, they are inherently limited. That is because the input image must be invertible, i.e., there must exist a latent code that reconstructs the image. 
This requirement is a severe limitation for tasks, such as conditional image generation, where the input image does not reside in the same StyleGAN domain.
To overcome this limitation we propose using our encoder together with the pretrained StyleGAN generator as a complete image-to-image translation framework. In this formulation, input images are directly encoded into the desired output latents which are then fed into StyleGAN to generate the desired output images. This allows one to utilize StyleGAN for image-to-image translation even when the input and output images are not from the same domain.

While many previous approaches to solving image-to-image translation tasks involve dedicated architectures specific for solving a single problem, we follow the spirit of pix2pix~\cite{isola2017image} and define a generic framework able to solve a wide range of tasks, all using the same architecture. Besides the simplification of the training process, as no adversary discriminator needs to be trained, using a pretrained StyleGAN generator offers several intriguing advantages over previous works. For example, many image-to-image architectures explicitly feed the generator with residual feature maps from the encoder~\cite{isola2017image, wang2018high}, creating a strong locality bias~\cite{richardson2020unsupervised}. In contrast, our generator is governed only by the styles with no direct spatial input. 
Another notable advantage of the intermediate style representation is the inherent support for multi-modal synthesis for ambiguous tasks such as image generation from sketches, segmentation maps, or low-resolution images. In such tasks, the generated styles can be resampled to create variations of the output image with no change to the architecture or training process. In a sense, our method performs \textit{pixel2style2pixel} translation, as every image is first encoded into style vectors and then into an image, and is therefore dubbed \textit{pSp}.

The main contributions of this paper are: (i)  A novel StyleGAN encoder able to directly encode real images into the $\mathcal{W+}$ latent domain; and (ii) A new methodology for utilizing a pretrained StyleGAN generator to solve image-to-image translation tasks.

\section{Related Work}

\begin{figure*}
    \centering
    \setlength{\belowcaptionskip}{-9pt}
    \includegraphics[width=\linewidth]{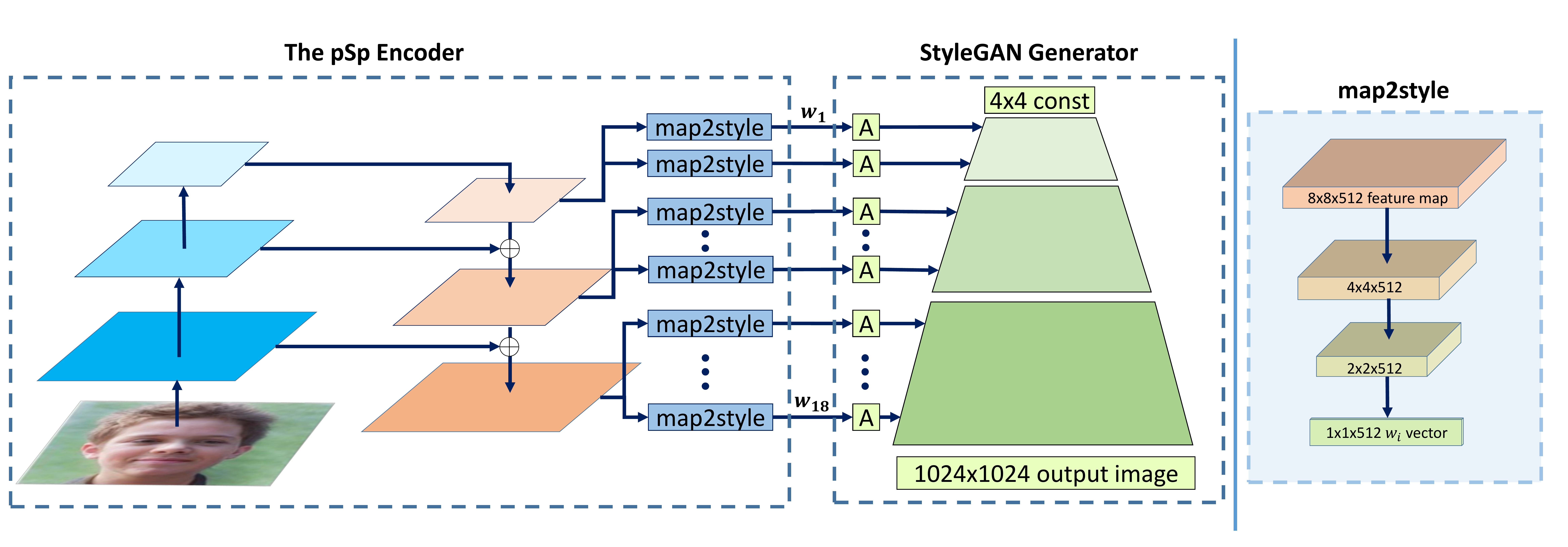}
    \captionof{figure}{Our pSp architecture. Feature maps are first extracted using a standard feature pyramid over a ResNet backbone. For each of the $18$ target styles, a small mapping network is trained to extract the learned styles from the corresponding feature map, where styles (0-2) are generated from the small feature map, (3-6) from the medium feature map, and (7-18) from the largest feature map. The mapping network, \textit{map2style}, is a small fully convolutional network, which gradually reduces spatial size using a set of 2-strided convolutions followed by LeakyReLU activations. Each generated $512$ vector, is fed into StyleGAN, starting from its matching affine transformation, $A$.}
    \label{fig:architecture}
\end{figure*}

\paragraph{\textit{\textbf{GAN Inversion.}}}
With the rapid evolution of GANs, many works have tried to understand and control their latent space. 
A specific task that has received substantial attention is \textit{GAN Inversion}, which was first introduced by Zhu \etal~\cite{zhu2016generative}. In this task, the latent vector from which a pretrained GAN most accurately reconstructs a given, known image, is sought. Motivated by its state-of-the-art image quality and latent space semantic richness, many recent works have used StyleGAN~\cite{karras2019style, karras2020analyzing} for this task. Generally, inversion methods either directly optimize the latent vector to minimize the error for the given image \cite{lipton2017precise,creswell2018inverting,abdal2019image2stylegan,abdal2020image2stylegan++}, train an encoder to map the given image to the latent space \cite{perarnau2016invertible, creswell2018inverting, pidhorskyi2020adversarial, guan2020collaborative, nitzan2020disentangling}, or use a hybrid approach combining both \cite{pbayliesstyleganencoder, zhu2020domain}. Typically, methods performing optimization are superior in reconstruction quality to a learned encoder mapping, but require a substantially longer time.
Unlike the above methods, our encoder can accurately and efficiently embed a given face image into the extended latent space $\mathcal{W}+$ with no further optimization.

\vspace{-0.2cm}
\paragraph{\textit{\textbf{Latent Space Manipulation.}}}

Recently, numerous papers have presented diverse methods for learning semantic edits of the latent code. One popular approach is to find linear directions that correspond to changes in a given binary labeled attribute, such as young $\leftrightarrow$ old, or no-smile $\leftrightarrow$ smile \cite{shen2020interpreting, goetschalckx2019ganalyze, denton2019detecting, Abdal:StyleFlow:Arxiv:2020}. 
Tewari~\etal \cite{tewari2020stylerig} utilize a pretrained 3DMM to learn semantic face edits in the latent space. 
Jahanian \etal \cite{jahanian2019steerability} find latent space paths that correspond to a specific transformation, such as zoom or rotation, in a self-supervised manner. 
Härkönen \etal \cite{harkonen2020ganspace} find useful paths in an unsupervised manner by using the principal component axes of an intermediate activation space. 
Collins \etal \cite{collins2020editing} perform local semantic editing by manipulating corresponding components of the latent code. 
These methods generally follow an ``\textit{invert first, edit later}'' procedure, where an image is first embedded into the latent space, and then its latent is edited in a semantically meaningful manner.  This differs from our approach which directly encodes input images into the corresponding output latents, allowing one to also handle inputs that do not reside in the StyleGAN domain.

\vspace{-0.2cm}
\paragraph{\textit{\textbf{Image-to-Image.}}}
Image-to-Image translation techniques aim at learning a conditional image generation function that maps an input image of a source domain to a corresponding image of a target domain. Isola \etal \cite{isola2017image} first introduced the use of conditional GANs to solve various image-to-image translation tasks. Since then, their work has been extended for many scenarios: high-resolution synthesis \cite{wang2018high}, unsupervised learning \cite{liu2017unsupervised, zhu2017unpaired, katzir2019cross, lira2020ganhopper}, 
multi-modal image synthesis \cite{zhu2017toward, huang2018multimodal, choi2020stargan}, 
and conditional image synthesis \cite{park2019semantic, li2019linestofacephoto, liu2019learning, zhu2020sean, chen2020deep}. The aforementioned works have constructed dedicated architectures, which require training the generator network and generally do not generalize to other translation tasks. This is in contrast to our method that uses the same architecture for solving a variety of tasks.

\section{The pSp Framework}
Our pSp framework builds upon the representative power of a pretrained StyleGAN generator and the $\mathcal{W+}$ latent space. To utilize this representation one needs a strong encoder that is able to match each input image to an accurate encoding in the latent domain. A simple technique to embed into this domain is directly encoding a given input image into $\mathcal{W+}$ using a single $512$-dimensional vector obtained from the last layer of the encoder network, thereby learning all $18$ style vectors together. However, such an architecture presents a strong bottleneck making it difficult to fully represent the finer details of the original image and is therefore limited in reconstruction quality.

In StyleGAN, the authors have shown that the different style inputs correspond to different levels of detail, which are roughly divided into three groups --- coarse, medium, and fine. Following this observation, in pSp we extend an encoder backbone with a feature pyramid~\cite{lin2017feature}, generating three levels of feature maps from which styles are extracted using a simple intermediate network --- map2style --- shown in Figure ~\ref{fig:architecture}. The styles, aligned with the hierarchical representation, are then fed into the generator in correspondence to their scale to generate the output image, thus completing the translation from input \textit{pixels} to output \textit{pixels}, through the intermediate \textit{style} representation. 
The complete architecture is illustrated in Figure~\ref{fig:architecture}. 

As in StyleGAN, we further define $\overline{\textbf{w}}$ to be the average style vector of the pretrained generator. Given an input image, $\textbf{x}$, the output of our model is then defined as 
\begin{equation*}
    pSp(\textbf{x}) := G(E(\textbf{x}) + \overline{\textbf{w}})
\end{equation*}
where $E(\cdot)$ and $G(\cdot)$ denote the encoder and StyleGAN generator, respectively. In this formulation, our encoder aims to learn the latent code with respect to the average style vector. We find that this results in better initialization.

\subsection{Loss Functions}
While the style-based translation is the core part of our framework, the choice of losses is also crucial. Our encoder is trained using a weighted combination of several objectives. 
First, we utilize the pixel-wise $\mathcal{L}_2$ loss,
\vspace{-1.15mm}
\begin{equation}
    \mathcal{L_{\text{2}}}\left ( \textbf{x} \right ) = || \textbf{x} - pSp(\textbf{x}) ||_2.
\end{equation}

\vspace{-1.15mm}
In addition, to learn perceptual similarities, we utilize the LPIPS~\cite{zhang2018unreasonable} loss, which has been shown to better preserve image quality~\cite{guan2020collaborative} compared to the more standard perceptual loss~\cite{johnson2016perceptual}:
\begin{equation}
    \mathcal{L_{\text{LPIPS}}}\left ( \textbf{x} \right ) = || F(\textbf{x}) - F(pSp(\textbf{x}))||_2,
\end{equation}
where $F(\cdot)$ denotes the perceptual feature extractor.

To encourage the encoder to output latent style vectors closer to the average latent vector, we additionally define the following regularization loss:
\begin{equation}
    \mathcal{L}_{\text{reg}}\left ( \textbf{x} \right ) = || E(\textbf{x}) - \overline{\textbf{w}}   ||_2.
\end{equation}

\vspace{-1.15mm}
Similar to the truncation trick introduced in StyleGAN, we find that adding this regularization in the training of our encoder improves image quality without harming the fidelity of our outputs, especially in some of the more ambiguous tasks explored below.

Finally, a common challenge when handling the specific task of encoding facial images is the preservation of the input identity. To tackle this, we incorporate a dedicated recognition loss measuring the cosine similarity between the output image and its source, 
\vspace{-1.25mm}
\begin{equation}
    \mathcal{L}_{\text{ID}}\left (\textbf{x} \right ) = 1-\left \langle R(\textbf{x}),R(pSp(\textbf{x}))) \right \rangle ,
\end{equation}
where $R$ is the pretrained ArcFace~\cite{deng2019arcface} network.

In summary, the total loss function is defined as
\begin{equation*}
\small
    \mathcal{L}(\textbf{x}) = 
    \lambda_1 \mathcal{L}_2(\textbf{x}) + 
    \lambda_2 \mathcal{L}_{\text{LPIPS}}(\textbf{x}) + 
    \lambda_3 \mathcal{L}_{\text{ID}}(\textbf{x}) + 
    \lambda_4 \mathcal{L}_{\text{reg}}(\textbf{x}),
\end{equation*}

where $\lambda_1$, $\lambda_2$, $\lambda_3$, $\lambda_4$ are constants defining the loss weights. This curated set of loss functions allows for more accurate encoding into StyleGAN compared to previous works and can be easily tuned for different encoding tasks according to their nature. Constants and other implementation details can be found in Appendix~\ref{Implement}.

\subsection{The Benefits of The StyleGAN Domain}\label{subsec:benefits}
\begin{figure}
    \centering
    \includegraphics[width=0.94\linewidth]{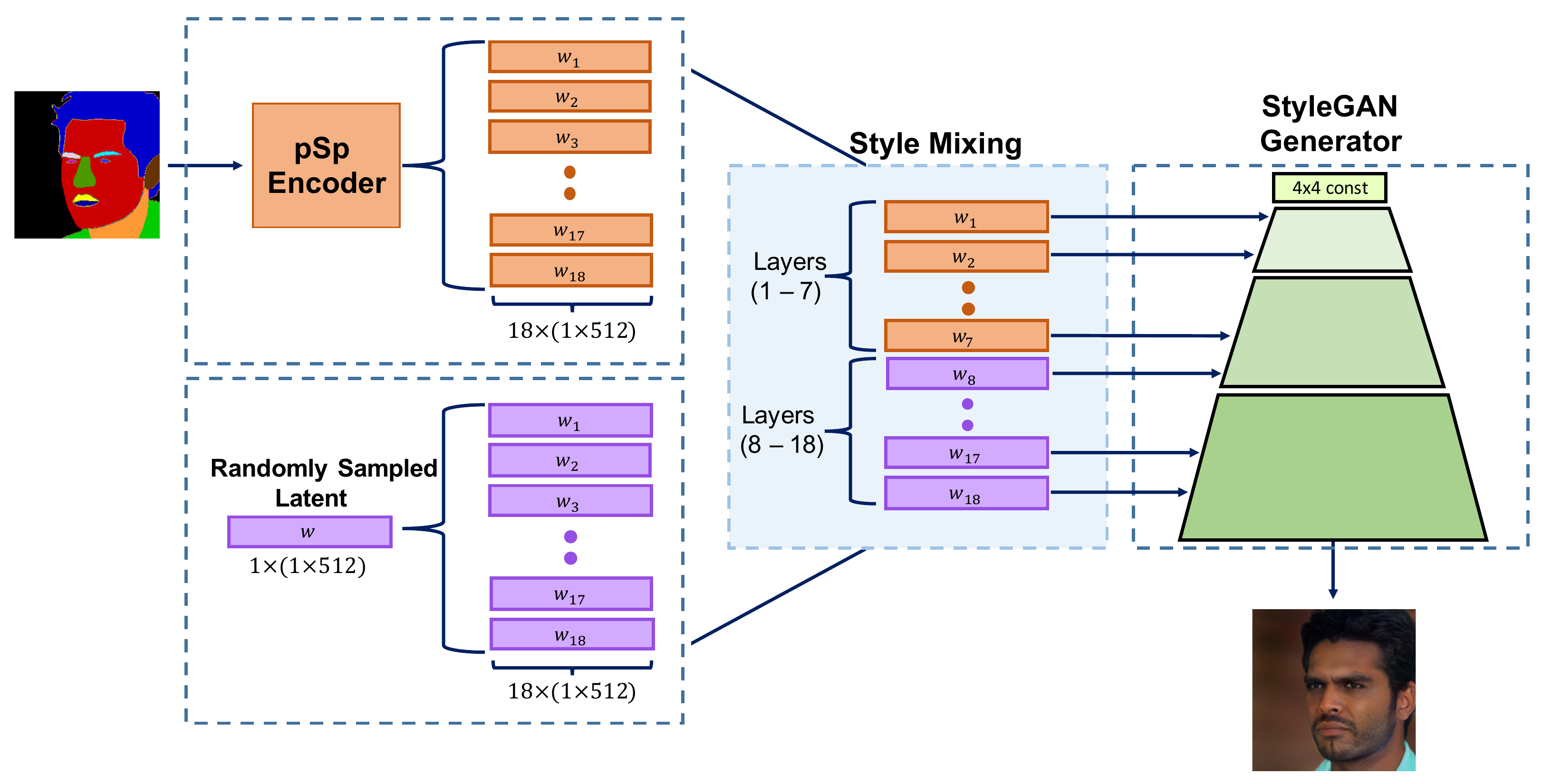}
    \captionof{figure}{Style-mixing for multi-modal generation.}
    \label{fig:style_mixing_explanation}   
    \vspace{-0.195in}
\end{figure}

The translation between images through the \textit{style} domain differentiates pSp from many standard image-to-image translation frameworks, as it makes our model operate \textit{globally} instead of \textit{locally}, without requiring pixel-to-pixel correspondence. This is a desired property as it has been shown that the locality bias limits current methods when handling non-local transformations~\cite{richardson2020unsupervised}.
Additionally, previous works~\cite{karras2019style, collins2020editing} have demonstrated that the disentanglement of semantic objects learned by StyleGAN is due to its layer-wise representation. This ability to independently manipulate semantic attributes leads to another desired property: the support for \textit{multi-modal synthesis}. As some translation tasks are ambiguous, where a single input image may correspond to several outputs, it is desirable to be able to sample these possible outputs. While this requires specialized changes in standard image-to-image architectures~\cite{zhu2017toward,huang2018multimodal}, our framework inherently supports this by simply sampling style vectors. 
In practice, this is done by randomly sampling a vector $\textbf{w}\in\mathbb{R}^{512}$ and generating a corresponding latent code in $\mathcal{W+}$ by replicating $\textbf{w}$. Style mixing is then performed by replacing select layers of the computed latent with those of the randomly generated latent, possibly with an $\alpha$ parameter for blending between the two styles. This approach is illustrated in Figure~\ref{fig:style_mixing_explanation}. 

\vspace{-0.125cm}
\section{Applications and Experiments}

\begin{figure*}
\setlength{\tabcolsep}{1pt}
\centering
{\small
    \begin{tabular}{c c c c c c c c c}
        \raisebox{0.25in}{\rotatebox[origin=t]{90}{Input}}&
        \includegraphics[width=0.11\textwidth]{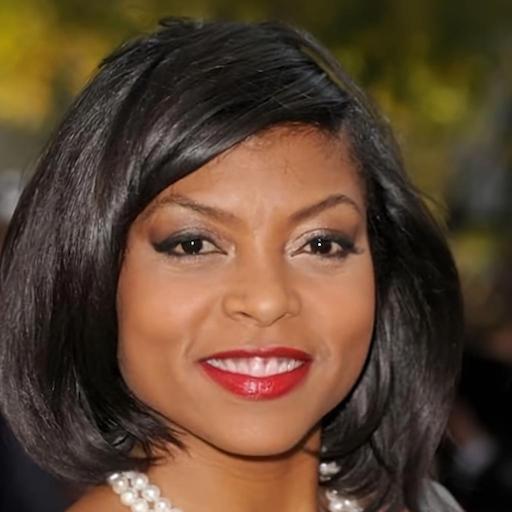}&
        \includegraphics[width=0.11\textwidth]{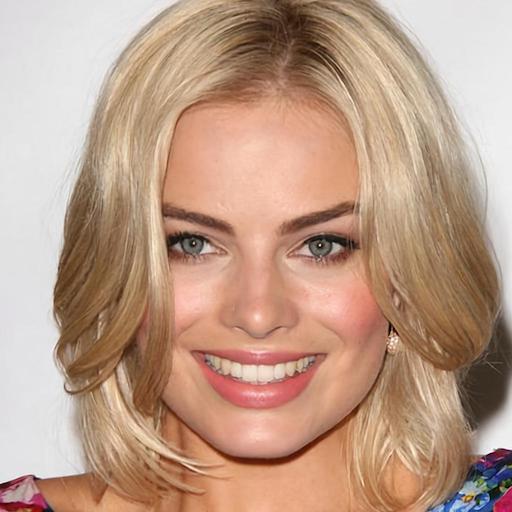}&        
        \includegraphics[width=0.11\textwidth]{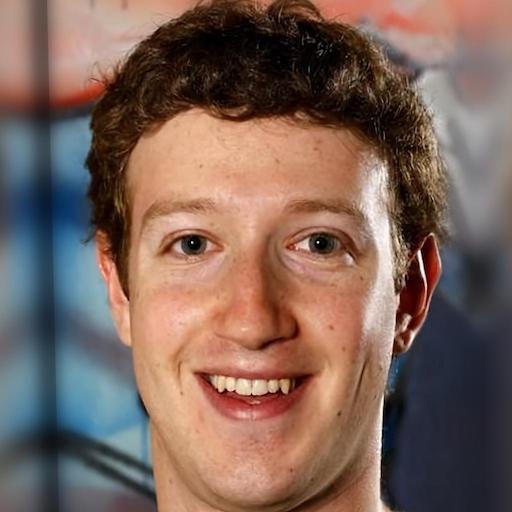}&
        \includegraphics[width=0.11\textwidth]{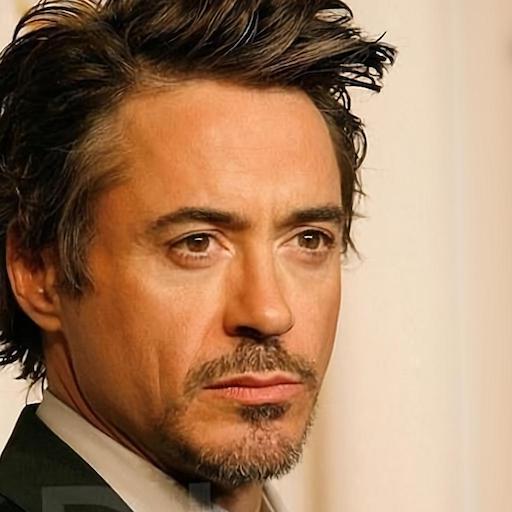}&
        \includegraphics[width=0.11\textwidth]{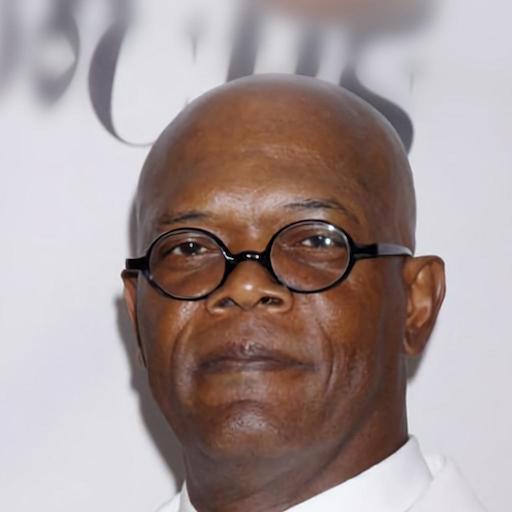}&
        \includegraphics[width=0.11\textwidth]{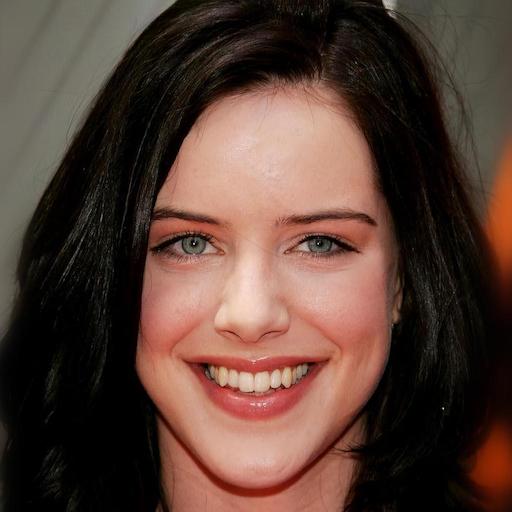}&        
        \includegraphics[width=0.11\textwidth]{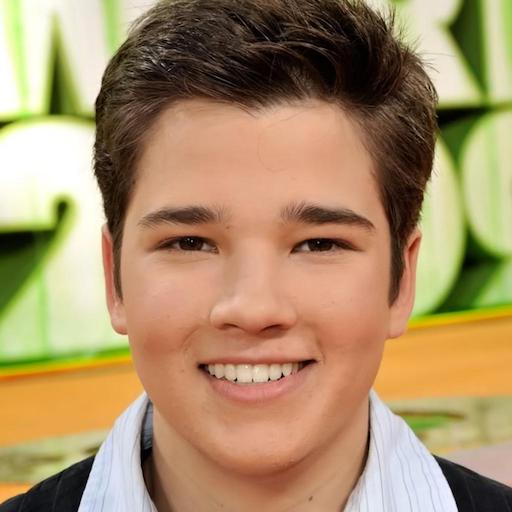}&
        \includegraphics[width=0.11\textwidth]{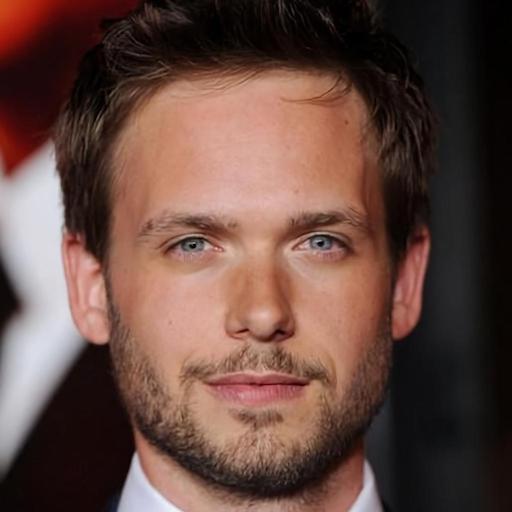} \tabularnewline
        \raisebox{0.3in}{\rotatebox[origin=t]{90}{ALAE~\cite{pidhorskyi2020adversarial}}}&
        \includegraphics[width=0.11\textwidth]{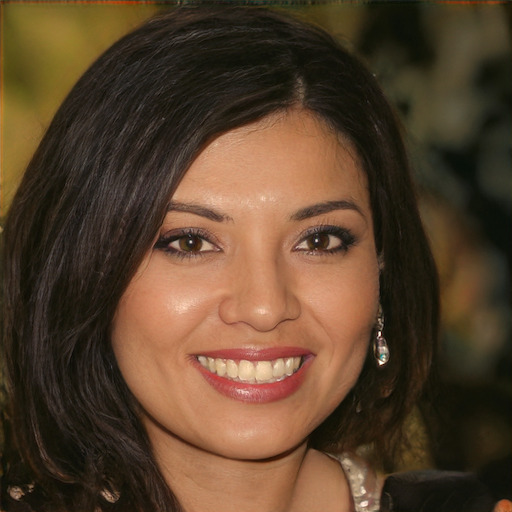}&
        \includegraphics[width=0.11\textwidth]{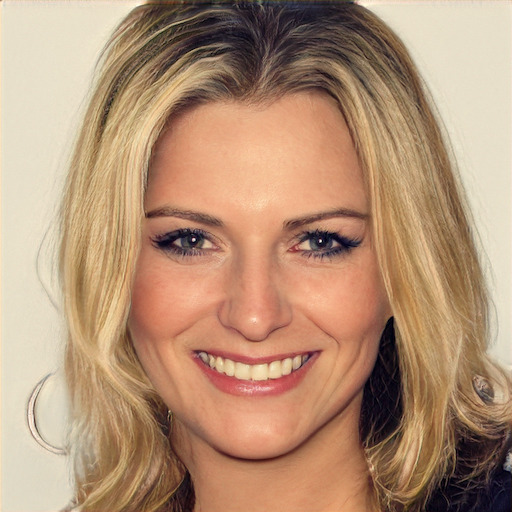}&        
        \includegraphics[width=0.11\textwidth]{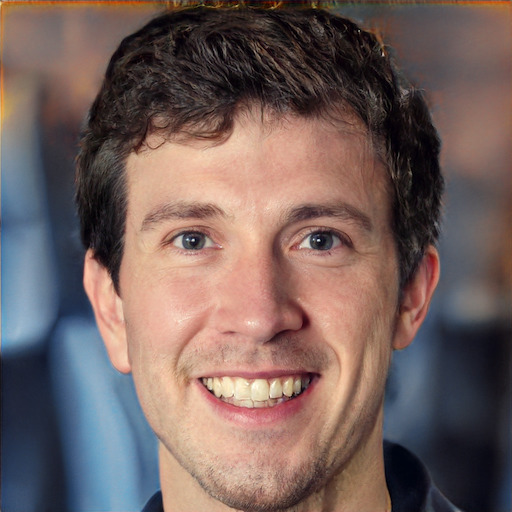}&
        \includegraphics[width=0.11\textwidth]{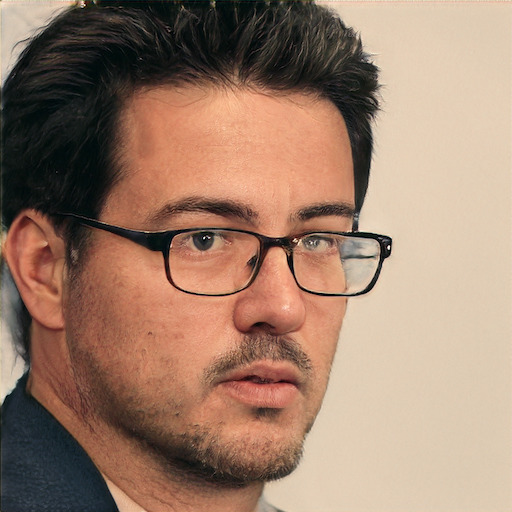}&
        \includegraphics[width=0.11\textwidth]{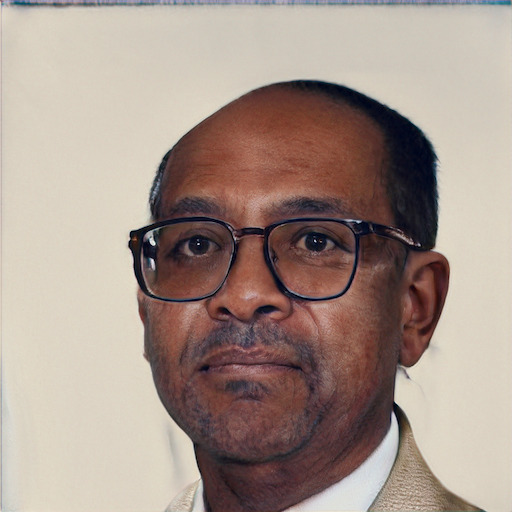}&
        \includegraphics[width=0.11\textwidth]{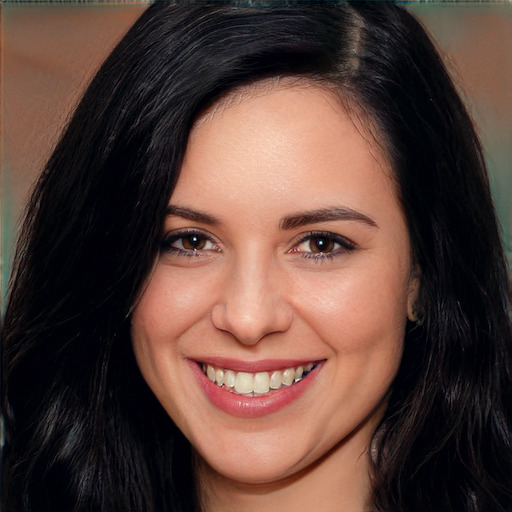}&        
        \includegraphics[width=0.11\textwidth]{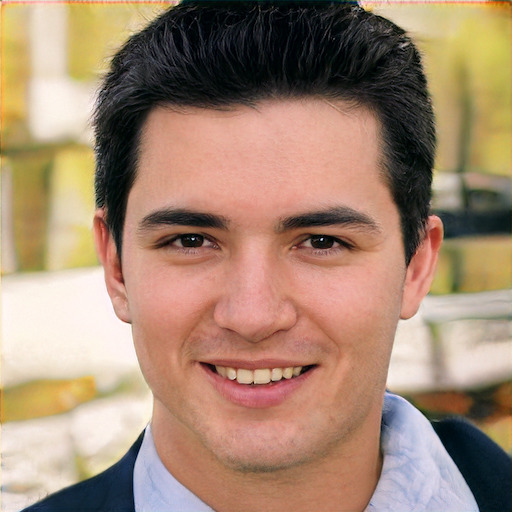}&
        \includegraphics[width=0.11\textwidth]{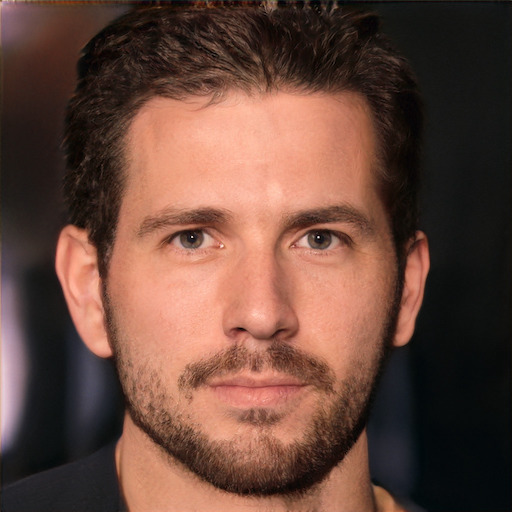} \tabularnewline
        \raisebox{0.29in}{\rotatebox[origin=t]{90}{IDInvert~\cite{zhu2020domain}}}&
        \includegraphics[width=0.11\textwidth]{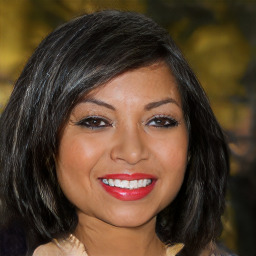}&
        \includegraphics[width=0.11\textwidth]{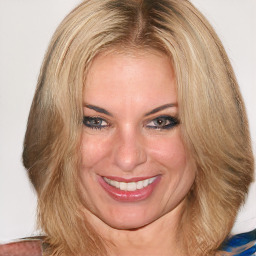}& 
        \includegraphics[width=0.11\textwidth]{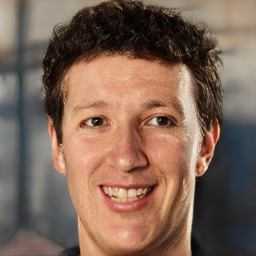}&
        \includegraphics[width=0.11\textwidth]{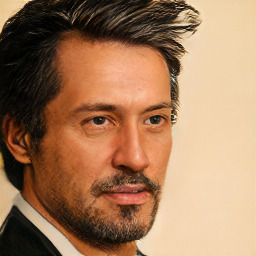}&
        \includegraphics[width=0.11\textwidth]{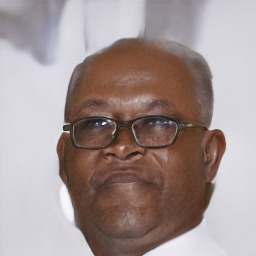}&
        \includegraphics[width=0.11\textwidth]{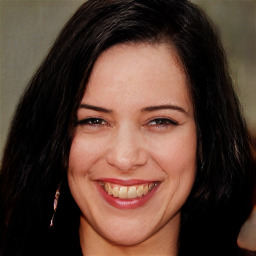}&      
        \includegraphics[width=0.11\textwidth]{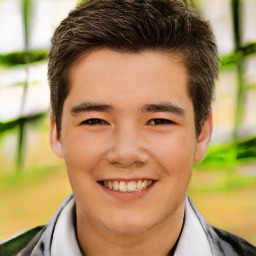}&
        \includegraphics[width=0.11\textwidth]{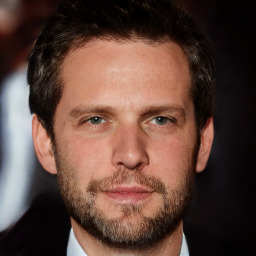}
        \tabularnewline
        \raisebox{0.25in}{\rotatebox[origin=t]{90}{pSp}}&
        \includegraphics[width=0.11\textwidth]{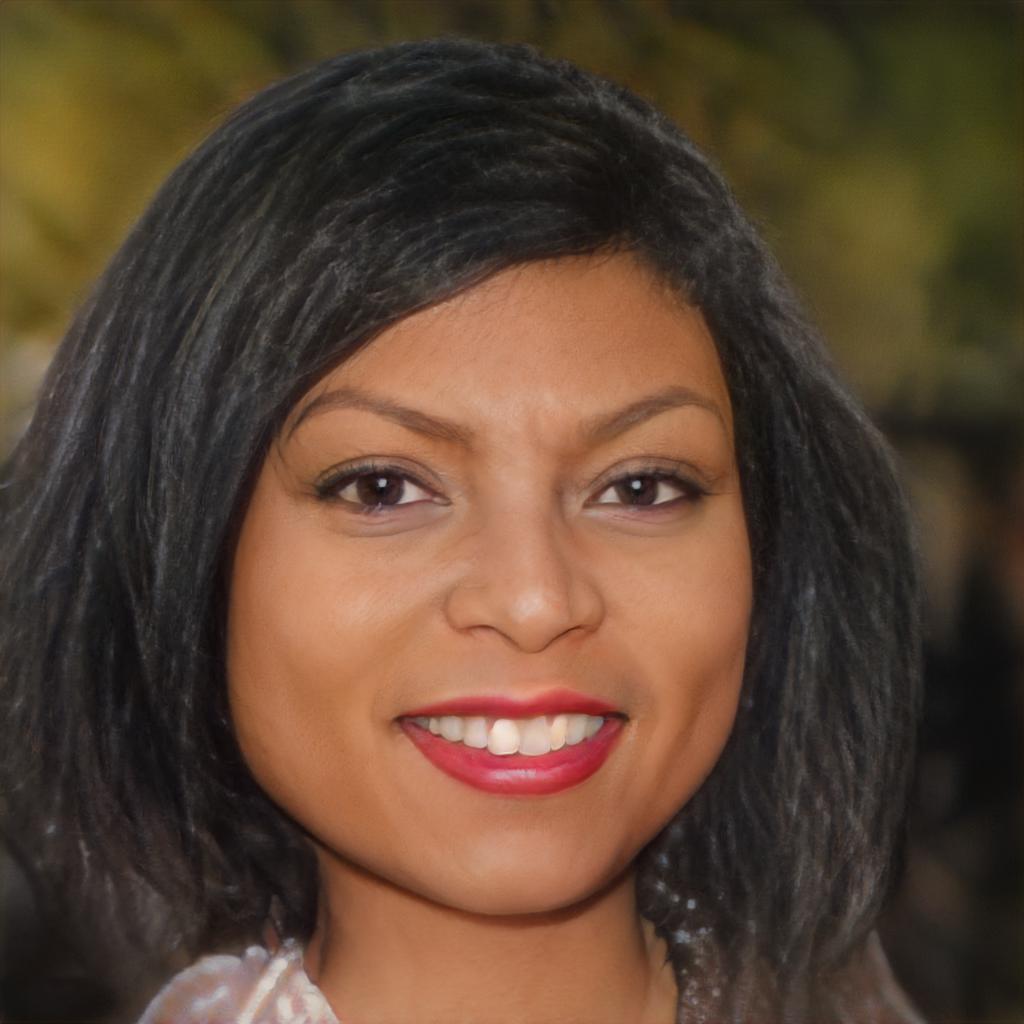}&
        \includegraphics[width=0.11\textwidth]{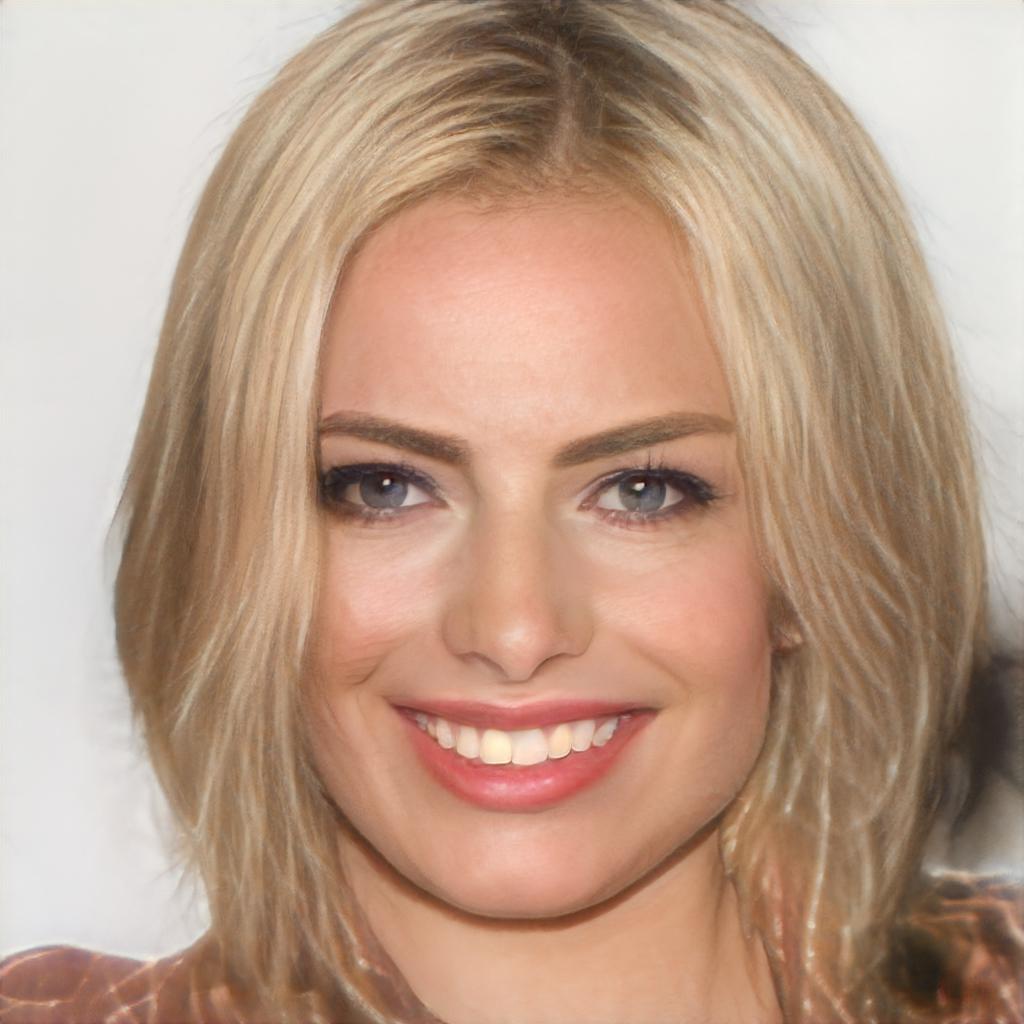}&        
        \includegraphics[width=0.11\textwidth]{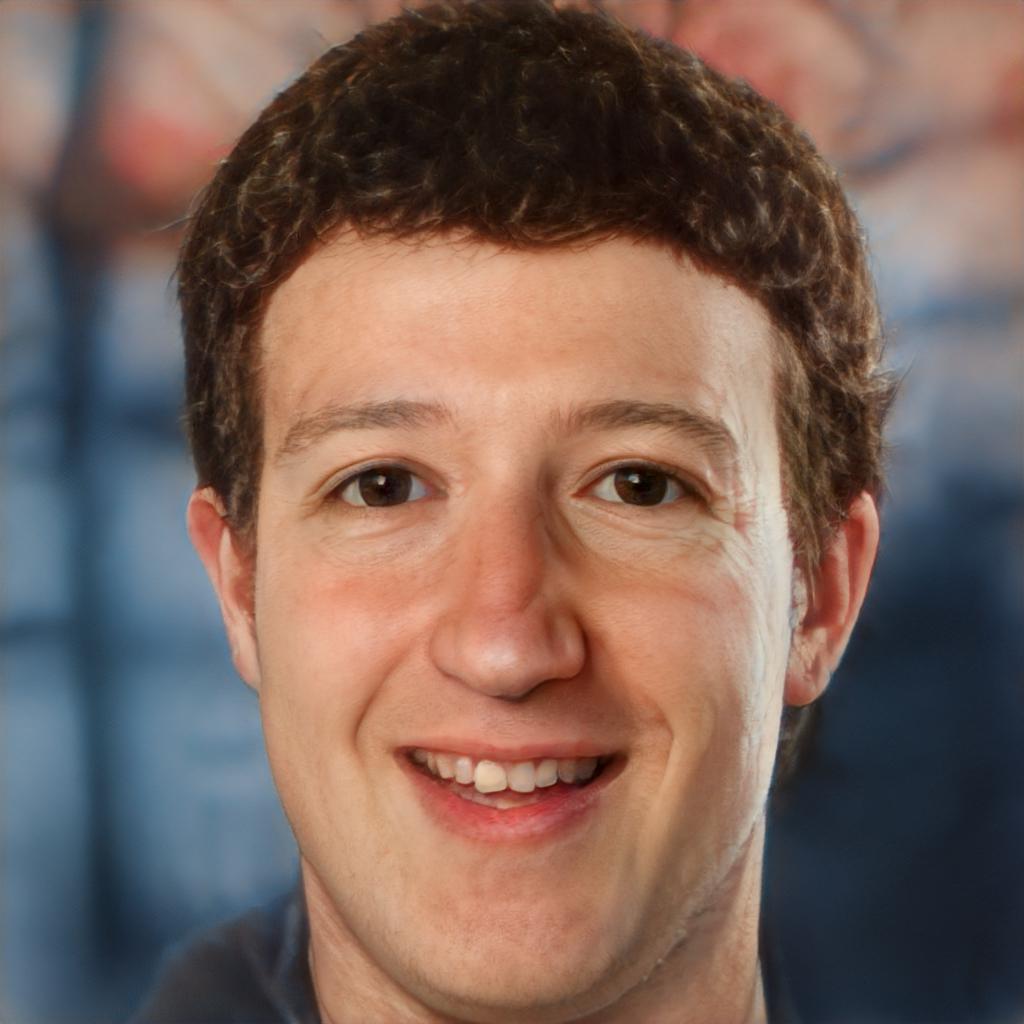}&
        \includegraphics[width=0.11\textwidth]{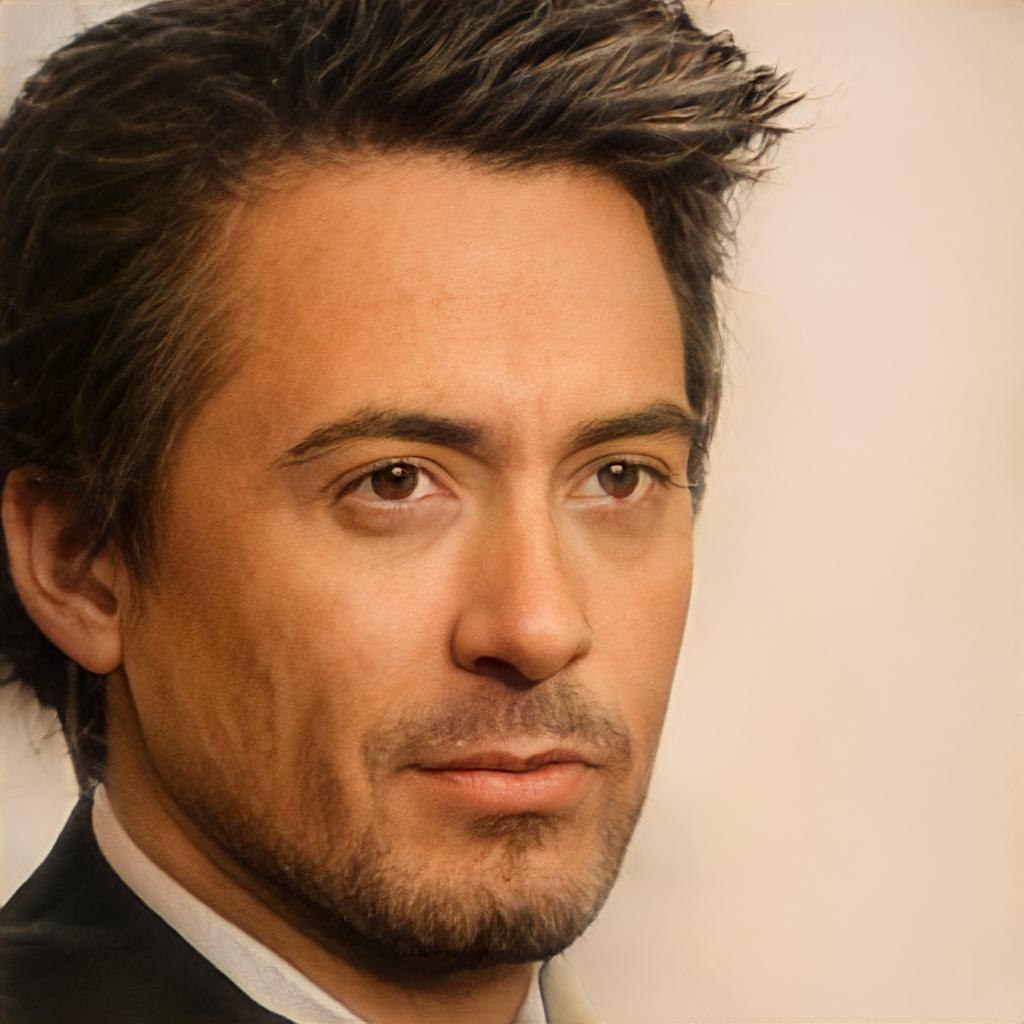}&
        \includegraphics[width=0.11\textwidth]{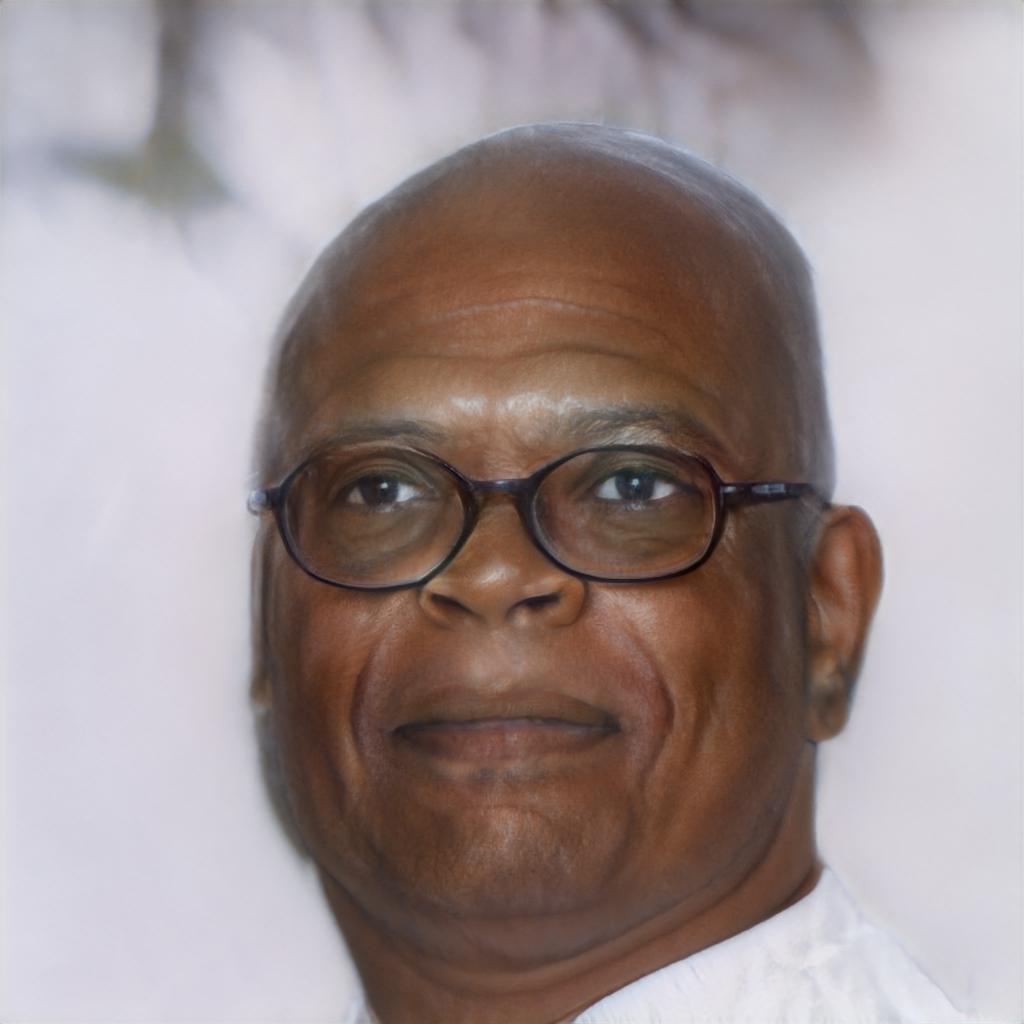}&
        \includegraphics[width=0.11\textwidth]{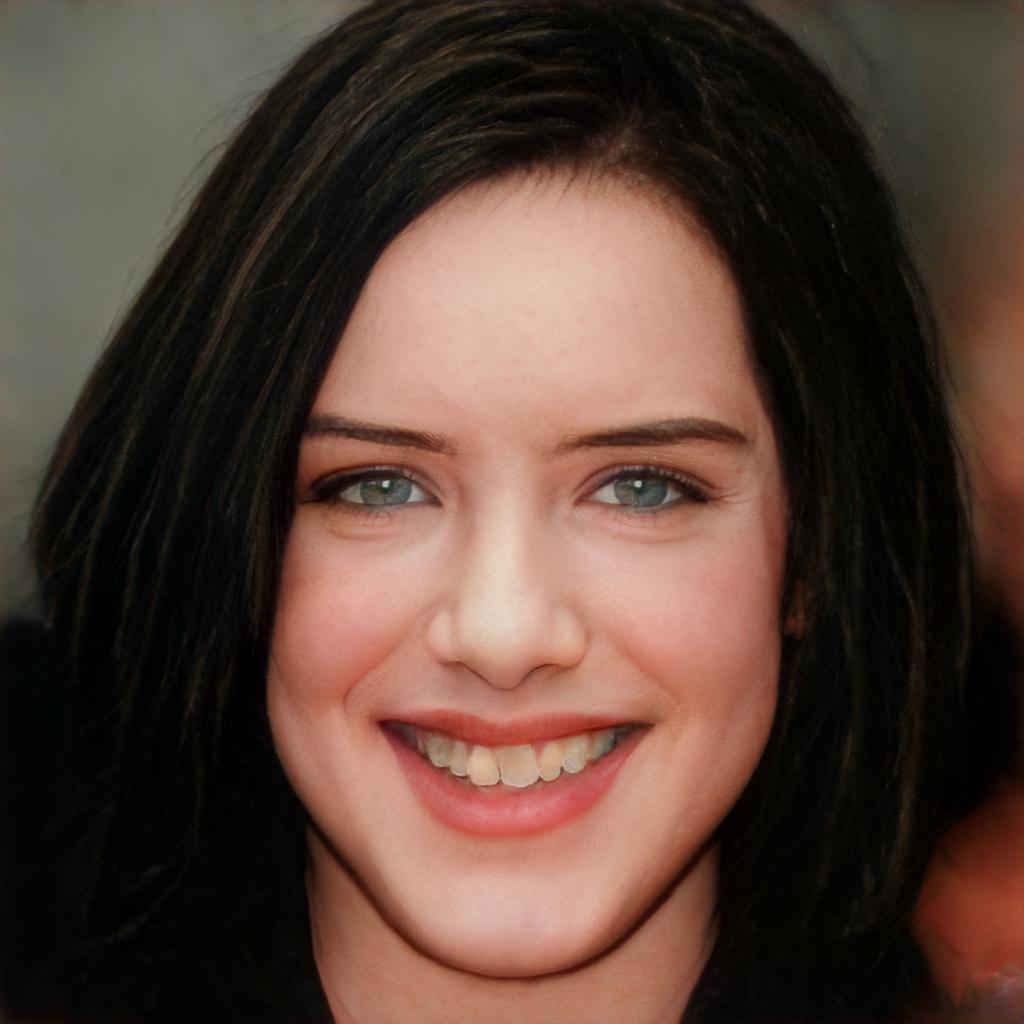}&        
        \includegraphics[width=0.11\textwidth]{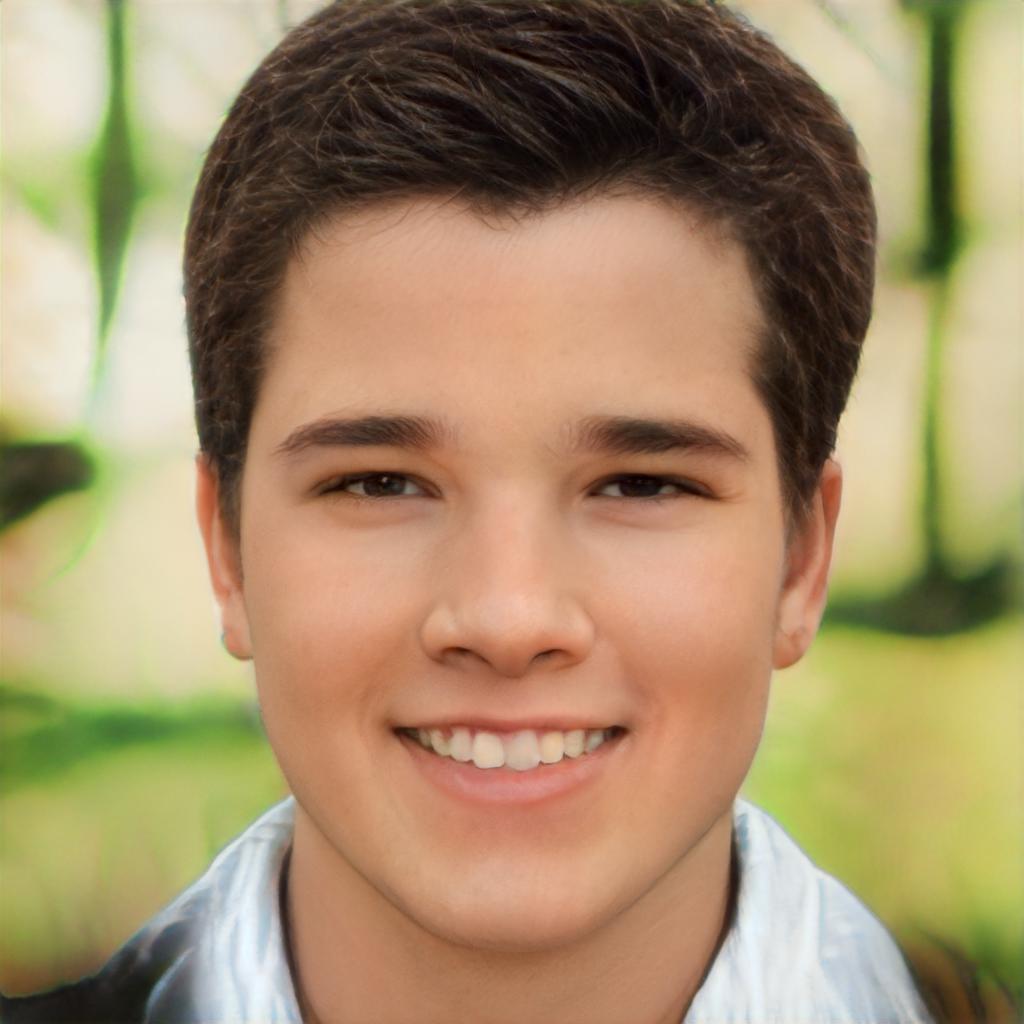}&
        \includegraphics[width=0.11\textwidth]{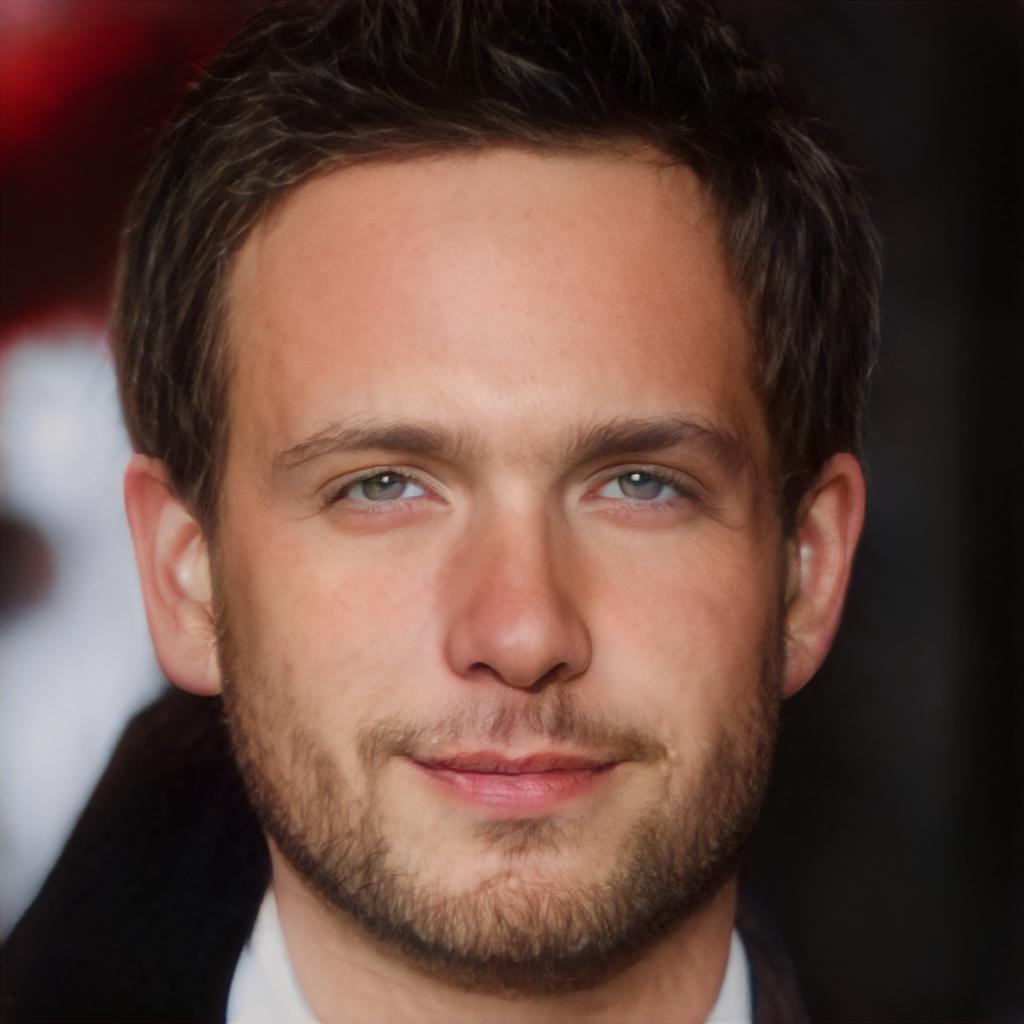} 
    \end{tabular}
    }
    \caption{Results of pSp for StyleGAN inversion compared to other encoders on CelebA-HQ.}
    \vspace{-0.25cm}
    \label{fig:encoder_compare}
\end{figure*}

To explore the effectiveness of our approach we evaluate pSp  on numerous image-to-image translation tasks. 

\subsection{StyleGAN Inversion}\label{inversion}
We start by evaluating the usage of the pSp framework for StyleGAN Inversion, that is, finding the latent code of real images in the latent domain. We compare our method to the optimization technique from Karras \etal~\cite{karras2020analyzing}, the ALAE encoder~\cite{pidhorskyi2020adversarial} and to the encoder from IDInvert~\cite{zhu2020domain}. The ALAE method proposes a StyleGAN-based autoencoder, where the encoder is trained alongside the generator to generate latent codes. In IDInvert, images are embedded into the latent domain of a pretrained StyleGAN by first encoding the image into $\mathcal{W+}$ and then directly optimizing over the generated image to tune the latent. For a fair comparison, we compare with IDInvert where no further optimization is performed after encoding.

\vspace{-0.25cm}
\paragraph{\textit{\textbf{Results.}}} 
Figure ~\ref{fig:encoder_compare} shows a  qualitative comparison between the methods. One can see that the ALAE method, operating in the $\mathcal{W}$ domain, cannot accurately reconstruct the input images. While IDInvert~\cite{zhu2020domain} better preserves the image attributes, it still fails to accurately preserve identity and the finer details of the input image. In contrast, our method is able to preserve identity while also reconstructing fine details such as lighting, hairstyle, and glasses.  

\begin{figure}
    \setlength{\tabcolsep}{1pt}
    \centering
        {\small
        \begin{tabular}{c c c c}
                \includegraphics[width=0.11\textwidth]{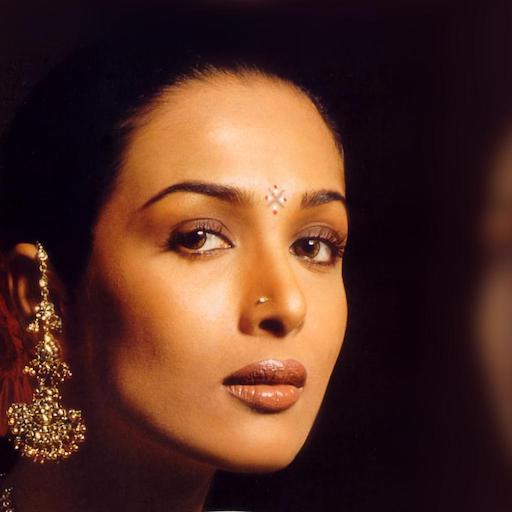}&
                \includegraphics[width=0.11\textwidth]{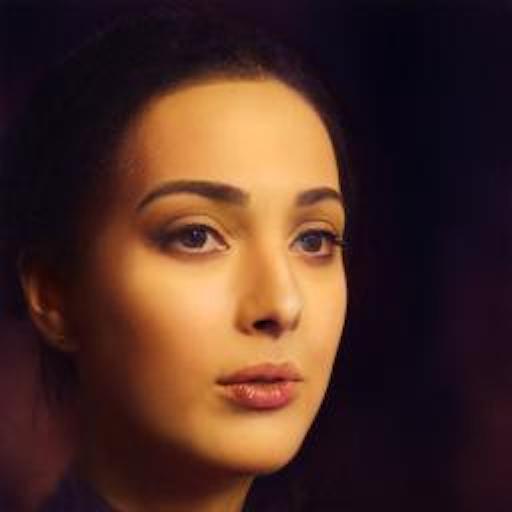}&
                \includegraphics[width=0.11\textwidth]{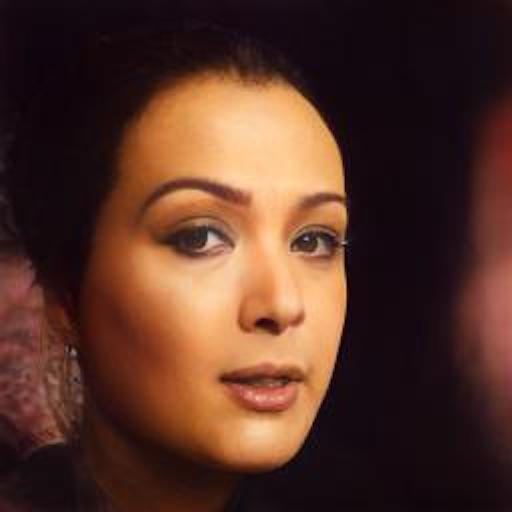}&
                \includegraphics[width=0.11\textwidth]{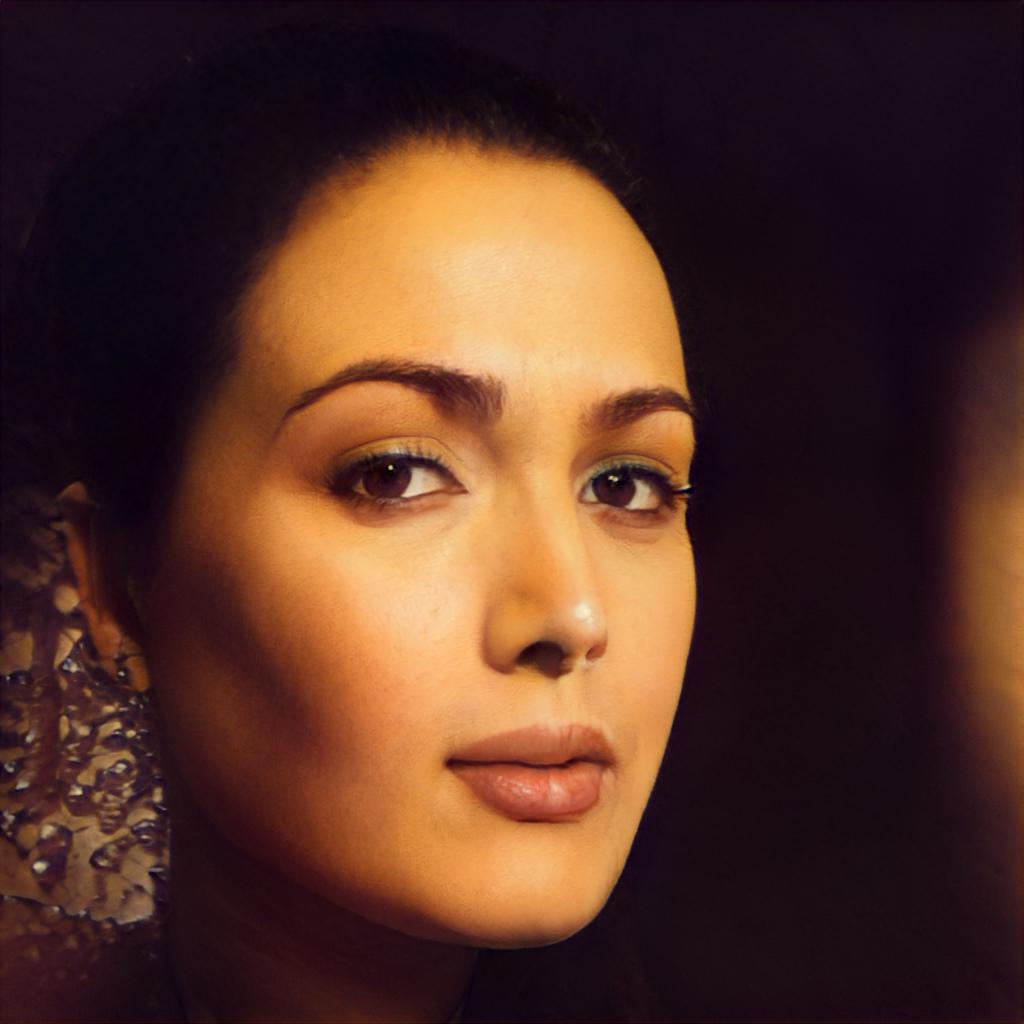}        
                \tabularnewline
                \includegraphics[width=0.11\textwidth]{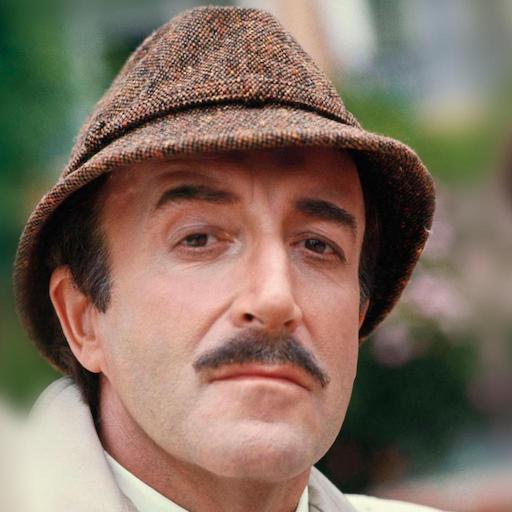}&
                \includegraphics[width=0.11\textwidth]{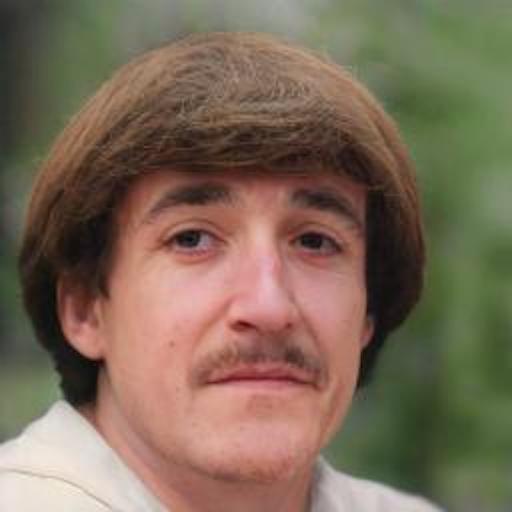}&
                \includegraphics[width=0.11\textwidth]{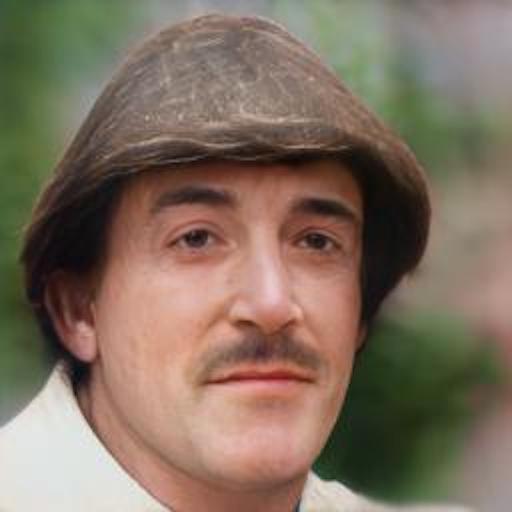}&
                \includegraphics[width=0.11\textwidth]{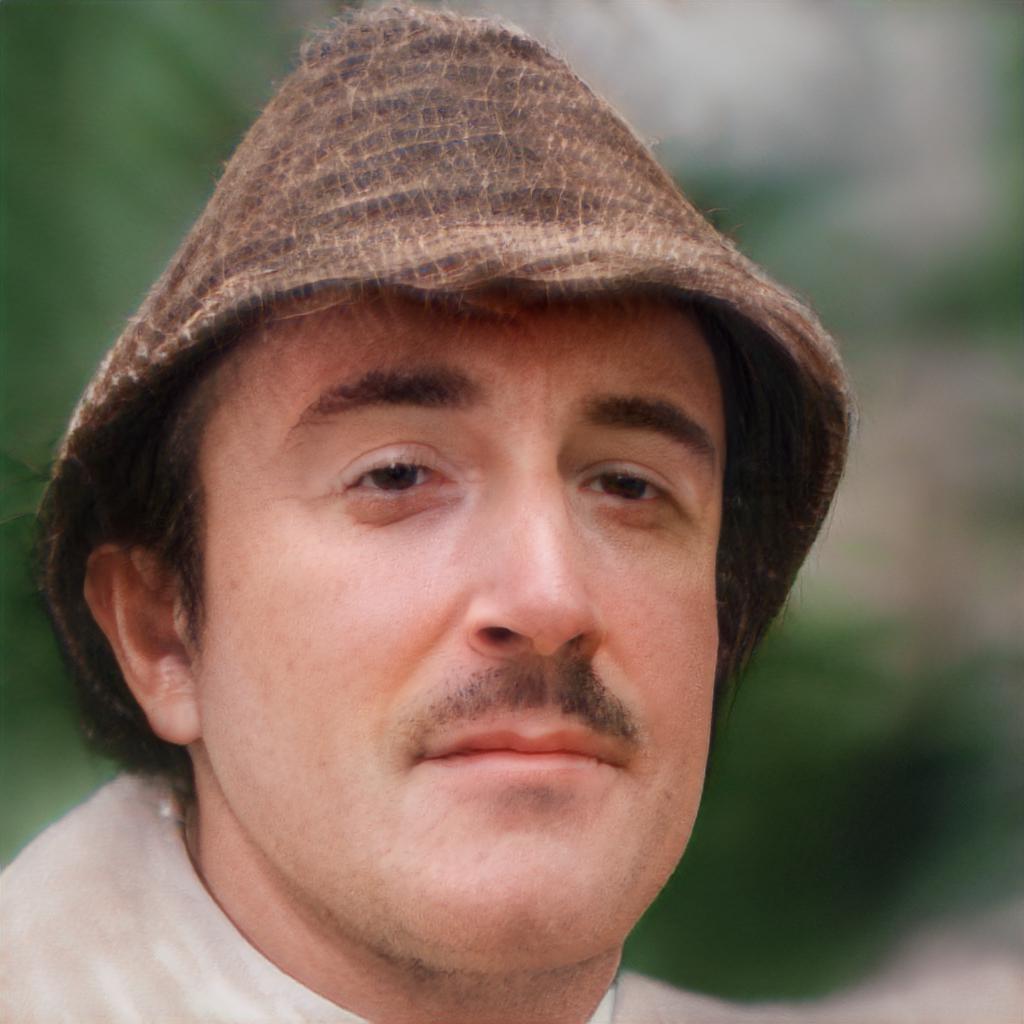}        
                \tabularnewline
                \includegraphics[width=0.11\textwidth]{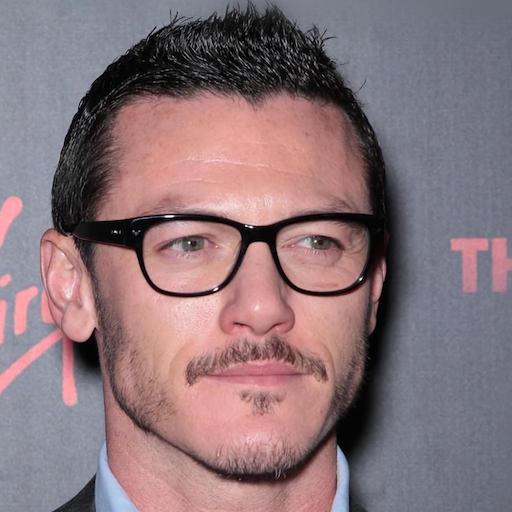}&
                \includegraphics[width=0.11\textwidth]{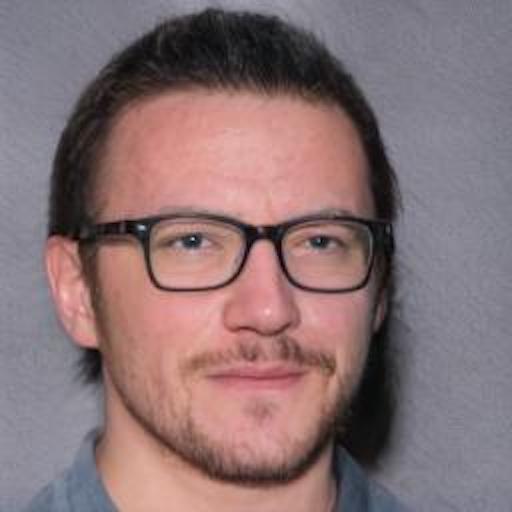}&
                \includegraphics[width=0.11\textwidth]{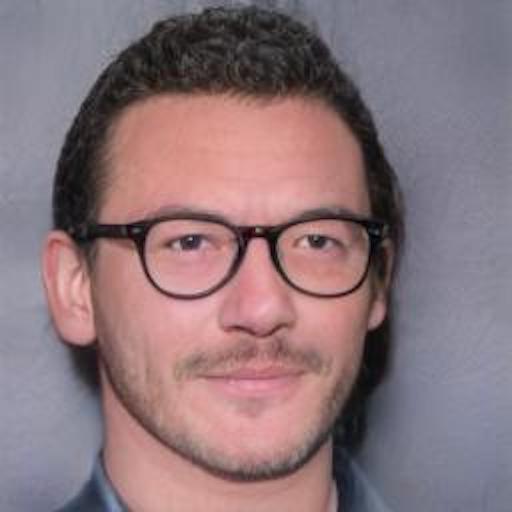}&
                \includegraphics[width=0.11\textwidth]{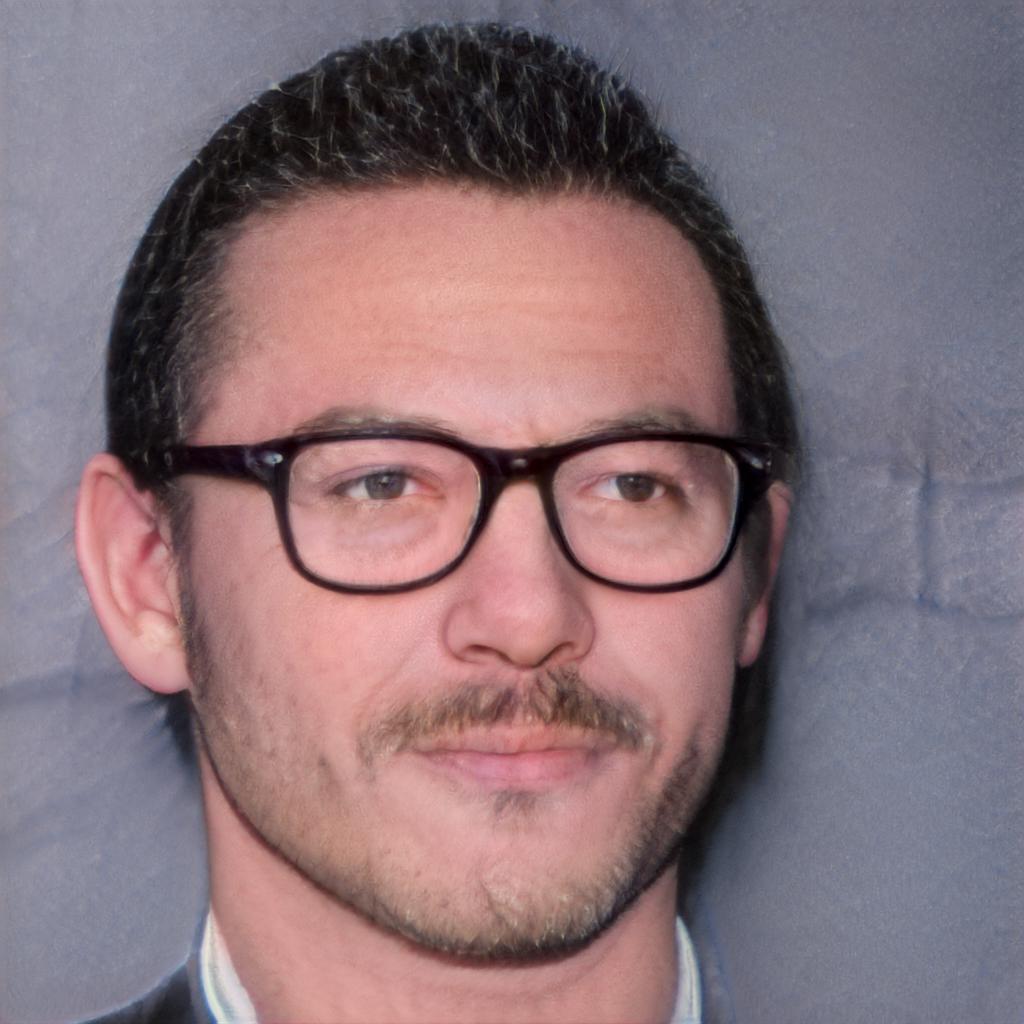}
                \tabularnewline
                Input & $\mathcal{W}$ & Naive $\mathcal{W+}$ & pSp
            \end{tabular}
            }
        \vspace{0.1cm}
        \caption{Ablation of the pSp encoder over CelebA-HQ.}
        \label{fig:encoder_ablation}
    \end{figure}

\begin{figure}
\setlength{\tabcolsep}{2pt}
\centering
    {\small
          \begin{tabular}{c c c}
                \includegraphics[width=0.126\textwidth]{images/encoder_2/2611_gt.jpg}&
                \includegraphics[width=0.126\textwidth]{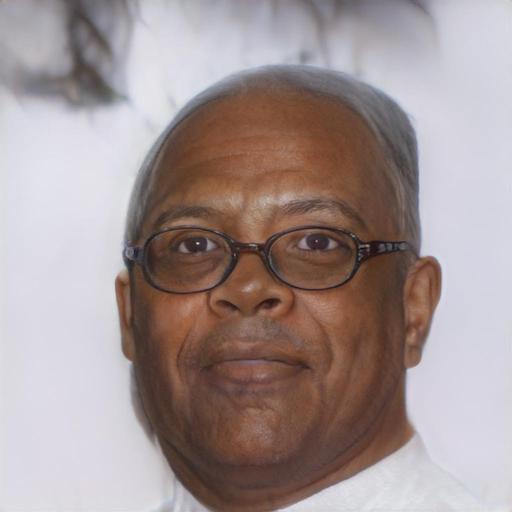}&
                \includegraphics[width=0.126\textwidth]{images/encoder_2/2611_ours.jpg}
                \tabularnewline
                \includegraphics[width=0.126\textwidth]{images/encoder_2/2468_gt.jpg}&
                \includegraphics[width=0.126\textwidth]{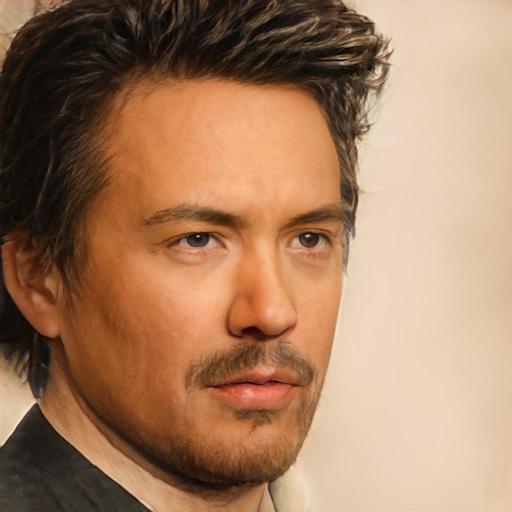}&
                \includegraphics[width=0.126\textwidth]{images/encoder_2/2468_ours.jpg}
                \tabularnewline
                Input & pSp w/o ID &  pSp w/ ID
            \end{tabular}
            }
    \vspace{0.1cm}
    \caption{The importance of identity loss.}
    \label{fig:identity}
\end{figure}

Next, we conduct an ablation study to analyze the effectiveness of the pSp architecture. We compare our architecture to two simpler variations. First, we define an encoder generating a $512$-dimensional style vector in the $\mathcal{W}$ latent domain, extracted from the last layer of the encoder network. We then expand this and define an encoder with an additional layer to transform the $512$-dimensional feature vector to a full $18\times512$ $\mathcal{W+}$ vector. Figure~\ref{fig:encoder_ablation} shows that while this simple extension into $\mathcal{W+}$ significantly improves the results, it still cannot preserve the finer details generated by our architecture. 
In Figure~\ref{fig:identity} we show the importance of the identity loss in the reconstruction task.

Finally, Table~\ref{tb:qouant} presents a quantitative evaluation measuring the different inversion methods. Compared to other encoders, pSp is able to better preserve the original images in terms of both perceptual similarity and identity. To make sure the similarity score is independent of our loss function, we utilize the  CurricularFace~\cite{huang2020curricularface} method for evaluation.

\begin{table}
            \setlength{\tabcolsep}{1.5pt}
            \centering
            \begin{tabular}{l c c c c}
            \toprule
            Method & $\uparrow$ Similarity & $\downarrow$ LPIPS & $\downarrow$ MSE & $\downarrow$ Runtime \\
            \midrule
            Karras \etal~\cite{karras2020analyzing} &
            \multicolumn{1}{c}{$0.77$} &
            \multicolumn{1}{c}{$0.11$} &
            \multicolumn{1}{c}{$0.02$} &
            \multicolumn{1}{c}{$182.1$} \\ 
            ALAE~\cite{pidhorskyi2020adversarial} &
            \multicolumn{1}{c}{$0.06$} &
            \multicolumn{1}{c}{$0.32$} &
            \multicolumn{1}{c}{$0.15$} &
            \multicolumn{1}{c}{$0.207$} \\ 
            IDInvert~\cite{zhu2020domain} & 
            \multicolumn{1}{c}{$0.18$} &
            \multicolumn{1}{c}{$0.22$} &
            \multicolumn{1}{c}{$0.06$} &
            \multicolumn{1}{c}{0.032} \\
            \midrule
            $\mathcal{W}$ Encoder &
            \multicolumn{1}{c}{$0.35$} &
            \multicolumn{1}{c}{$0.23$} &
            \multicolumn{1}{c}{$0.06$} &
            \multicolumn{1}{c}{$0.064$} \\
            Naive $\mathcal{W+}$ & 
            \multicolumn{1}{c}{$0.49$} &
            \multicolumn{1}{c}{$0.19$} &
            \multicolumn{1}{c}{$0.04$} &
            \multicolumn{1}{c}{$0.064$}  \\
            pSp w/o ID & 
            \multicolumn{1}{c}{$0.19$} &  
            \multicolumn{1}{c}{$0.17$} & 
            \multicolumn{1}{c}{$0.03$} &
            \multicolumn{1}{c}{$0.105$} \\
            pSp & 
            \multicolumn{1}{c}{0.56} &  
            \multicolumn{1}{c}{0.17} & 
            \multicolumn{1}{c}{0.03} &
            \multicolumn{1}{c}{0.105} \\
            \bottomrule
            \end{tabular}
            \vspace{0.1cm}
            \caption{Quantitative results for image reconstruction.}
            \vspace{-0.1cm}
            \label{tb:qouant}
\end{table}

\begin{figure}
        \setlength{\tabcolsep}{1pt}
        \centering
        {\small
            \begin{tabular}{c c c c}
                \includegraphics[width=0.11\textwidth]{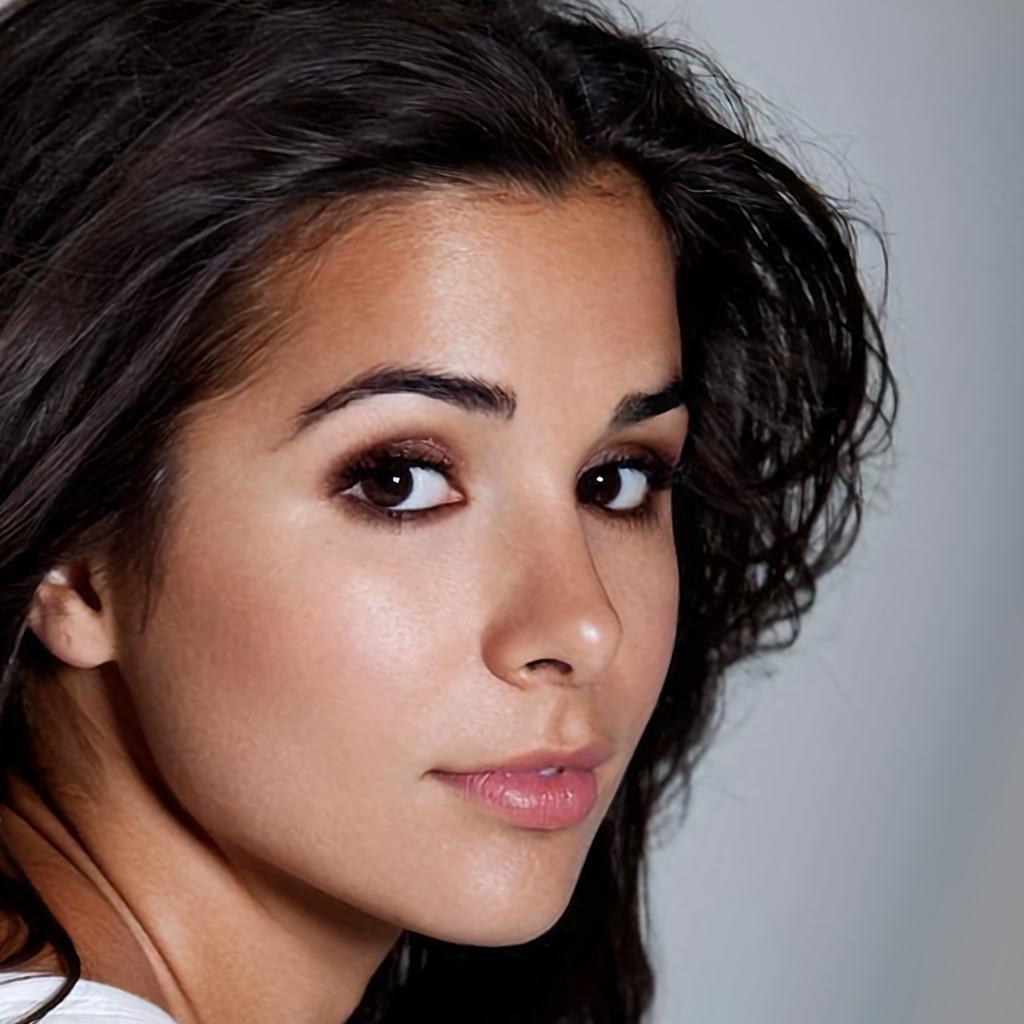}&
                \includegraphics[width=0.11\textwidth]{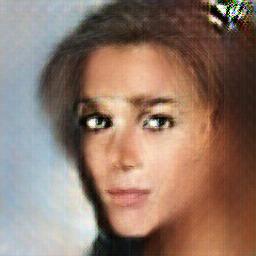}&        
                \includegraphics[width=0.11\textwidth]{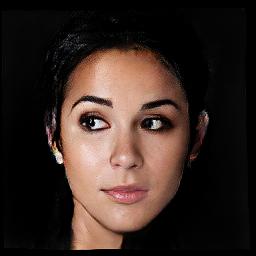}
                &
                \includegraphics[width=0.11\textwidth]{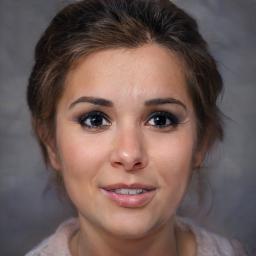}
                \tabularnewline
                
                \includegraphics[width=0.11\textwidth]{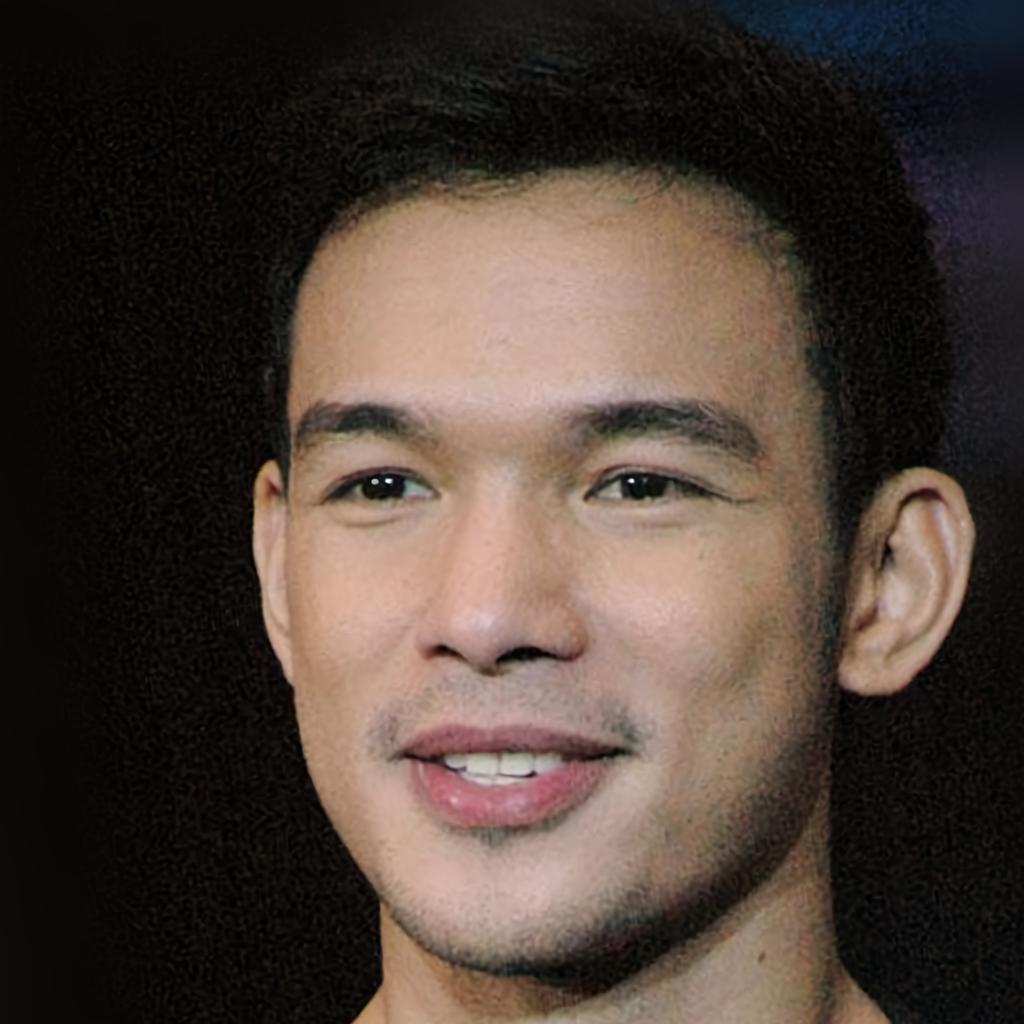}&
                \includegraphics[width=0.11\textwidth]{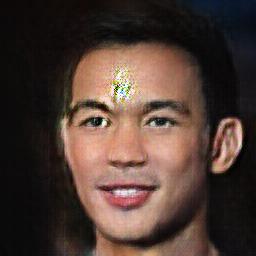}&        
                \includegraphics[width=0.11\textwidth]{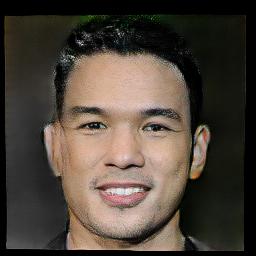}
                &
                \includegraphics[width=0.11\textwidth]{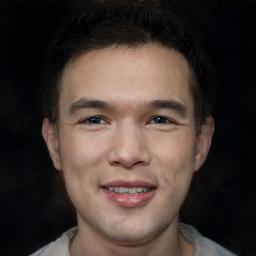}
                \tabularnewline
                
                 \includegraphics[width=0.11\textwidth]{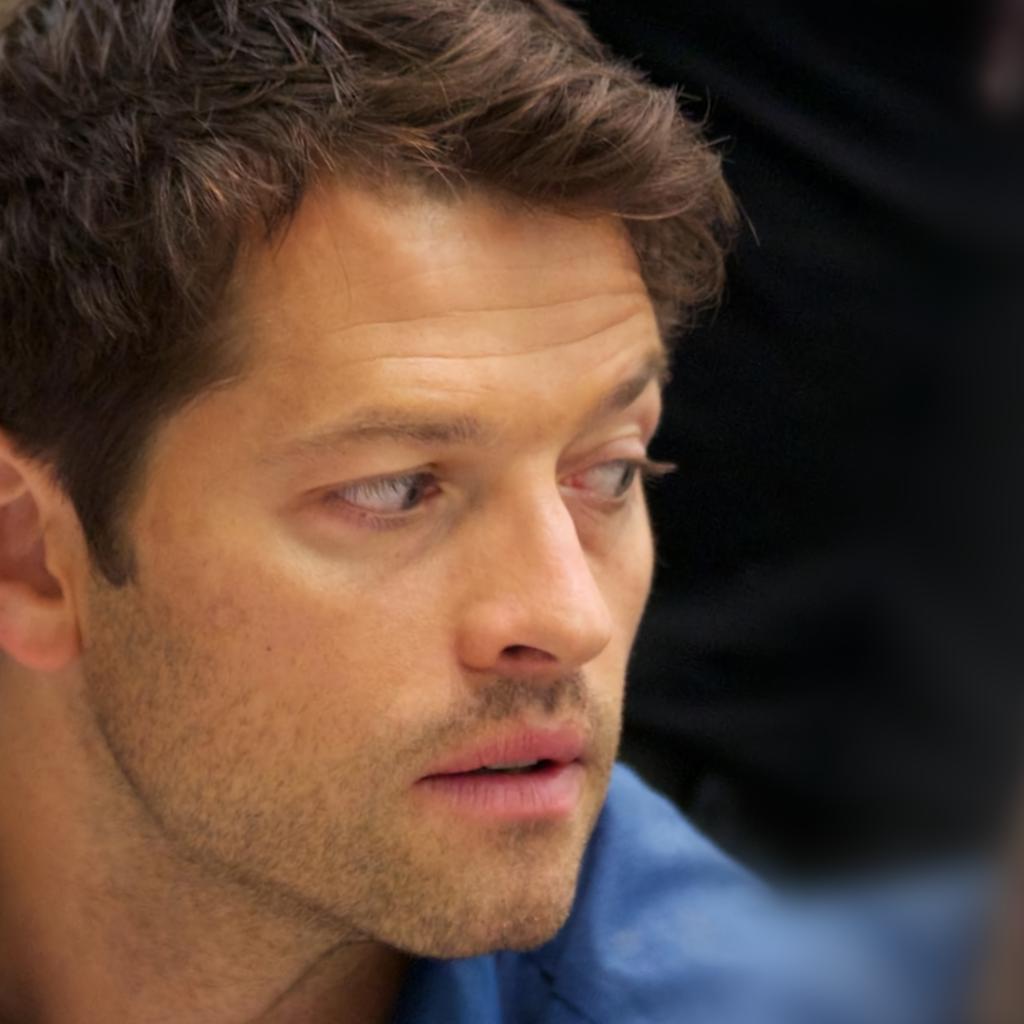}&
                \includegraphics[width=0.11\textwidth]{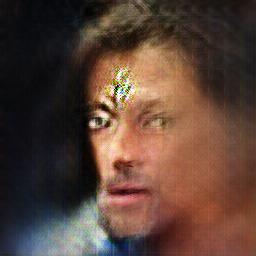}&        
                \includegraphics[width=0.11\textwidth]{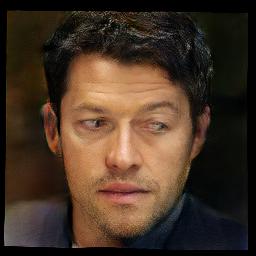}
                &
                \includegraphics[width=0.11\textwidth]{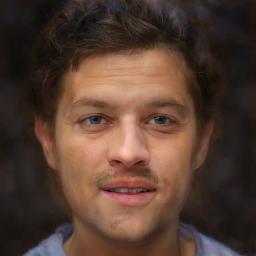}
                \tabularnewline
                
                 \includegraphics[width=0.11\textwidth]{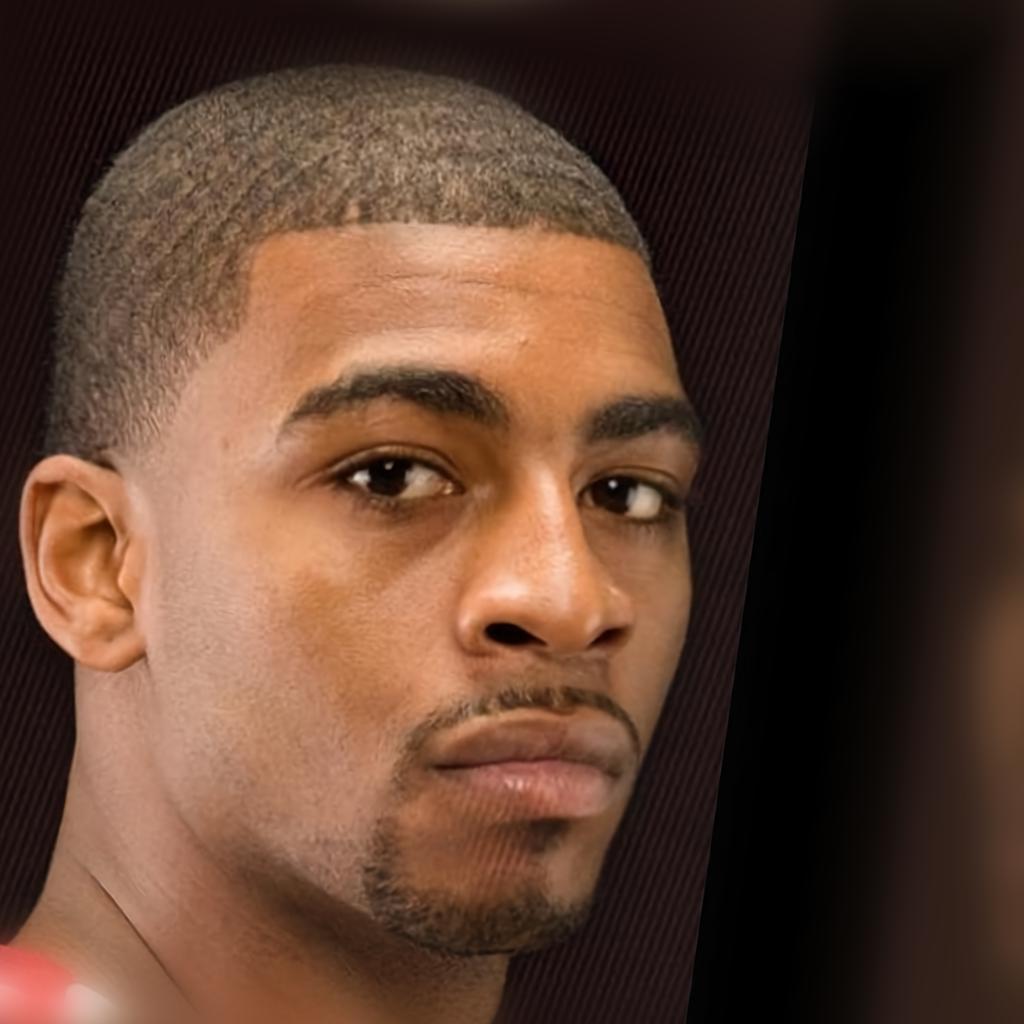}&
                \includegraphics[width=0.11\textwidth]{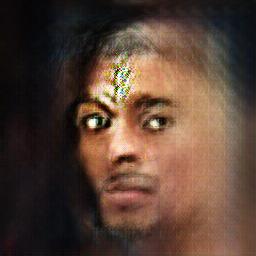}&        
                \includegraphics[width=0.11\textwidth]{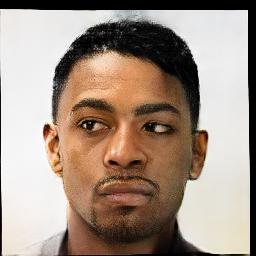}
                &
                \includegraphics[width=0.11\textwidth]{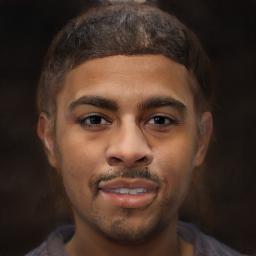}
                \tabularnewline
                
                 \includegraphics[width=0.11\textwidth]{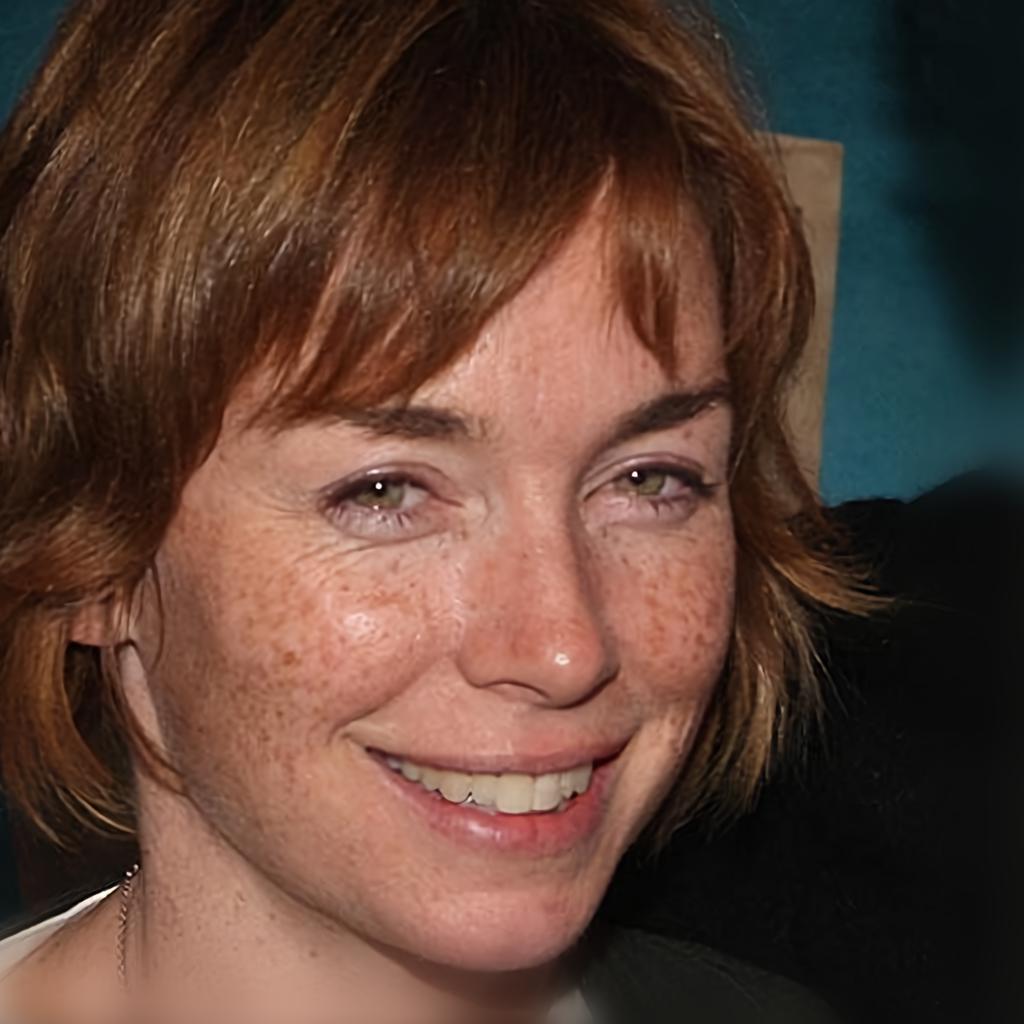}&
                \includegraphics[width=0.11\textwidth]{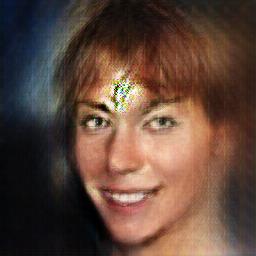}&        
                \includegraphics[width=0.11\textwidth]{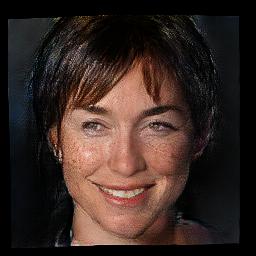}
                &
                \includegraphics[width=0.11\textwidth]{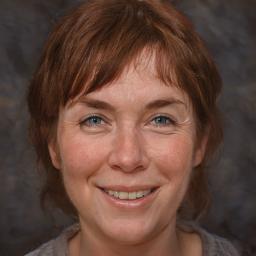}
                \tabularnewline
                
                 \includegraphics[width=0.11\textwidth]{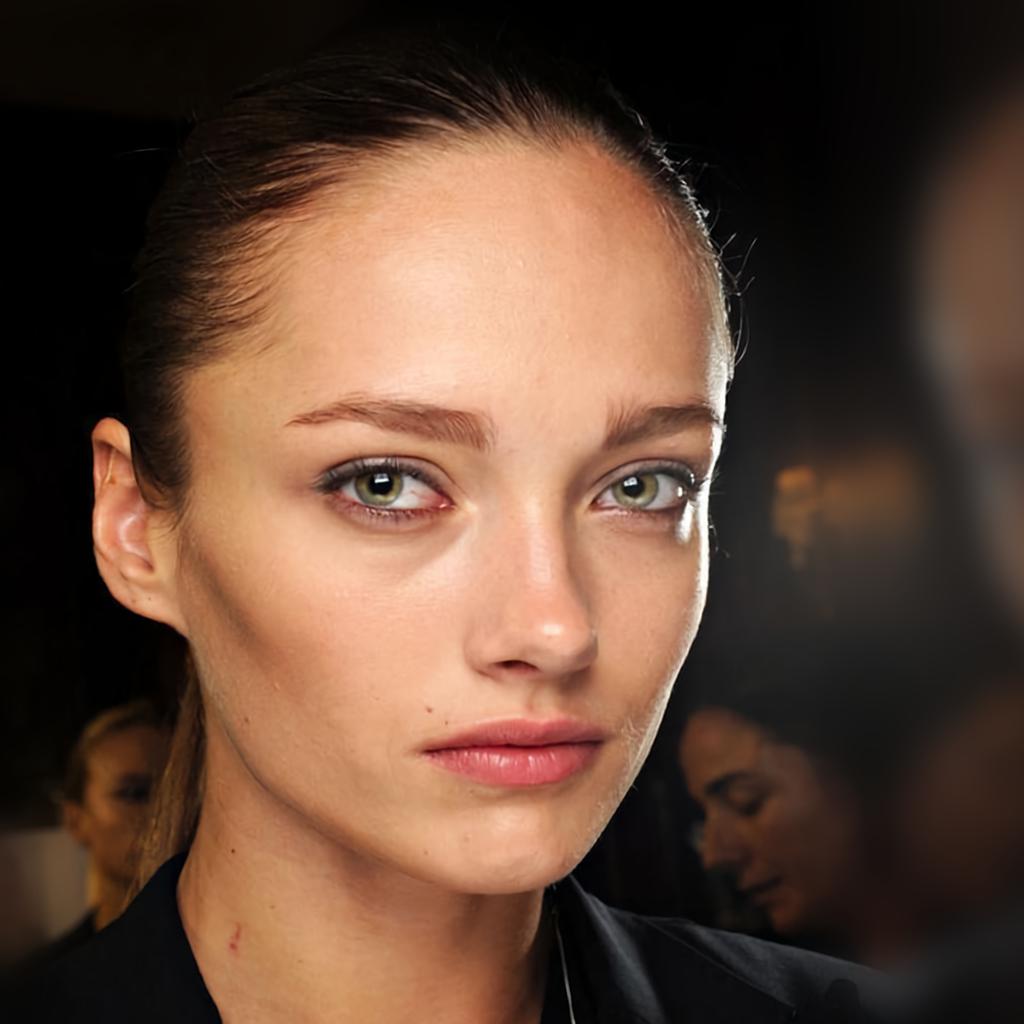}&
                \includegraphics[width=0.11\textwidth]{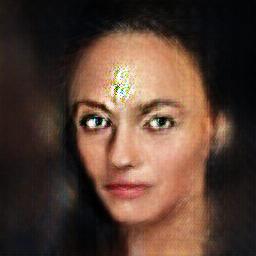}&        
                \includegraphics[width=0.11\textwidth]{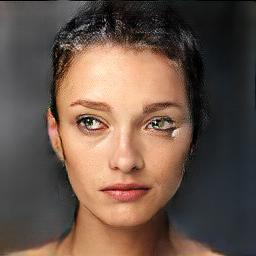}
                &
                \includegraphics[width=0.11\textwidth]{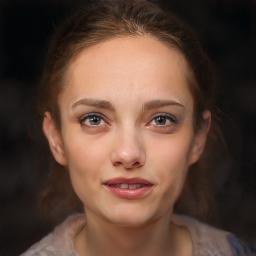}
                \tabularnewline

                Input & pix2pixHD & R\&R & pSp
            \end{tabular}
            }
            \vspace{0.1cm}
            \caption{Comparison of face frontalization methods.}
            \vspace{-0.3cm}
            \label{fig:front_compare}
        \end{figure}

\vspace{-0.1cm}
\subsection{Face Frontalization} \label{app-frontalization}
Face frontalization is a challenging task for image-to-image translation frameworks due to the required non-local transformations and the lack of paired training data. RotateAndRender (R\&R)~\cite{zhou2020rotate} overcome this challenge by incorporating a geometric 3D alignment process before the translation process. Alternatively, we show that our style-based translation mechanism is able to overcome these challenges, even when trained with no labeled data.

\vspace{-0.1cm}
\paragraph{\textit{\textbf{Methodology.}}} For this task, training is the same as the encoder formulation with two important changes.
First, we randomly flip the target image during training, effectively forcing the model to output an image that is close to both the original image and the mirrored one. The underlying idea behind this augmentation is that it guides the model to converge to a fixed frontal pose.
Next, we increase $\mathcal{L}_{\text{ID}}$ and decrease the $\mathcal{L}_2$ and $\mathcal{L}_{\text{LPIPS}}$ losses for the outer part of the image. This change is based on the fact that for frontalization we are less interested in preserving the background region compared to the face region and the facial identity.

\begin{table}
    \setlength{\tabcolsep}{4.2pt}
    \centering
    \begin{tabular}{l c c c c c}
    \toprule
    Method & \multicolumn{4}{c}{$\uparrow$ Similarity} &  $\downarrow$ Runtime \\
     & $\ang{90}$ & $\ang{70}$ & $\ang{50}$ & $\ang{30}$ \\
    \midrule
    R\&R & 
    \multicolumn{1}{c}{\boldmath{$0.34$}} &
    \multicolumn{1}{c}{\boldmath{$0.56$}} &
    \multicolumn{1}{c}{\boldmath{$0.66$}} &
    \multicolumn{1}{c}{\boldmath{$0.7$}} &
    \multicolumn{1}{c}{$1.5s$} \\
    \midrule
    pSp &
    \multicolumn{1}{c}{$0.32$} &
    \multicolumn{1}{c}{$0.52$} &
    \multicolumn{1}{c}{$0.60$} &
    \multicolumn{1}{c}{$0.63$} &
    \multicolumn{1}{c}{\boldmath{$0.1s$}} \\
    \bottomrule
    
    \end{tabular}
    \vspace{0.1cm}
    \caption{Results for Face Frontalization on the FEI Face Database split by rotation angle of the face in the input.}
    \vspace{-0.2cm}
    \label{tb:frontalization_quantitative}
\end{table}

\paragraph{\textit{\textbf{Results.}}} Results are illustrated in Figure~\ref{fig:front_compare}. When trained with the same data and methodology, pix2pixHD is unable to converge to satisfying results as it is much more dependent on the correspondence between the input and output pairs. Conversely, our method is able to handle the task successfully, generating realistic frontal faces, which are comparable to the more involved R\&R approach. This shows the benefit of using a pretrained StyleGAN for image translation, as it allows us to achieve visually-pleasing results even with weak supervision. 
Table~\ref{tb:frontalization_quantitative} provides a quantitative evaluation on the FEI Database \cite{THOMAZ2010902}. While R\&R outperforms pSp, our simple approach provides a fast and elegant alternative, without requiring specialized modules, such as R\&R's 3DMM fitting and inpainting steps.  

\begin{figure*}[t]
    \begin{minipage}{0.44\textwidth}
    \setlength{\tabcolsep}{1pt}
    \centering
    {\small
        \begin{tabular}{c c c c}
            \includegraphics[width=0.22\textwidth]{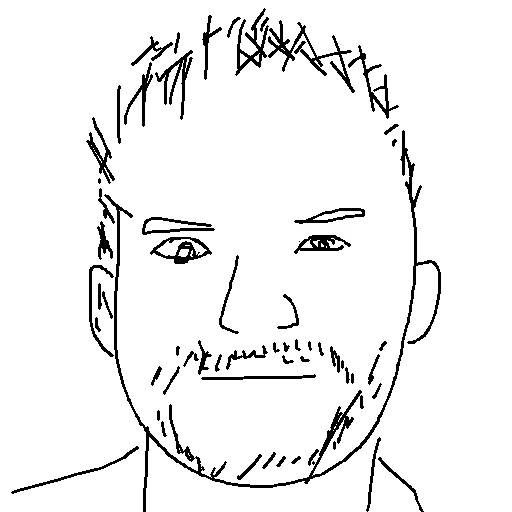}&
            \includegraphics[width=0.22\textwidth]{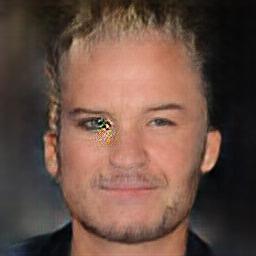}&
            \includegraphics[width=0.22\textwidth]{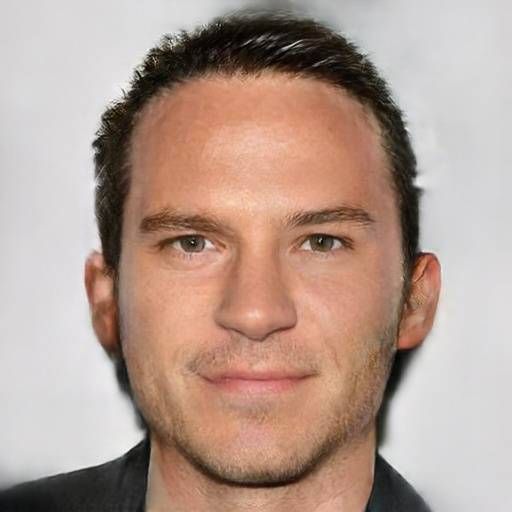}&
            \includegraphics[width=0.22\textwidth]{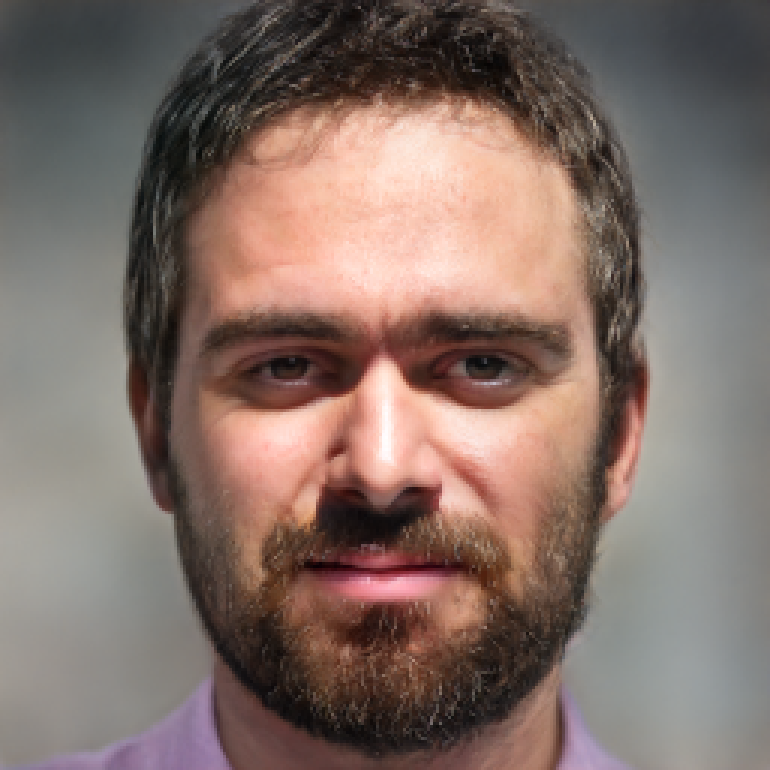}
            \vspace{-0.15em}
            \tabularnewline
            \includegraphics[width=0.22\textwidth]{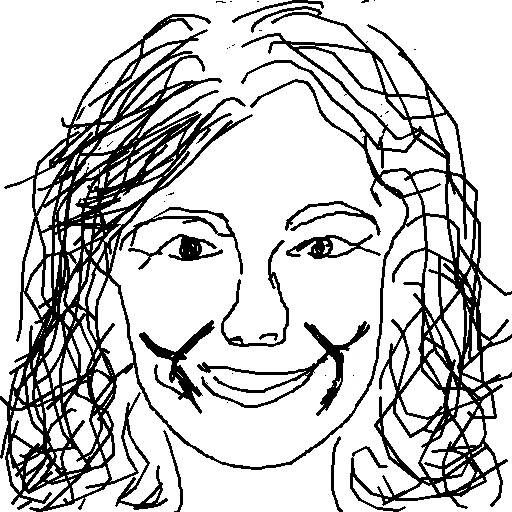}&
            \includegraphics[width=0.22\textwidth]{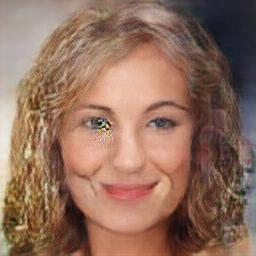}&
            \includegraphics[width=0.22\textwidth]{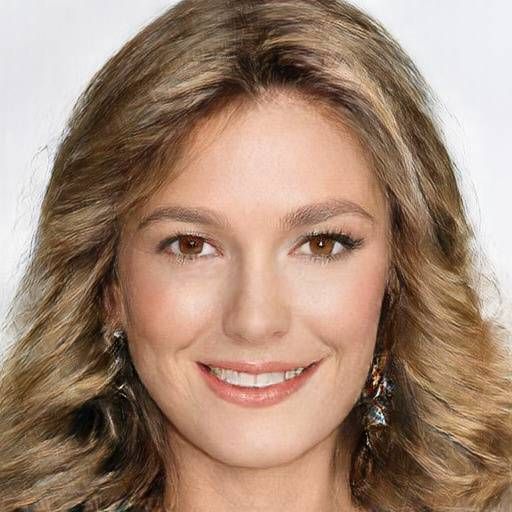}&
            \includegraphics[width=0.22\textwidth]{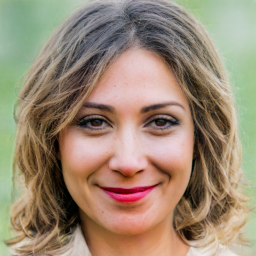}
            \vspace{-0.15em}
            \tabularnewline
            \includegraphics[width=0.22\textwidth]{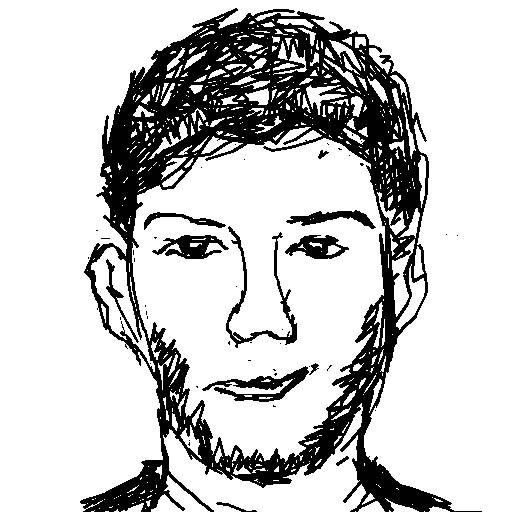}&
            \includegraphics[width=0.22\textwidth]{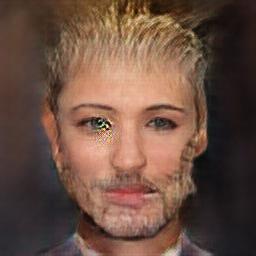}&
            \includegraphics[width=0.22\textwidth]{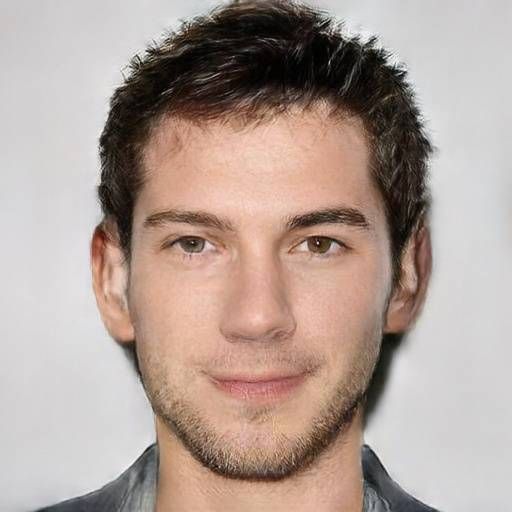}&
            \includegraphics[width=0.22\textwidth]{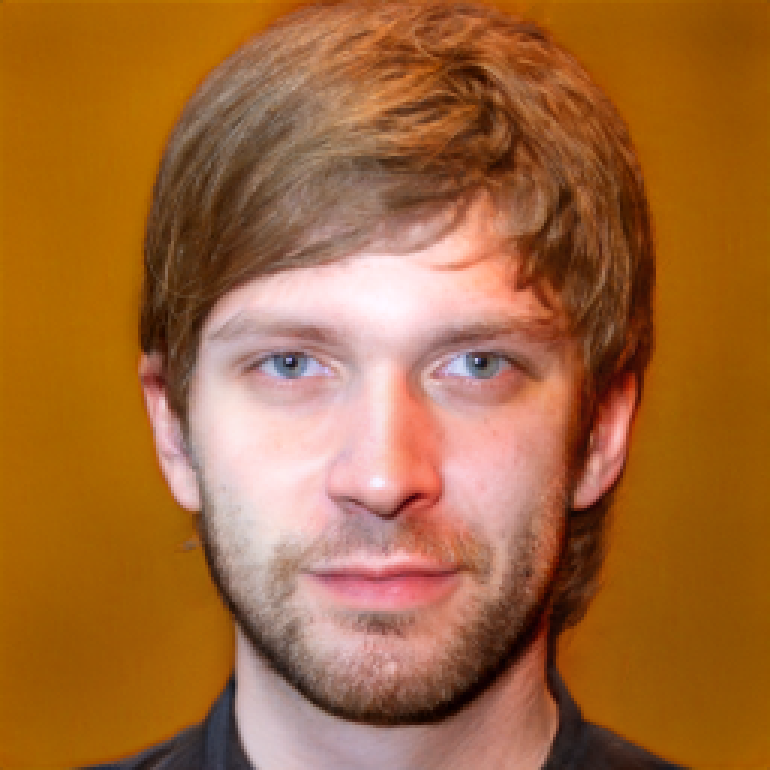}
            \vspace{-0.15em}
            \tabularnewline
            \includegraphics[width=0.22\textwidth]{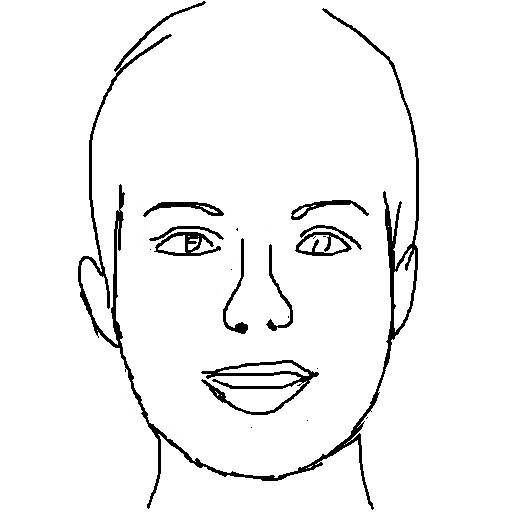}&
            \includegraphics[width=0.22\textwidth]{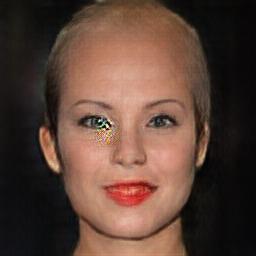}&
            \includegraphics[width=0.22\textwidth]{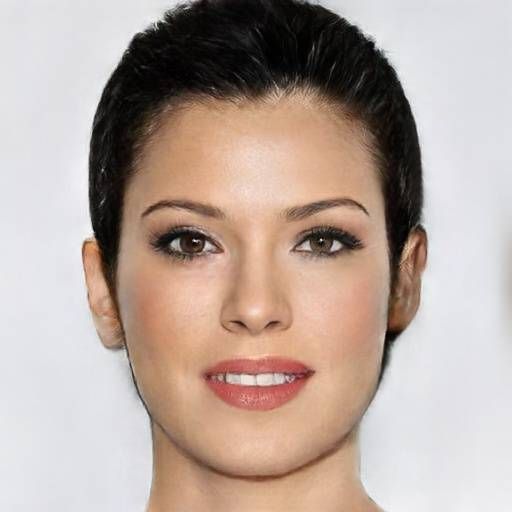}&
            \includegraphics[width=0.22\textwidth]{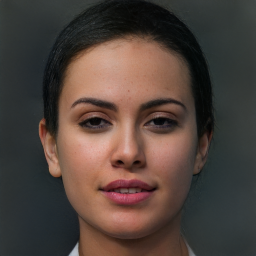}
            \tabularnewline
            Input & pix2pixHD & DeepFace & pSp
        \end{tabular}
        }
        \vspace{0.1cm}
        \caption{Comparison of sketches presented in DeepFaceDrawing~\cite{chen2020deep}.}
        \vspace{-0.2cm}
        \label{fig:sketch_comparison_deepfacedrawing}
\end{minipage}%
    \hspace{1.2em}
    \begin{minipage}{0.55\textwidth}
        \setlength{\tabcolsep}{1pt}
        {\small
        \begin{tabular}{c c c c c}
            \includegraphics[width=0.176\textwidth]{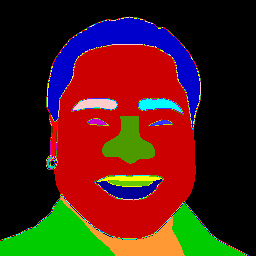}&
            \includegraphics[width=0.176\textwidth]{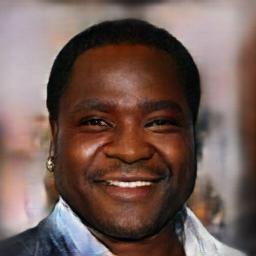}&
            \includegraphics[width=0.176\textwidth]{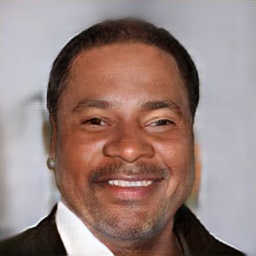}&
            \includegraphics[width=0.176\textwidth]{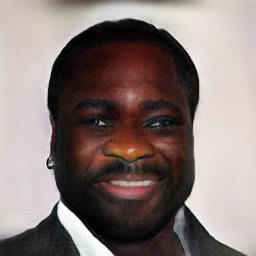}&
            \includegraphics[width=0.176\textwidth]{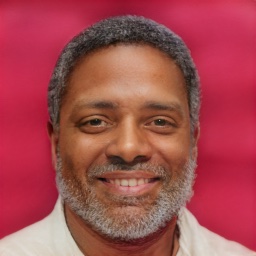}
            \vspace{-0.15em}
            \tabularnewline
            \includegraphics[width=0.176\textwidth]{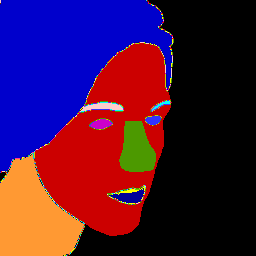}&
            \includegraphics[width=0.176\textwidth]{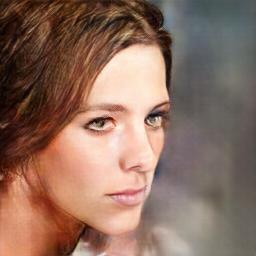}&
            \includegraphics[width=0.176\textwidth]{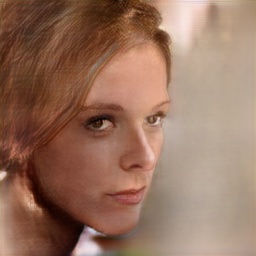}&
            \includegraphics[width=0.176\textwidth]{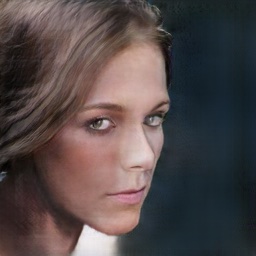}&
            \includegraphics[width=0.176\textwidth]{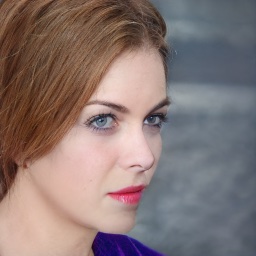}
            \vspace{-0.15em}
            \tabularnewline
            \includegraphics[width=0.176\textwidth]{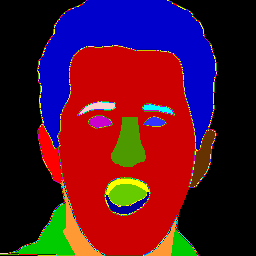}&
            \includegraphics[width=0.176\textwidth]{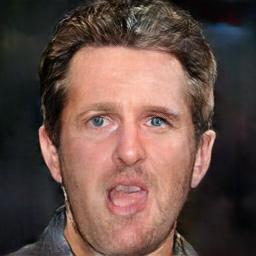}&
            \includegraphics[width=0.176\textwidth]{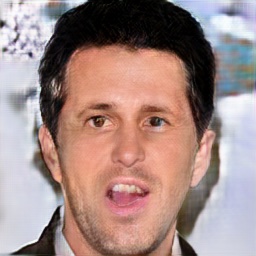}&
            \includegraphics[width=0.176\textwidth]{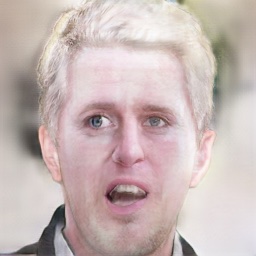}&
            \includegraphics[width=0.176\textwidth]{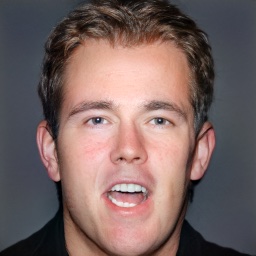}
            \vspace{-0.15em}
            \tabularnewline
            \includegraphics[width=0.176\textwidth]{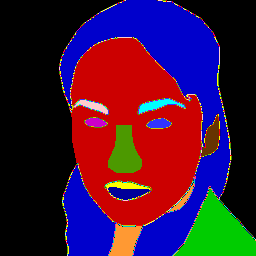}&
            \includegraphics[width=0.176\textwidth]{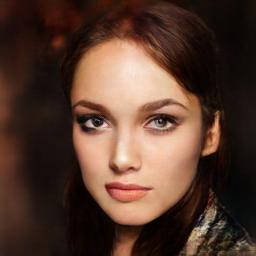}&
            \includegraphics[width=0.176\textwidth]{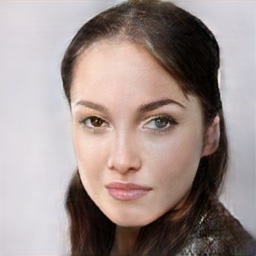}&
            \includegraphics[width=0.176\textwidth]{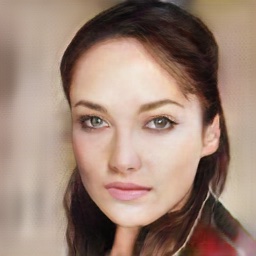}&
            \includegraphics[width=0.176\textwidth]{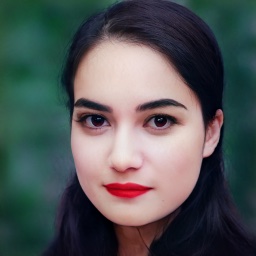}
            \tabularnewline
            Input & pix2pixHD & SPADE & CC\_FPSE & pSp
        \end{tabular}
        }
        \vspace{0.1cm}
        \caption{Comparisons to other label-to-image methods.}
        \vspace{-0.1cm}
        \label{fig:seg_comparison}
        \end{minipage}
        \vspace{-0.2cm}
\end{figure*}

\subsection{Conditional Image Synthesis}
Conditional image synthesis aims at generating photo-realistic images conditioned on certain input types. In this section, our pSp architecture is tested on two conditional image generation tasks: generating high-quality face images from sketches and semantic segmentation maps. We demonstrate that, with only minimal changes, our encoder successfully utilizes the expressiveness of StyleGAN to generate high-quality and diverse outputs. 

\vspace{-0.2cm}
\paragraph{\textit{\textbf{Methodology and details.}}} The training of the two conditional generation tasks is similar to that of the encoder, where the input is the conditioned image and the target is the corresponding real image. To generate multiple images at inference time we perform style-mixing on the fine-level features, taking layers (1-7) from the latent code of the input image and layers (8-18) from a randomly drawn $\textbf{w}$ vector.

\vspace{-0.1cm}
\subsubsection{Face From Sketch}
Common approaches for sketch-to-image synthesis incorporate hard constraints that require pixel-wise correspondence between the input sketch and generated image, making them ill-suited when given incomplete, sparse sketches. DeepFaceDrawing \cite{chen2020deep} address this using a set of dedicated mapping networks. We show that pSp provides a simple alternative to past approaches. As there are currently no publicly available datasets representative of hand-drawn face sketches, we elect to construct our own dataset, which we describe in Appendix~\ref{datasets}.

\paragraph{\textit{\textbf{Results.}}} Figure \ref{fig:sketch_comparison_deepfacedrawing} compares the results of our method to those of pix2pixHD and DeepFaceDrawing. As no code release is available for DeepFaceDrawing, we compare directly with sketches and results published in their paper. 
While DeepFaceDrawing obtain more visually pleasing results compared to pix2pixHD, they are still limited in their diversity. Conversely, although our model is trained on a different dataset, we are still able to generalize well to their sketches. Notably, we observe our ability to obtain more diverse outputs that better retain finer details (e.g. facial hair). Additional results, including those on non-frontal sketches are provided in the Appendix.  

\vspace{-0.2cm}
\subsubsection{Face from Segmentation Map}
\vspace{-0.075cm}
Here, we evaluate using pSp for synthesizing face images from segmentation maps. In addition to pix2pixHD, we compare our approach to two additional state-of-the-art label-to-image methods: SPADE~\cite{park2019semantic}, and CC\_FPSE~\cite{liu2019learning}, both of which are based on pix2pixHD.

\vspace{-0.2cm}
\paragraph{\textit{\textbf{Results.}}} In Figure~\ref{fig:seg_comparison} we provide a visual comparison of the competing approaches on the CelebAMask-HQ dataset containing 19 semantic categories. As the competing methods are based on pix2pixHD, the results of all three suffer from similar artifacts. Conversely, our approach is able to generate high-quality outputs across a wide range of inputs of various poses and expressions. 
Additionally, using our multi-modal technique, pSp can easily generate various possible outputs with the same pose and attributes but varying fine styles for a single input semantic map or sketch image. We provide examples in Figure~\ref{teaser} with additional results in the Appendix.

\vspace{-0.1cm}
\paragraph{\textit{\textbf{Human Perceptual Study.}}}
We additionally perform a human evaluation to compare the visual quality of each method presented above. Each worker is given two images synthesized by different methods on the same input and is given an unlimited time to select which output looks more realistic. Each of our three workers reviews approximately $2,800$ pairs for each task, resulting in over $8,400$ human judgements for each method. Table \ref{tb:human_evaluation} shows that pSp significantly outperforms the other respective methods in both synthesis tasks.

\begin{table}
    \setlength{\tabcolsep}{4.2pt}
    \centering
    \begin{tabular}{l c c c}
    \toprule
    Task & pix2pixHD & SPADE & CC\_FPSE \\
    \midrule
    Segmentation &
    \multicolumn{1}{c}{$94.72\%$} &
    \multicolumn{1}{c}{$95.25\%$} &
    \multicolumn{1}{c}{$93.06\%$} \\ 
    Sketch &
    \multicolumn{1}{c}{$93.34\%$} &
    \multicolumn{1}{c}{N/A} &
    \multicolumn{1}{c}{N/A} \\
    \bottomrule
    \end{tabular}
    \vspace{0.1cm}
    \caption{Human evaluation results on CelebA-HQ for conditional image synthesis tasks. Each cell denotes the percentage of users who favored pSp over the listed method.}
    \vspace{-0.25cm}
    \label{tb:human_evaluation}
\end{table}

\subsection{Extending to Other Applications}
Besides the applications presented above, we have found pSp to be applicable to a wide variety of additional tasks with minimal changes to the training process. Specifically, we present samples of super-resolution and inpainting results using pSp in Figure~\ref{teaser} with further details and results presented in Appendix~\ref{application_details}. For both tasks, paired data is generated and training is performed in a supervised fashion. Additionally, we show multi-modal support for super-resolution via style-mixing on medium-level features and evaluate pSp on several image editing tasks, including image interpolation and local patch editing.

\subsection{Going Beyond the Facial Domain}
In this section we show that our pSp framework can be trained to solve the various tasks explored above without relying on the advantages provided by the identity loss in the facial domain. While our method does require a pretrained StyleGAN generator, recent works~\cite{Karras2020ada, robb2020fsgan} have shown that such a generator can be easily trained with significantly fewer examples than required in the past. 

Figure~\ref{fig:additional_domains} shows the results on the AFHQ Cat and AFHQ Dog datasets~\cite{choi2020stargan} for the StyleGAN inversion and sketch-to-image tasks. For these tasks, we use a pretrained StyleGAN-ADA~\cite{Karras2020ada} model for each of the two domains and train our pSp encoder using only the $\mathcal{L_{\text{2}}}$, $\mathcal{L_{\text{LPIPS}}}$, and $\mathcal{L_{\text{reg}}}$ losses with the same $\lambda$ values as those used for the facial domain. As shown, we are able to generalize well to the examined domains, obtaining high-quality, accurate reconstruction results while also supporting multi-modal synthesis via our style-mixing approach. The accompanying Appendix provides additional results for super-resolution and inpainting on these domains.

\begin{figure}
    \setlength{\tabcolsep}{1pt}
    \centering
        \begin{tabular}{c c}
            \includegraphics[width=0.225\textwidth]{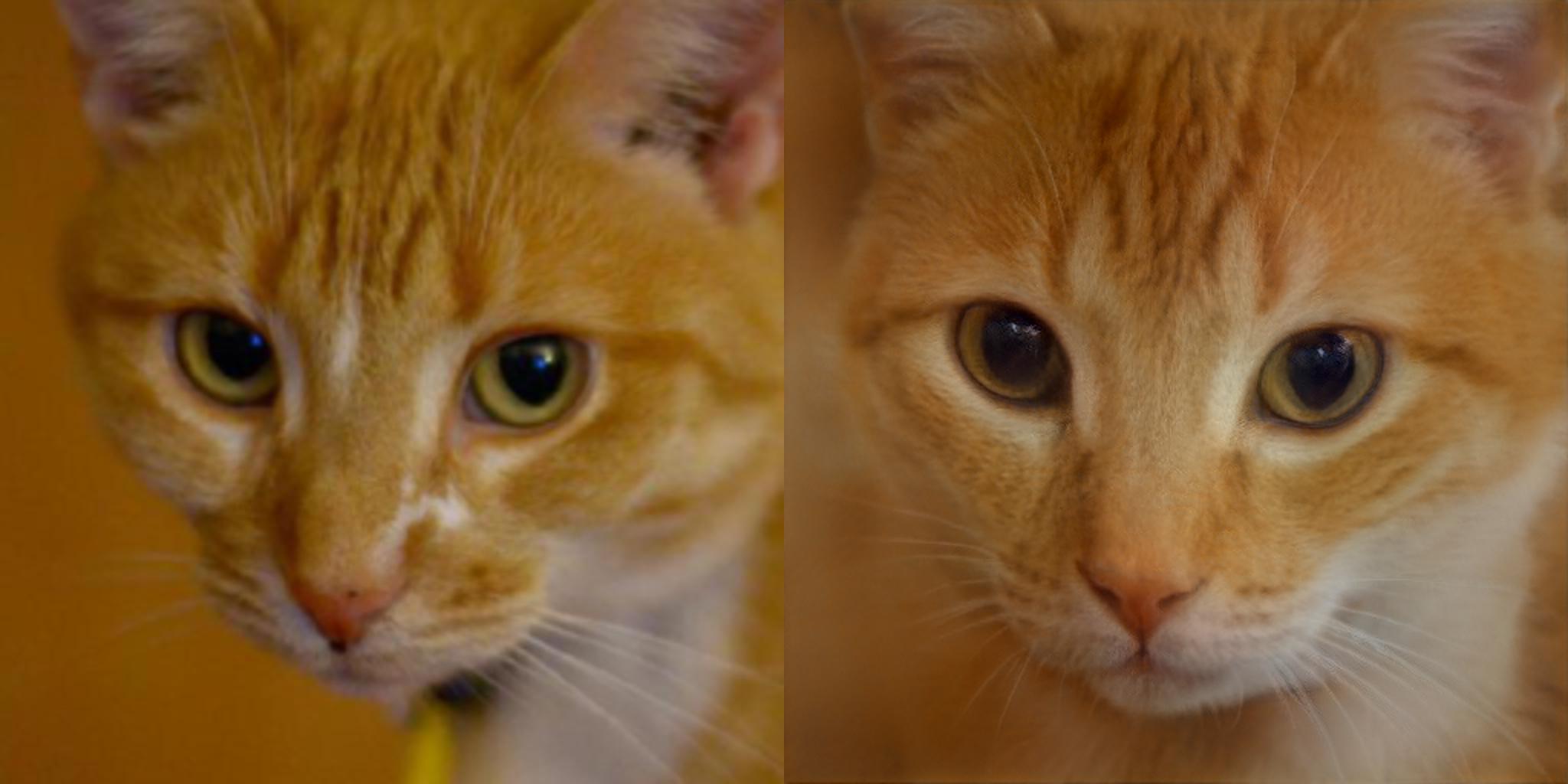}&
            \includegraphics[width=0.225\textwidth]{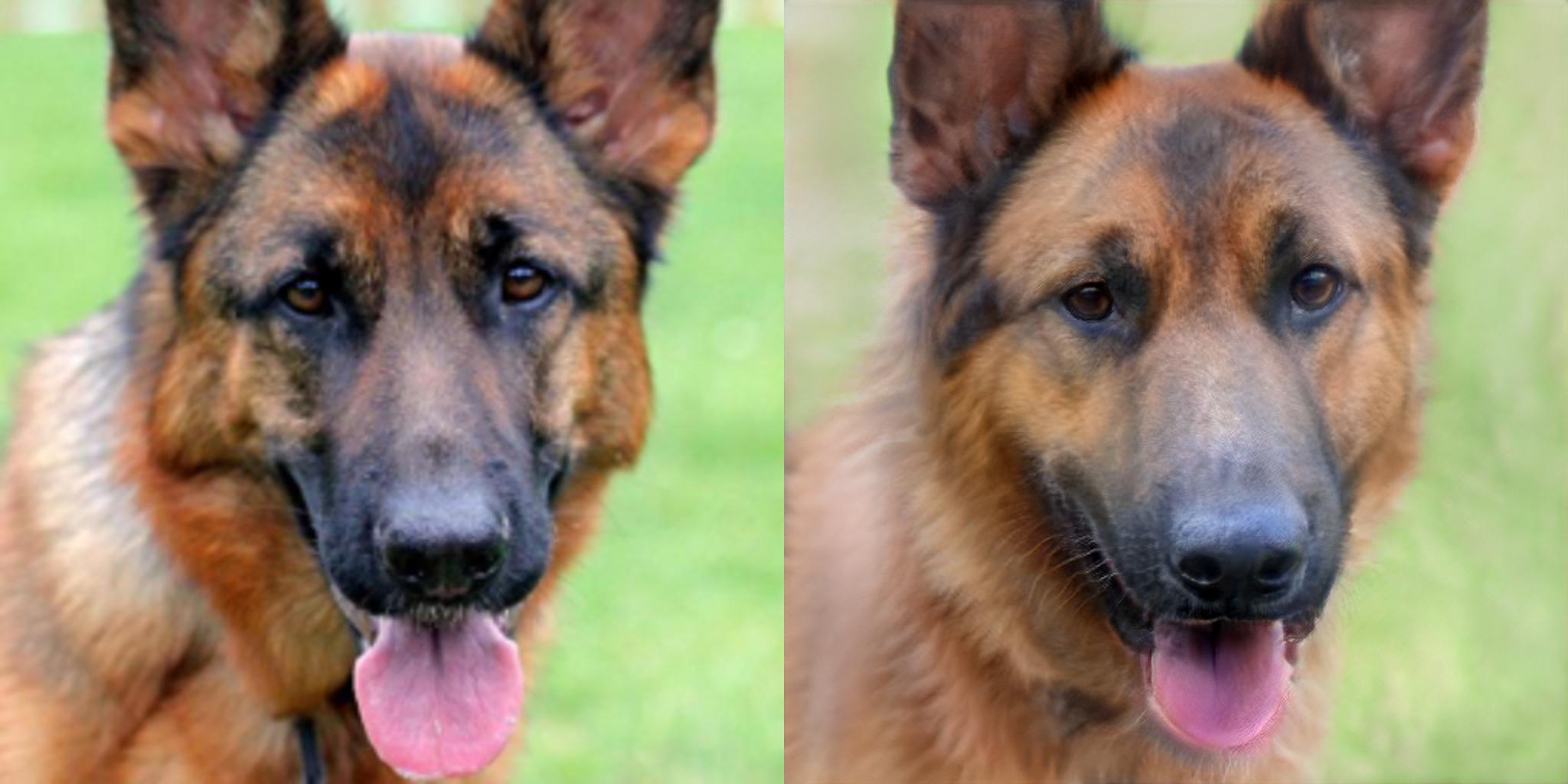}
            \tabularnewline
            \includegraphics[width=0.225\textwidth]{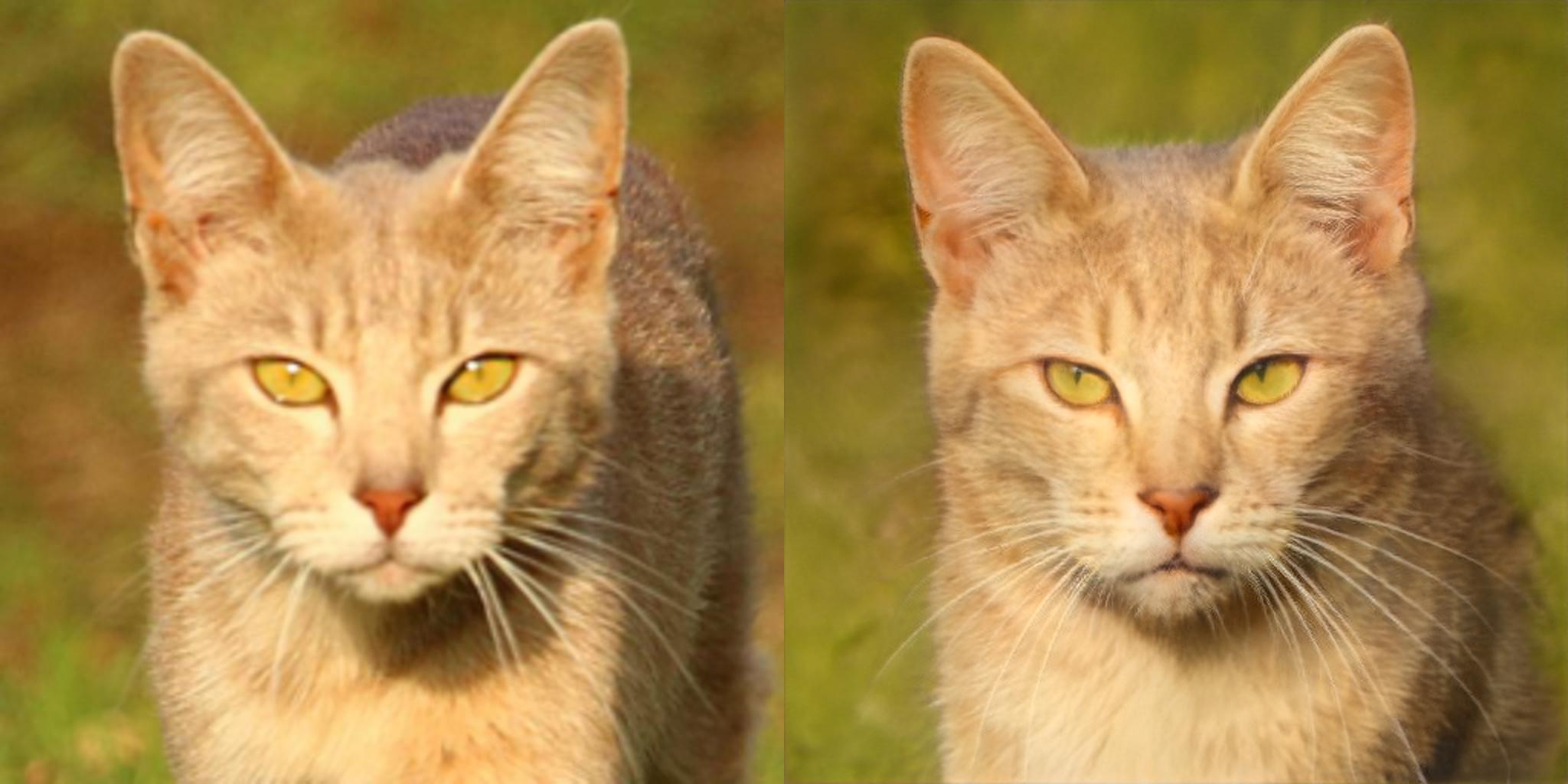}&
            \includegraphics[width=0.225\textwidth]{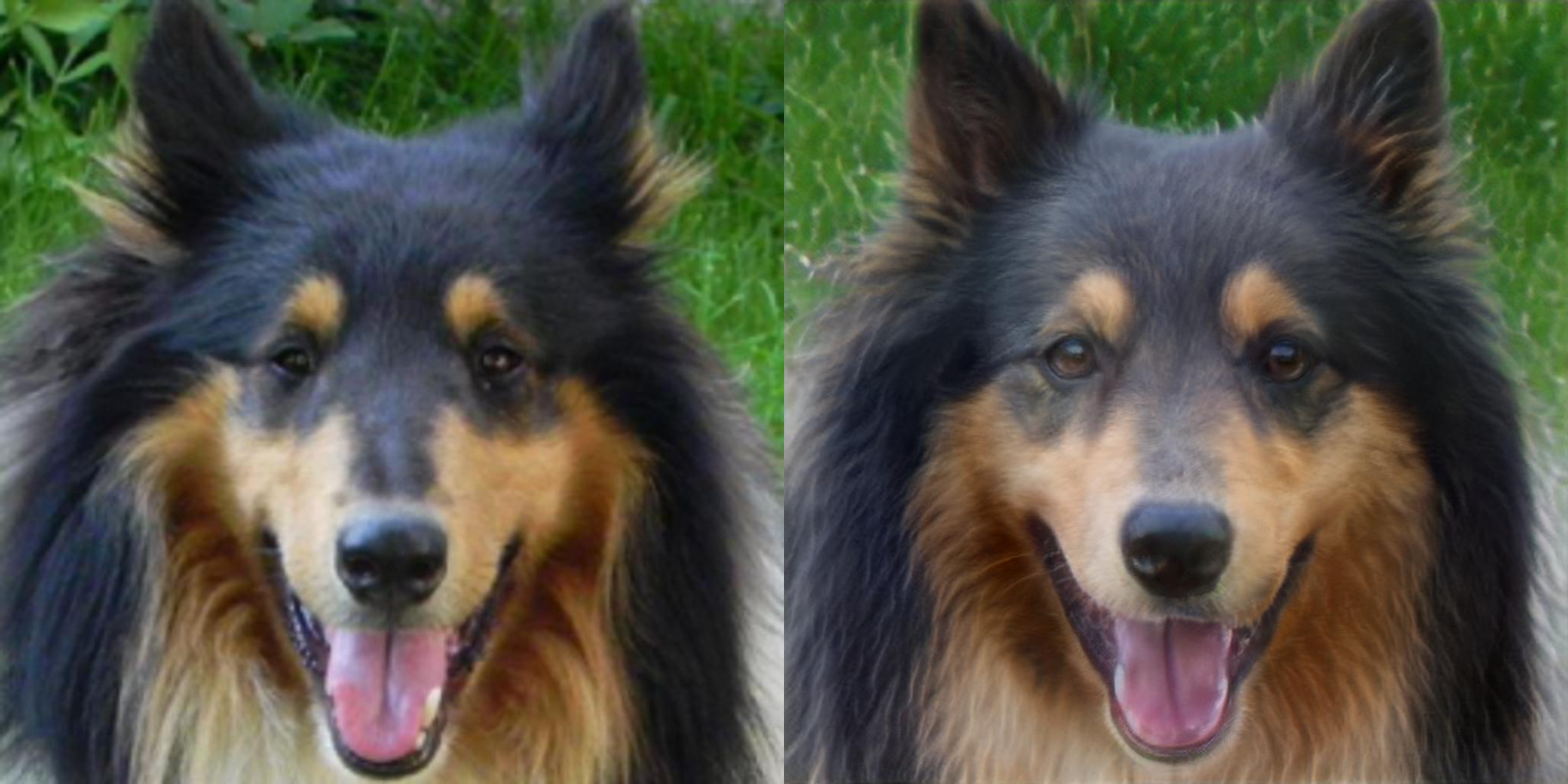}
            \tabularnewline
            \includegraphics[width=0.225\textwidth]{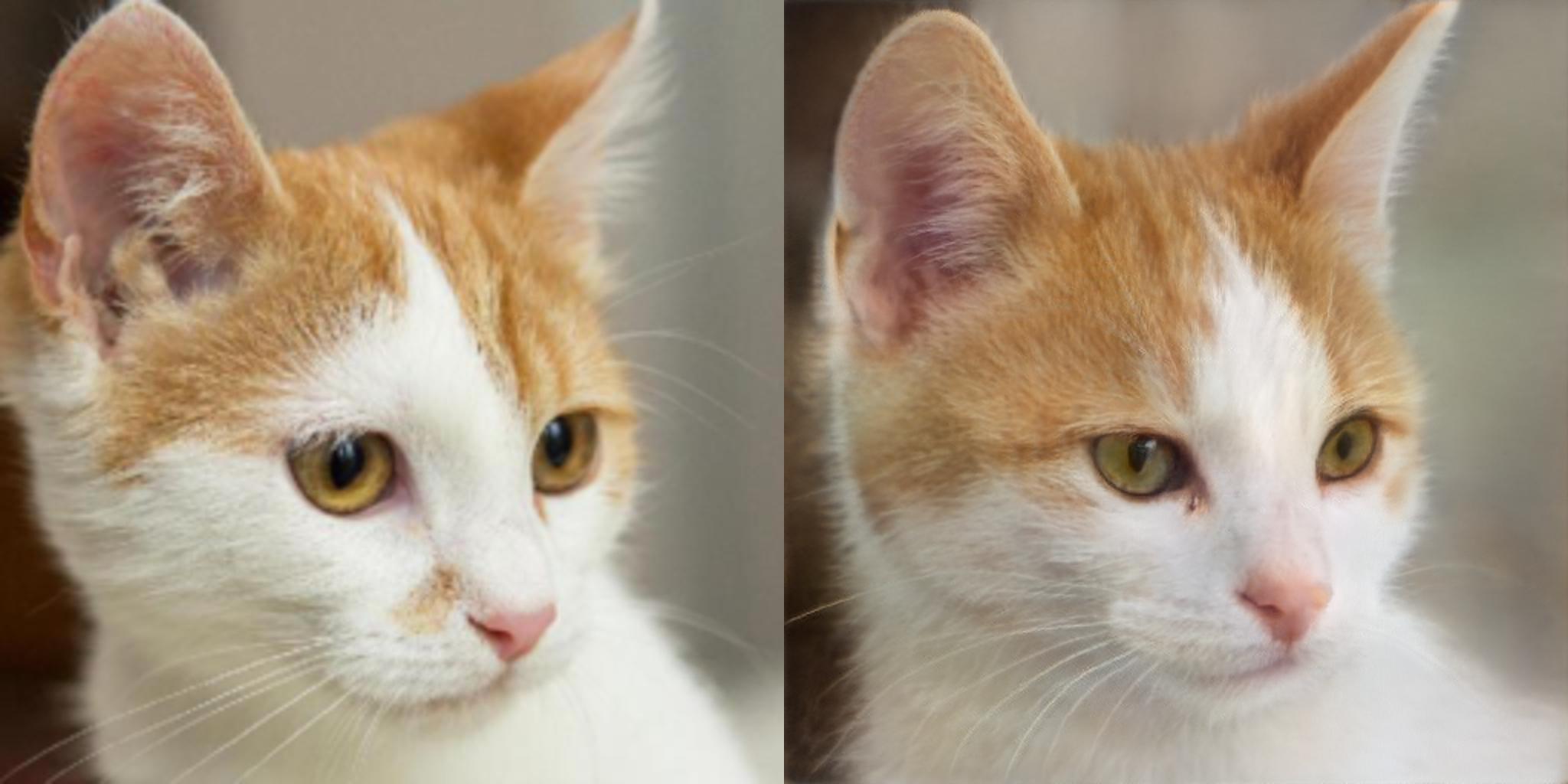}&
            \includegraphics[width=0.225\textwidth]{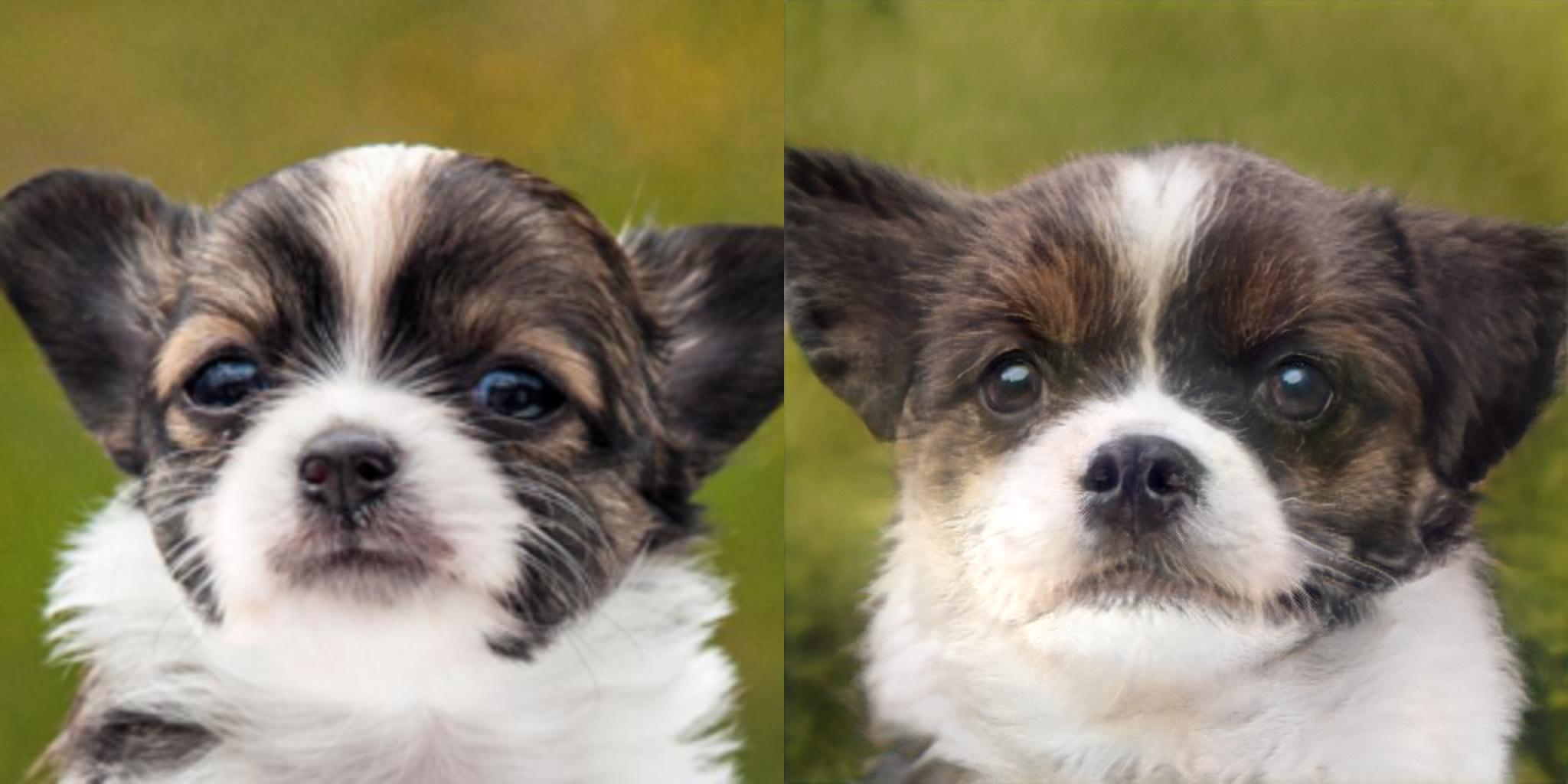}
            \tabularnewline
        \end{tabular}
        {\small
        \begin{tabular}{c}
            StyleGAN Inversion and Reconstruction
        \end{tabular}
        }
        {\small
        \begin{tabular}{c}
            \includegraphics[width=0.45\textwidth]{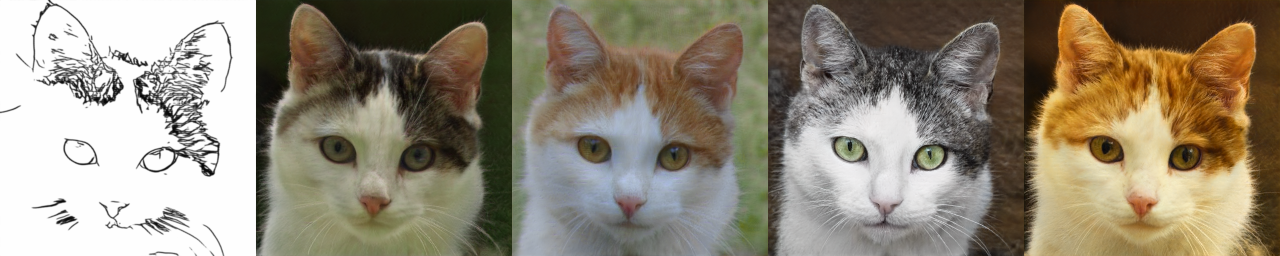}
            \tabularnewline
            \includegraphics[width=0.45\textwidth]{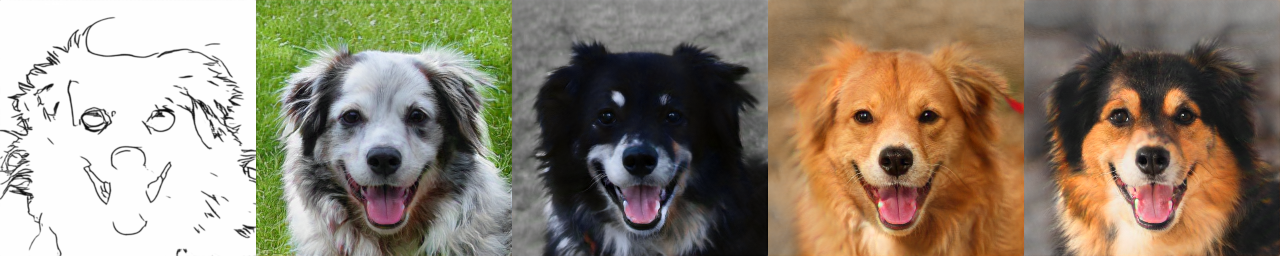}
            \tabularnewline
            Image Generation from Sketches
        \end{tabular}
        }
    \vspace{0.1cm}
    \caption{Results of pSp on the AFHQ Dataset for StyleGAN Inversion and the sketch-to-image tasks. For reconstruction, the input (left) is shown alongside the reconstructed output (right). For sketch-to-image, multiple outputs are generated via style-mixing.}
    \label{fig:additional_domains}
\end{figure}

\begin{figure}[h!]
    \setlength{\tabcolsep}{1pt}
    \centering
        \begin{tabular}{c c c c}
            \includegraphics[width=0.11\textwidth]{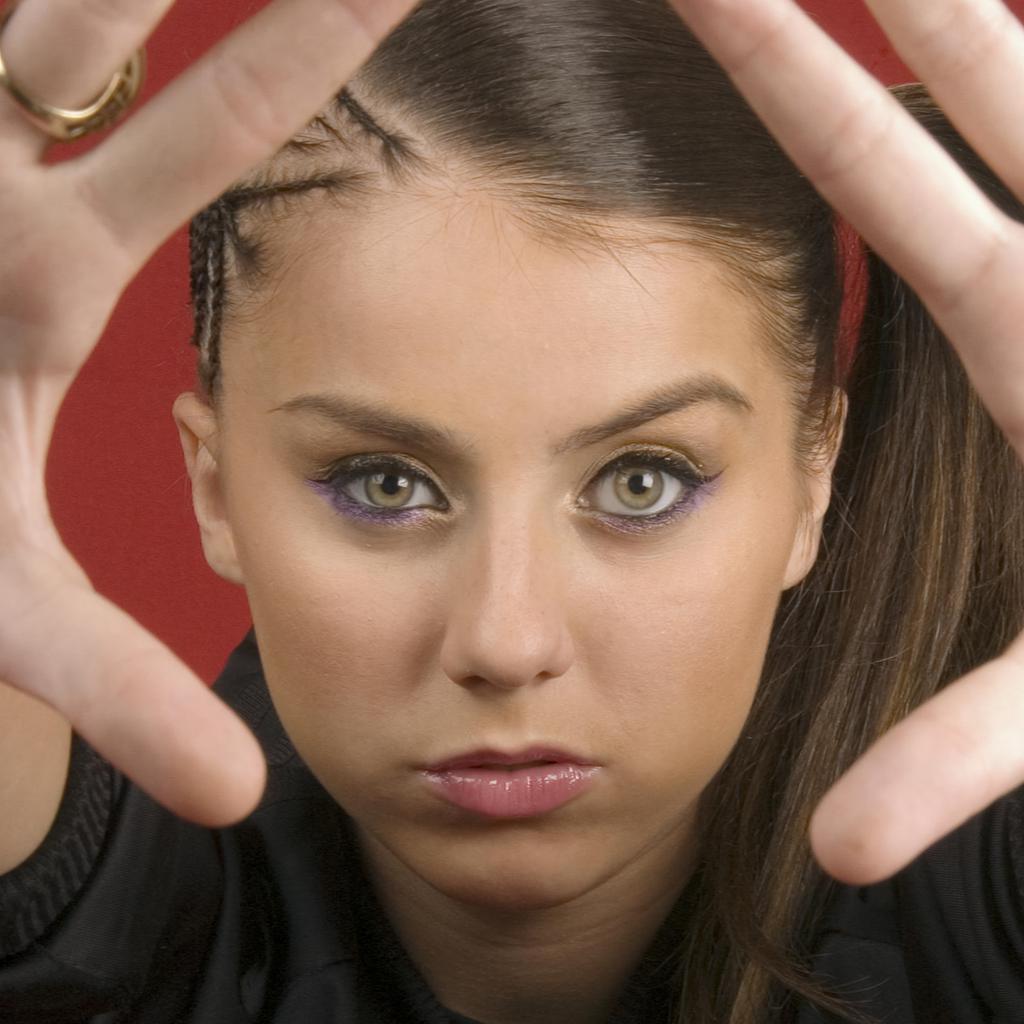}&
            \includegraphics[width=0.11\textwidth]{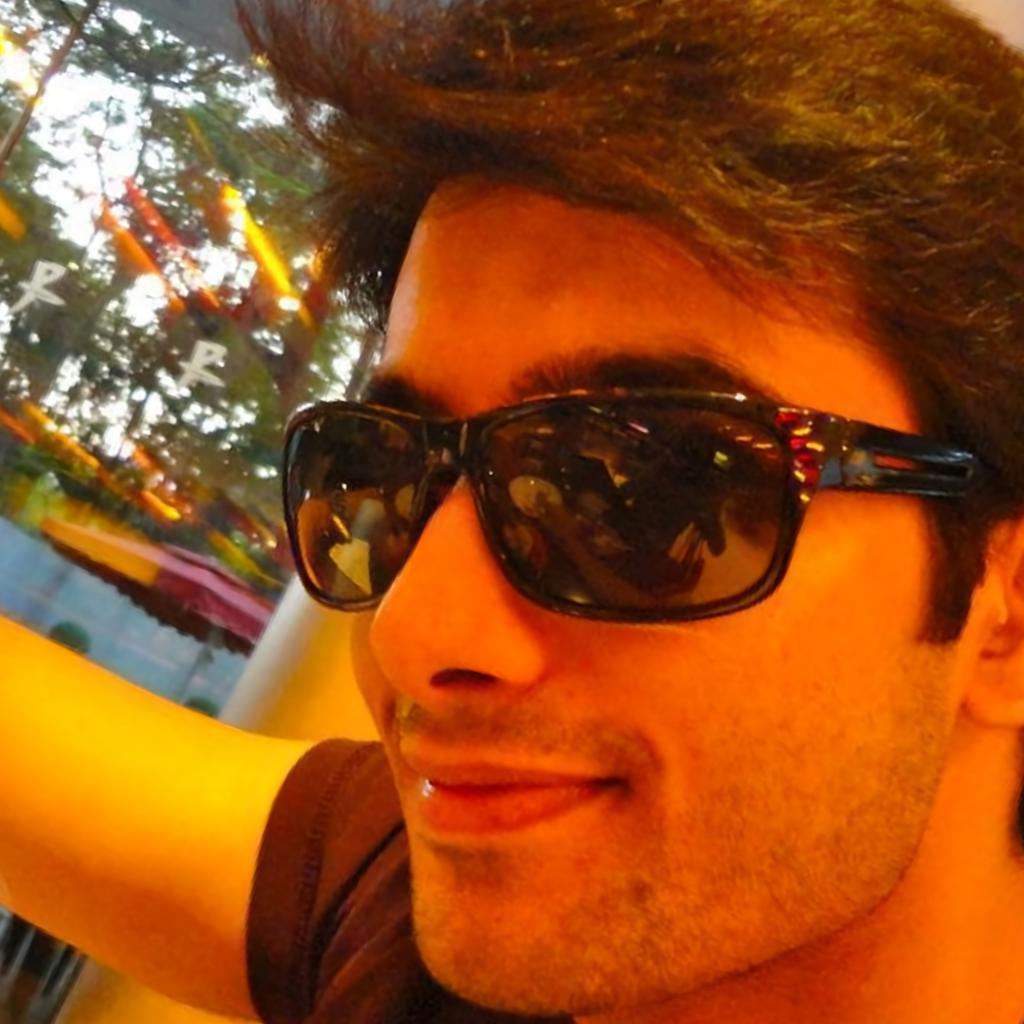}&        
            \includegraphics[width=0.11\textwidth]{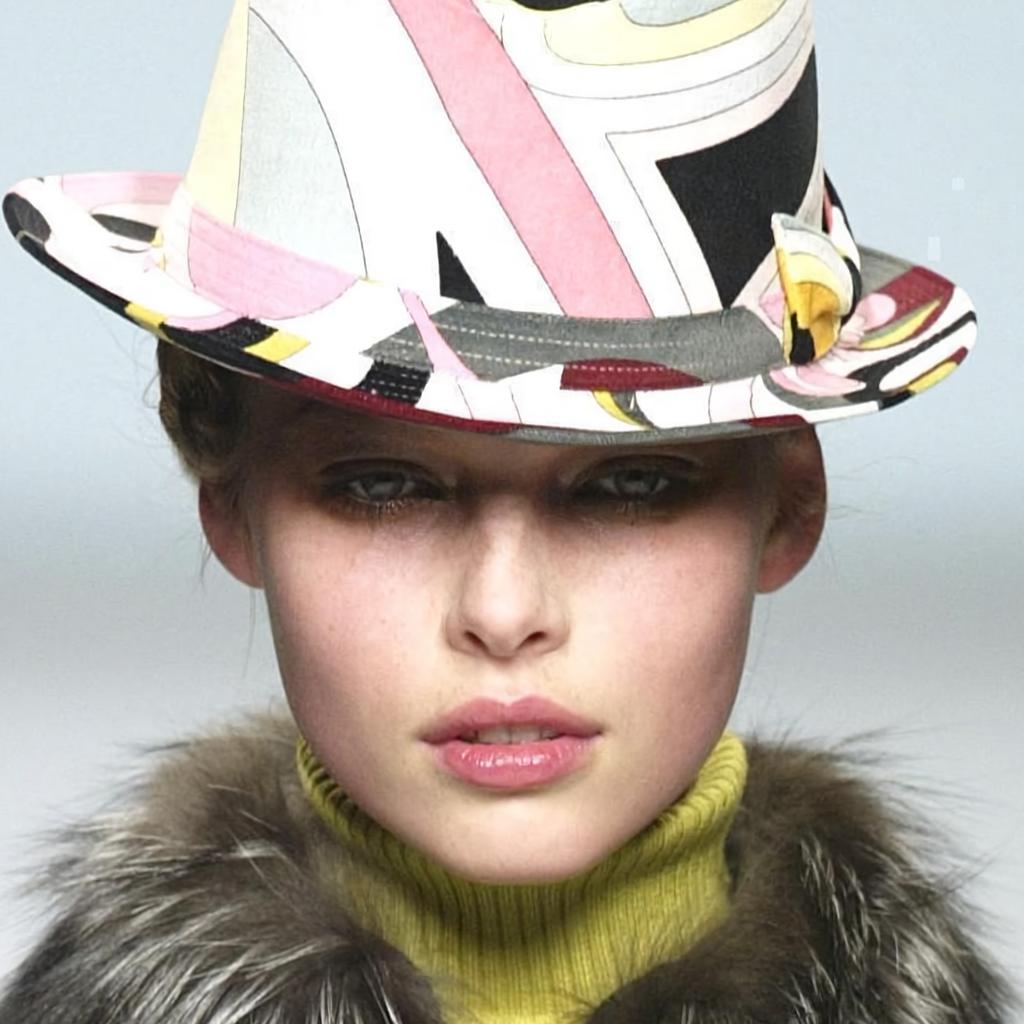}&
            \includegraphics[width=0.11\textwidth]{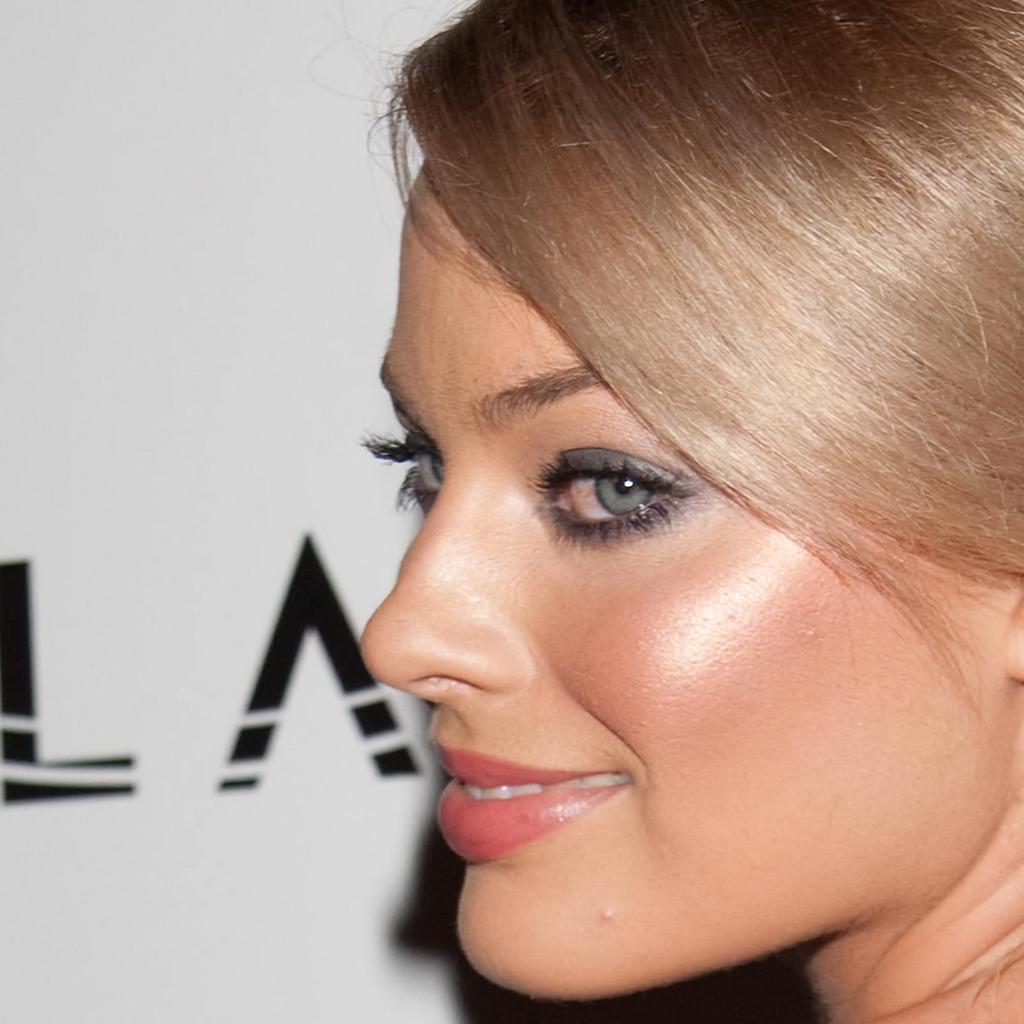}
            \tabularnewline
            \includegraphics[width=0.11\textwidth]{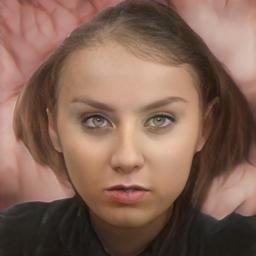}&
            \includegraphics[width=0.11\textwidth]{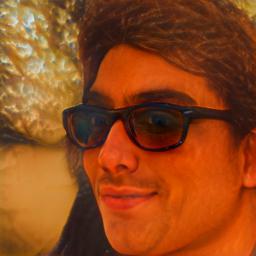}&        
            \includegraphics[width=0.11\textwidth]{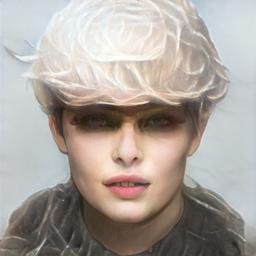}&
            \includegraphics[width=0.11\textwidth]{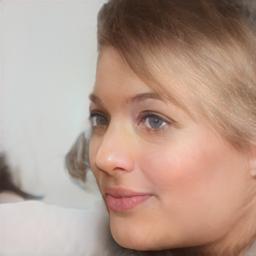}
        \end{tabular}
        \vspace{0.1cm}
        \caption{Challenging cases for StyleGAN Inversion.}
        \label{fig:encoder_limit}
        \vspace{-0.3cm}
    \end{figure}

\section{Discussion}
Although our suggested framework for image-to-image translation achieves compelling results in various applications, it has some inherent assumptions that should be considered.
First, the high-quality images that are generated by utilizing the pretrained StyleGAN come with a cost --- the method is limited to images that can be generated by StyleGAN. Thus, generating faces which are not close to frontal, or have certain expressions may be challenging if such examples were not available when training the StyleGAN model. Also, the global approach of pSp, although advantageous for many tasks, does introduce a challenge in preserving finer details of the input image, such as earrings or background details. This is especially significant in tasks such as inpainting or super-resolution where standard image-to-image architectures can simply propagate local information. Figure~\ref{fig:encoder_limit} presents some examples of such reconstruction failures.

\section{Conclusion}
In this work, we propose a novel encoder architecture that can be used to directly map a real image into the $\mathcal{W+}$ latent space with no optimization required. 
There, styles are extracted in a hierarchical fashion and fed into the corresponding inputs of a fixed StyleGAN generator. 
Combining our encoder with a StyleGAN decoder, we present a generic framework for solving various image-to-image translation tasks, all using the same architecture. 
Notably, in contrast to the ``\textit{invert first, edit later}'' approach of previous StyleGAN encoders, we show pSp can be used to directly encode these translation tasks into StyleGAN, thereby supporting input images that do not reside in the StyleGAN domain.
Additionally, differing from previous works that typically rely on dedicated architectures for solving a single translation task, we show pSp to be capable of solving a wide variety of problems, requiring only minimal changes to the training losses and methodology.
We hope that the ease-of-use of our approach will encourage further research into utilizing StyleGAN for real image-to-image translation tasks.

{\small
\bibliographystyle{ieee_fullname}
\bibliography{egbib}

\begin{thebibliography}{10}\itemsep=-1pt

\bibitem{abdal2019image2stylegan}
Rameen Abdal, Yipeng Qin, and Peter Wonka.
\newblock Image2stylegan: How to embed images into the stylegan latent space?
\newblock In {\em Proceedings of the IEEE international conference on computer
  vision}, pages 4432--4441, 2019.

\bibitem{abdal2020image2stylegan++}
Rameen Abdal, Yipeng Qin, and Peter Wonka.
\newblock Image2stylegan++: How to edit the embedded images?
\newblock In {\em Proceedings of the IEEE/CVF Conference on Computer Vision and
  Pattern Recognition}, pages 8296--8305, 2020.

\bibitem{Abdal:StyleFlow:Arxiv:2020}
Rameen Adbal, Pie Zhu, Niloy {J. Mitra}, and Peter Wonka.
\newblock Styleflow: Attribute-conditioned exploration of stylegan-generated
  images using conditional continuous normalizing flows.
\newblock {\em arXiv preprint arXiv:}, 2020.

\bibitem{pbayliesstyleganencoder}
{Baylies}.
\newblock stylegan-encoder.
\newblock \url{https://github.com/pbaylies/stylegan-encoder}, 2019.
\newblock Accessed: April 2020.

\bibitem{chen2020deep}
Shu-Yu Chen, Wanchao Su, Lin Gao, Shihong Xia, and Hongbo Fu.
\newblock {DeepFaceDrawing}: Deep generation of face images from sketches.
\newblock {\em ACM Transactions on Graphics (Proceedings of ACM SIGGRAPH
  2020)}, 39(4):72:1--72:16, 2020.

\bibitem{choi2020stargan}
Yunjey Choi, Youngjung Uh, Jaejun Yoo, and Jung-Woo Ha.
\newblock Stargan v2: Diverse image synthesis for multiple domains.
\newblock In {\em Proceedings of the IEEE/CVF Conference on Computer Vision and
  Pattern Recognition}, pages 8188--8197, 2020.

\bibitem{collins2020editing}
Edo Collins, Raja Bala, Bob Price, and Sabine Susstrunk.
\newblock Editing in style: Uncovering the local semantics of gans.
\newblock In {\em Proceedings of the IEEE/CVF Conference on Computer Vision and
  Pattern Recognition}, pages 5771--5780, 2020.

\bibitem{creswell2018inverting}
Antonia Creswell and Anil~Anthony Bharath.
\newblock Inverting the generator of a generative adversarial network.
\newblock {\em IEEE transactions on neural networks and learning systems},
  30(7):1967--1974, 2018.

\bibitem{deng2019arcface}
Jiankang Deng, Jia Guo, Niannan Xue, and Stefanos Zafeiriou.
\newblock Arcface: Additive angular margin loss for deep face recognition.
\newblock In {\em Proceedings of the IEEE Conference on Computer Vision and
  Pattern Recognition}, pages 4690--4699, 2019.

\bibitem{denton2019detecting}
Emily Denton, Ben Hutchinson, Margaret Mitchell, and Timnit Gebru.
\newblock Detecting bias with generative counterfactual face attribute
  augmentation.
\newblock {\em arXiv preprint arXiv:1906.06439}, 2019.

\bibitem{goetschalckx2019ganalyze}
Lore Goetschalckx, Alex Andonian, Aude Oliva, and Phillip Isola.
\newblock Ganalyze: Toward visual definitions of cognitive image properties,
  2019.

\bibitem{guan2020collaborative}
Shanyan Guan, Ying Tai, Bingbing Ni, Feida Zhu, Feiyue Huang, and Xiaokang
  Yang.
\newblock Collaborative learning for faster stylegan embedding.
\newblock {\em arXiv preprint arXiv:2007.01758}, 2020.

\bibitem{harkonen2020ganspace}
Erik H{\"a}rk{\"o}nen, Aaron Hertzmann, Jaakko Lehtinen, and Sylvain Paris.
\newblock Ganspace: Discovering interpretable gan controls.
\newblock {\em arXiv preprint arXiv:2004.02546}, 2020.

\bibitem{huang2018multimodal}
Xun Huang, Ming-Yu Liu, Serge Belongie, and Jan Kautz.
\newblock Multimodal unsupervised image-to-image translation.
\newblock In {\em Proceedings of the European Conference on Computer Vision
  (ECCV)}, pages 172--189, 2018.

\bibitem{huang2020curricularface}
Yuge Huang, Yuhan Wang, Ying Tai, Xiaoming Liu, Pengcheng Shen, Shaoxin Li,
  Jilin Li, and Feiyue Huang.
\newblock Curricularface: adaptive curriculum learning loss for deep face
  recognition.
\newblock In {\em Proceedings of the IEEE/CVF Conference on Computer Vision and
  Pattern Recognition}, pages 5901--5910, 2020.

\bibitem{isola2017image}
Phillip Isola, Jun-Yan Zhu, Tinghui Zhou, and Alexei~A Efros.
\newblock Image-to-image translation with conditional adversarial networks.
\newblock In {\em Proceedings of the IEEE conference on computer vision and
  pattern recognition}, pages 1125--1134, 2017.

\bibitem{jahanian2019steerability}
Ali Jahanian, Lucy Chai, and Phillip Isola.
\newblock On the "steerability" of generative adversarial networks.
\newblock {\em arXiv preprint arXiv:1907.07171}, 2019.

\bibitem{johnson2016perceptual}
Justin Johnson, Alexandre Alahi, and Li Fei-Fei.
\newblock Perceptual losses for real-time style transfer and super-resolution.
\newblock In {\em European conference on computer vision}, pages 694--711.
  Springer, 2016.

\bibitem{karras2018progressive}
Tero Karras, Timo Aila, Samuli Laine, and Jaakko Lehtinen.
\newblock Progressive growing of {GAN}s for improved quality, stability, and
  variation.
\newblock In {\em International Conference on Learning Representations}, 2018.

\bibitem{Karras2020ada}
Tero Karras, Miika Aittala, Janne Hellsten, Samuli Laine, Jaakko Lehtinen, and
  Timo Aila.
\newblock Training generative adversarial networks with limited data.
\newblock In {\em Proc. NeurIPS}, 2020.

\bibitem{karras2019style}
Tero Karras, Samuli Laine, and Timo Aila.
\newblock A style-based generator architecture for generative adversarial
  networks.
\newblock In {\em Proceedings of the IEEE conference on computer vision and
  pattern recognition}, pages 4401--4410, 2019.

\bibitem{karras2020analyzing}
Tero Karras, Samuli Laine, Miika Aittala, Janne Hellsten, Jaakko Lehtinen, and
  Timo Aila.
\newblock Analyzing and improving the image quality of stylegan.
\newblock In {\em Proceedings of the IEEE/CVF Conference on Computer Vision and
  Pattern Recognition}, pages 8110--8119, 2020.

\bibitem{katzir2019cross}
Oren Katzir, Dani Lischinski, and Daniel Cohen-Or.
\newblock Cross-domain cascaded deep feature translation.
\newblock {\em arXiv}, pages arXiv--1906, 2019.

\bibitem{HelenFaces}
Vuong Le, Jonathan Brandt, Zhe Lin, Lubomir Bourdev, and Thomas~S. Huang.
\newblock Interactive facial feature localization.
\newblock In Andrew Fitzgibbon, Svetlana Lazebnik, Pietro Perona, Yoichi Sato,
  and Cordelia Schmid, editors, {\em Computer Vision -- ECCV 2012}, pages
  679--692, Berlin, Heidelberg, 2012. Springer Berlin Heidelberg.

\bibitem{li2019linestofacephoto}
Yuhang Li, Xuejin Chen, Feng Wu, and Zheng-Jun Zha.
\newblock Linestofacephoto: Face photo generation from lines with conditional
  self-attention generative adversarial networks.
\newblock In {\em Proceedings of the 27th ACM International Conference on
  Multimedia}, pages 2323--2331, 2019.

\bibitem{lin2017feature}
Tsung-Yi Lin, Piotr Doll{\'a}r, Ross Girshick, Kaiming He, Bharath Hariharan,
  and Serge Belongie.
\newblock Feature pyramid networks for object detection.
\newblock In {\em Proceedings of the IEEE conference on computer vision and
  pattern recognition}, pages 2117--2125, 2017.

\bibitem{lipton2017precise}
Zachary~C Lipton and Subarna Tripathi.
\newblock Precise recovery of latent vectors from generative adversarial
  networks.
\newblock {\em arXiv preprint arXiv:1702.04782}, 2017.

\bibitem{lira2020ganhopper}
Wallace Lira, Johannes Merz, Daniel Ritchie, Daniel Cohen-Or, and Hao Zhang.
\newblock Ganhopper: Multi-hop gan for unsupervised image-to-image translation.
\newblock {\em arXiv preprint arXiv:2002.10102}, 2020.

\bibitem{liu2019variance}
Liyuan Liu, Haoming Jiang, Pengcheng He, Weizhu Chen, Xiaodong Liu, Jianfeng
  Gao, and Jiawei Han.
\newblock On the variance of the adaptive learning rate and beyond.
\newblock {\em arXiv preprint arXiv:1908.03265}, 2019.

\bibitem{liu2017unsupervised}
Ming-Yu Liu, Thomas Breuel, and Jan Kautz.
\newblock Unsupervised image-to-image translation networks.
\newblock In {\em Advances in neural information processing systems}, pages
  700--708, 2017.

\bibitem{liu2019learning}
Xihui Liu, Guojun Yin, Jing Shao, Xiaogang Wang, et~al.
\newblock Learning to predict layout-to-image conditional convolutions for
  semantic image synthesis.
\newblock In {\em Advances in Neural Information Processing Systems}, pages
  570--580, 2019.

\bibitem{Menon_2020_CVPR}
Sachit Menon, Alexandru Damian, Shijia Hu, Nikhil Ravi, and Cynthia Rudin.
\newblock Pulse: Self-supervised photo upsampling via latent space exploration
  of generative models.
\newblock In {\em Proceedings of the IEEE/CVF Conference on Computer Vision and
  Pattern Recognition (CVPR)}, June 2020.

\bibitem{nitzan2020disentangling}
Yotam Nitzan, Amit Bermano, Yangyan Li, and Daniel Cohen-Or.
\newblock Disentangling in latent space by harnessing a pretrained generator.
\newblock {\em arXiv preprint arXiv:2005.07728}, 2020.

\bibitem{park2019semantic}
Taesung Park, Ming-Yu Liu, Ting-Chun Wang, and Jun-Yan Zhu.
\newblock Semantic image synthesis with spatially-adaptive normalization.
\newblock In {\em Proceedings of the IEEE Conference on Computer Vision and
  Pattern Recognition}, pages 2337--2346, 2019.

\bibitem{perarnau2016invertible}
Guim Perarnau, Joost Van De~Weijer, Bogdan Raducanu, and Jose~M {\'A}lvarez.
\newblock Invertible conditional gans for image editing.
\newblock {\em arXiv preprint arXiv:1611.06355}, 2016.

\bibitem{pidhorskyi2020adversarial}
Stanislav Pidhorskyi, Donald~A Adjeroh, and Gianfranco Doretto.
\newblock Adversarial latent autoencoders.
\newblock In {\em Proceedings of the IEEE/CVF Conference on Computer Vision and
  Pattern Recognition}, pages 14104--14113, 2020.

\bibitem{richardson2020unsupervised}
Eitan Richardson and Yair Weiss.
\newblock The surprising effectiveness of linear unsupervised image-to-image
  translation.
\newblock {\em ArXiv}, abs/2007.12568, 2020.

\bibitem{robb2020fsgan}
Esther Robb, Wen-Sheng Chu, Abhishek Kumar, and Jia-Bin Huang.
\newblock Few-shot adaptation of generative adversarial networks.
\newblock {\em arXiv}, 2020.

\bibitem{shen2020interpreting}
Yujun Shen, Jinjin Gu, Xiaoou Tang, and Bolei Zhou.
\newblock Interpreting the latent space of gans for semantic face editing.
\newblock In {\em Proceedings of the IEEE/CVF Conference on Computer Vision and
  Pattern Recognition}, pages 9243--9252, 2020.

\bibitem{sketch-simpf}
Edgar Simo-Serra, Satoshi Iizuka, Kazuma Sasaki, and Hiroshi Ishikawa.
\newblock Learning to simplify: fully convolutional networks for rough sketch
  cleanup.
\newblock {\em ACM Transactions on Graphics}, 35:1--11, 07 2016.

\bibitem{tewari2020stylerig}
Ayush Tewari, Mohamed Elgharib, Gaurav Bharaj, Florian Bernard, Hans-Peter
  Seidel, Patrick P{\'e}rez, Michael Zollh{\"o}fer, and Christian Theobalt.
\newblock Stylerig: Rigging stylegan for 3d control over portrait images.
\newblock {\em arXiv preprint arXiv:2004.00121}, 2020.

\bibitem{THOMAZ2010902}
Carlos~Eduardo Thomaz and Gilson~Antonio Giraldi.
\newblock A new ranking method for principal components analysis and its
  application to face image analysis.
\newblock {\em Image and Vision Computing}, 28(6):902 -- 913, 2010.

\bibitem{wang2018high}
Ting-Chun Wang, Ming-Yu Liu, Jun-Yan Zhu, Andrew Tao, Jan Kautz, and Bryan
  Catanzaro.
\newblock High-resolution image synthesis and semantic manipulation with
  conditional gans.
\newblock In {\em Proceedings of the IEEE conference on computer vision and
  pattern recognition}, pages 8798--8807, 2018.

\bibitem{yang2019semantic}
Ceyuan Yang, Yujun Shen, and Bolei Zhou.
\newblock Semantic hierarchy emerges in deep generative representations for
  scene synthesis, 2019.

\bibitem{zhang2019lookahead}
Michael Zhang, James Lucas, Jimmy Ba, and Geoffrey~E Hinton.
\newblock Lookahead optimizer: k steps forward, 1 step back.
\newblock In {\em Advances in Neural Information Processing Systems}, pages
  9597--9608, 2019.

\bibitem{zhang2018unreasonable}
Richard Zhang, Phillip Isola, Alexei~A Efros, Eli Shechtman, and Oliver Wang.
\newblock The unreasonable effectiveness of deep features as a perceptual
  metric.
\newblock In {\em Proceedings of the IEEE conference on computer vision and
  pattern recognition}, pages 586--595, 2018.

\bibitem{zhou2020rotate}
Hang Zhou, Jihao Liu, Ziwei Liu, Yu Liu, and Xiaogang Wang.
\newblock Rotate-and-render: Unsupervised photorealistic face rotation from
  single-view images.
\newblock In {\em Proceedings of the IEEE/CVF Conference on Computer Vision and
  Pattern Recognition}, pages 5911--5920, 2020.

\bibitem{zhu2020domain}
Jiapeng Zhu, Yujun Shen, Deli Zhao, and Bolei Zhou.
\newblock In-domain gan inversion for real image editing.
\newblock {\em arXiv preprint arXiv:2004.00049}, 2020.

\bibitem{zhu2016generative}
Jun-Yan Zhu, Philipp Kr{\"a}henb{\"u}hl, Eli Shechtman, and Alexei~A Efros.
\newblock Generative visual manipulation on the natural image manifold.
\newblock In {\em European conference on computer vision}, pages 597--613.
  Springer, 2016.

\bibitem{zhu2017unpaired}
Jun-Yan Zhu, Taesung Park, Phillip Isola, and Alexei~A Efros.
\newblock Unpaired image-to-image translation using cycle-consistent
  adversarial networks.
\newblock In {\em Proceedings of the IEEE international conference on computer
  vision}, pages 2223--2232, 2017.

\bibitem{zhu2017toward}
Jun-Yan Zhu, Richard Zhang, Deepak Pathak, Trevor Darrell, Alexei~A Efros,
  Oliver Wang, and Eli Shechtman.
\newblock Toward multimodal image-to-image translation.
\newblock In {\em Advances in neural information processing systems}, pages
  465--476, 2017.

\bibitem{zhu2020sean}
Peihao Zhu, Rameen Abdal, Yipeng Qin, and Peter Wonka.
\newblock Sean: Image synthesis with semantic region-adaptive normalization.
\newblock In {\em Proceedings of the IEEE/CVF Conference on Computer Vision and
  Pattern Recognition}, pages 5104--5113, 2020.

\end{thebibliography}
}

\newpage
\appendix
\section{Implementation Details} \label{Implement}

For our backbone network we use the ResNet-IR architecture from~\cite{deng2019arcface} pretrained on face recognition, which accelerated convergence. We use a \textit{fixed} StyleGAN2 generator trained on the FFHQ~\cite{karras2019style} dataset. That is, only the pSp encoder network is trained on the given translation task. For all applications, the input image resolution is $256\times256$, where the generated $1024\times1024$ output is resized before being fed into the loss functions. Specifically for $\mathcal{L}_{\text{ID}}$, the images are cropped around the face region and resized to $112\times122$ before being fed into the recognition network. For training, we use the Ranger optimizer, a combination of Rectified Adam~\cite{liu2019variance} with the Lookahead technique~\cite{zhang2019lookahead}, with a constant learning rate of $0.001$.  Only horizontal flips are used as augmentations. All experiments are performed using a single NVIDIA Tesla P40 GPU.  

For the StyleGAN inversion task, the $\lambda$ values are set as $\lambda_1=1$, $\lambda_2=0.8$, and $\lambda_3=0.1$. For face frontalization, we increase the weight of the $\mathcal{L}_{\text{ID}}$, setting $\lambda_3=1$ and decrease the $\mathcal{L}_{\text{2}}$ and $\mathcal{L}_{\text{LPIPS}}$ loss functions, setting $\lambda_1=0.01$, $\lambda_2=0.8$ over the inner part of the face and $\lambda_1=0.001$, $\lambda_2=0.08$ elsewhere. Additionally, the constants used in the conditional image synthesis tasks are identical to those used in the inversion task except for the omission of the identity loss (i.e. $\lambda_3=0$). Finally, $\lambda_4$ is set to $0.005$ in all applications except for the StyleGAN inversion task, which does not utilize the regularization loss.

\section{Dataset Details}\label{datasets}
We conduct our experiments on the CelebA-HQ dataset~\cite{karras2018progressive}, which contains 30,000 high-quality images. We use a standard train-test split of the dataset, resulting in approximately 24,000 training images. The FFHQ dataset from~\cite{karras2019style}, which contains 70,000 face images, is used for the StyleGAN inversion and face frontalization tasks. 

For the generation of real images from sketches, we construct a dataset representative of hand-drawn sketches using the CelebA-HQ dataset. Given an input image, we first apply a ``pencil sketch" filter which retains most facial details of the original image while removing the remaining noise. We then apply the sketch-simplification method by \cite{sketch-simpf}, resulting in images resembling hand-drawn sketches. The same approach is also used for generating the sketch images on the AFHQ Cat and AFHQ Dog datasets \cite{choi2020stargan}.

\begin{figure}
    \setlength{\tabcolsep}{1pt}
    \centering
        \begin{tabular}{c c c c c}
    
            \includegraphics[width=0.088\textwidth]{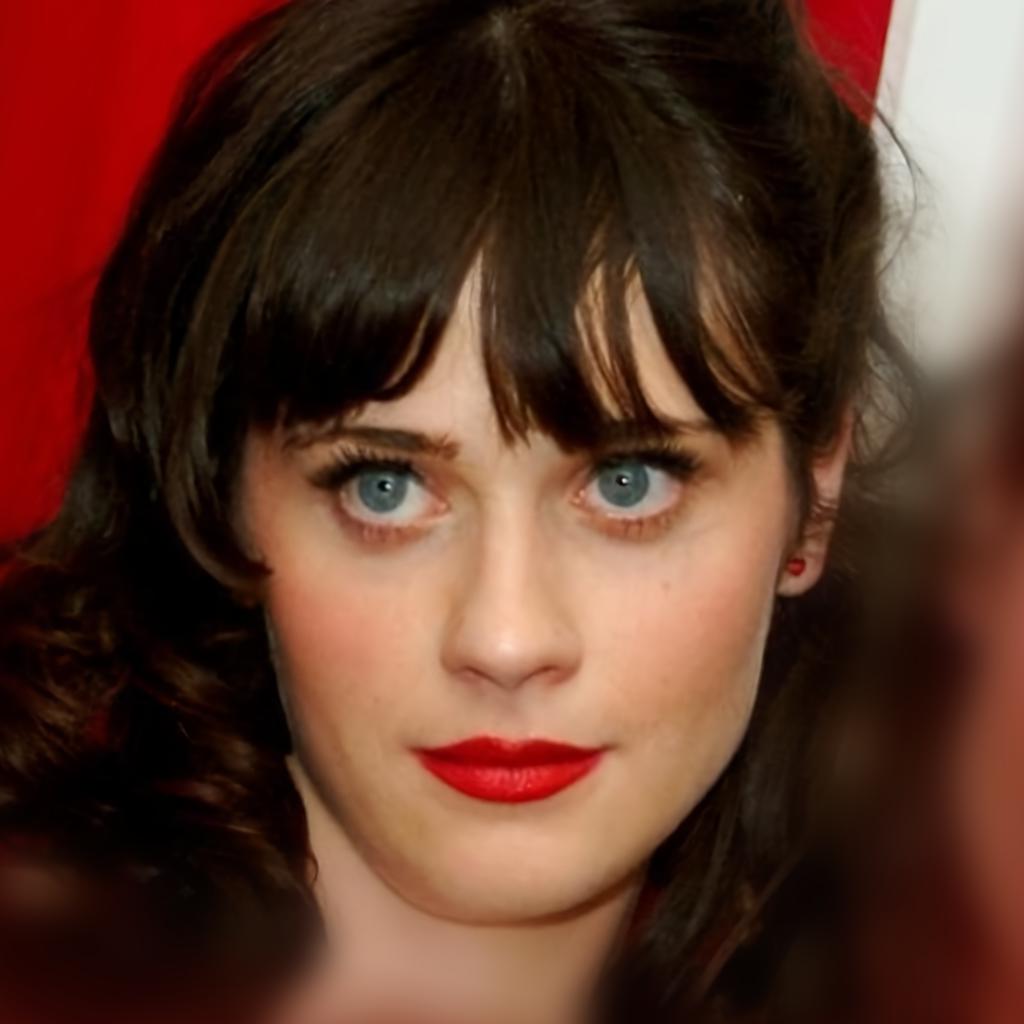}&
            \includegraphics[width=0.088\textwidth]{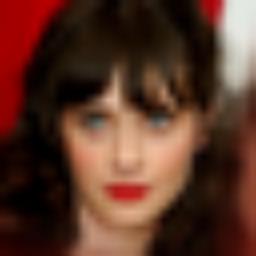}&
            \includegraphics[width=0.088\textwidth]{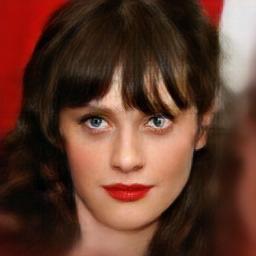}&
            \includegraphics[width=0.088\textwidth]{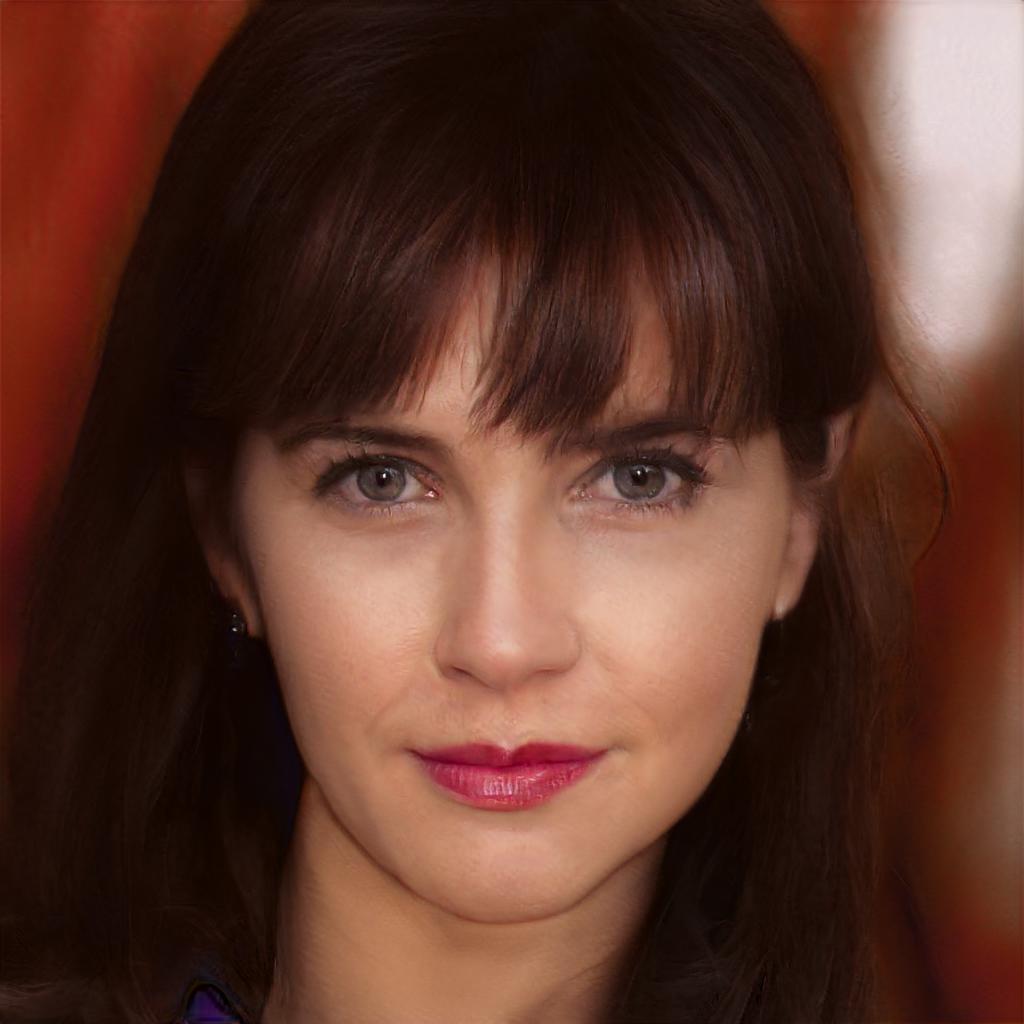}&
            \includegraphics[width=0.088\textwidth]{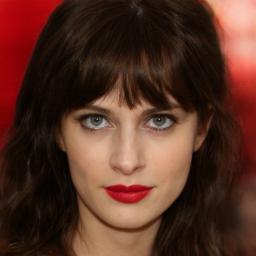}
            \tabularnewline
    
            \includegraphics[width=0.088\textwidth]{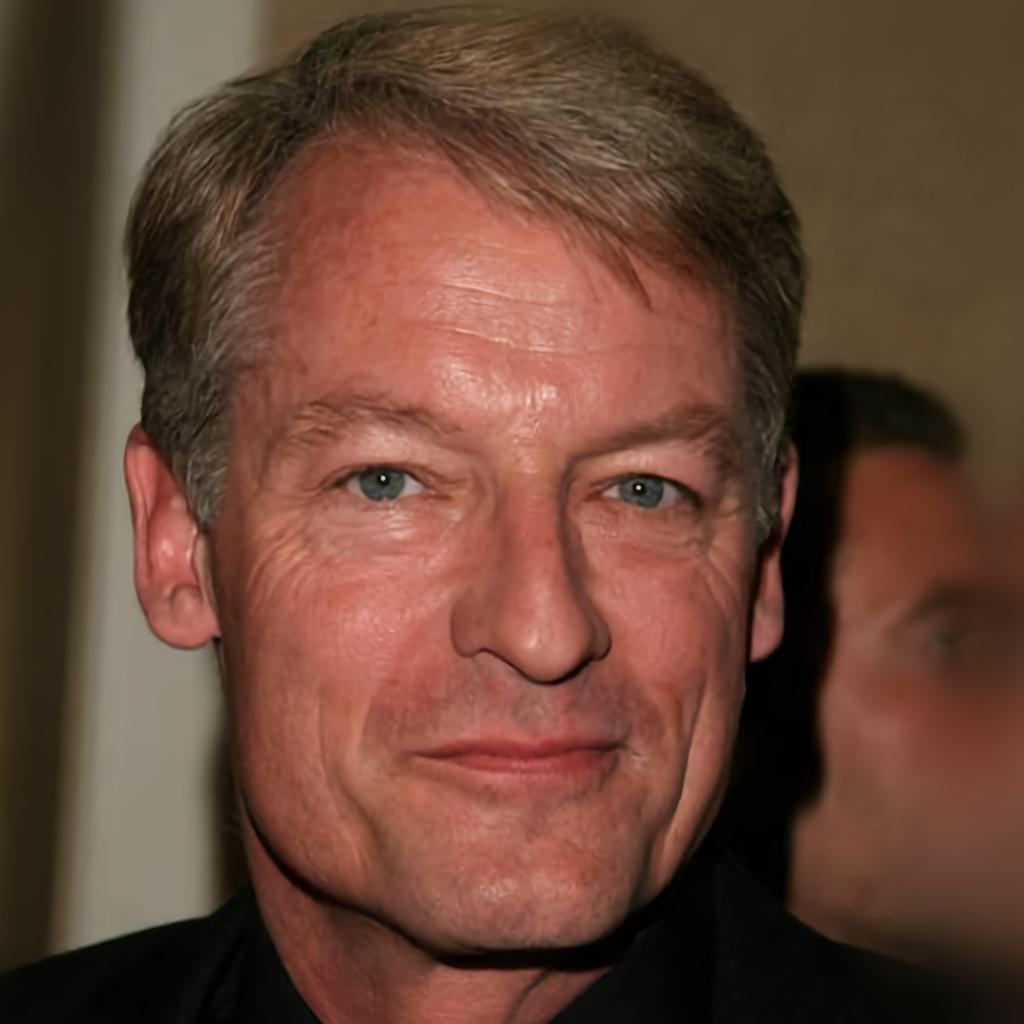}&
            \includegraphics[width=0.088\textwidth]{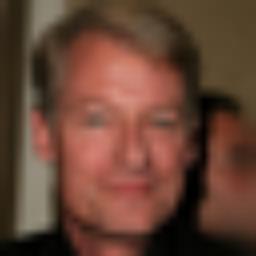}&
            \includegraphics[width=0.088\textwidth]{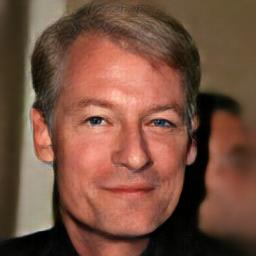}&
            \includegraphics[width=0.088\textwidth]{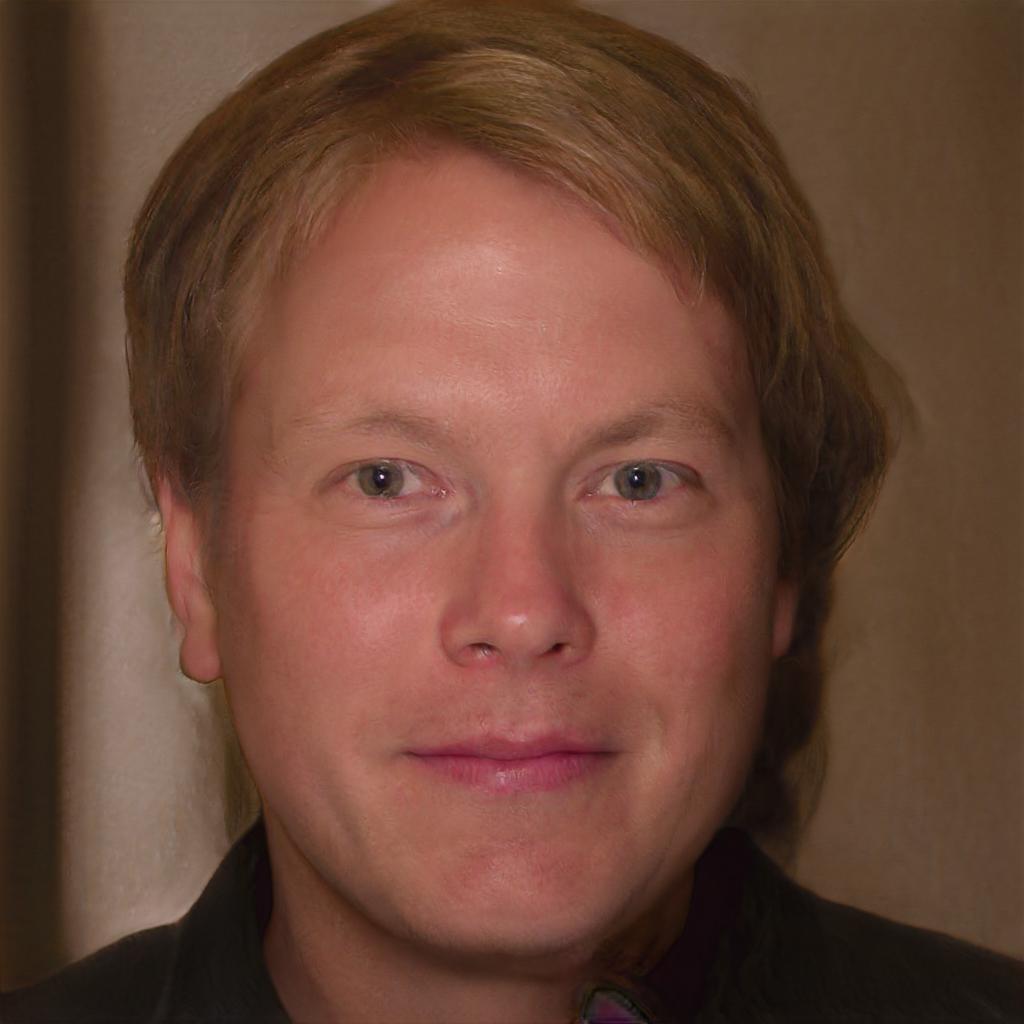}&
            \includegraphics[width=0.088\textwidth]{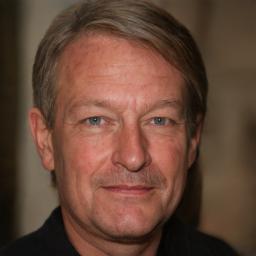}
            \tabularnewline
    
            \includegraphics[width=0.088\textwidth]{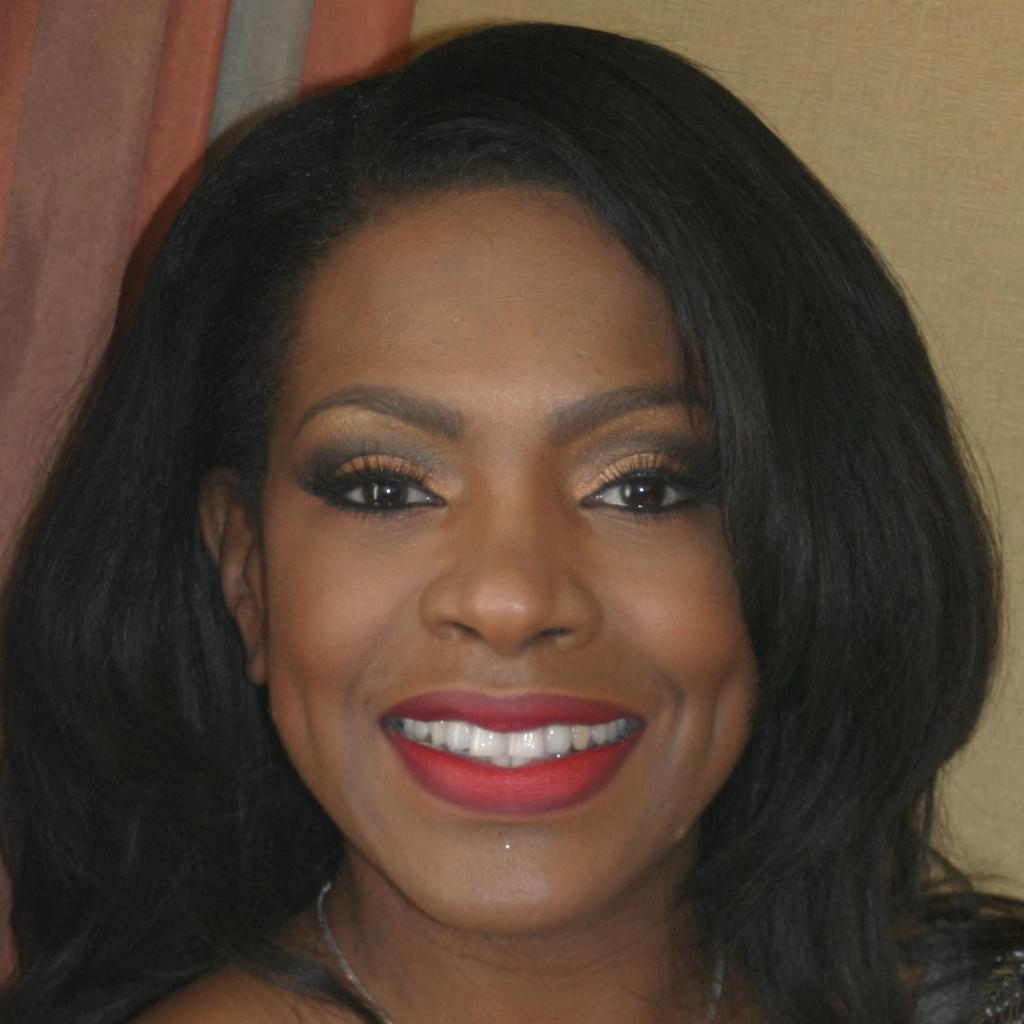}&
            \includegraphics[width=0.088\textwidth]{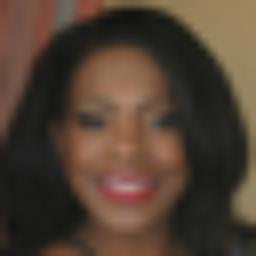}&
            \includegraphics[width=0.088\textwidth]{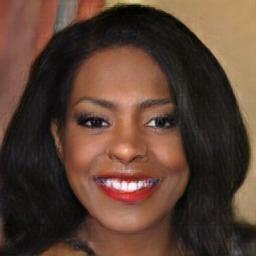}&
            \includegraphics[width=0.088\textwidth]{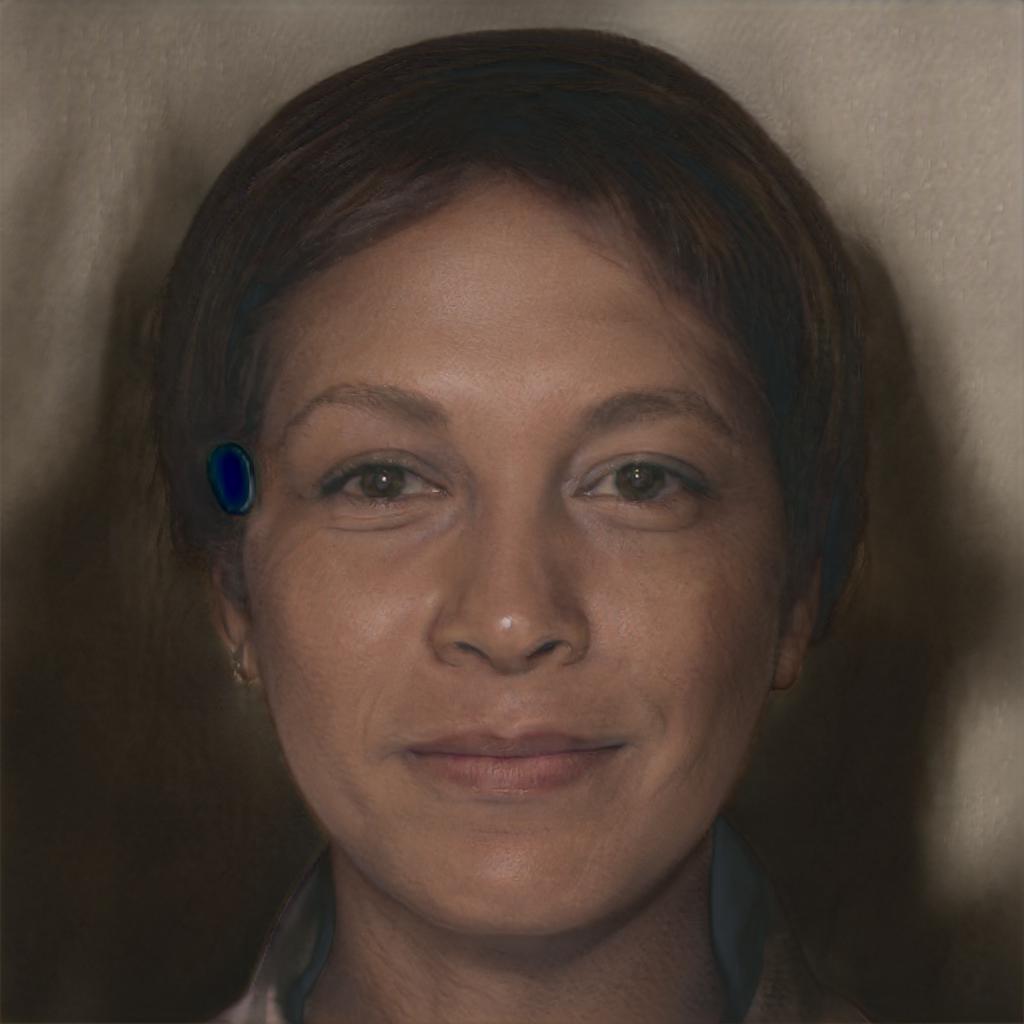}&
            \includegraphics[width=0.088\textwidth]{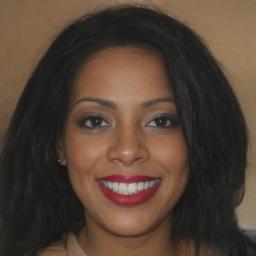}
            \tabularnewline
            Original & LR $\times8$ & pix2pixHD & PULSE & pSp
            
        \end{tabular}
        \vspace{0.1cm}
        \caption{Comparison of super-resolution approaches with $\times8$ down-sampling on the  CelebA-HQ~\cite{karras2018progressive} test set.}
    \label{fig:super_res_8}
\end{figure}

\begin{figure}
    \setlength{\tabcolsep}{1pt}
    \centering
        \begin{tabular}{c c c c c}
            \includegraphics[width=0.088\textwidth]{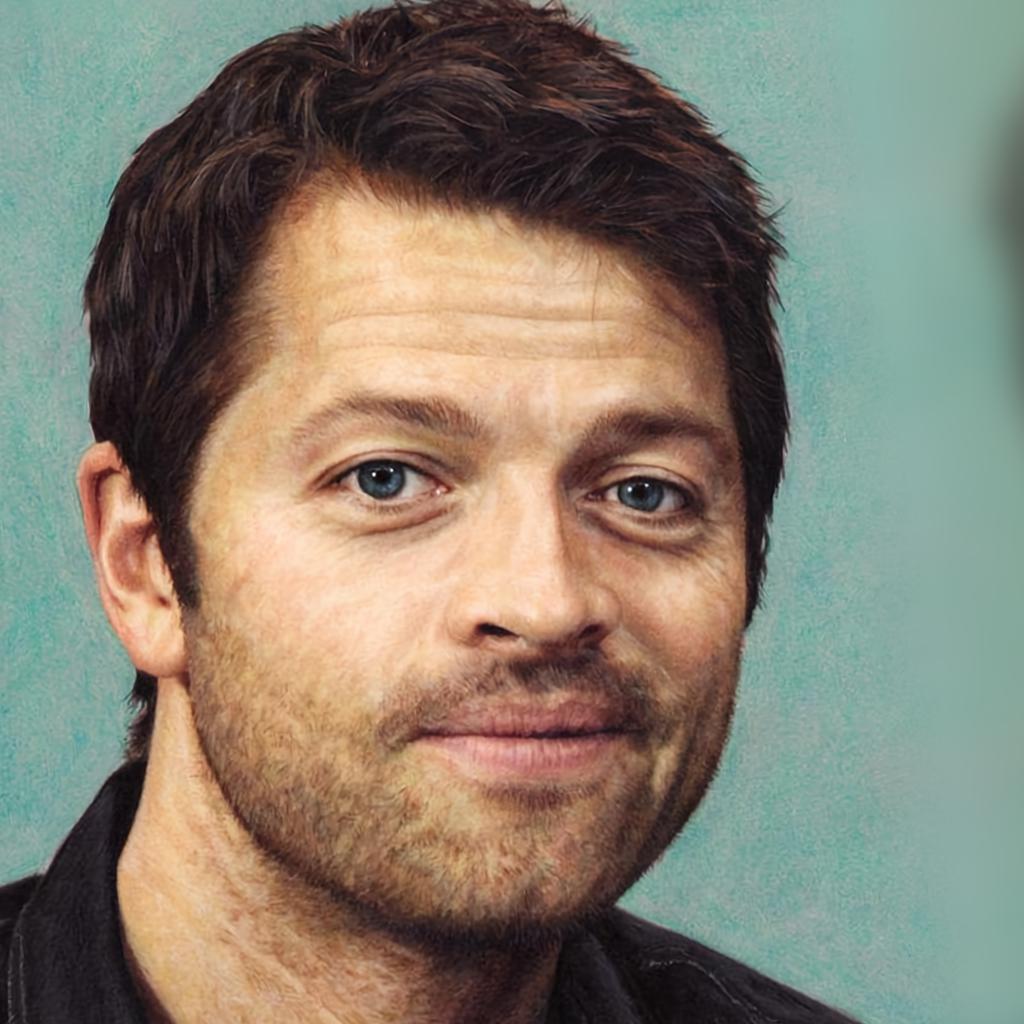}&
            \includegraphics[width=0.088\textwidth]{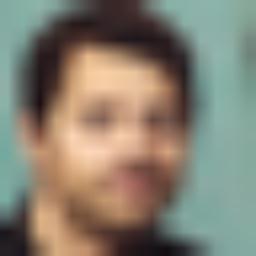}&
            \includegraphics[width=0.088\textwidth]{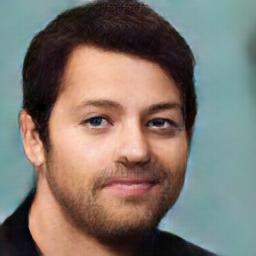}&
            \includegraphics[width=0.088\textwidth]{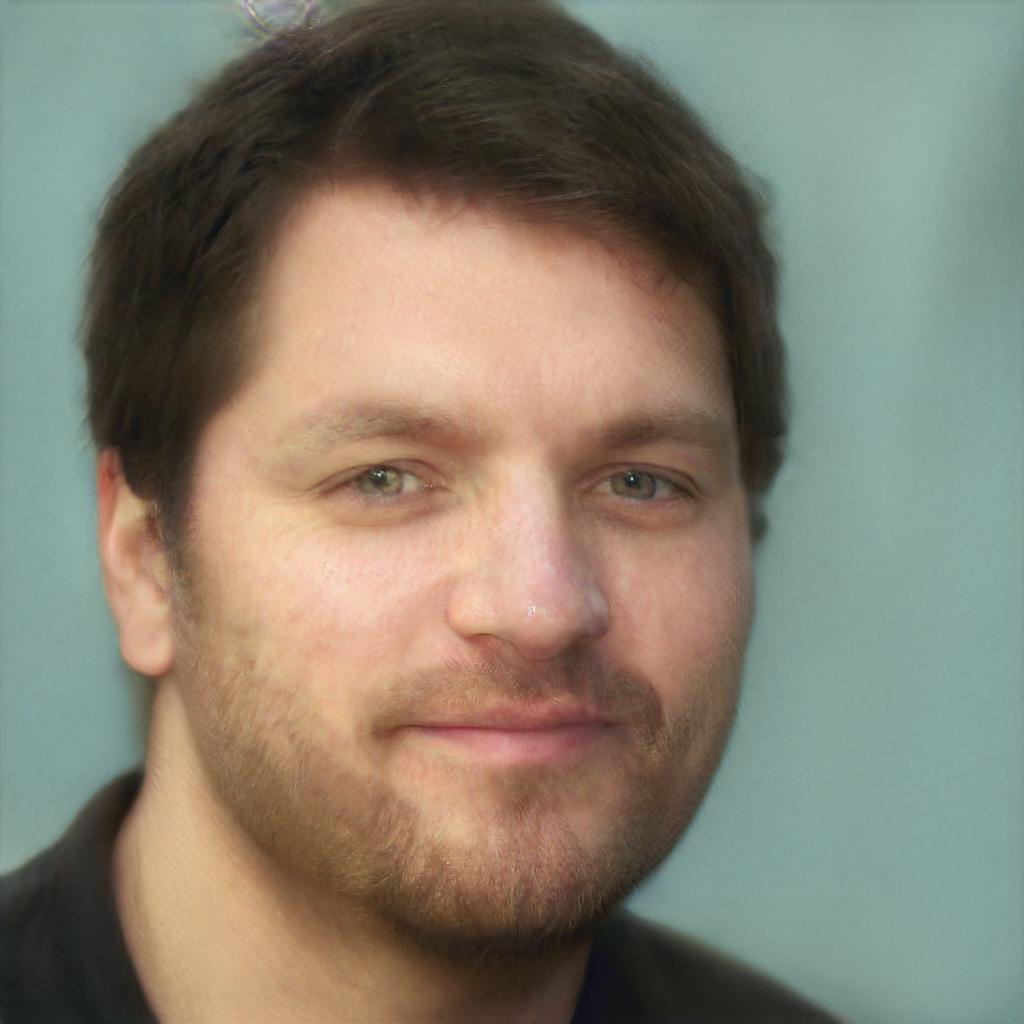}&
            \includegraphics[width=0.088\textwidth]{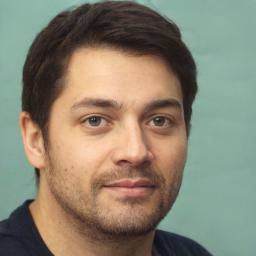}
            \tabularnewline
            
            \includegraphics[width=0.088\textwidth]{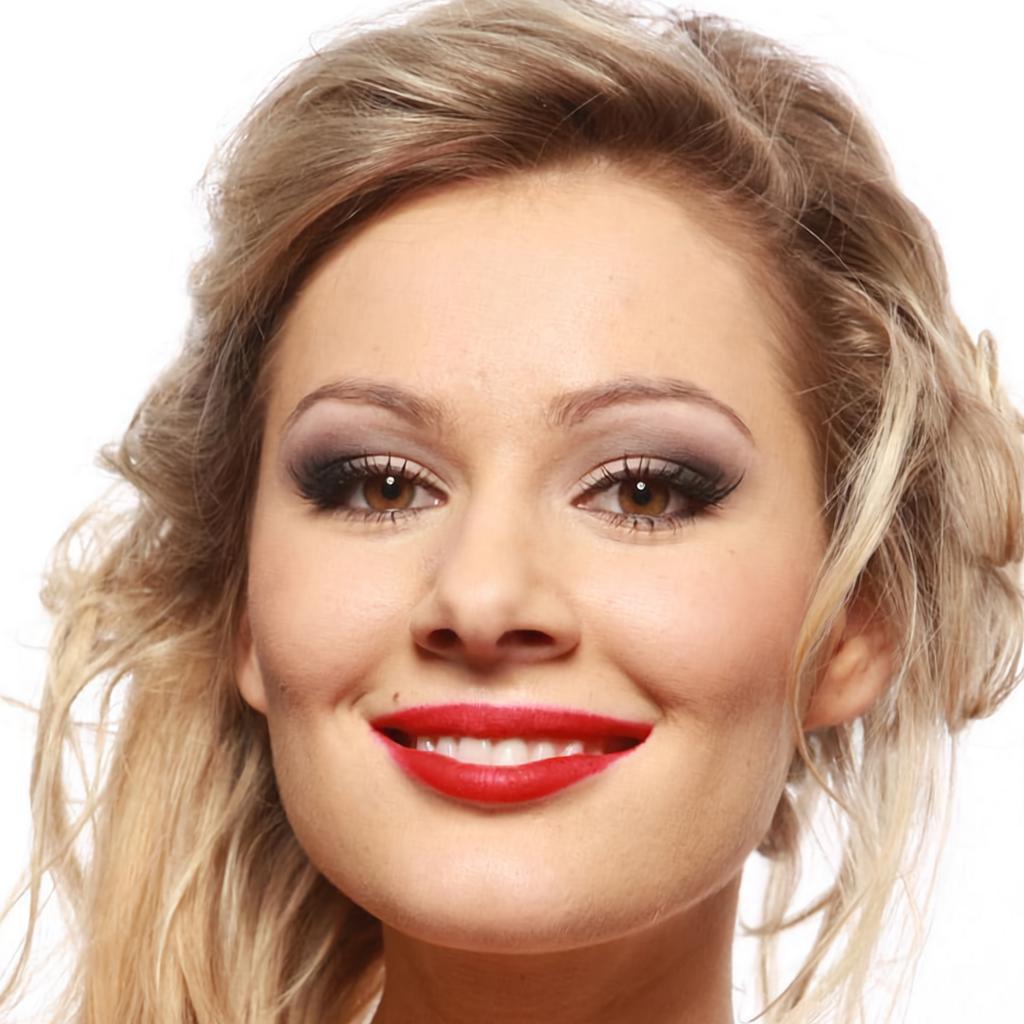}&
            \includegraphics[width=0.088\textwidth]{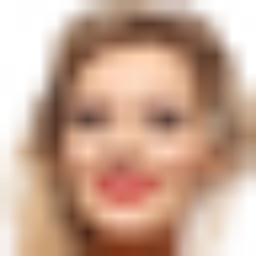}&
            \includegraphics[width=0.088\textwidth]{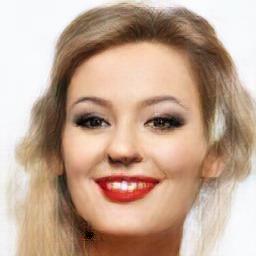}&
            \includegraphics[width=0.088\textwidth]{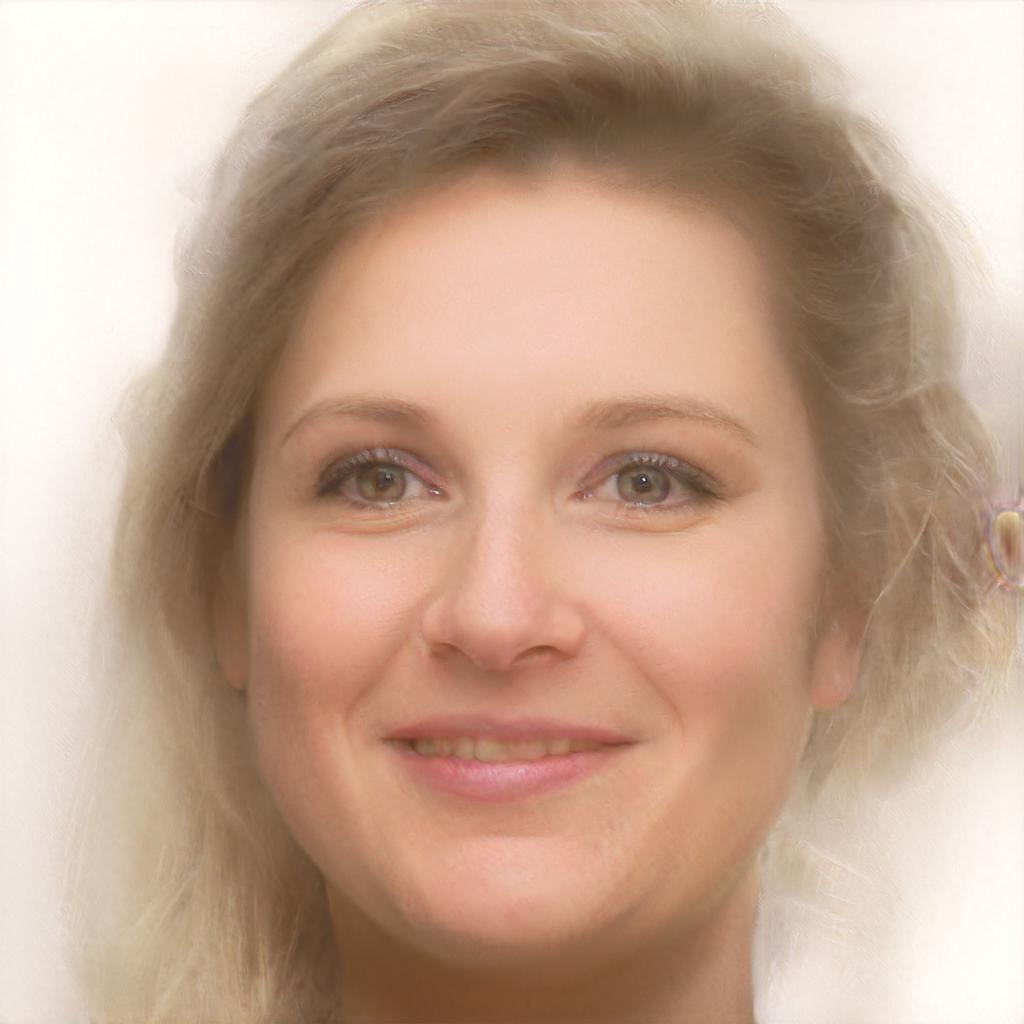}&
            \includegraphics[width=0.088\textwidth]{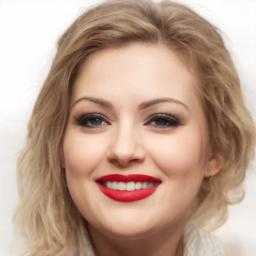}
            \tabularnewline
            
            \includegraphics[width=0.088\textwidth]{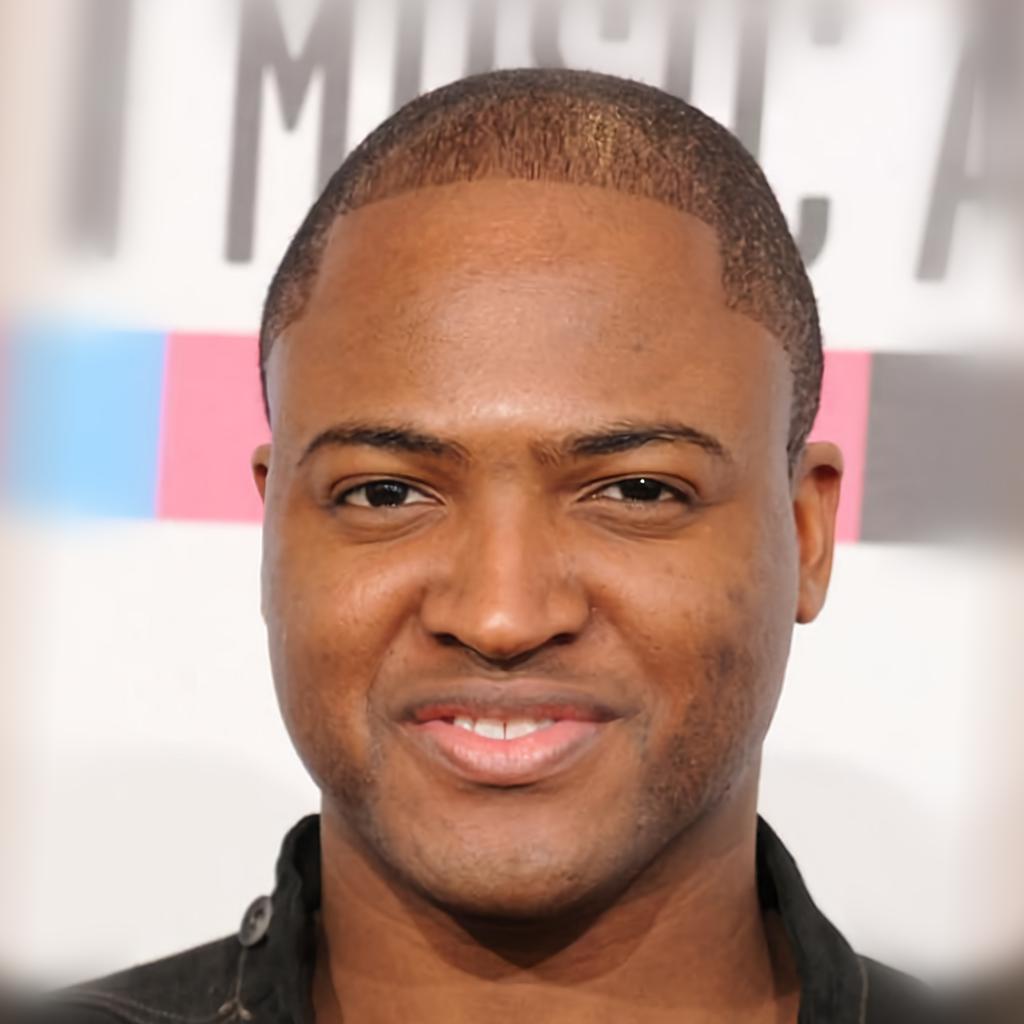}&
            \includegraphics[width=0.088\textwidth]{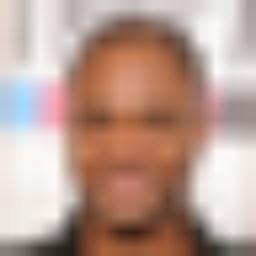}&
            \includegraphics[width=0.088\textwidth]{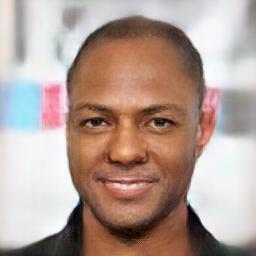}&
            \includegraphics[width=0.088\textwidth]{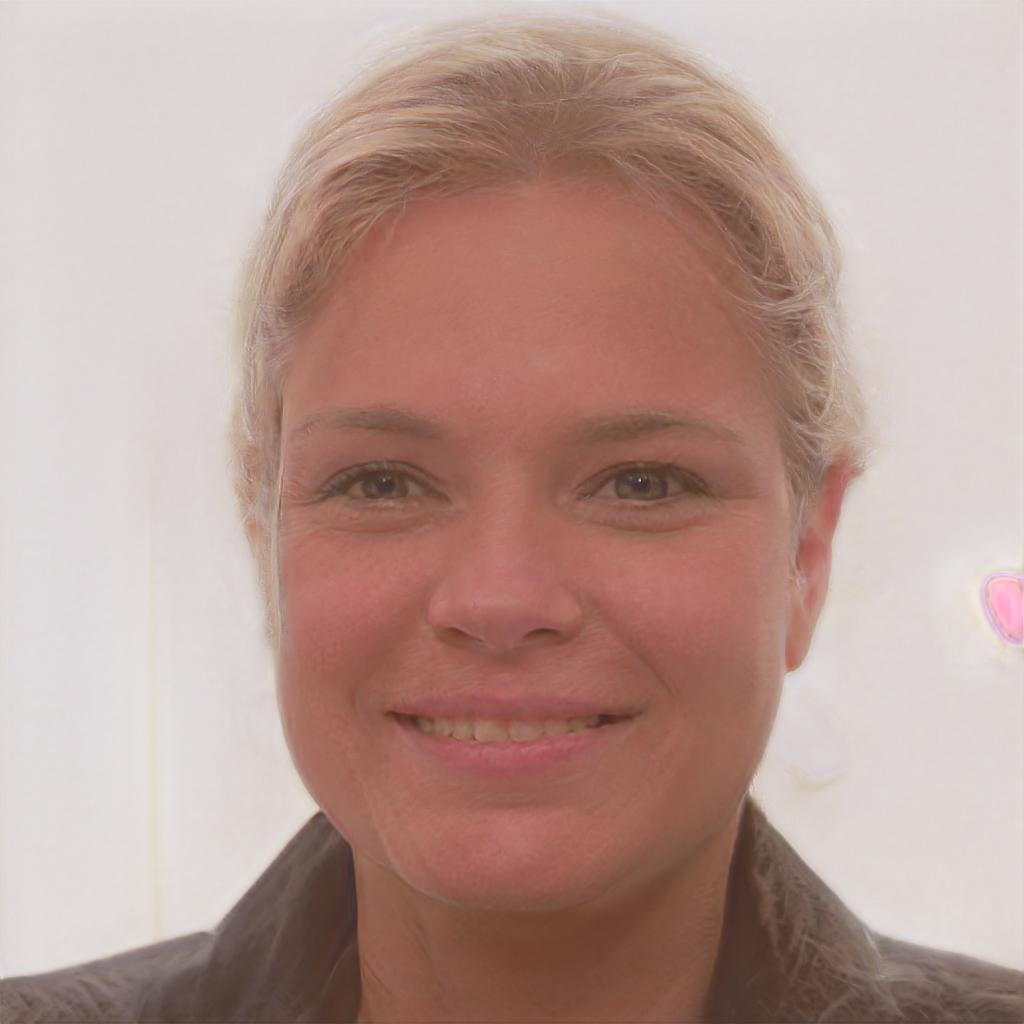}&
            \includegraphics[width=0.088\textwidth]{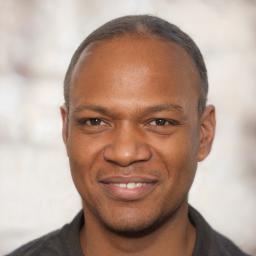}
            \tabularnewline
            Original & LR $\times16$ & pix2pixHD & PULSE & pSp
        \end{tabular}
        \vspace{0.1cm}
        \caption{Comparison of super-resolution approaches with $\times16$ down-sampling on the  CelebA-HQ~\cite{karras2018progressive} test set.}
        \label{fig:super_res_16}
\end{figure}

\section{Application Details}\label{application_details}
\subsection{Super Resolution}
In super resolution, the pSp framework is used to construct high-resolution (HR) images from corresponding low-resolution (LR) input images. PULSE \cite{Menon_2020_CVPR} approaches this task in an unsupervised manner by traversing the HR image manifold in search of an image that downsamples to the input LR image. 

\paragraph{\textit{\textbf{Methodology and details.}}} We train both our model and pix2pixHD~\cite{wang2018high} in a supervised fashion, where for each input we perform random bi-cubic down-sampling of $\times 1$ (i.e. no down-sampling), $\times2, \times4, \times8$, $\times16$, or $\times32$ and set the original, full resolution image as the target.

\begin{figure}
    \setlength{\tabcolsep}{1pt}
    \centering
        \begin{tabular}{c c c c}
            \includegraphics[width=0.11\textwidth]{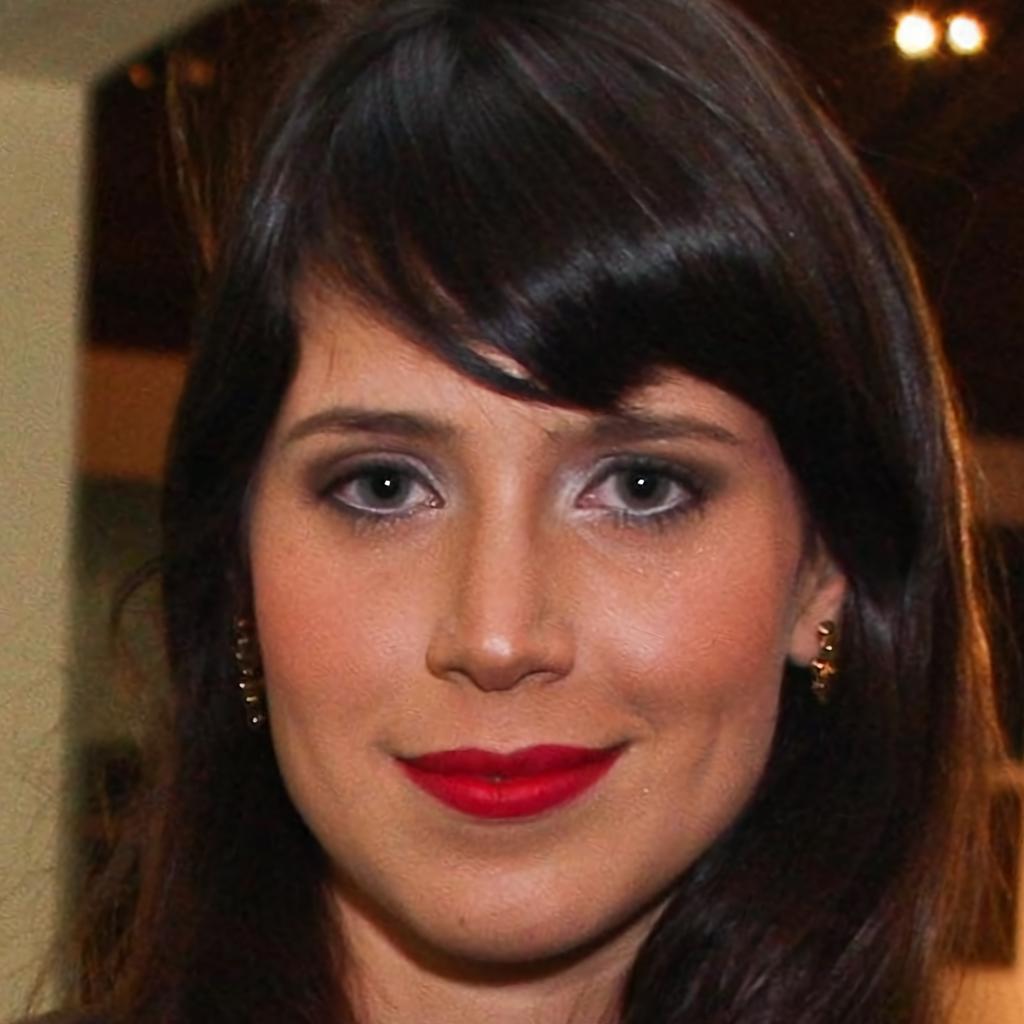}&
            \includegraphics[width=0.11\textwidth]{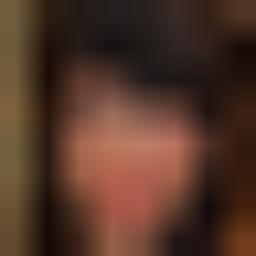}&
            \includegraphics[width=0.11\textwidth]{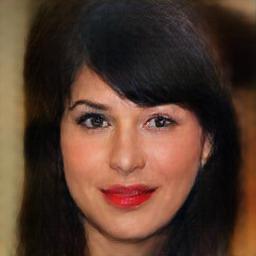}&
            \includegraphics[width=0.11\textwidth]{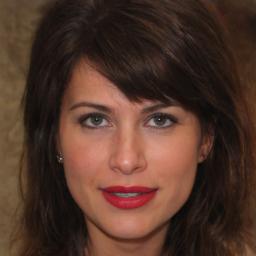}
            \tabularnewline

            \includegraphics[width=0.11\textwidth]{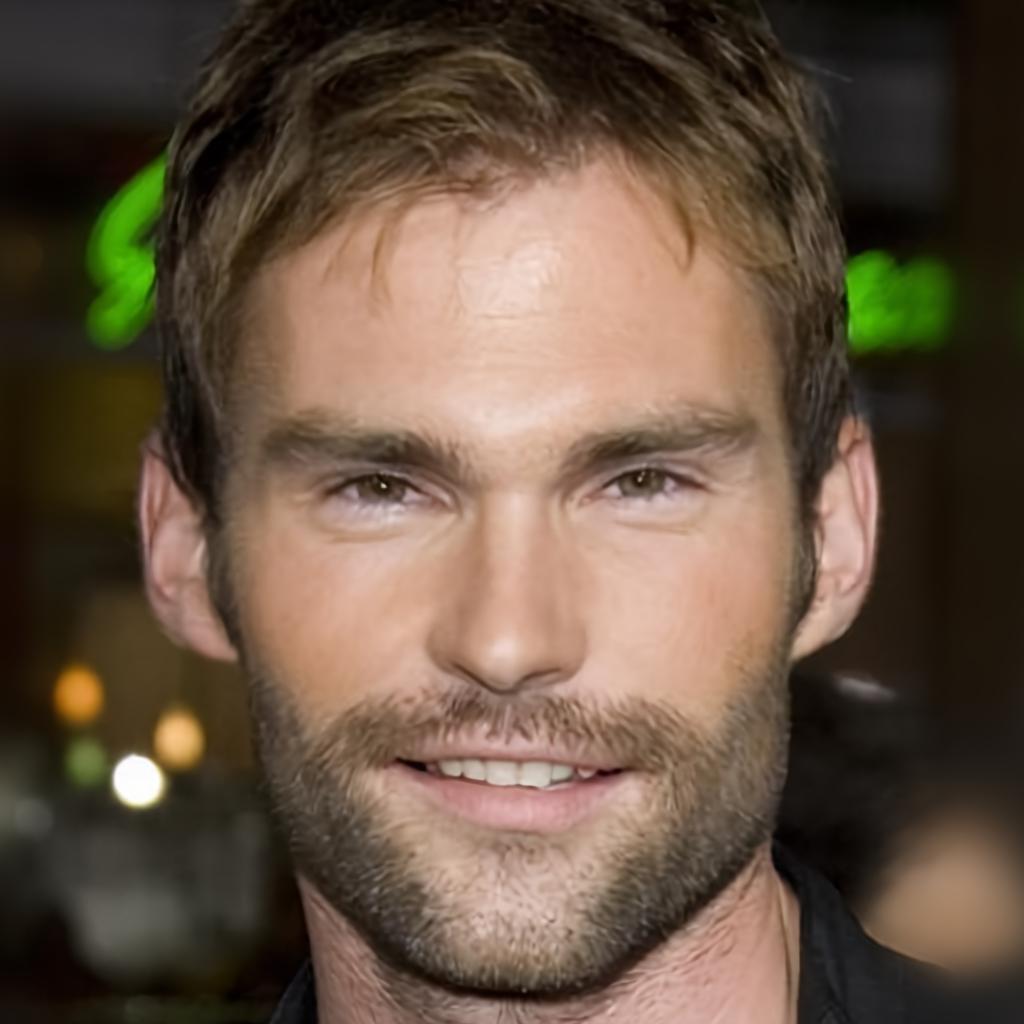}&
            \includegraphics[width=0.11\textwidth]{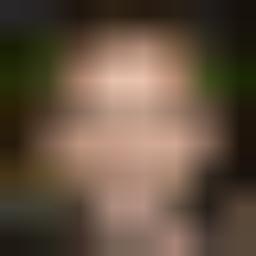}&
            \includegraphics[width=0.11\textwidth]{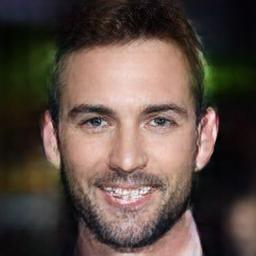}&
            \includegraphics[width=0.11\textwidth]{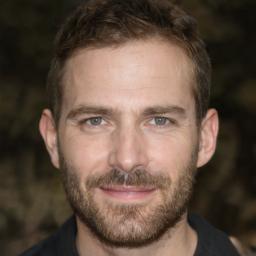}
            \tabularnewline
            
            \includegraphics[width=0.11\textwidth]{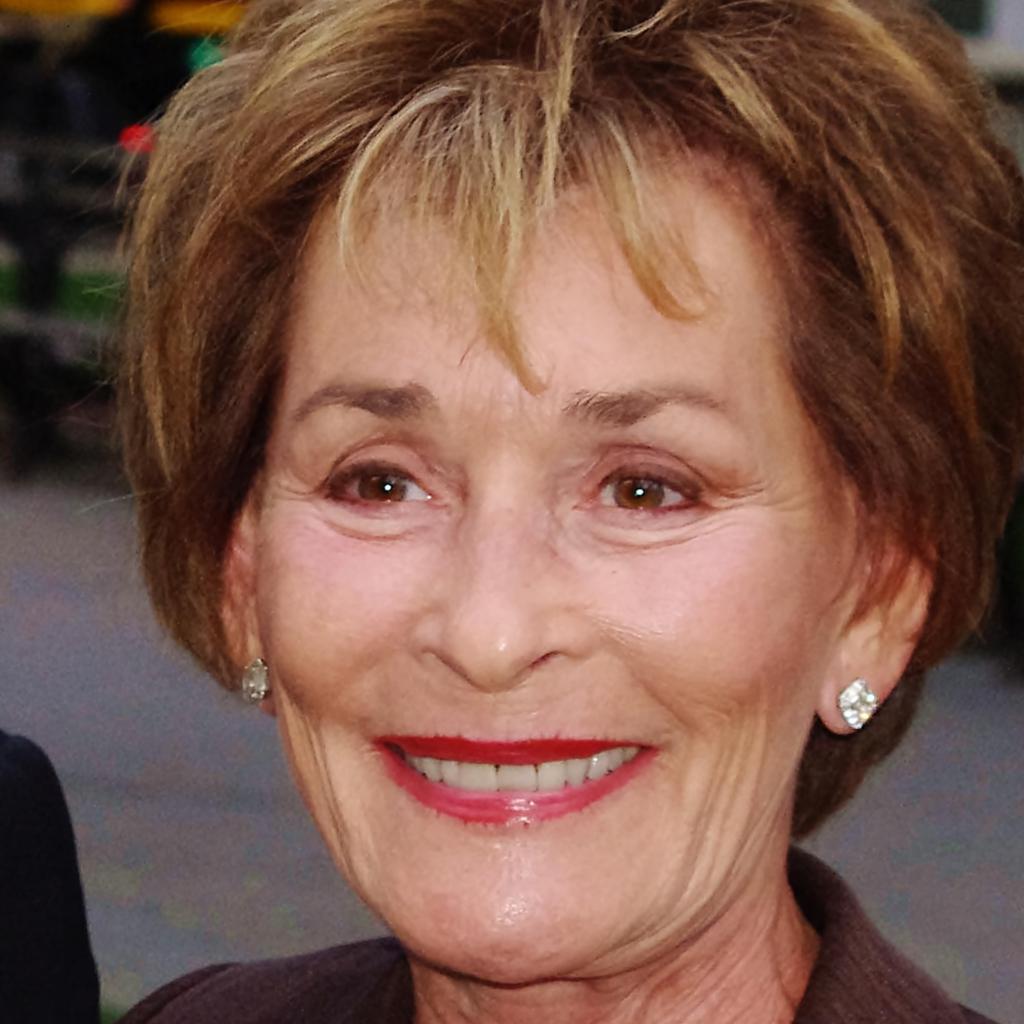}&
            \includegraphics[width=0.11\textwidth]{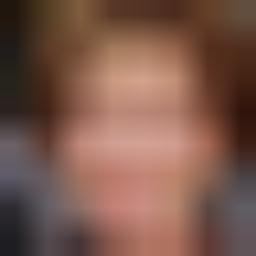}&
            \includegraphics[width=0.11\textwidth]{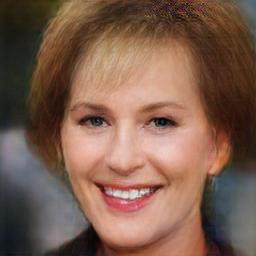}&
            \includegraphics[width=0.11\textwidth]{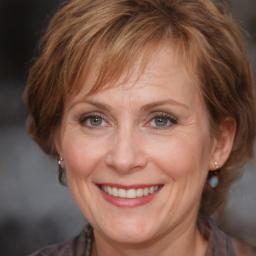}
            \tabularnewline
            Original & LR $\times32$ & pix2pixHD & pSp
        \end{tabular}
        \vspace{0.1cm}
        \caption{Comparison of super-resolution approaches with $\times32$ down-sampling on the  CelebA-HQ~\cite{karras2018progressive} test set.}
        \label{fig:super_res_32}
\end{figure}

\begin{figure}
    \setlength{\tabcolsep}{1pt}
    \centering
    \begin{tabular}{c}
        \includegraphics[width=0.47\textwidth]{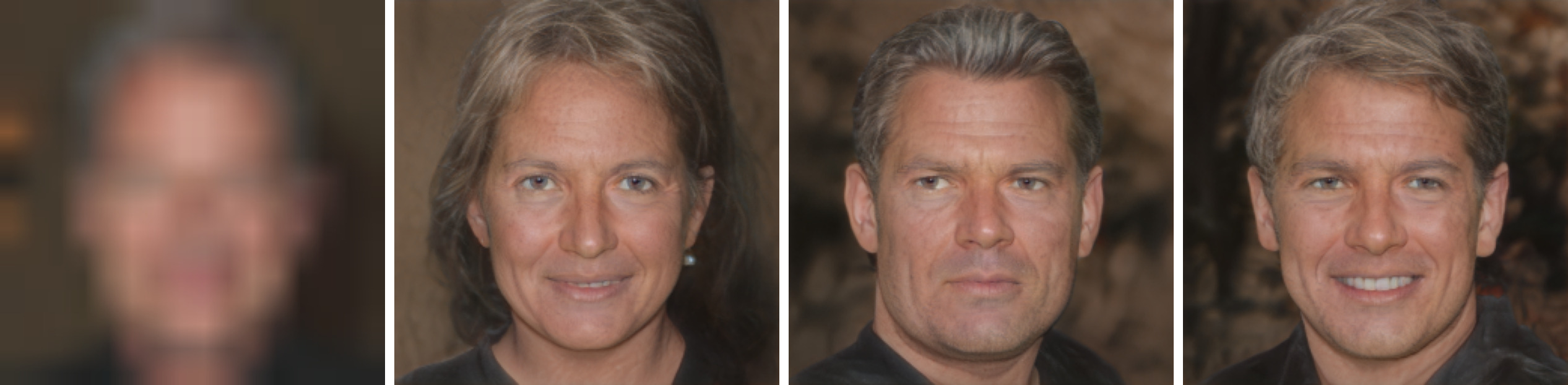}
        \tabularnewline
        \includegraphics[width=0.47\textwidth]{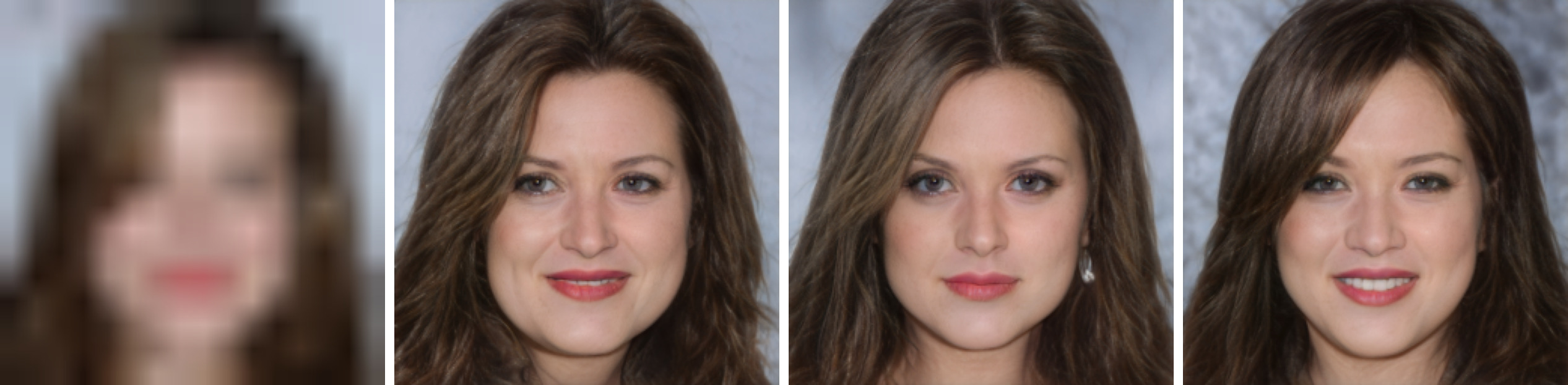}
        \tabularnewline
        \includegraphics[width=0.47\textwidth]{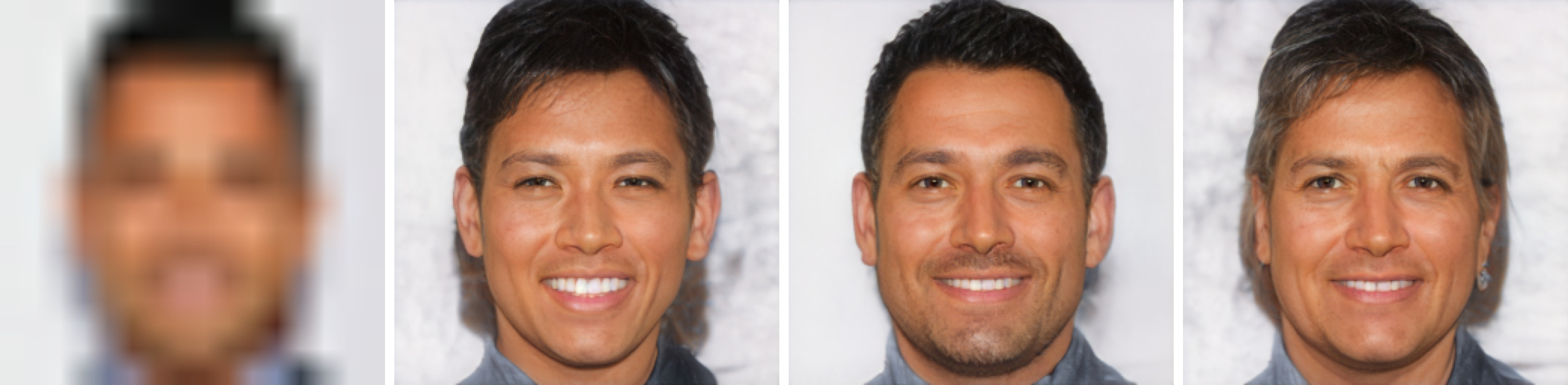}
        \tabularnewline
    \end{tabular}
    \caption{Multi-modal synthesis for super-resolution using pSp with style-mixing.}
    \label{fig:super_res_style_mixing}
\end{figure}

\paragraph{\textit{\textbf{Results.}}} Figures \ref{fig:super_res_8}-\ref{fig:super_res_32} demonstrates the visual quality of the resulting images from our method along with those of the previous approaches. Although PULSE is able to achieve very high-quality results due to their usage of StyleGAN to generate images, they are unable to accurately reconstruct the original image even when performing down-sampling of $\times8$ to a resolution of $32\times32$. By learning a pixel-wise correspondence between the LR and HR images, pix2pixHD is able to obtain satisfying results even when down-sampled to a resolution of $16\times16$ (i.e. $\times16$ down-sampling). However, visually, their results appear less photo-realistic.  Contrary to these previous works, we are able to obtain high-quality results even when down-sampling to resolutions of $16\times16$ and $8\times8$.
Finally, in Figure~\ref{fig:super_res_style_mixing} we generate multiple outputs for a given LR image using our multi-modal technique by performing style-mixing with a randomly sampled $\textbf{w}$ vector on layers (4-7) with an $\alpha$ value of $0.5$. Doing so alters medium-level styles that mainly control facial features.

\begin{figure}
    \setlength{\tabcolsep}{1pt}
    \centering
    \begin{tabular}{c c c c c}
        \raisebox{0.25in}{\rotatebox[origin=t]{90}{Input}}&
        \includegraphics[width=0.11\textwidth]{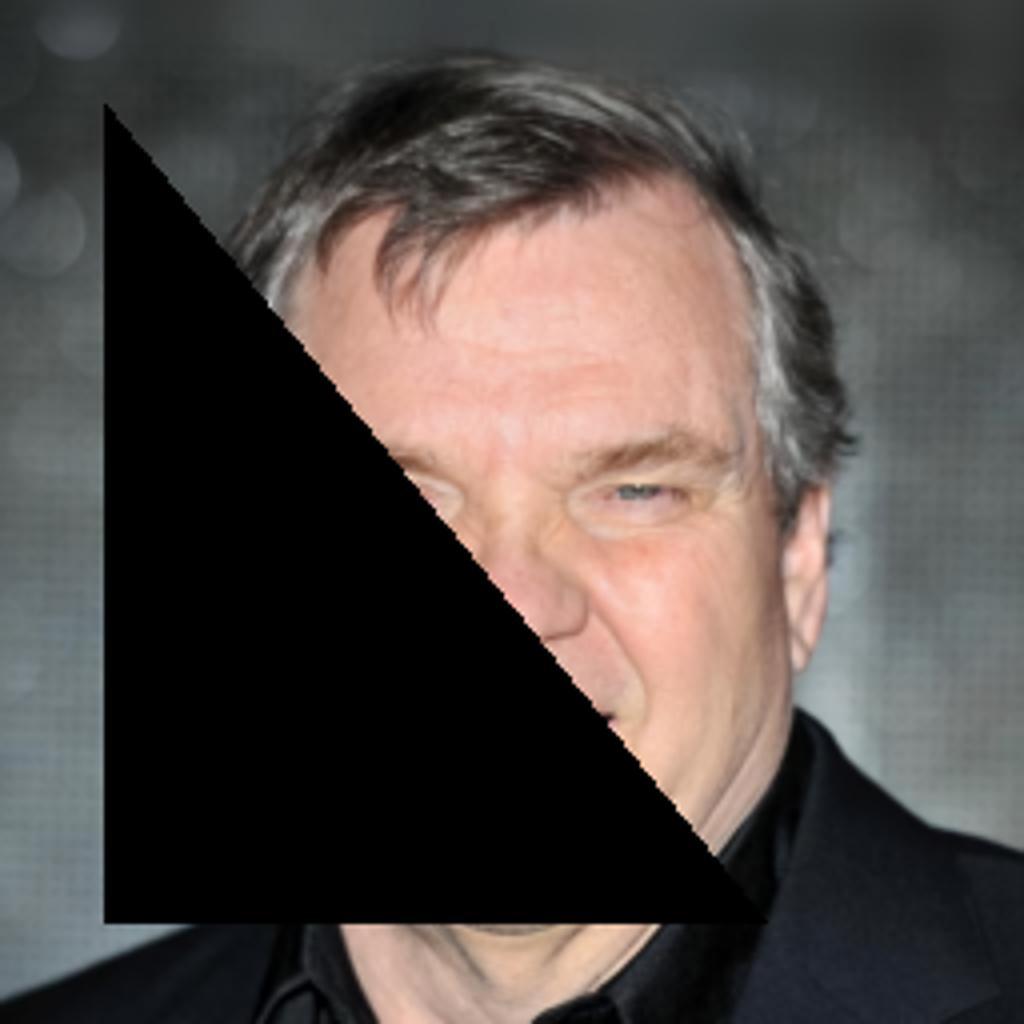}&
        \includegraphics[width=0.11\textwidth]{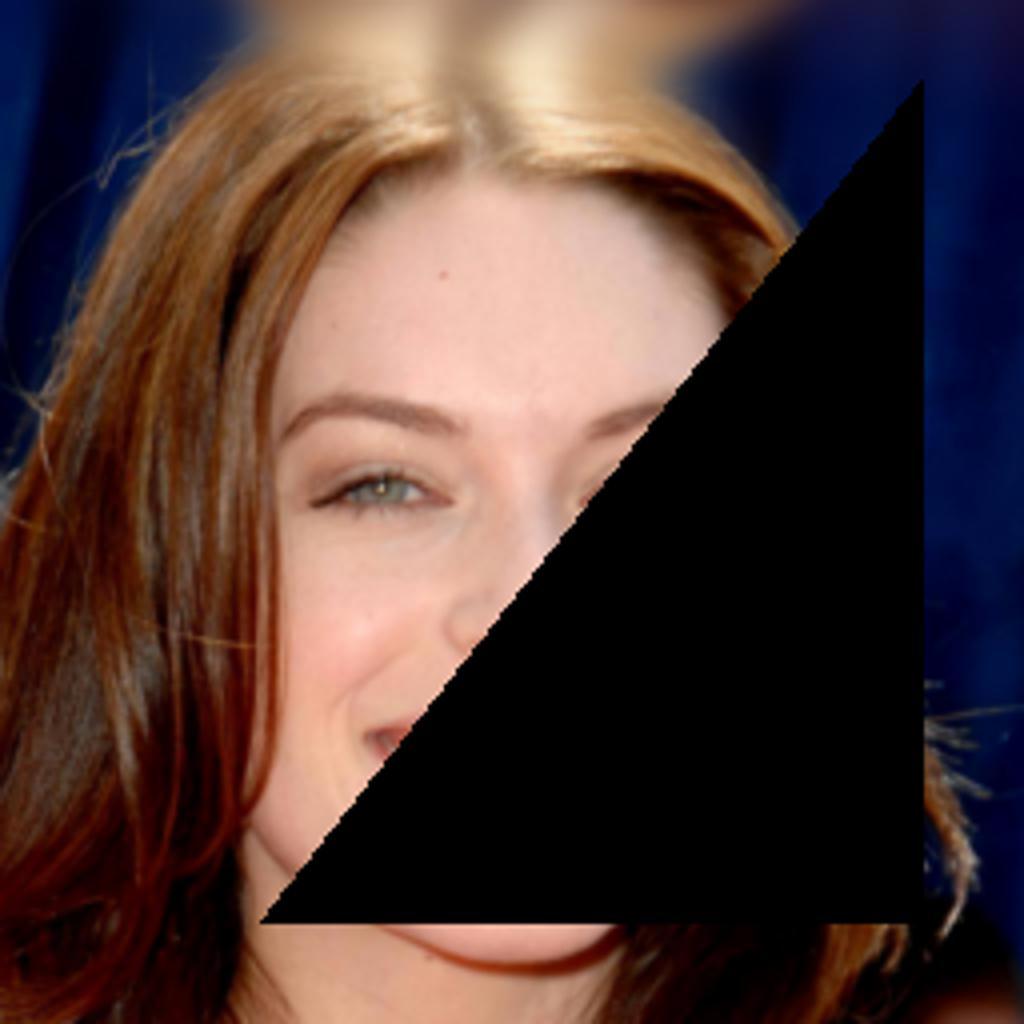}&
        \includegraphics[width=0.11\textwidth]{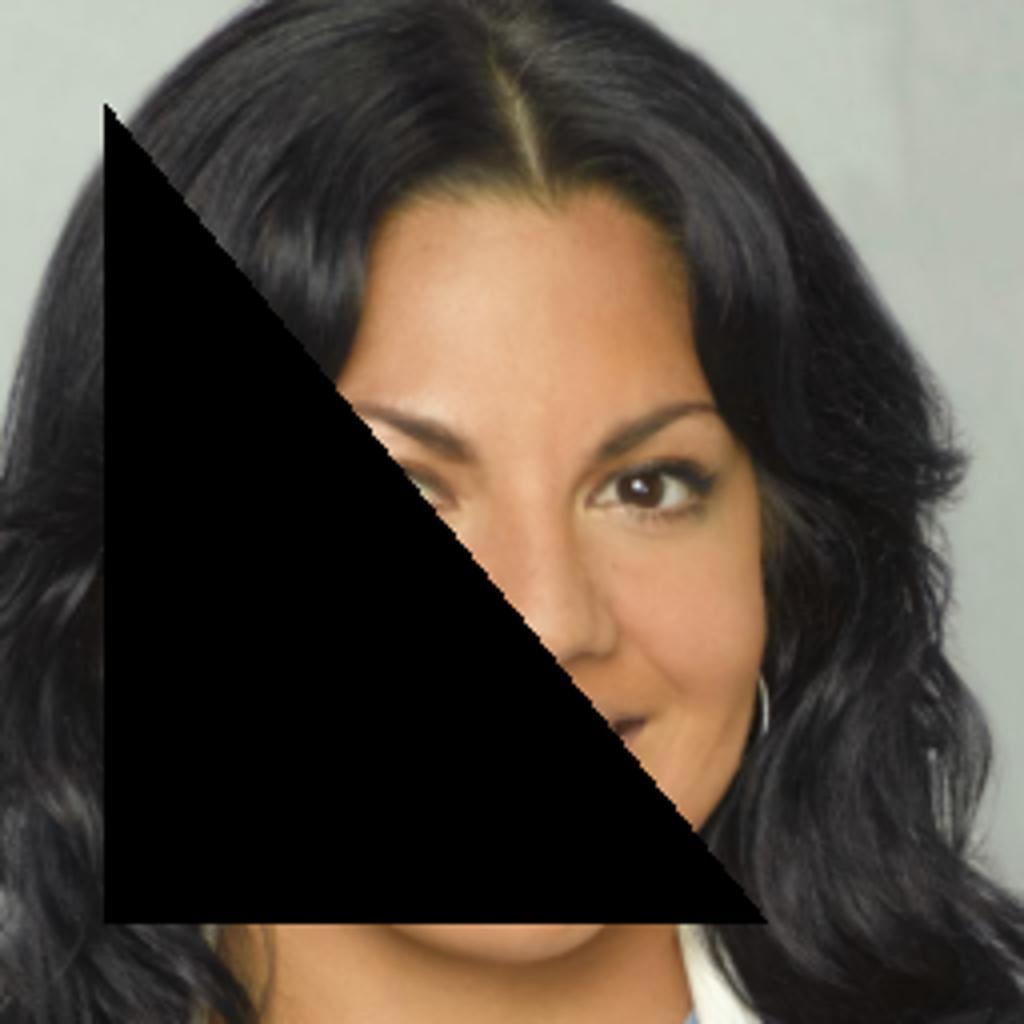}&
        \includegraphics[width=0.11\textwidth]{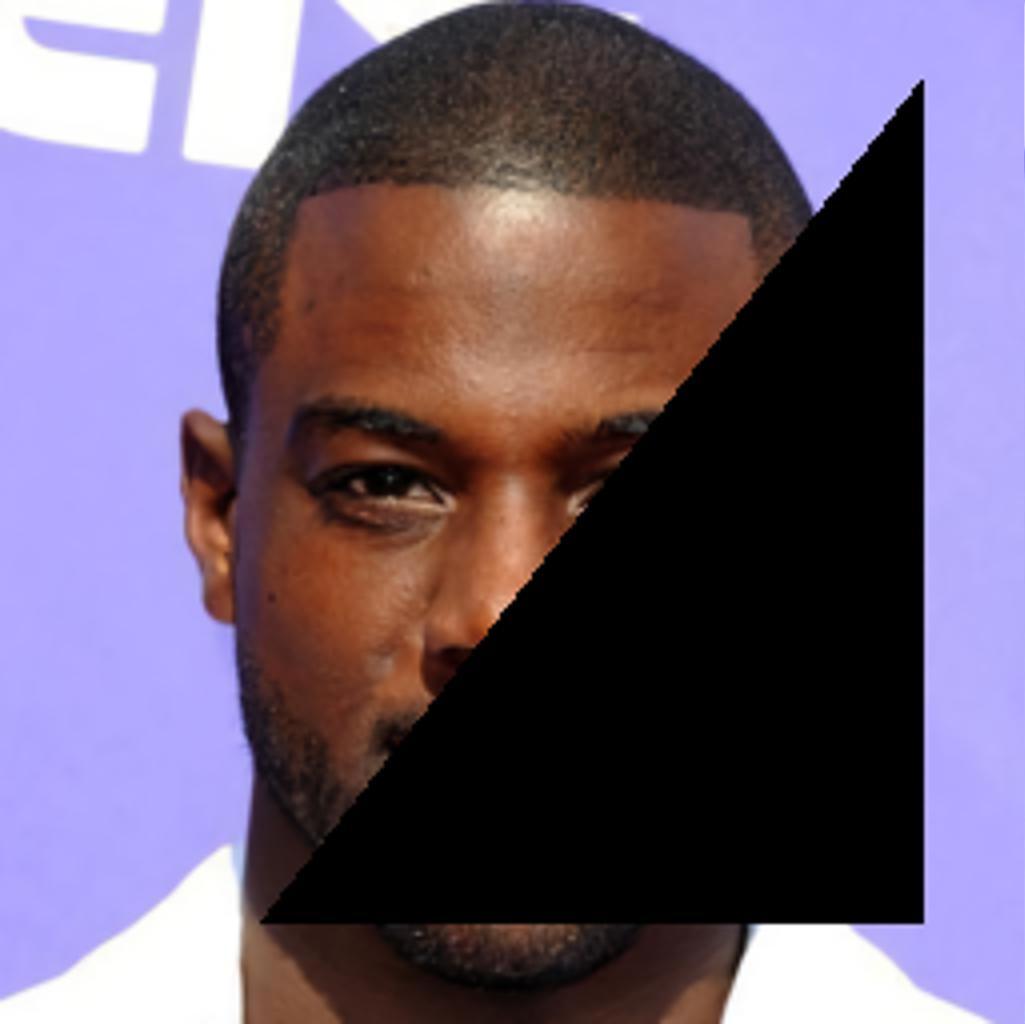}
        \tabularnewline
        \raisebox{0.3in}{\rotatebox[origin=t]{90}{pix2pixHD}}&
        \includegraphics[width=0.11\textwidth]{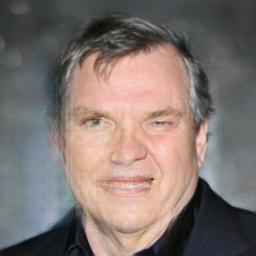}&
        \includegraphics[width=0.11\textwidth]{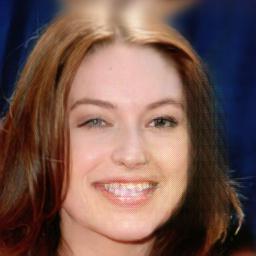}&
        \includegraphics[width=0.11\textwidth]{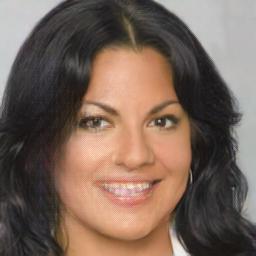}&
        \includegraphics[width=0.11\textwidth]{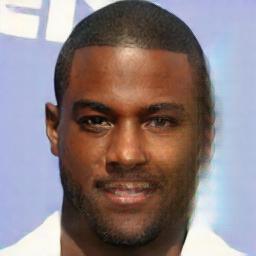}
        \tabularnewline
        \raisebox{0.3in}{\rotatebox[origin=t]{90}{pSp}}&
        \includegraphics[width=0.11\textwidth]{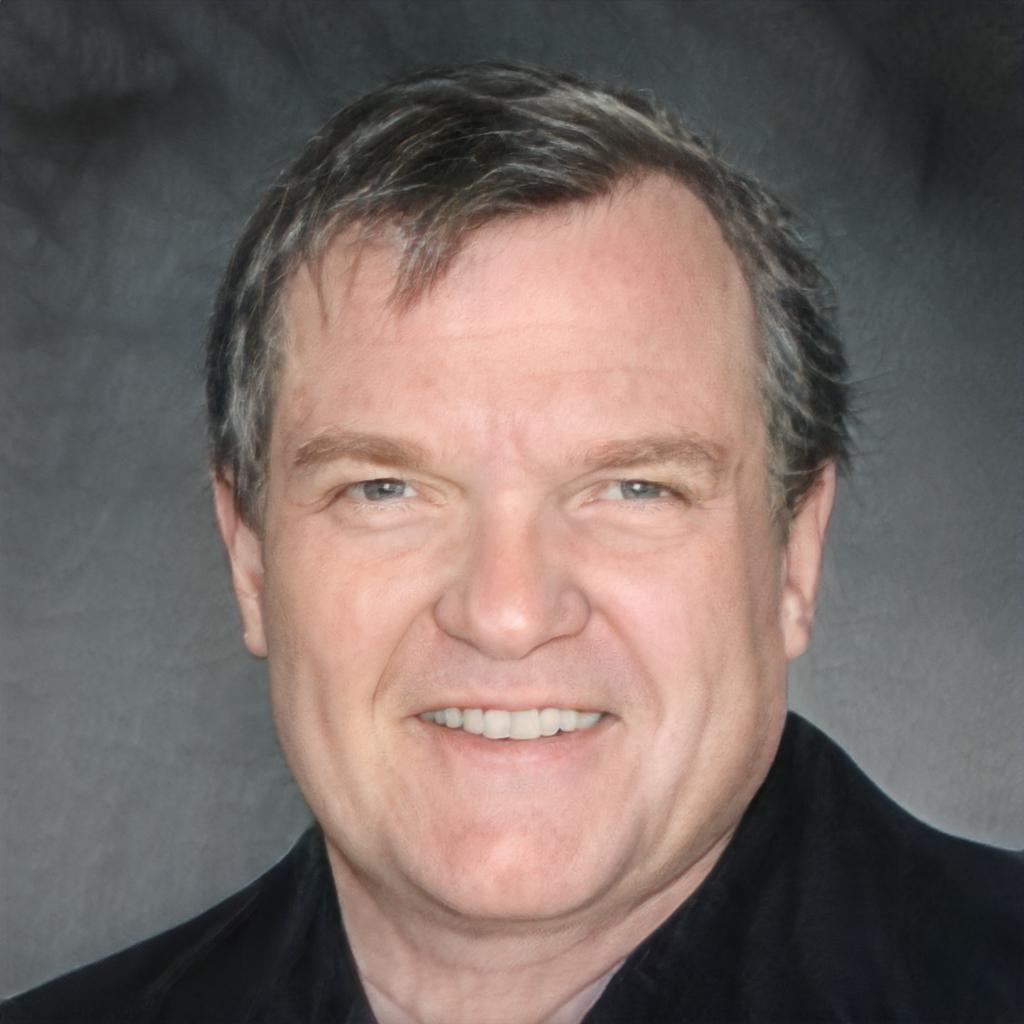}&
        \includegraphics[width=0.11\textwidth]{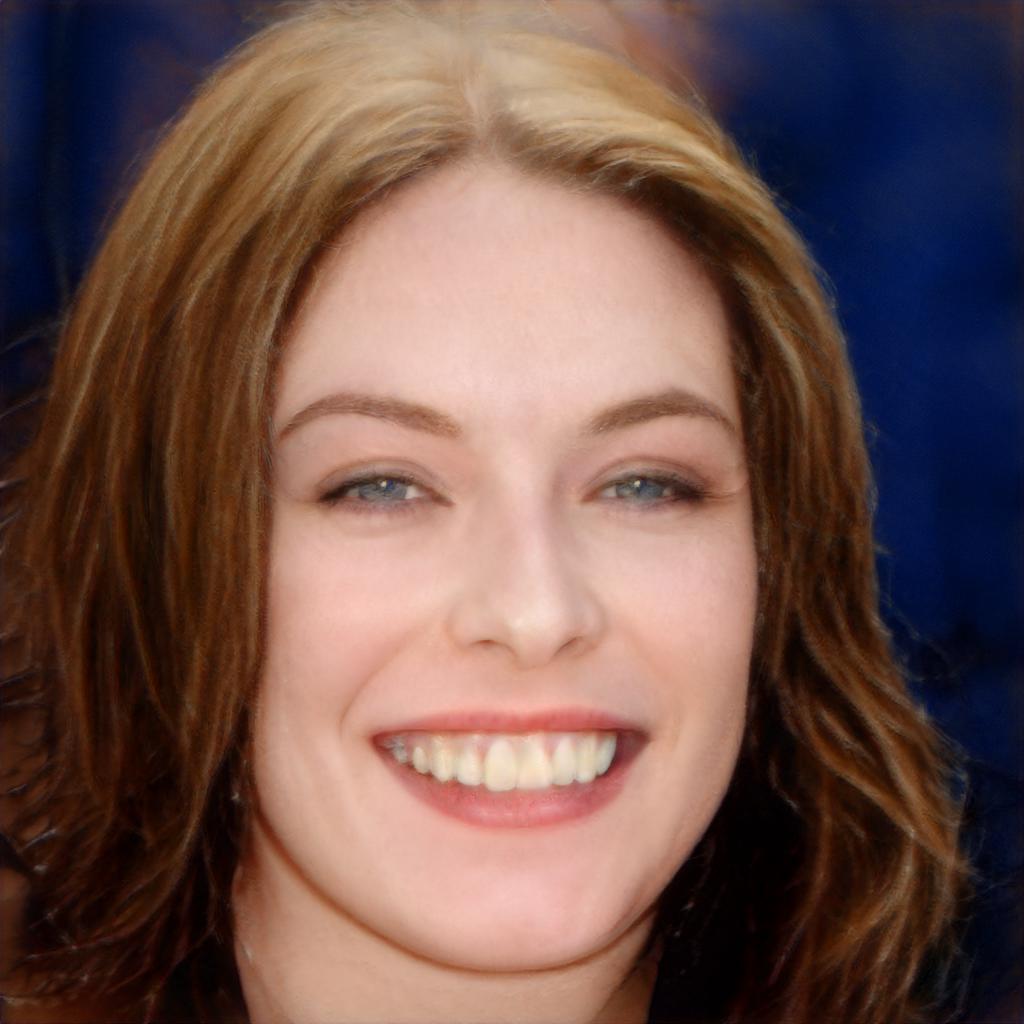}&
        \includegraphics[width=0.11\textwidth]{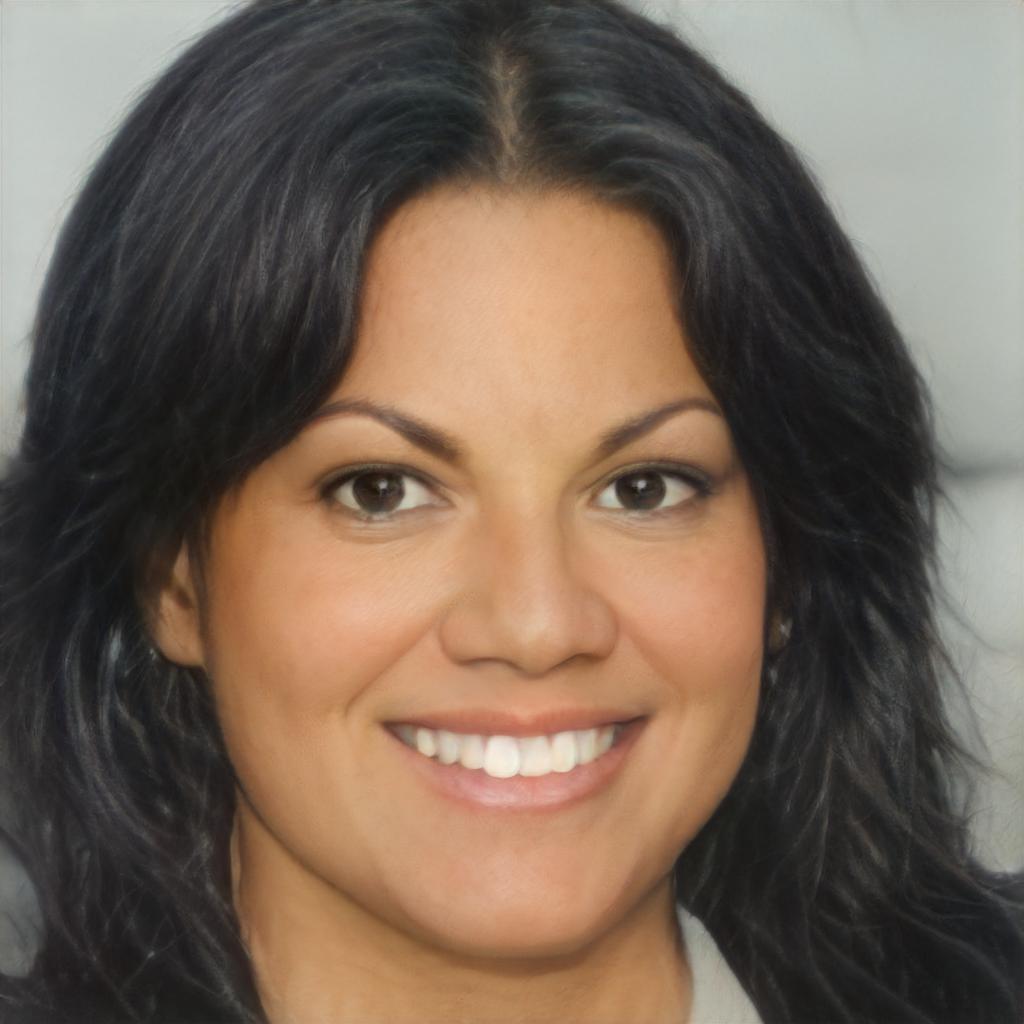}&
        \includegraphics[width=0.11\textwidth]{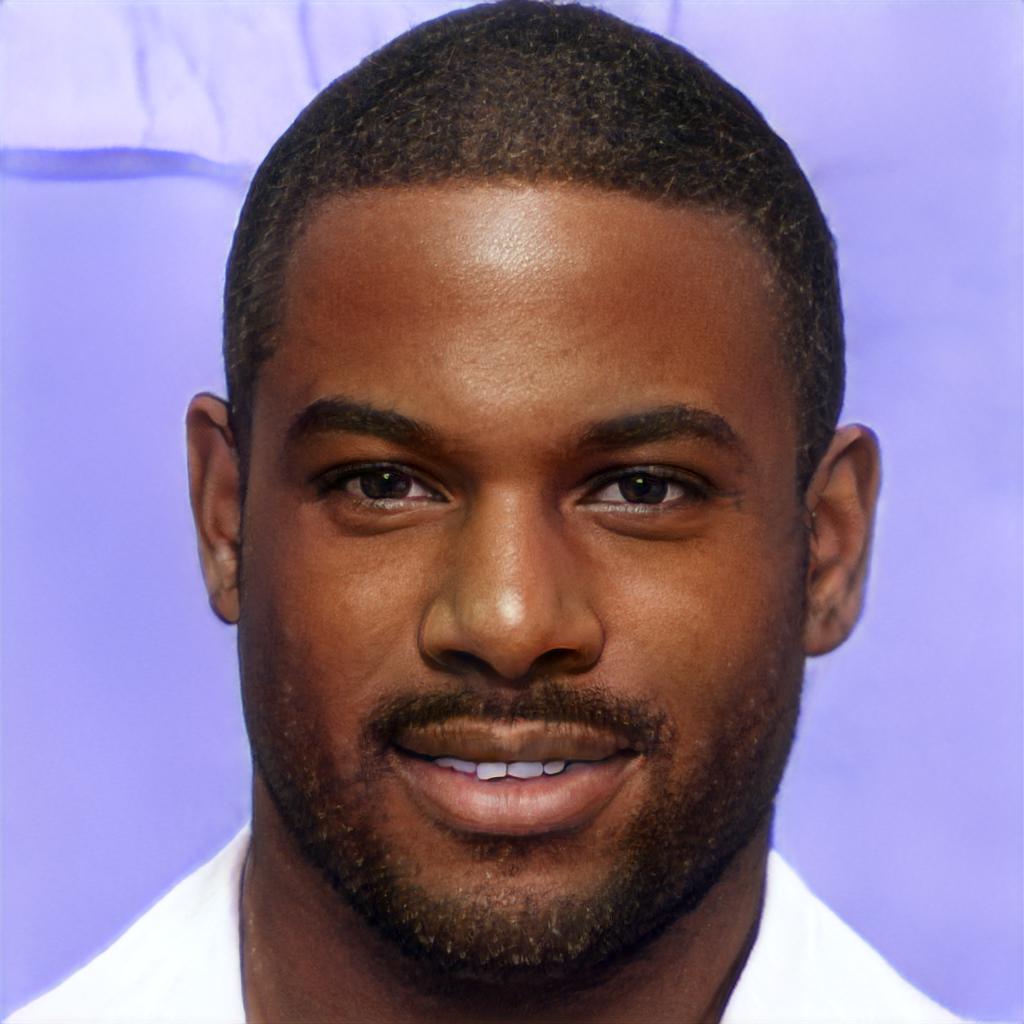}
        \tabularnewline
    \end{tabular}
    \caption{Image inpainting results using pSp and pix2pixHD~\cite{wang2018high} on the CelebA-HQ~\cite{karras2018progressive} test set.}
    \label{fig:inpainting}
\end{figure}

\begin{figure*}
    \setlength{\tabcolsep}{1pt}
    \centering
        \begin{tabular}{c c c c}
            \includegraphics[width=0.187\textwidth]{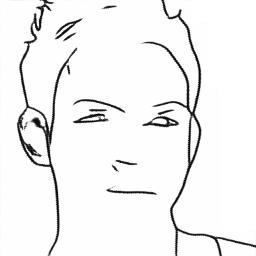}&
            \includegraphics[width=0.187\textwidth]{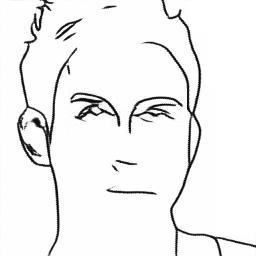}&
            \includegraphics[width=0.187\textwidth]{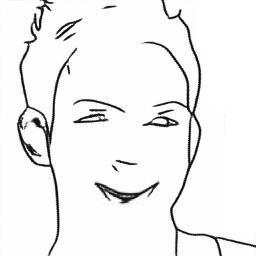}&
            \includegraphics[width=0.187\textwidth]{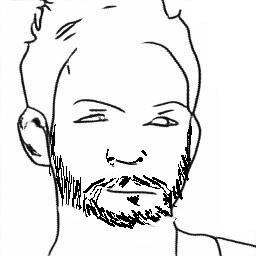}
            \tabularnewline
            \includegraphics[width=0.187\textwidth]{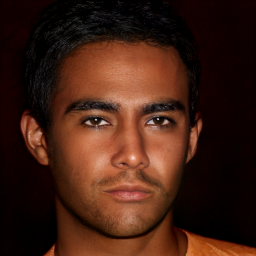}&
            \includegraphics[width=0.187\textwidth]{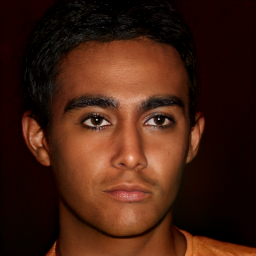}&
            \includegraphics[width=0.187\textwidth]{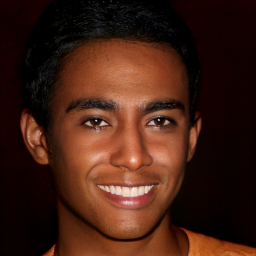}&
            \includegraphics[width=0.187\textwidth]{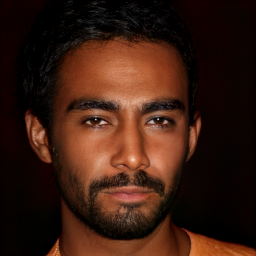}
            \tabularnewline
        \end{tabular}
    \caption{Sketch-Based Local Editing}
    \label{fig:sketch_local_edit}
\end{figure*}

\begin{figure*}
    \setlength{\tabcolsep}{1pt}
    \centering
        \begin{tabular}{c c c}
            \includegraphics[width=0.25\textwidth]{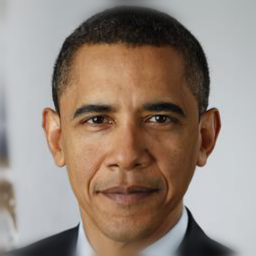}&
            \includegraphics[width=0.25\textwidth]{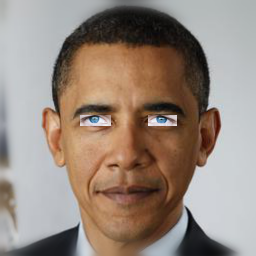}&
            \includegraphics[width=0.25\textwidth]{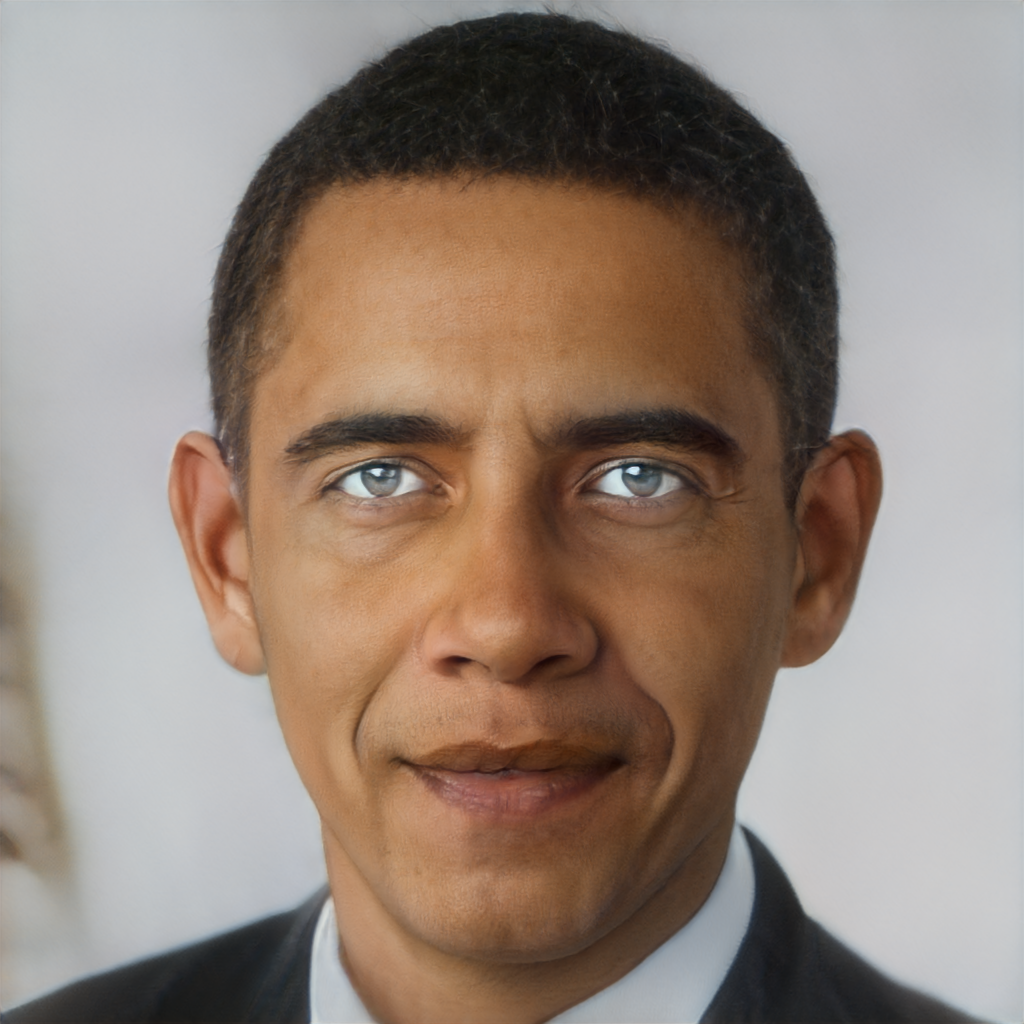}
            \tabularnewline
            \includegraphics[width=0.25\textwidth]{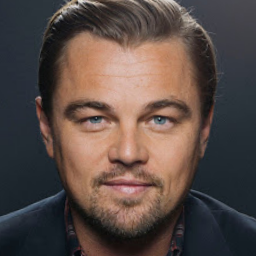}&
            \includegraphics[width=0.25\textwidth]{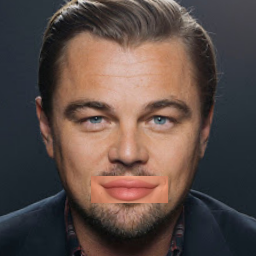}&
            \includegraphics[width=0.25\textwidth]{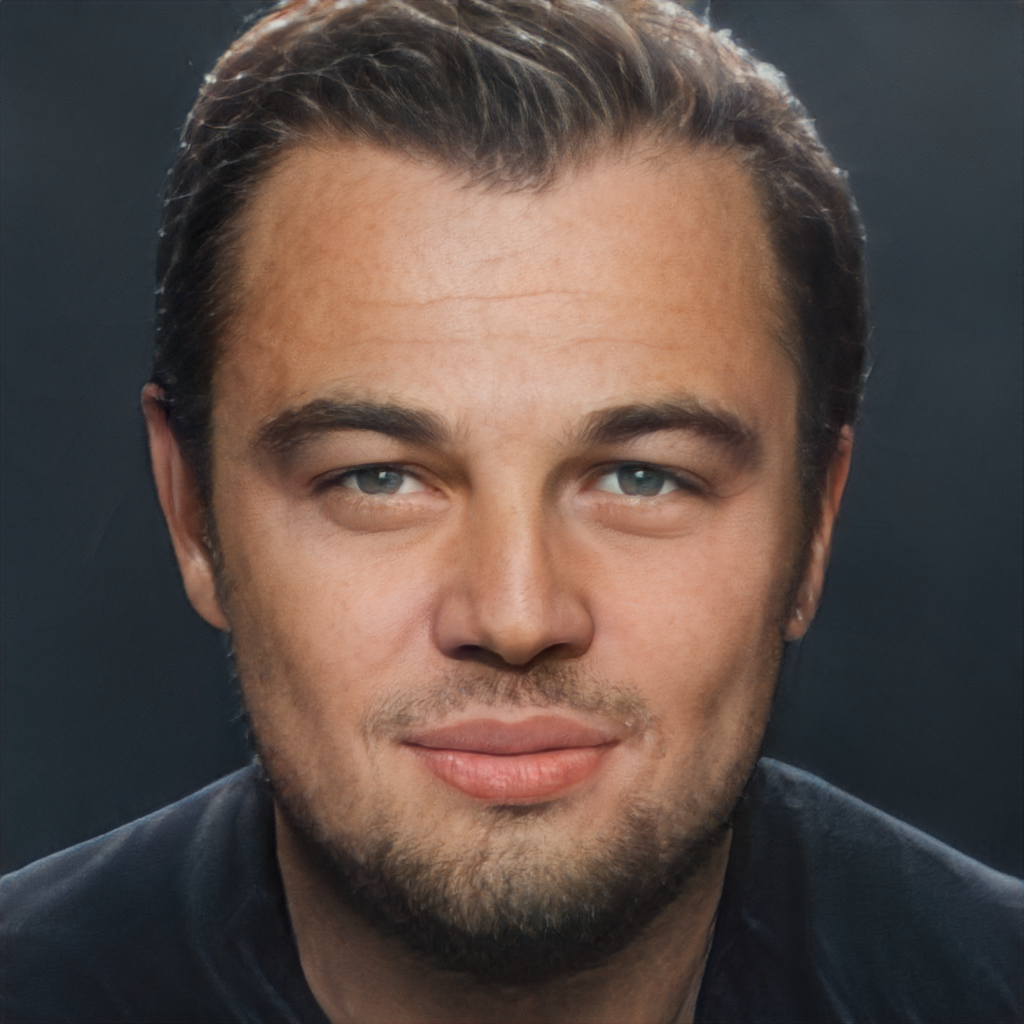}
            \tabularnewline
        \end{tabular}
    \caption{Local patch editing results using pSp on real images.}
    \label{fig:real_local_edit}
\end{figure*}

\begin{figure*}
    \setlength{\tabcolsep}{1pt}
    \centering
    \begin{tabular}{c}
        \includegraphics[width=\textwidth]{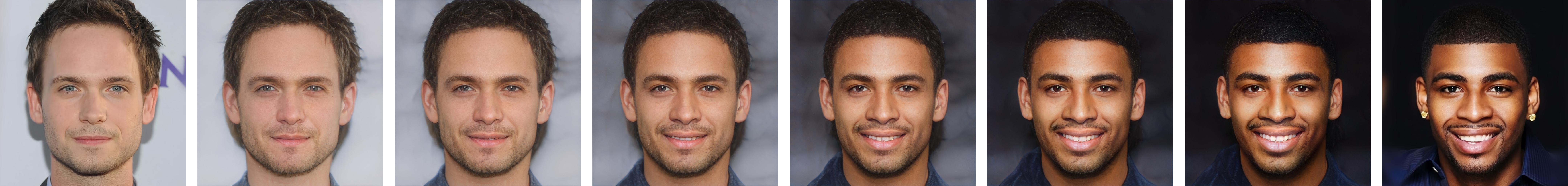}
        \tabularnewline
        \includegraphics[width=\textwidth]{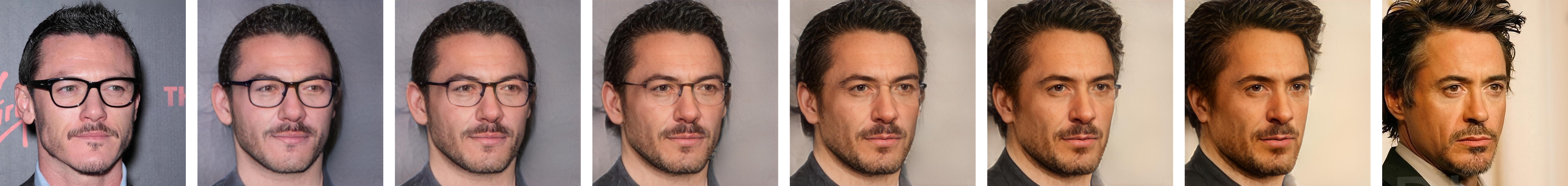}
        \tabularnewline
    \end{tabular}
    \caption{Image interpolation results using pSp on the CelebA-HQ~\cite{karras2018progressive} test set.}
    \label{fig:interpolation}
\end{figure*}

\subsection{Inpainting}~\label{app-inpainting} 
In the task of inpainting we wish to reconstruct missing or occluded regions in a given image. Due to their local nature, pix2pix~\cite{isola2017image} and other local-based translation methods, have shown success in tackling this problem as they can simply propagate non-occluded regions. 

\paragraph{Methodology and details} We train both pSp and pix2pixHD~\cite{wang2018high} in a supervised fashion, where each input image is occluded with a symmetric triangular mask. 

\paragraph{Results} Figure~\ref{fig:inpainting} presents results for both our method and pix2pixHD. As shown, due to the lack of information in the occluded regions, pix2pixHD is unable to accurately reconstruct the original image and incurs many artifacts. In contrast, since pSp is trained to encode images into realistic face latents, it is able to accurately reconstruct the occluded region, resulting in high-quality outputs with no artifacts. 

\subsection{Local Editing} 
Our framework allows for a simple approach to local image editing using a trained pSp encoder where altering specific attributes of an input sketch (e.g. eyes, smile) or segmentation map (e.g. hair) results in local edits of the generated images.
We can further extend this and perform local patch editing on real face images. As shown in Figure~\ref{fig:real_local_edit}, pSp is able to seamlessly merge the desired patch into the original image. 

\subsection{Face Interpolation} 
Given two real images one can obtain their respective latent codes $w_1,w_2\in \mathcal{W+}$ by feeding the images through our encoder. We can then naturally interpolate between the two images by computing their intermediate latent code $w'=\alpha w_1 + (1 - \alpha)w_2$ for $0\leq \alpha \leq 1$ and generate the corresponding image using the new code $w'$.

\begin{figure*}
    
    \centering
    \begin{subfigure}{0.75\textwidth}
    \setlength{\tabcolsep}{1pt}
    \centering
        \begin{tabular}{c c}
            \includegraphics[width=0.5\textwidth]{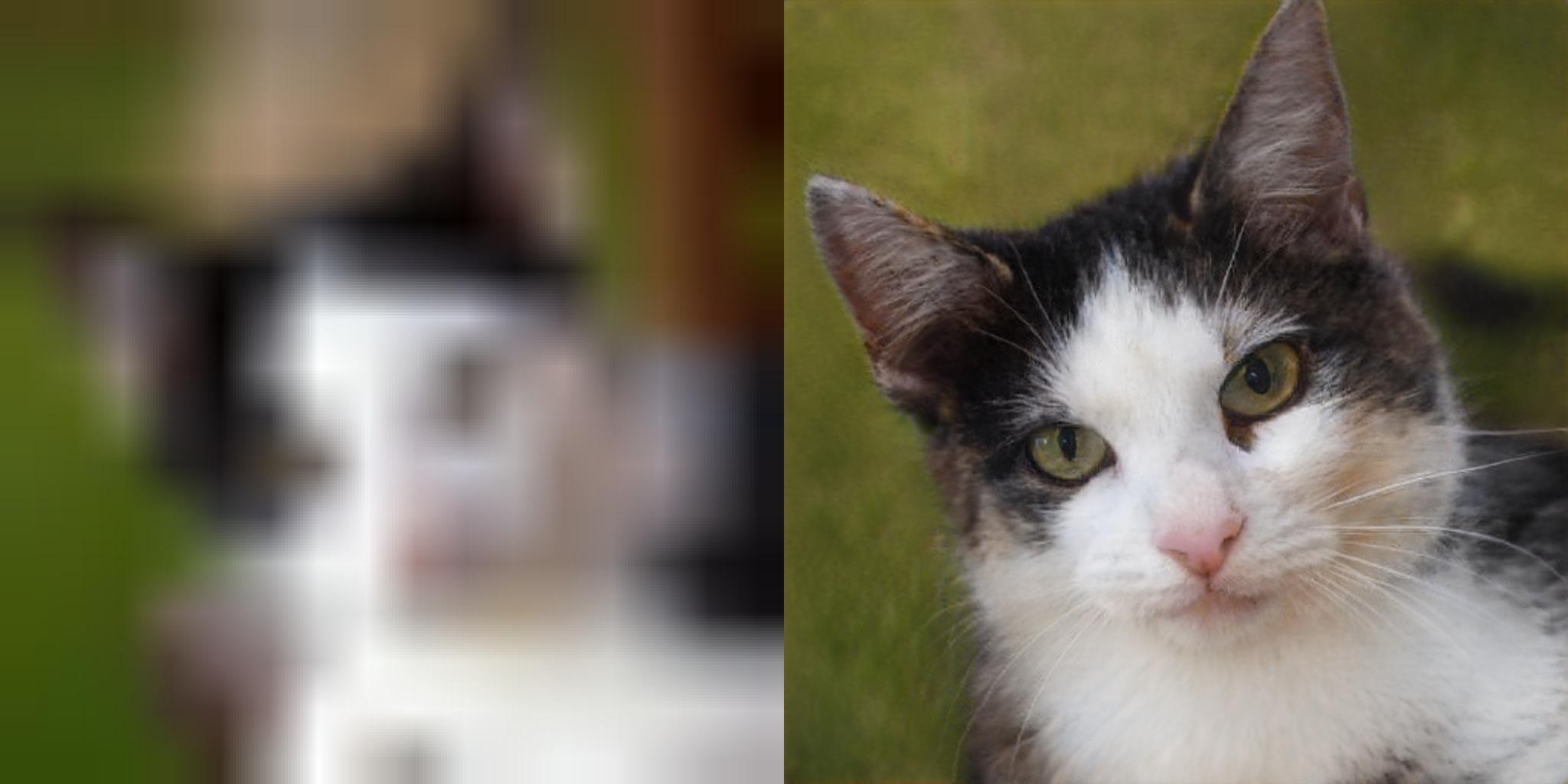}&
            \includegraphics[width=0.5\textwidth]{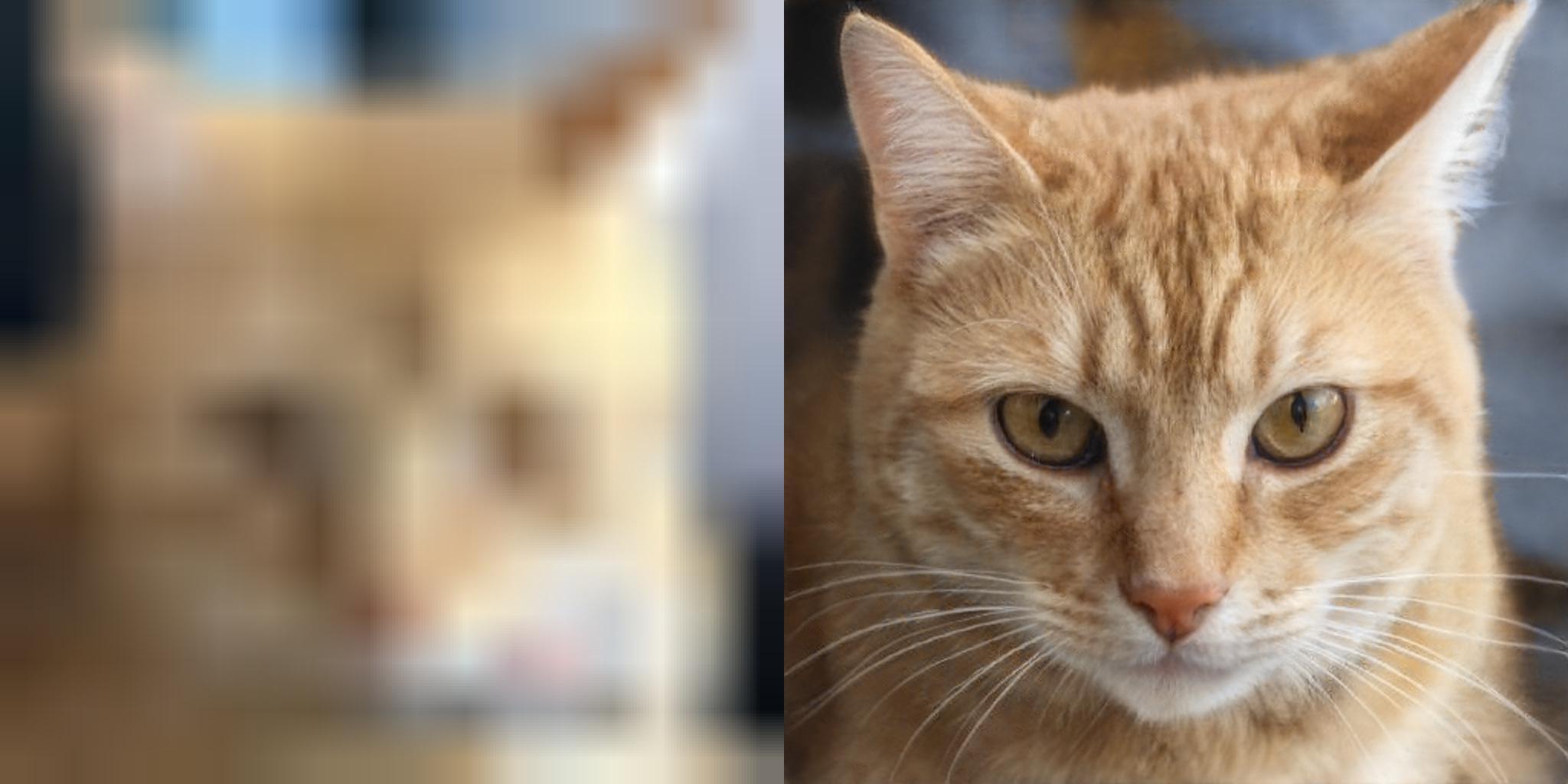}
            \tabularnewline
            \includegraphics[width=0.5\textwidth]{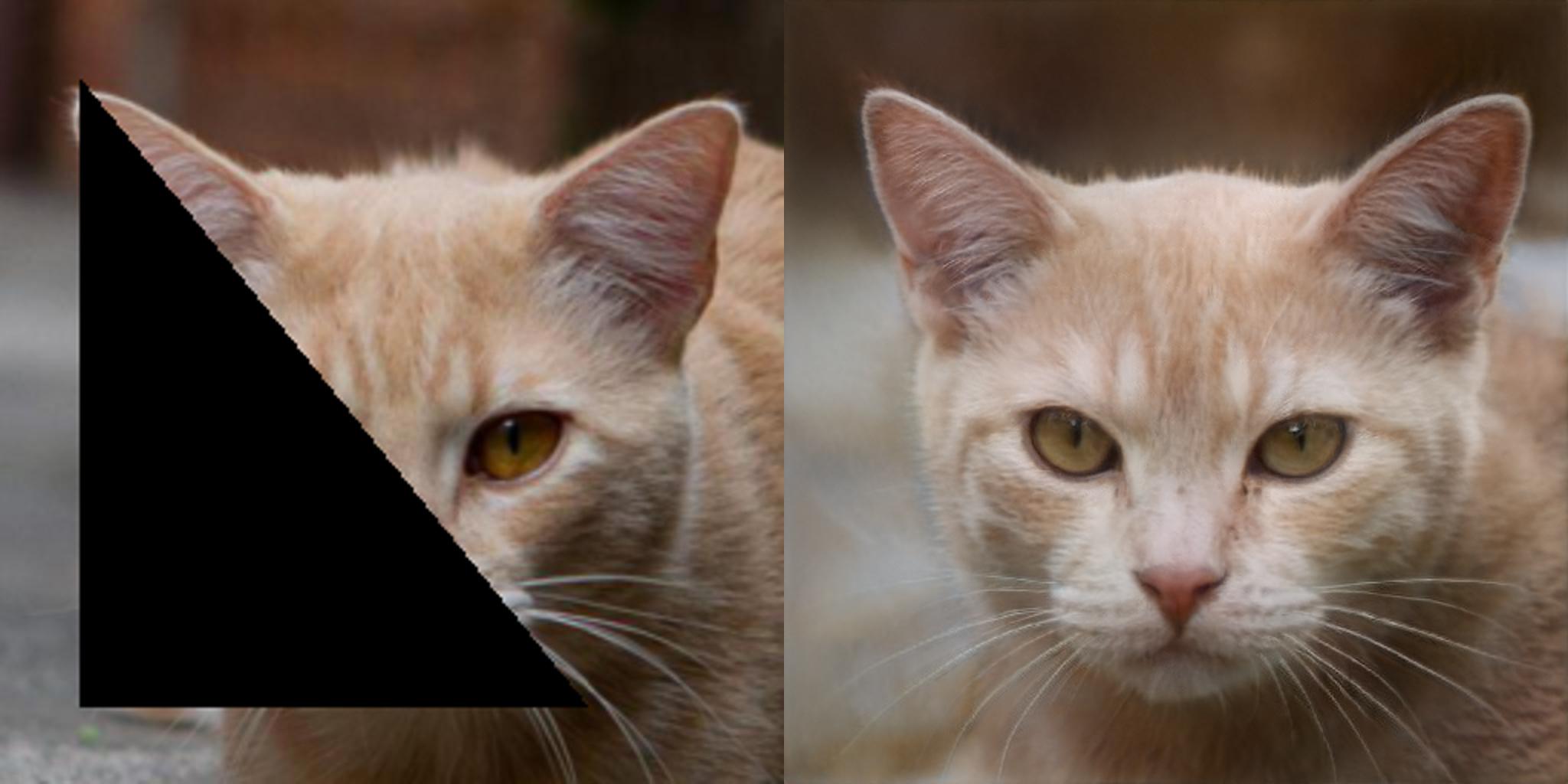}&
            \includegraphics[width=0.5\textwidth]{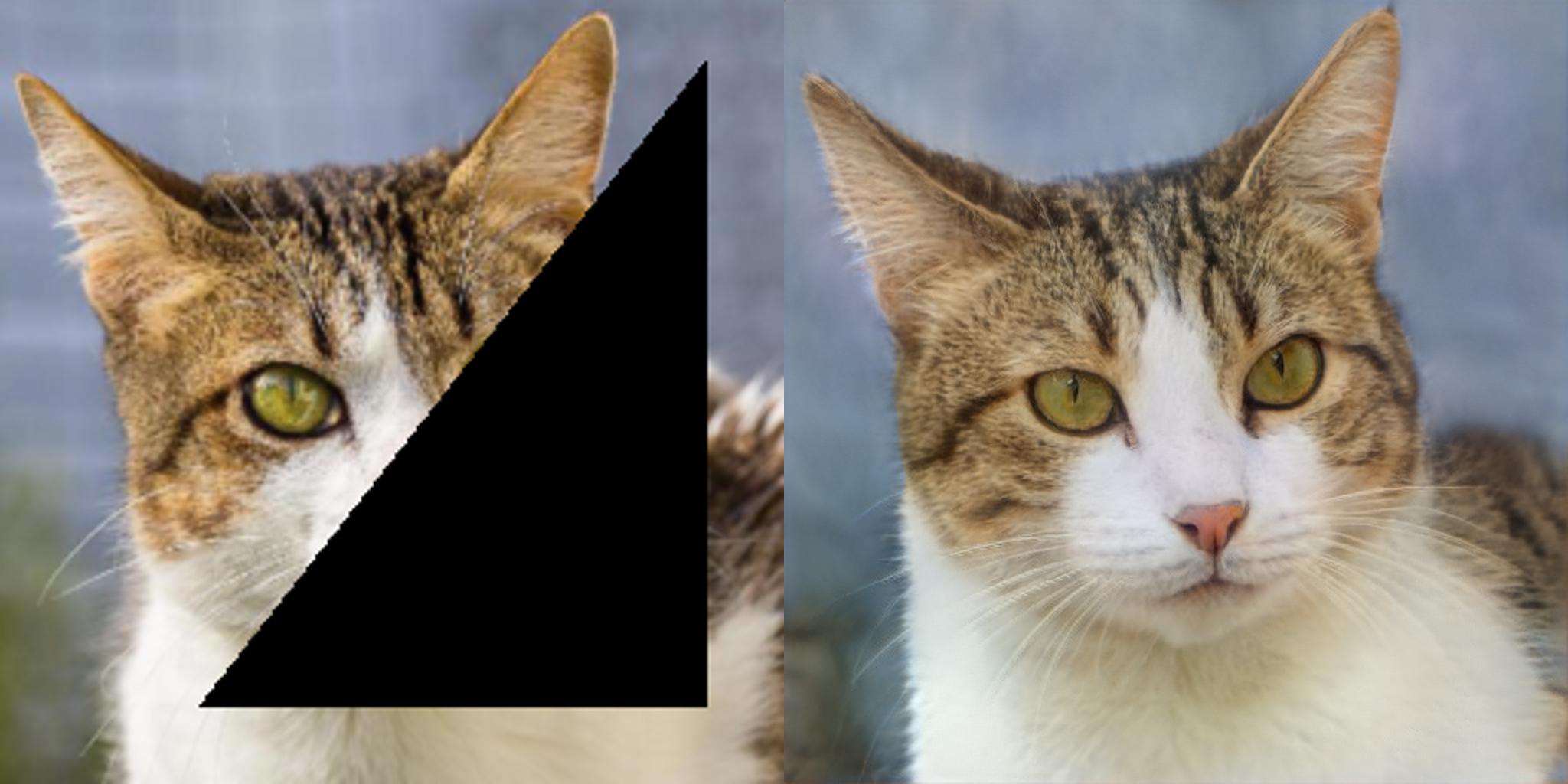}
            \tabularnewline
        \end{tabular}
        \begin{tabular}{c}
            \includegraphics[width=\textwidth]{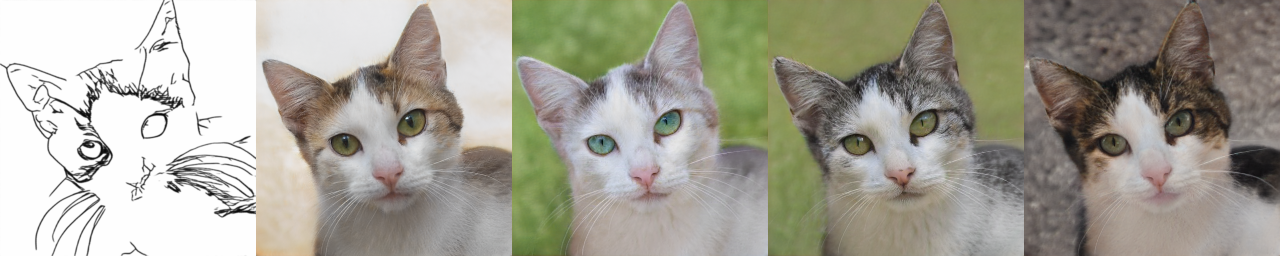}
        \end{tabular}
    \label{fig:afhq_cat}
\end{subfigure}
    \begin{subfigure}{0.75\textwidth}
    \setlength{\tabcolsep}{1pt}
    \centering
        \begin{tabular}{c c}
            \includegraphics[width=0.5\textwidth]{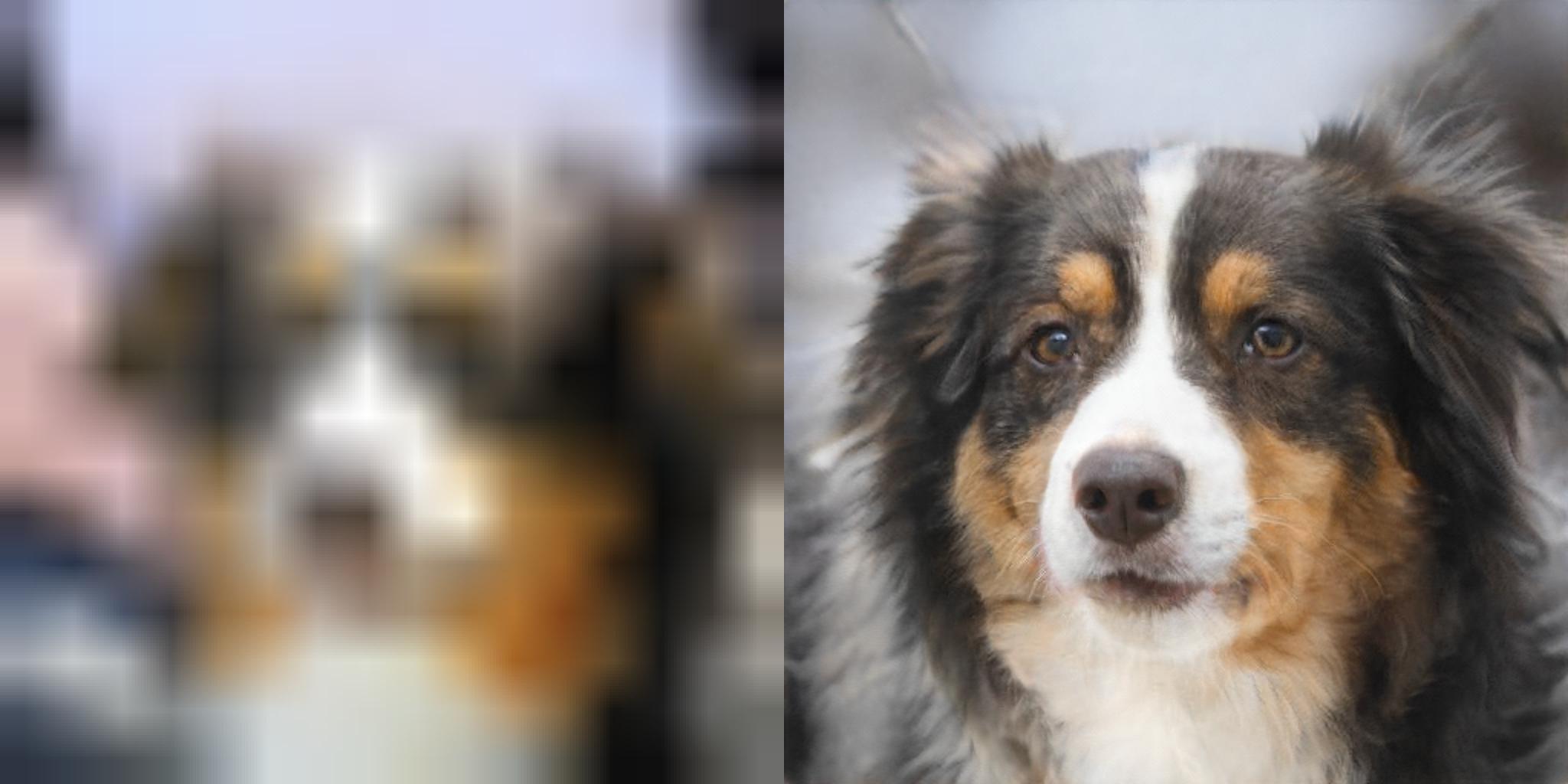}&
            \includegraphics[width=0.5\textwidth]{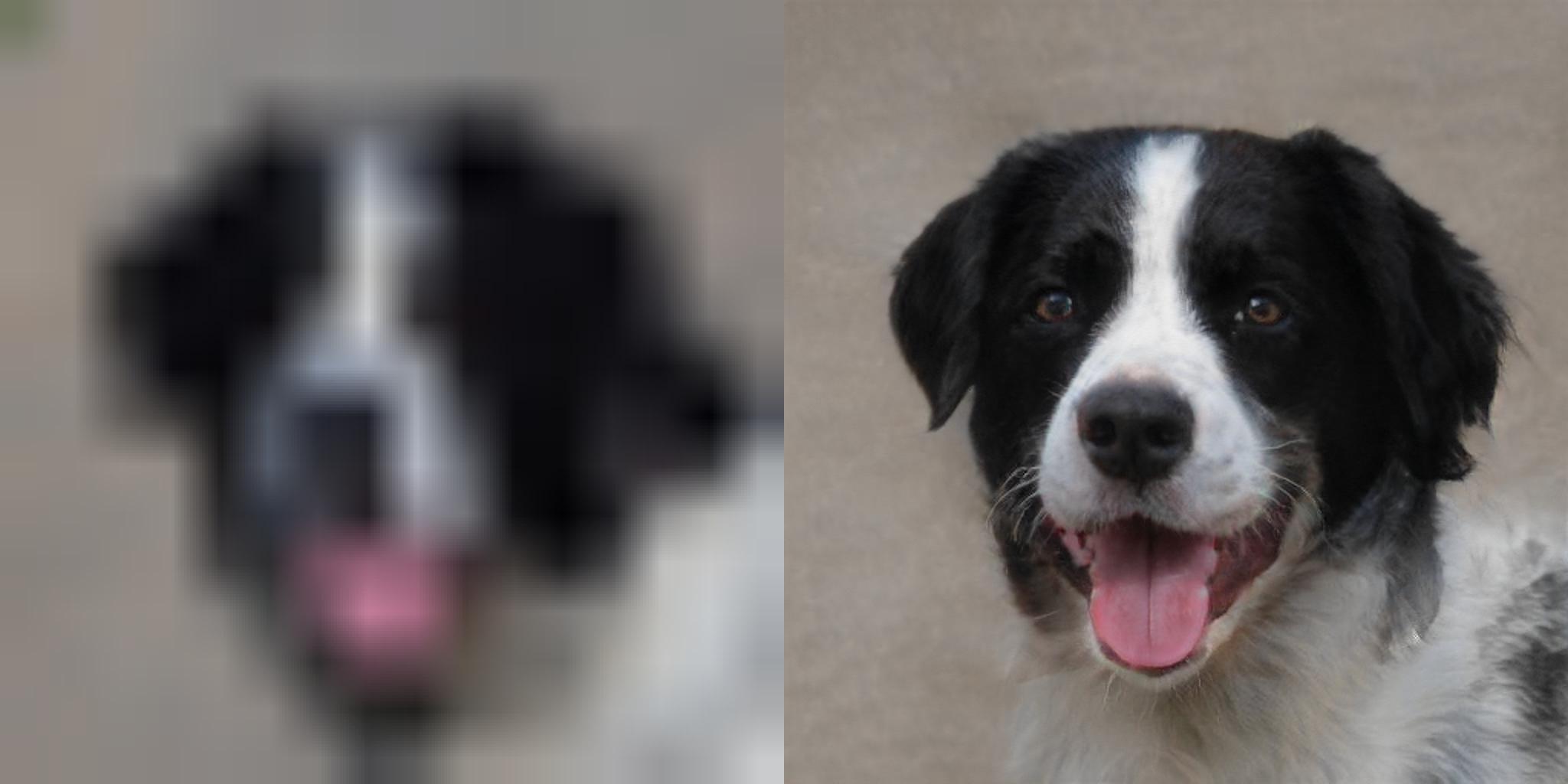}
            \tabularnewline
            \includegraphics[width=0.5\textwidth]{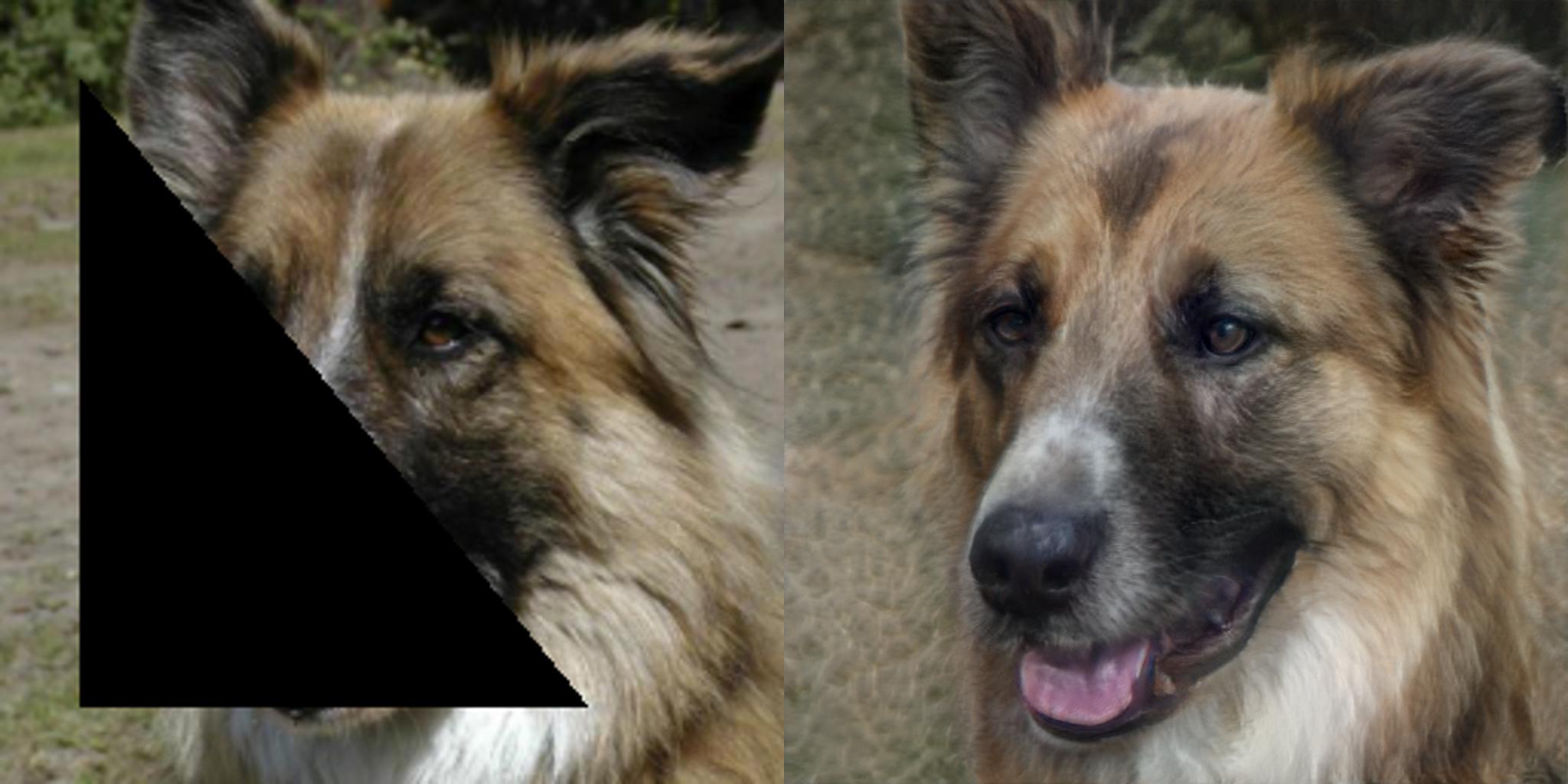}&
            \includegraphics[width=0.5\textwidth]{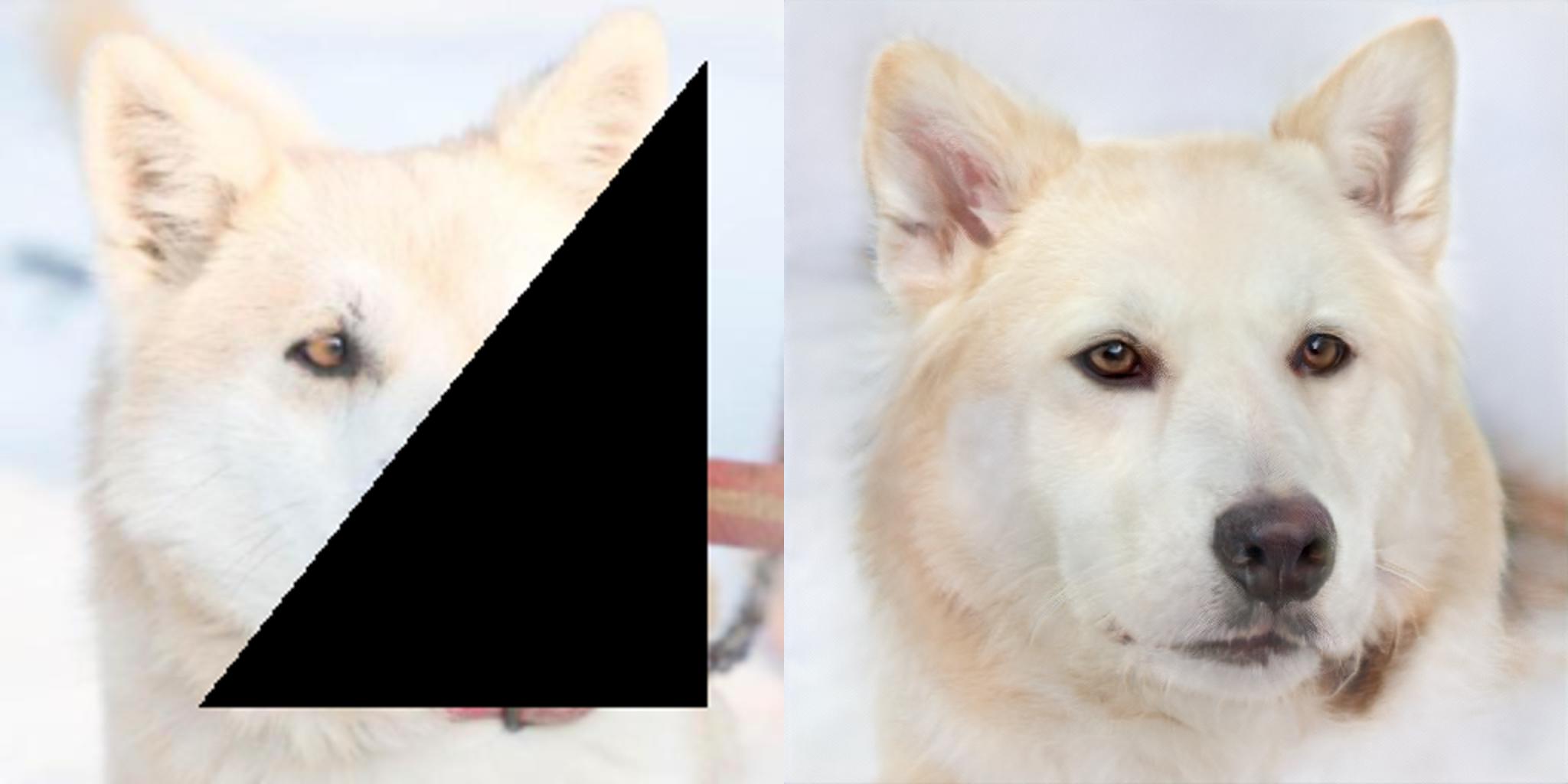}
            \tabularnewline
        \end{tabular}
        \begin{tabular}{c}
            \includegraphics[width=\textwidth]{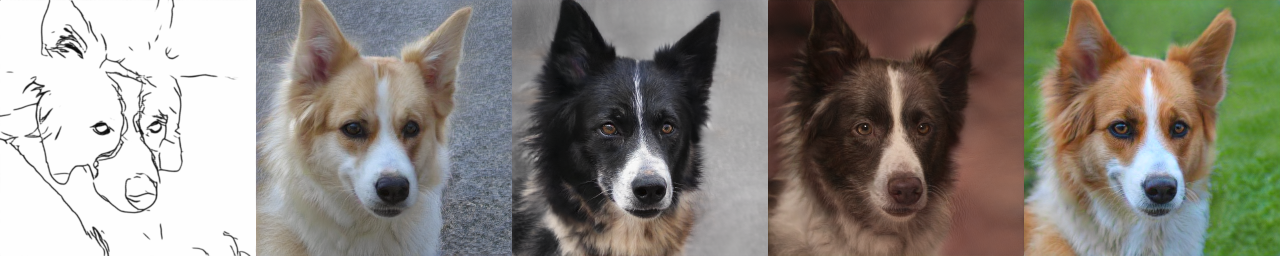}
            \tabularnewline
        \end{tabular}
    \label{fig:afhq_dog}
\end{subfigure}
    \caption{Results of pSp on the AFHQ Cat and Dog datasets~\cite{choi2020stargan} on super resolution, inpainting, and image generation from sketches.}     \label{fig:additional_domains}
\end{figure*}

\begin{figure*}[!htb]
\setlength{\tabcolsep}{1pt}
\centering
    \begin{tabular}{c c c c c}
        \includegraphics[width=0.175\textwidth]{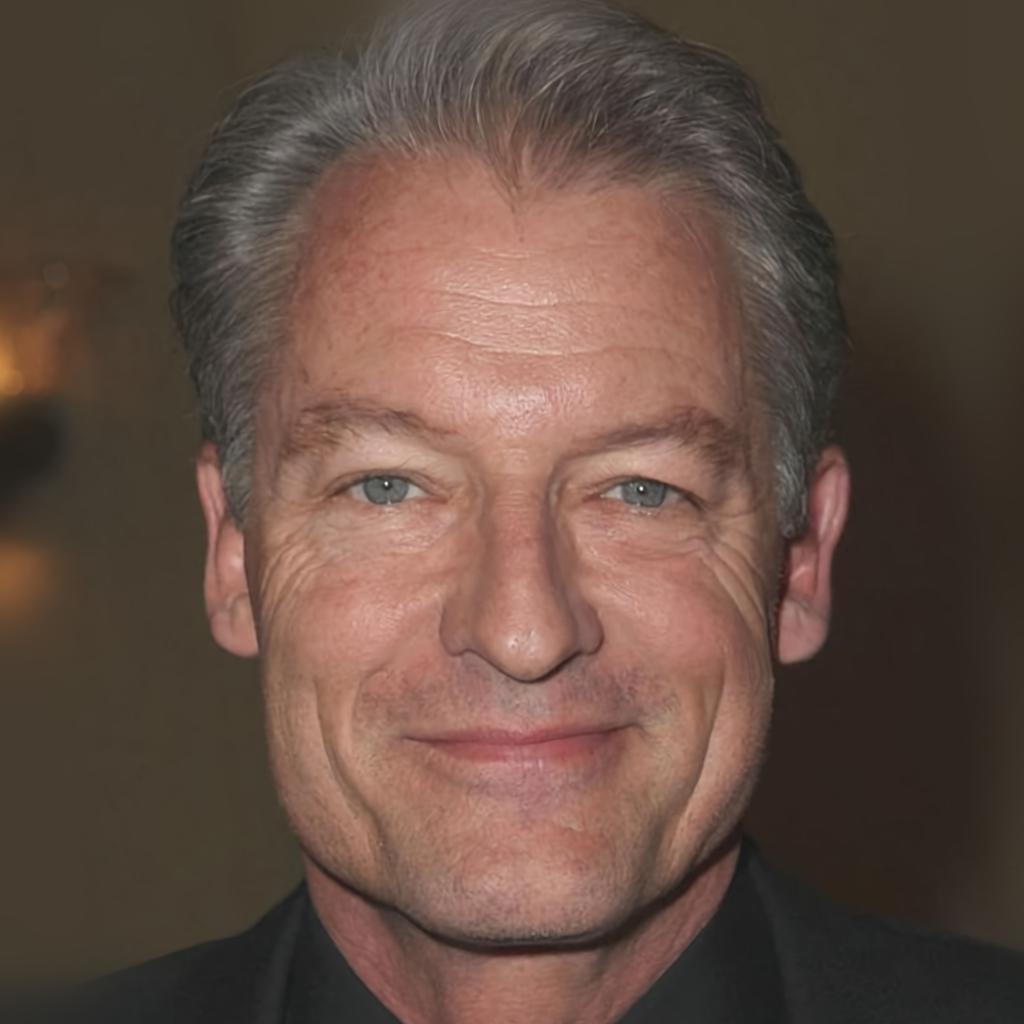}&
        \includegraphics[width=0.175\textwidth]{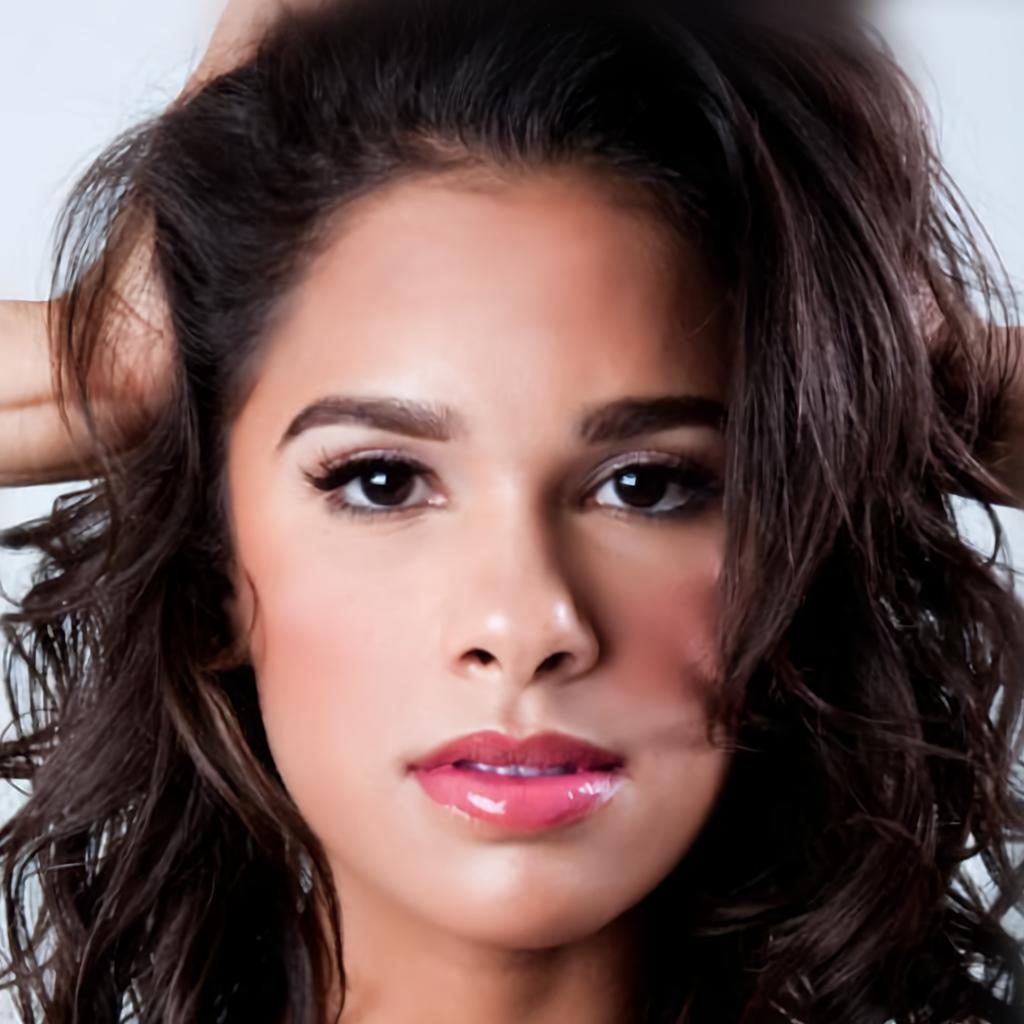}&
        \includegraphics[width=0.175\textwidth]{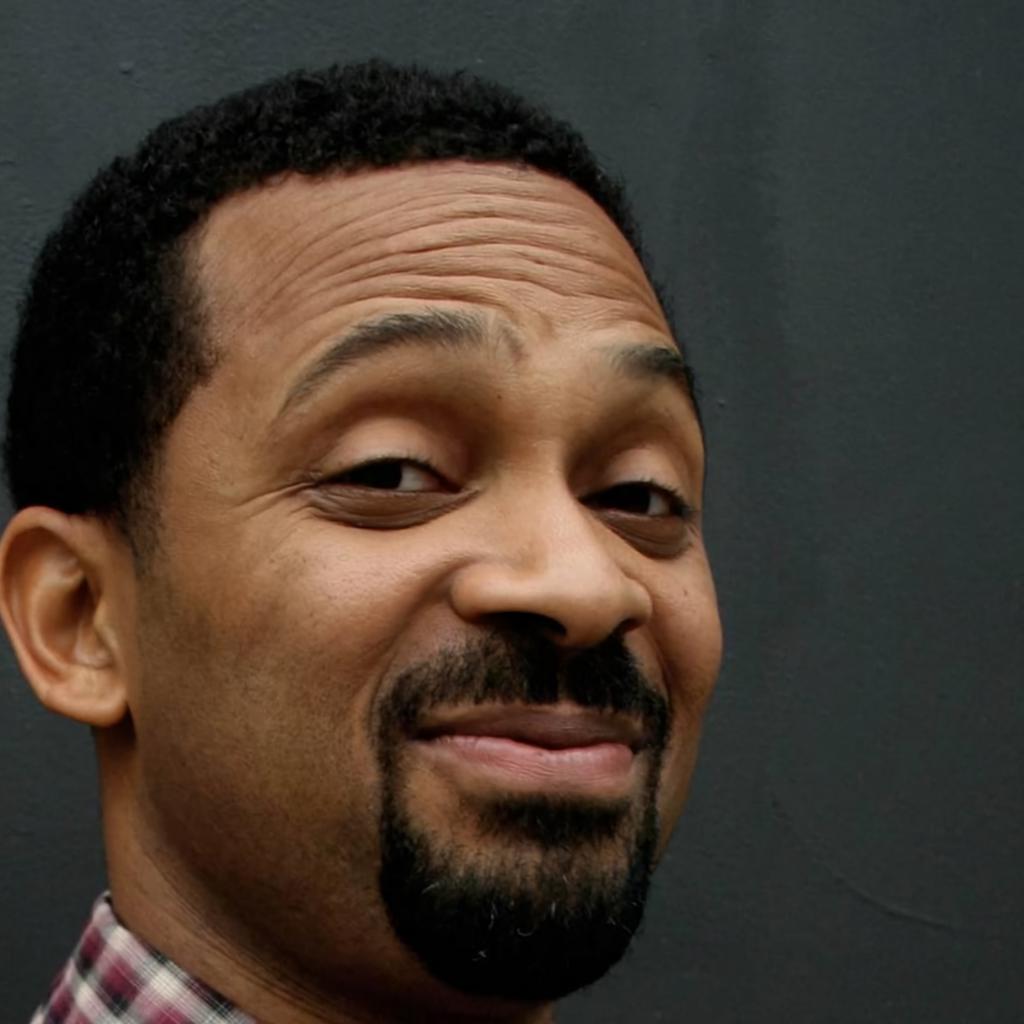}&
        \includegraphics[width=0.175\textwidth]{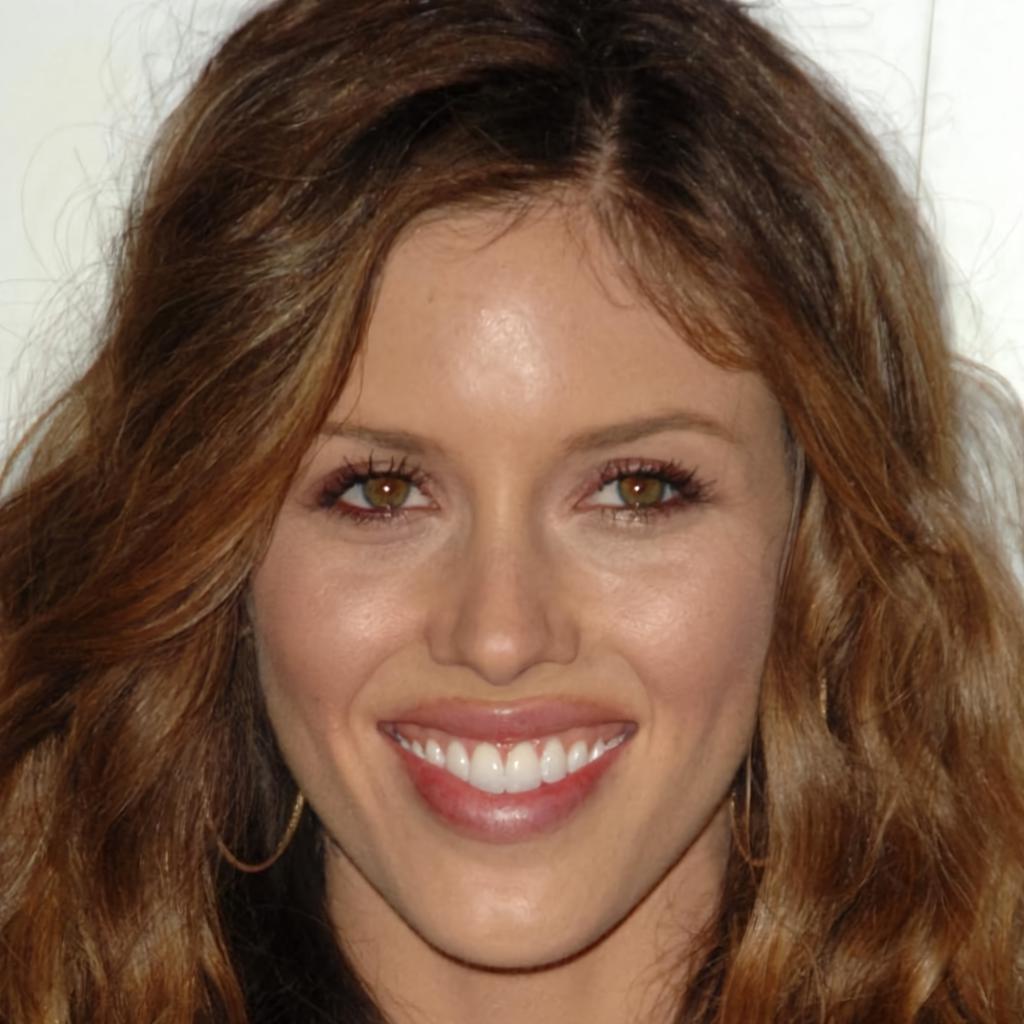}&
        \includegraphics[width=0.175\textwidth]{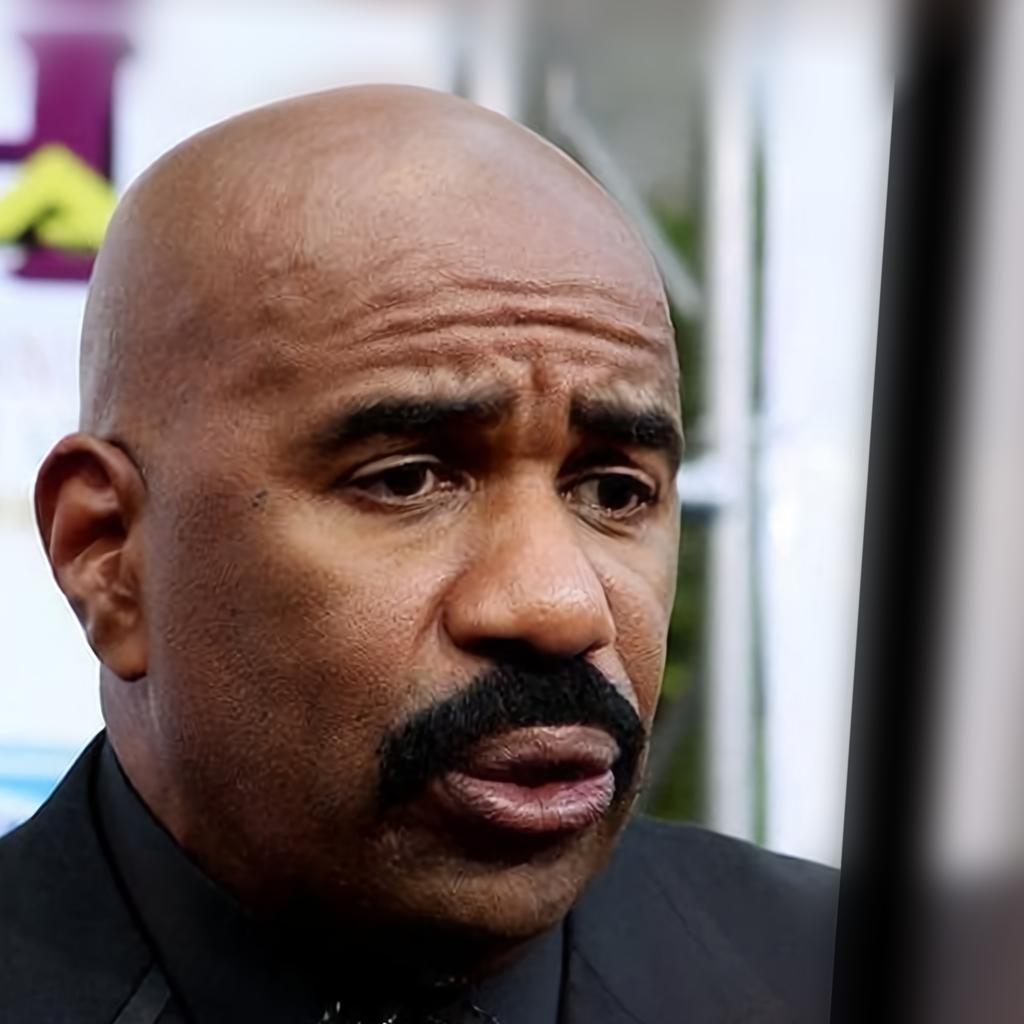}
        \tabularnewline
        \includegraphics[width=0.175\textwidth]{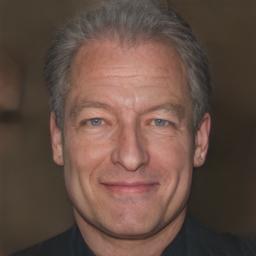}&
        \includegraphics[width=0.175\textwidth]{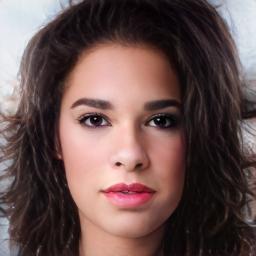}&
        \includegraphics[width=0.175\textwidth]{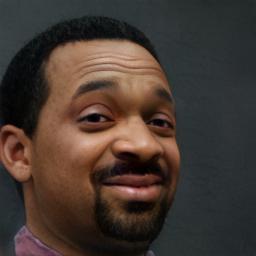}&
        \includegraphics[width=0.175\textwidth]{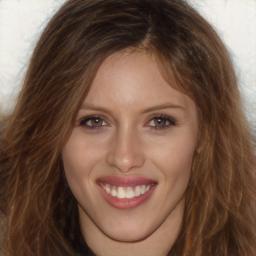}&
        \includegraphics[width=0.175\textwidth]{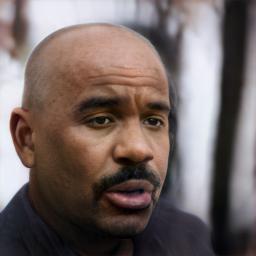}
        \tabularnewline
        \tabularnewline
        \includegraphics[width=0.175\textwidth]{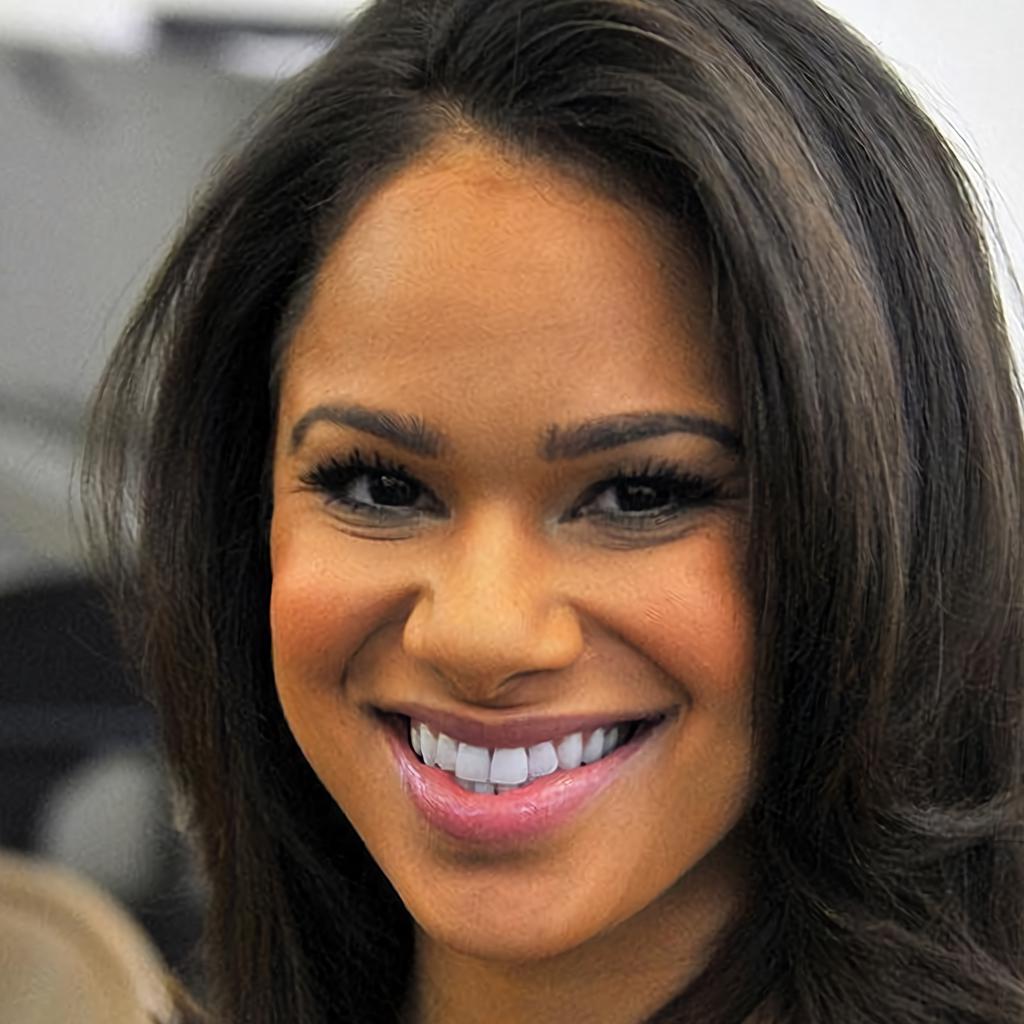}&
        \includegraphics[width=0.175\textwidth]{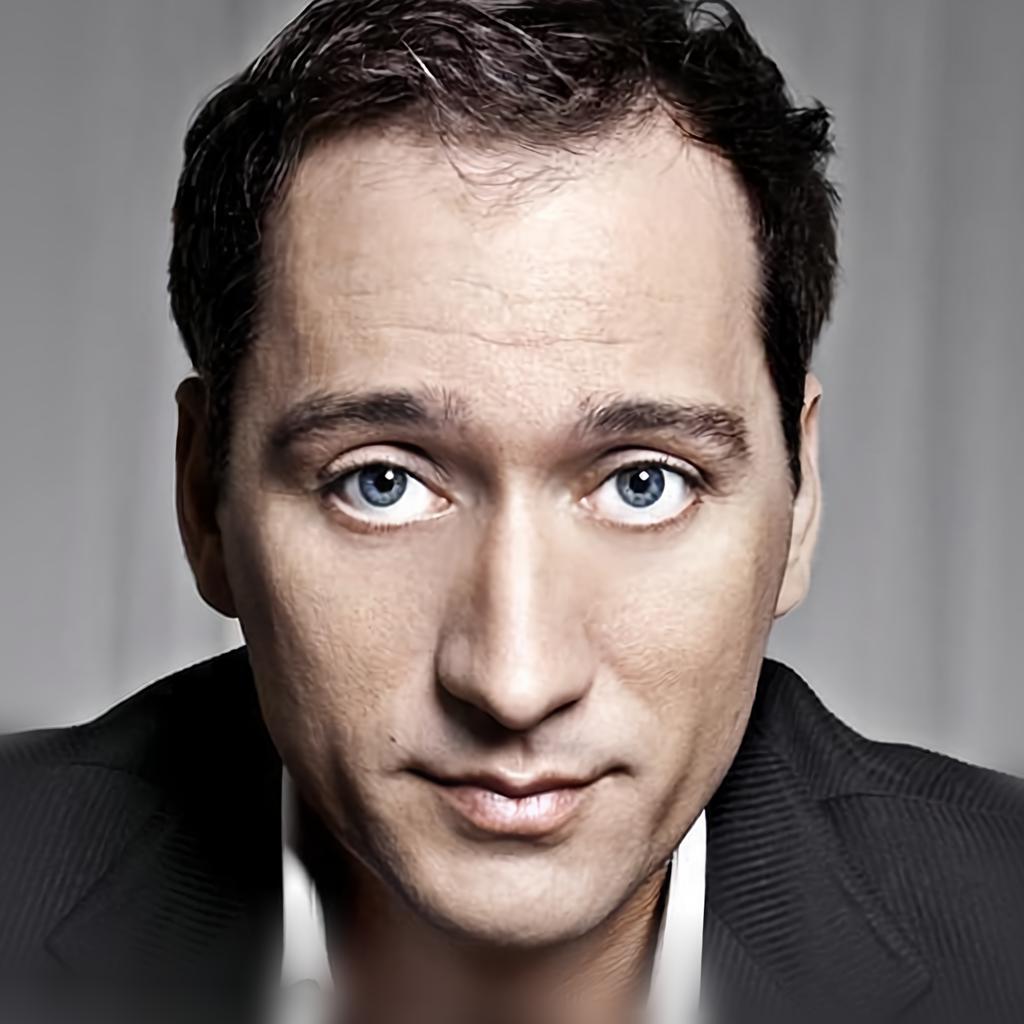}&
        \includegraphics[width=0.175\textwidth]{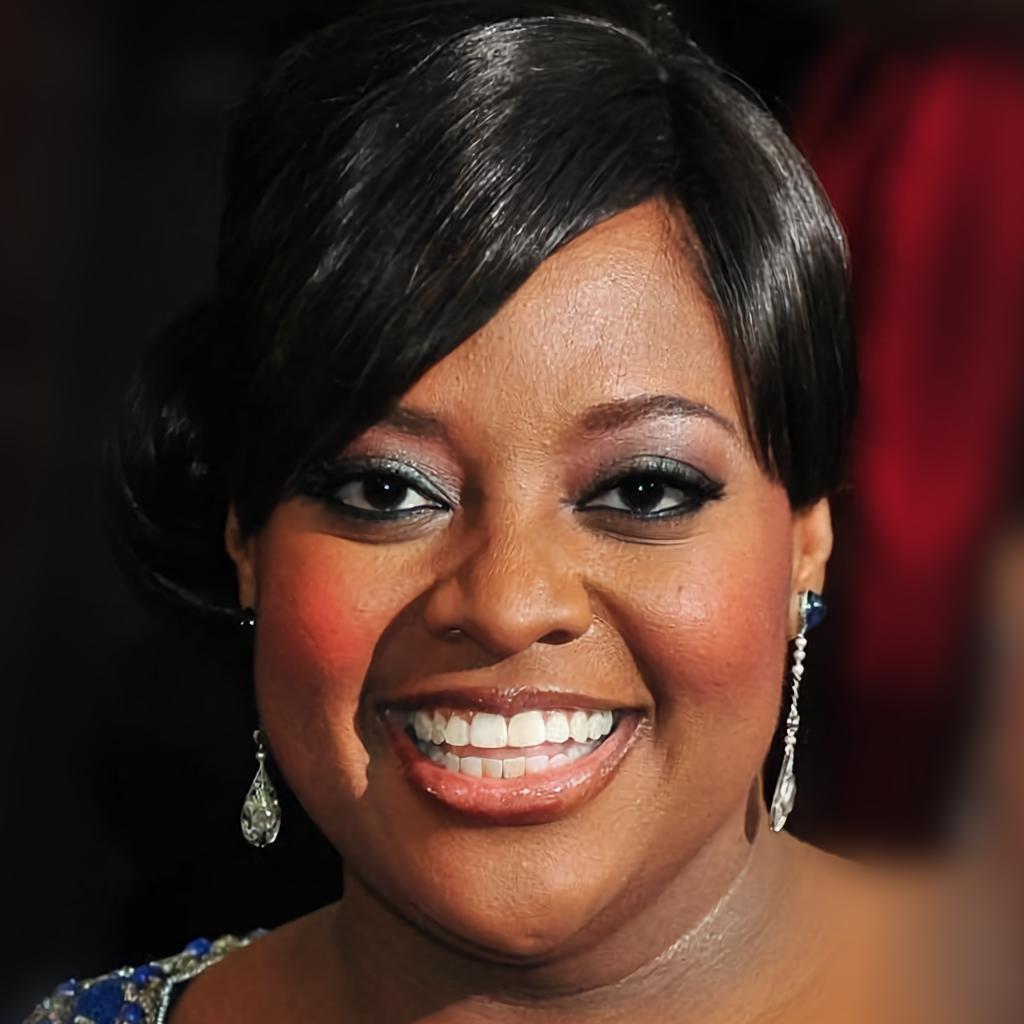}&
        \includegraphics[width=0.175\textwidth]{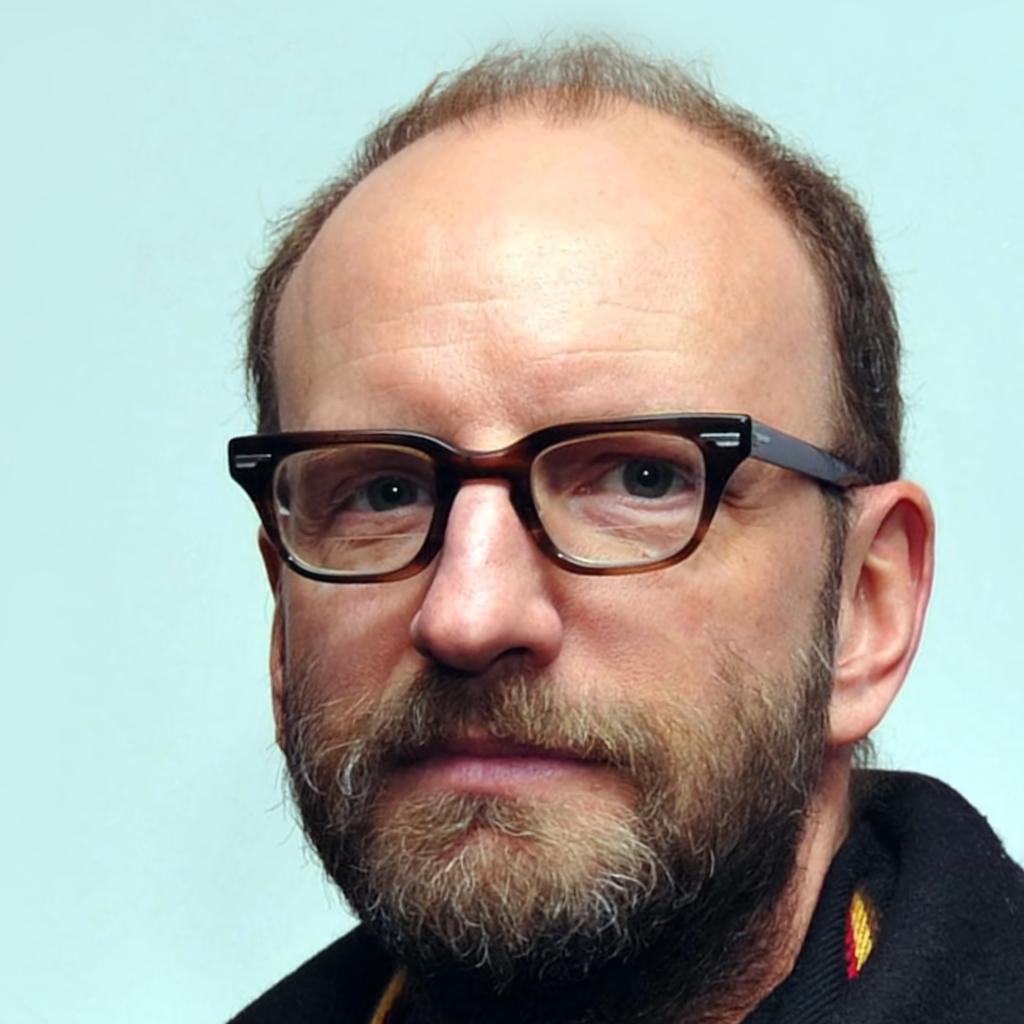}&
        \includegraphics[width=0.175\textwidth]{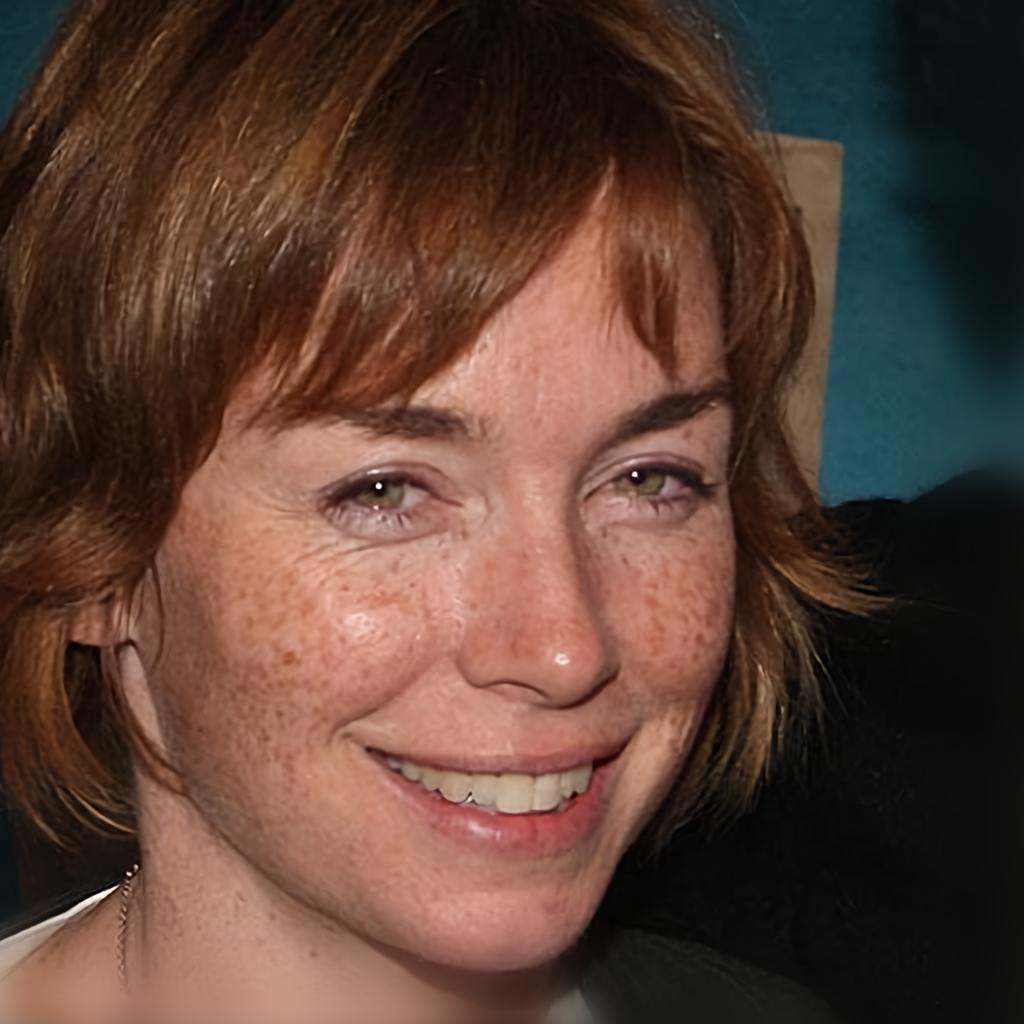}
        \tabularnewline
        \includegraphics[width=0.175\textwidth]{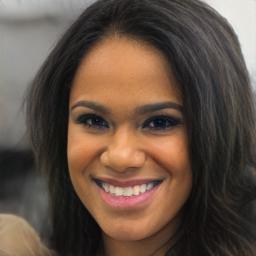}&
        \includegraphics[width=0.175\textwidth]{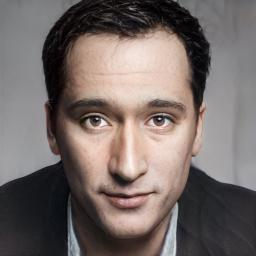}&
        \includegraphics[width=0.175\textwidth]{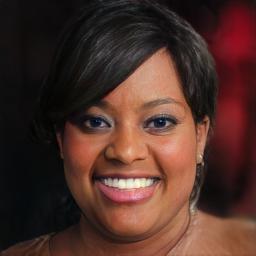}&
        \includegraphics[width=0.175\textwidth]{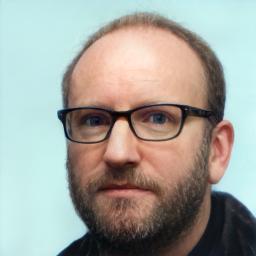}&
        \includegraphics[width=0.175\textwidth]{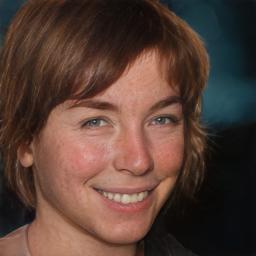}
        \tabularnewline
        \tabularnewline
        \includegraphics[width=0.175\textwidth]{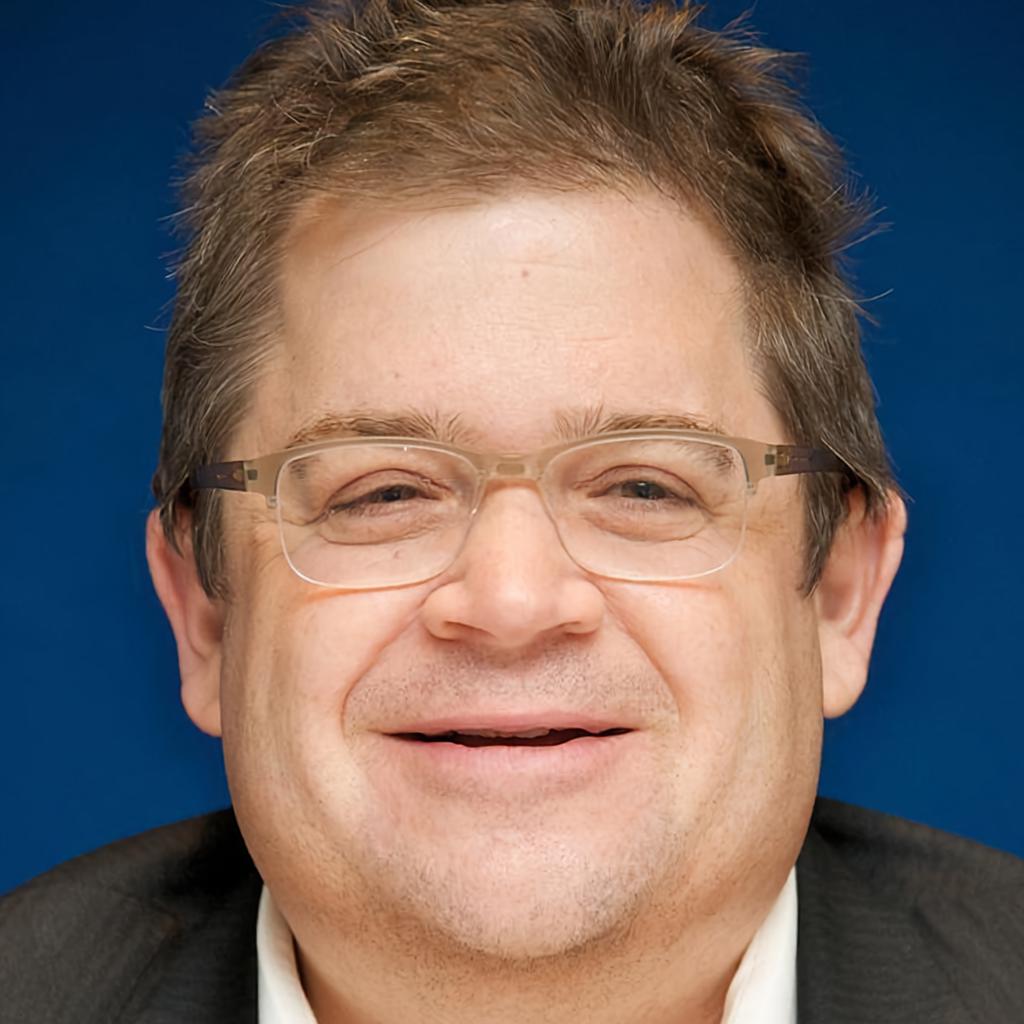}&
        \includegraphics[width=0.175\textwidth]{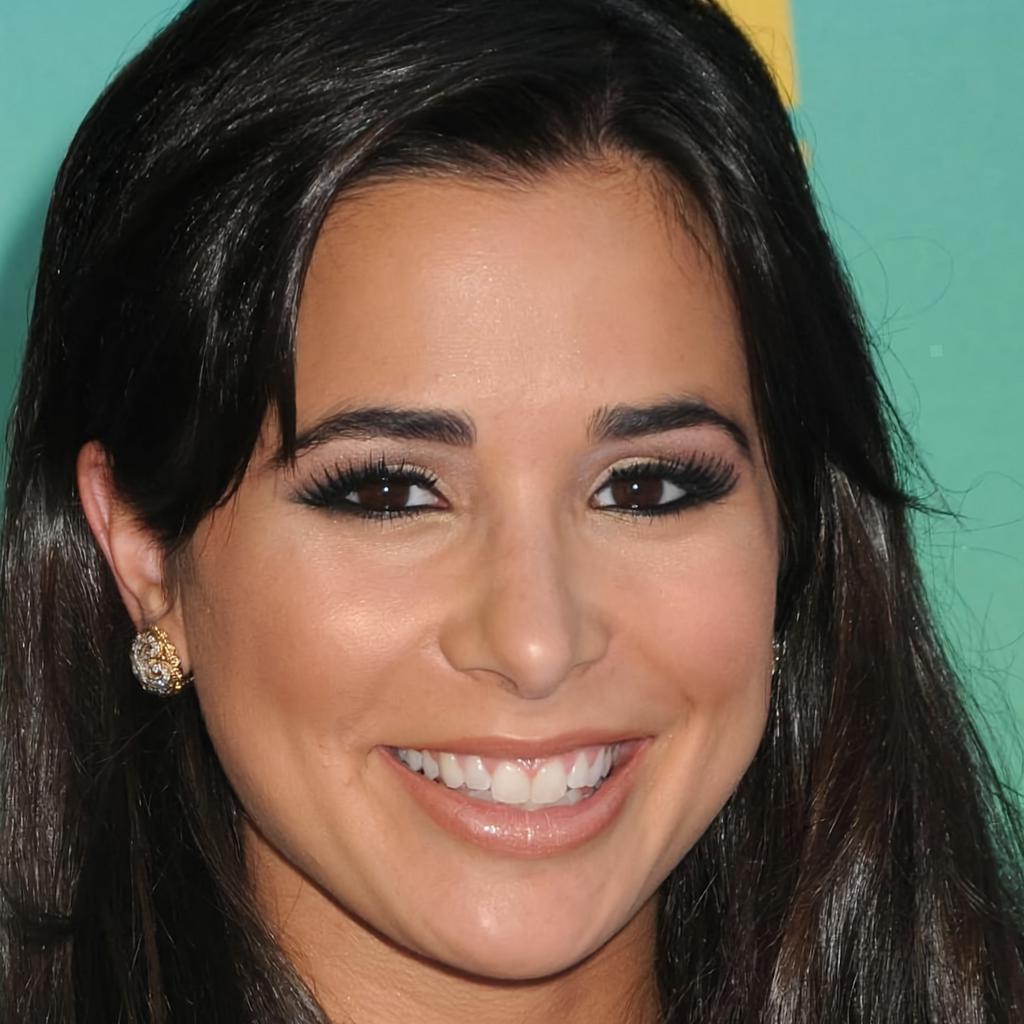}&
        \includegraphics[width=0.175\textwidth]{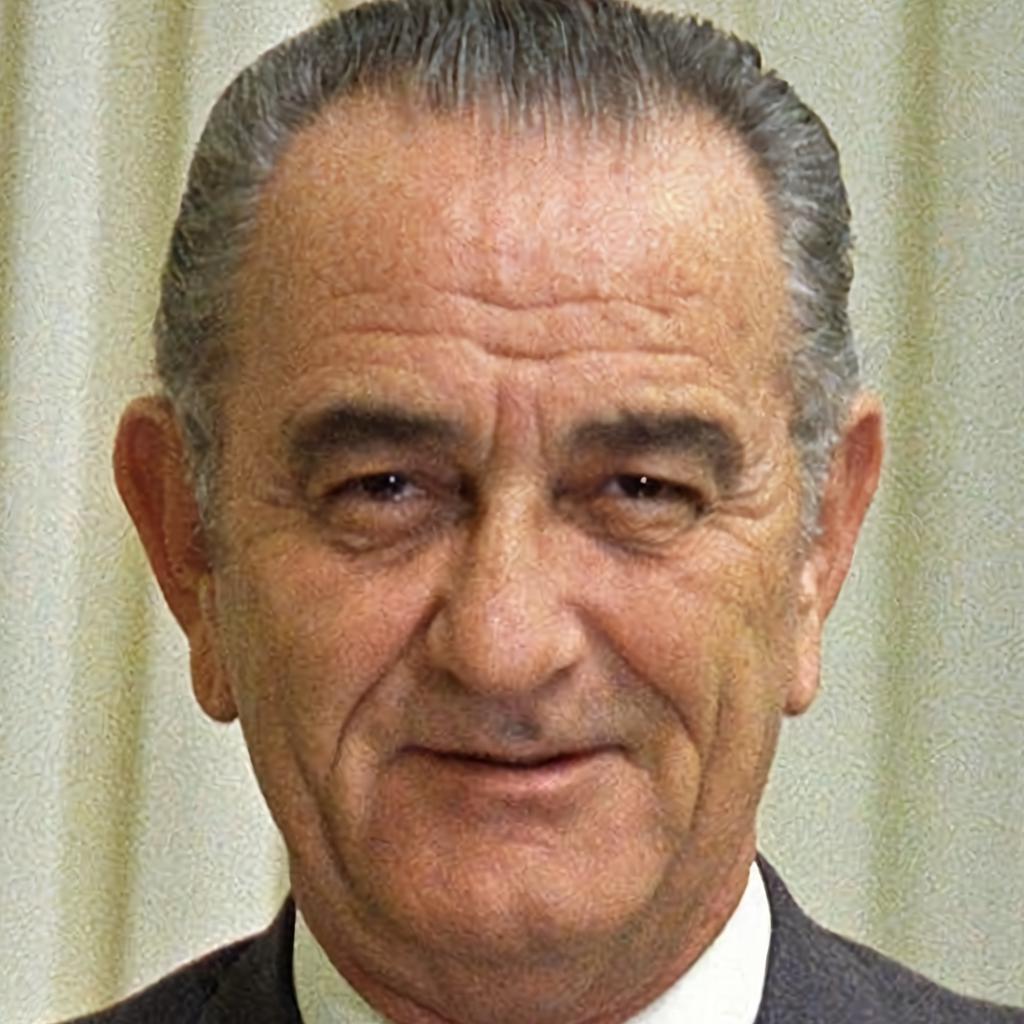}&
        \includegraphics[width=0.175\textwidth]{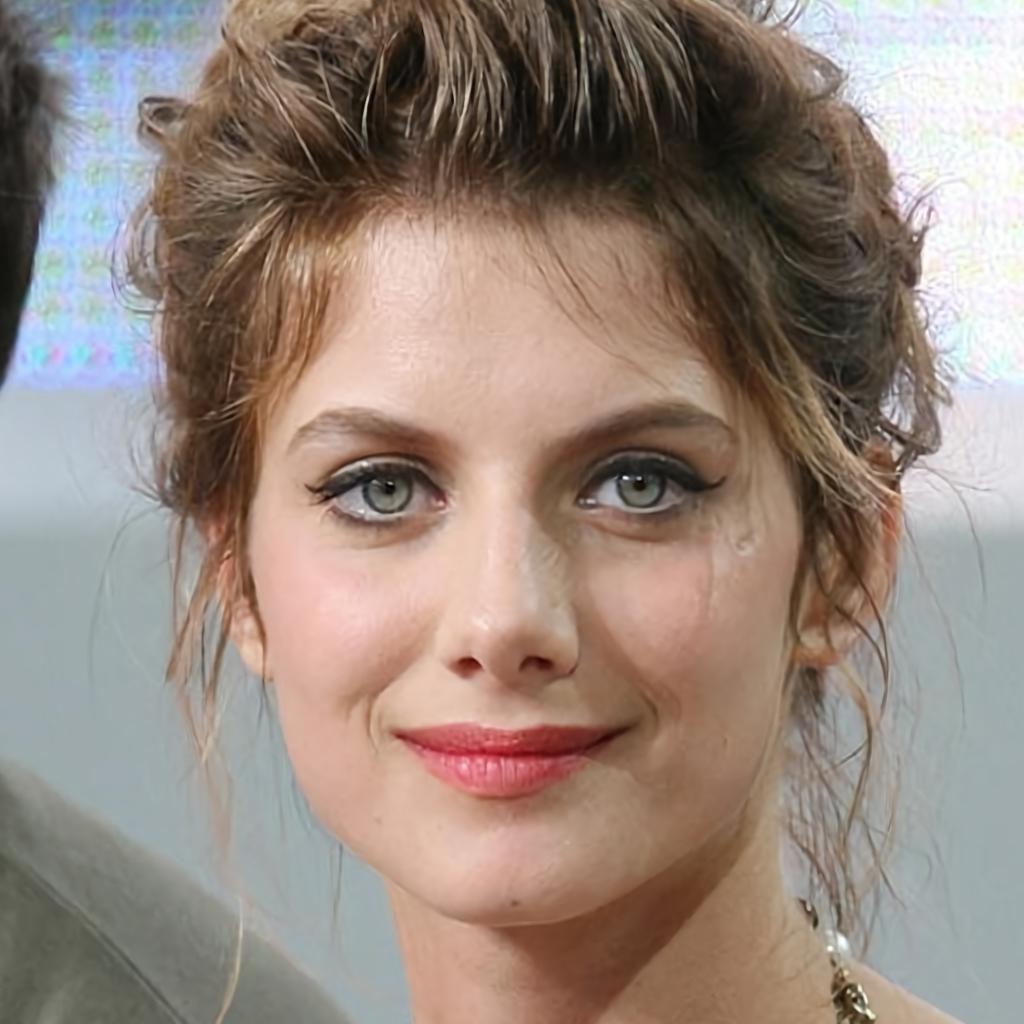}&
        \includegraphics[width=0.175\textwidth]{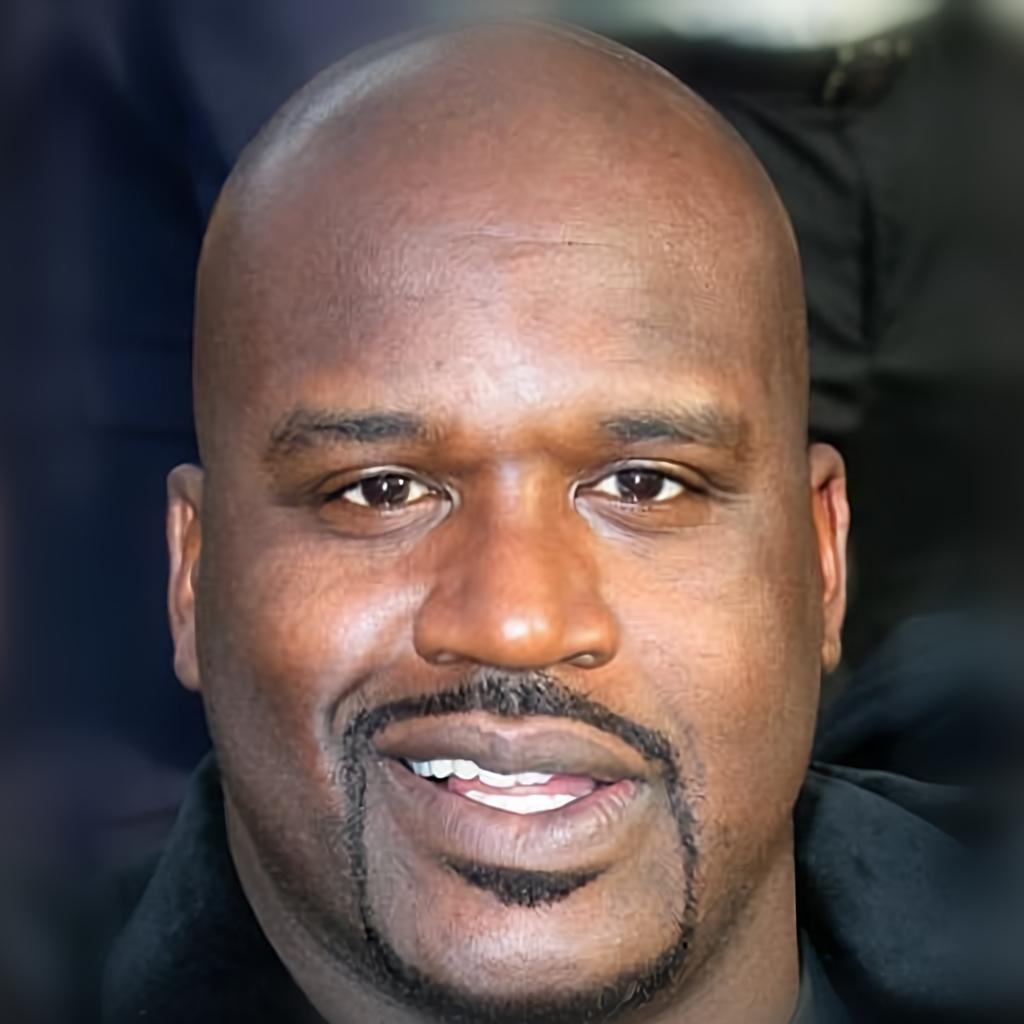}
        \tabularnewline
        \includegraphics[width=0.175\textwidth]{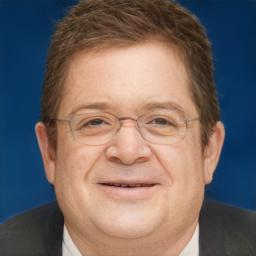}&
        \includegraphics[width=0.175\textwidth]{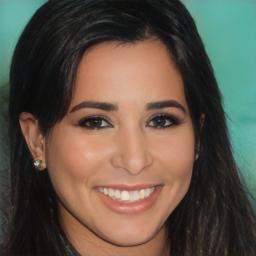}&
        \includegraphics[width=0.175\textwidth]{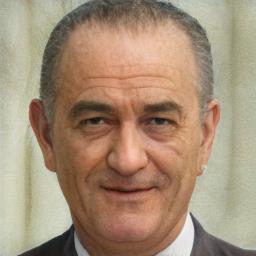}&
        \includegraphics[width=0.175\textwidth]{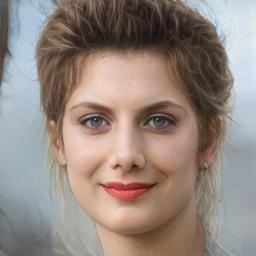}&
        \includegraphics[width=0.175\textwidth]{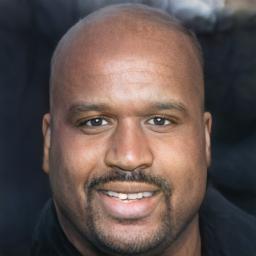}
    \end{tabular}
    \caption{Additional StyleGAN inversion results using pSp on the CelebA-HQ~\cite{karras2018progressive} test set.}
    \label{fig:additional_sketches}
\end{figure*}

\begin{figure*}[!htb]
\setlength{\tabcolsep}{1pt}
\centering
    \begin{tabular}{c c c c c}
        \includegraphics[width=0.175\textwidth]{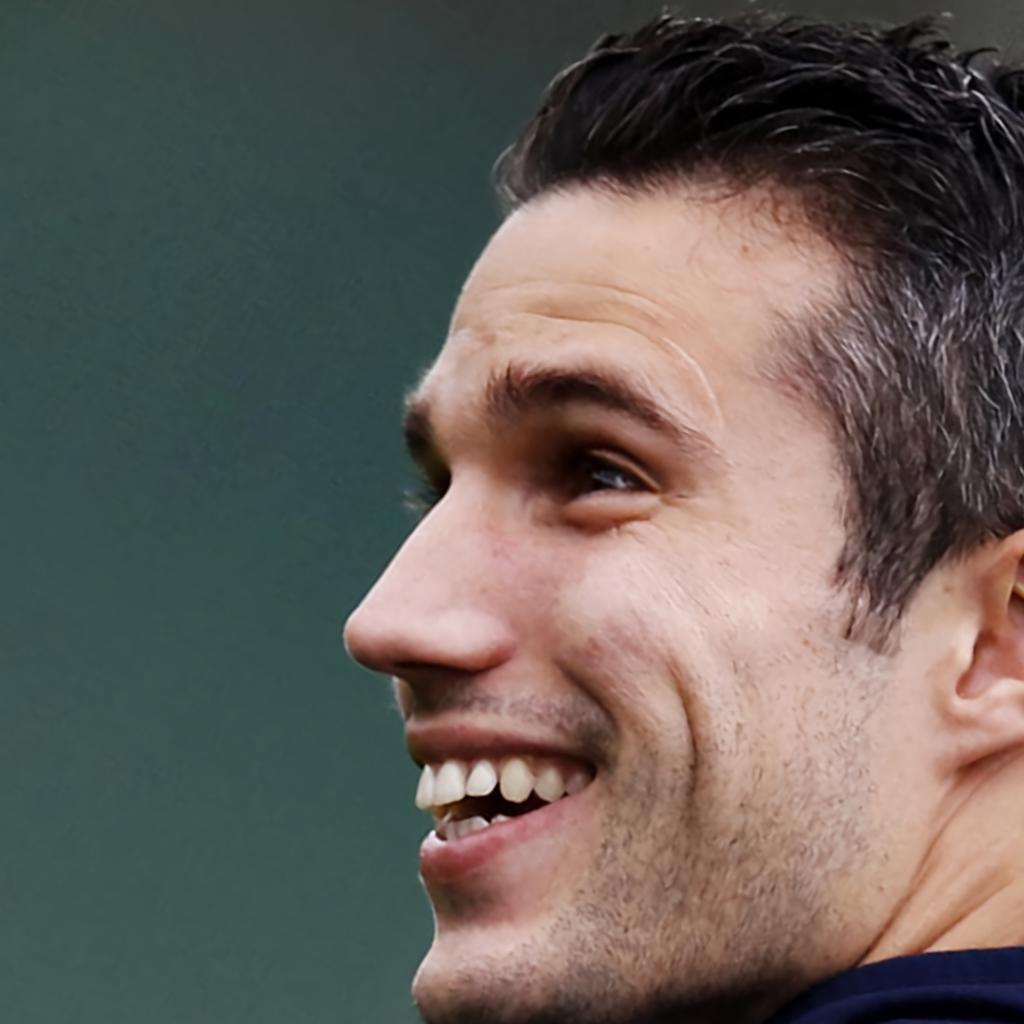}&
        \includegraphics[width=0.175\textwidth]{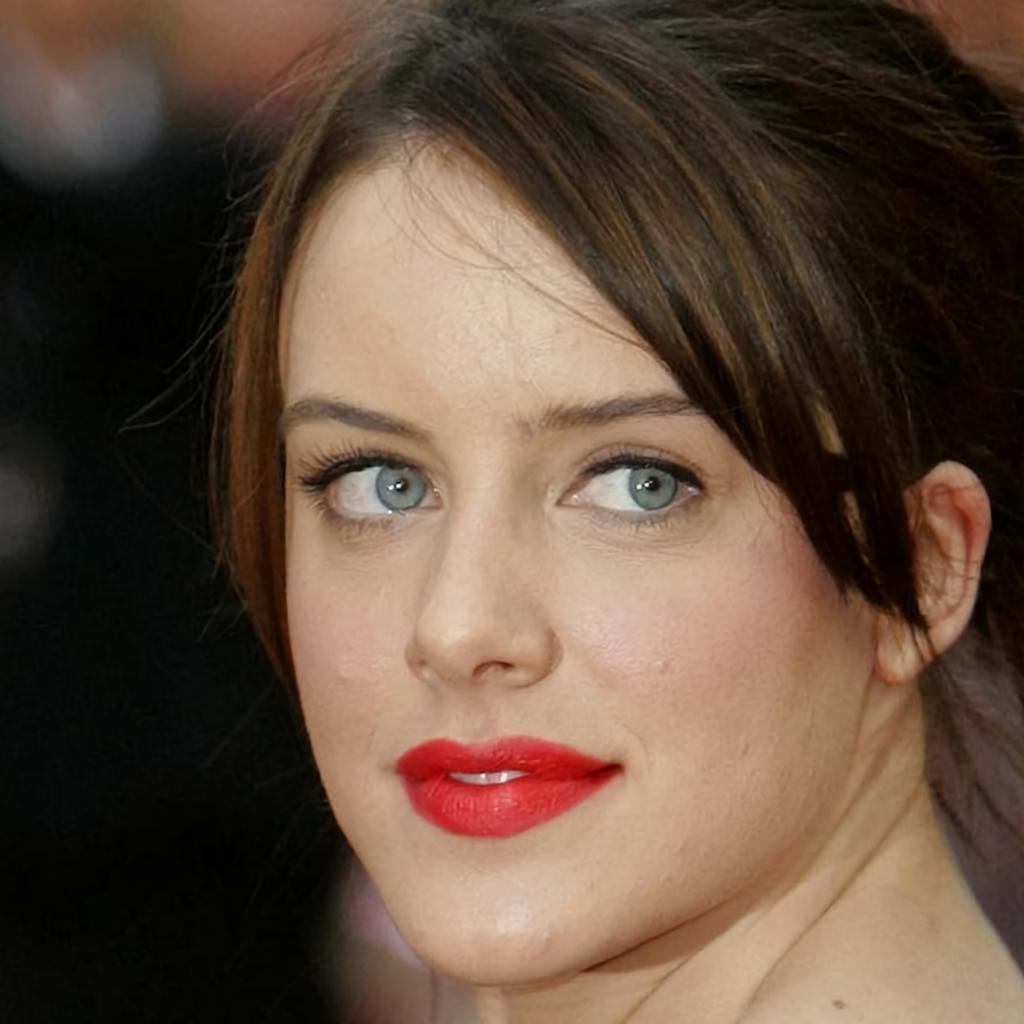}&
        \includegraphics[width=0.175\textwidth]{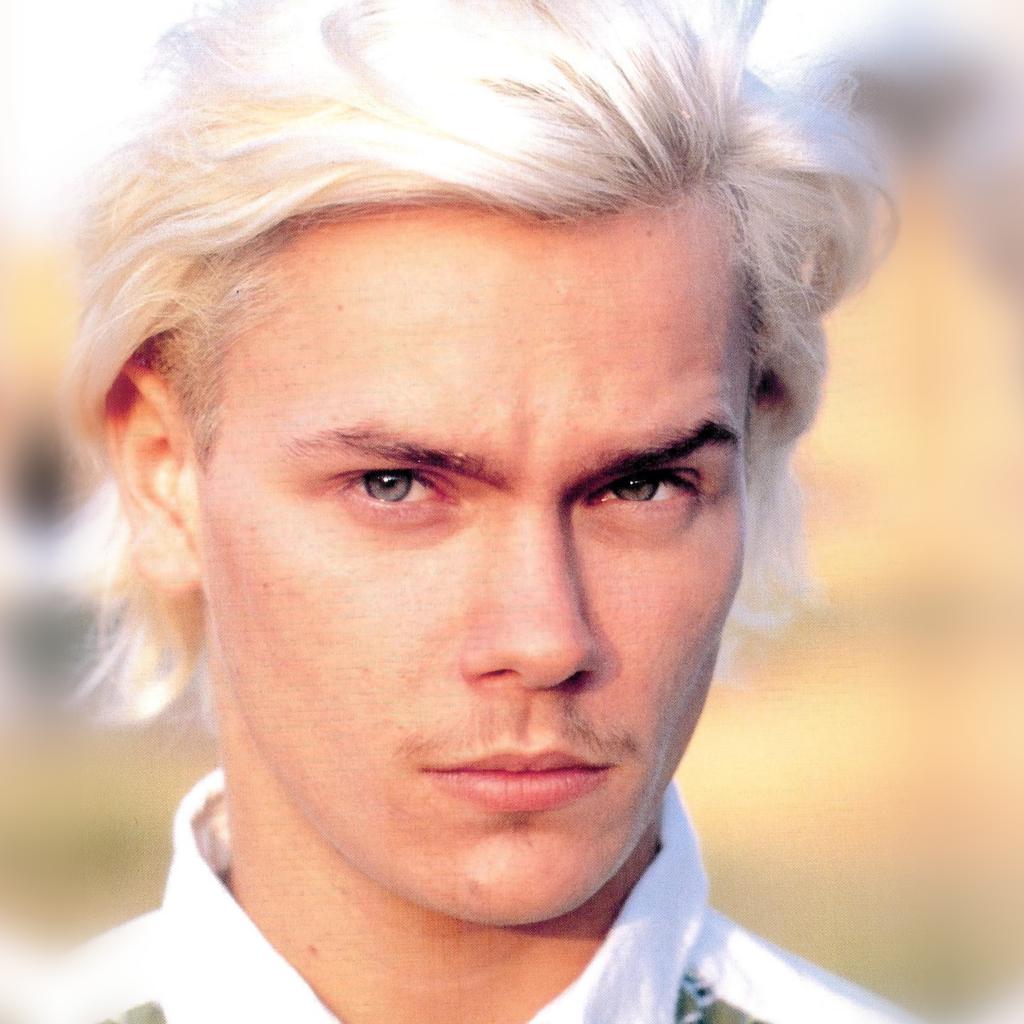}&
        \includegraphics[width=0.175\textwidth]{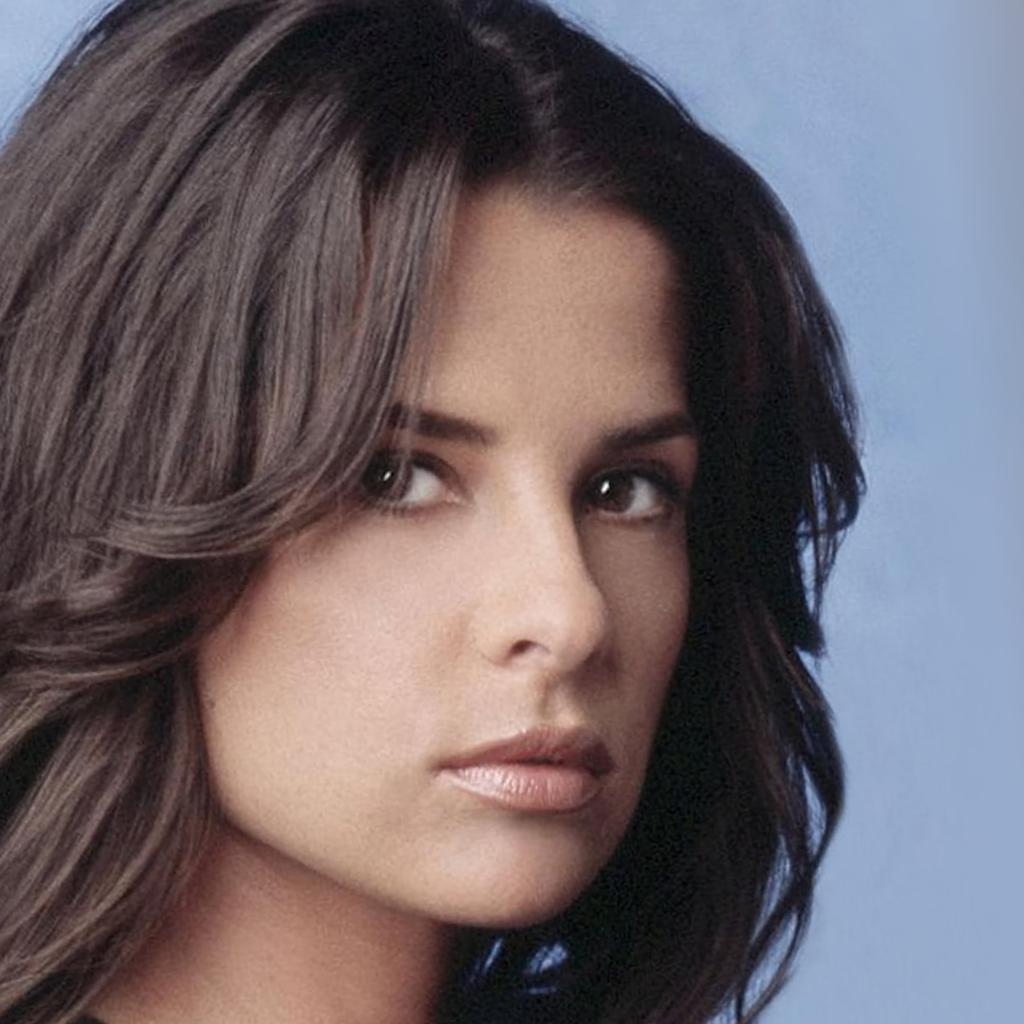}&
        \includegraphics[width=0.175\textwidth]{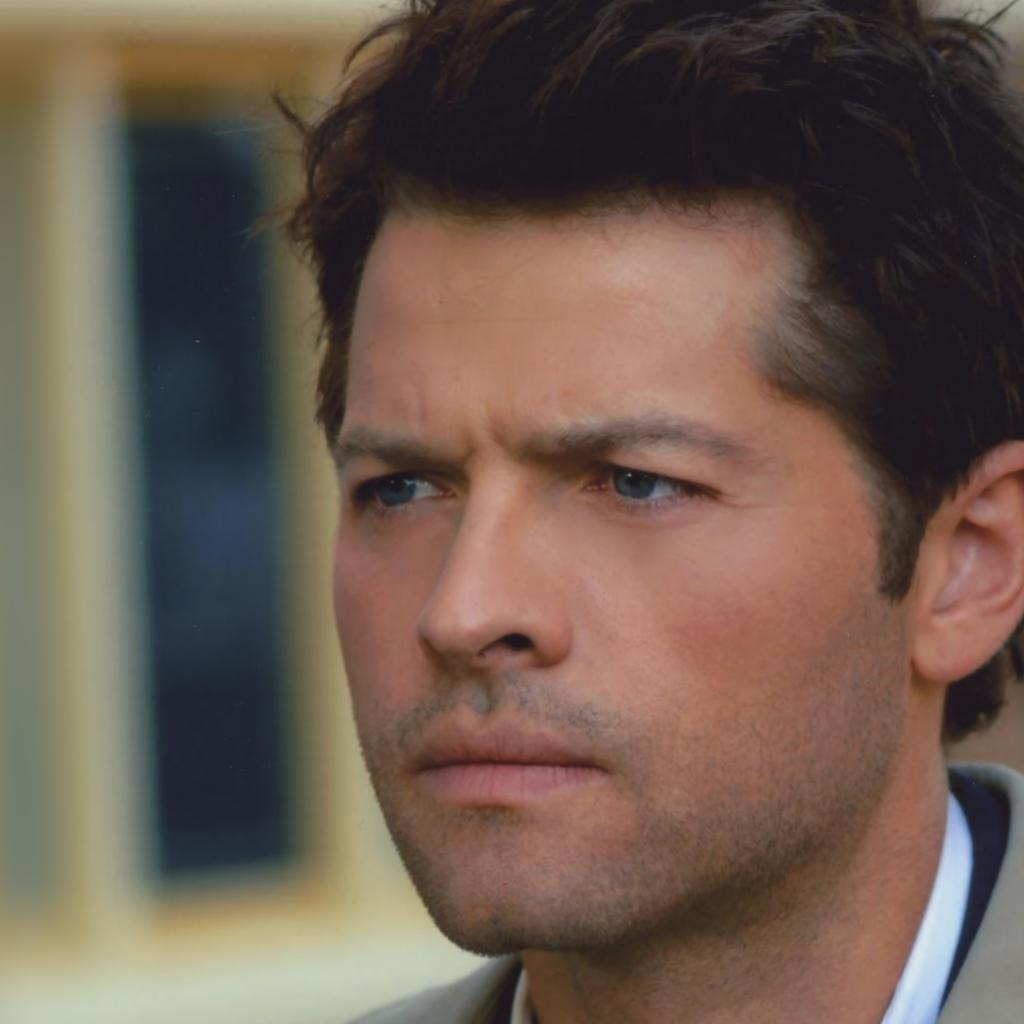}
        \tabularnewline
        \includegraphics[width=0.175\textwidth]{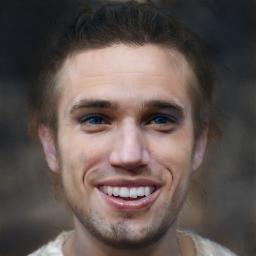}&
        \includegraphics[width=0.175\textwidth]{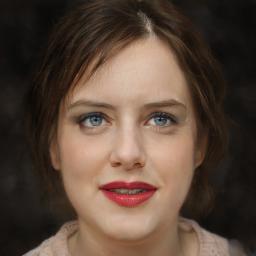}&
        \includegraphics[width=0.175\textwidth]{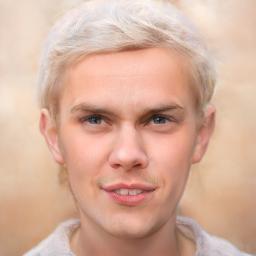}&
        \includegraphics[width=0.175\textwidth]{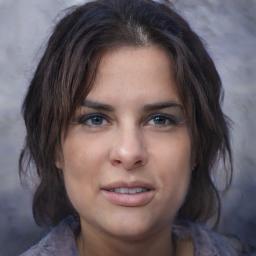}&
        \includegraphics[width=0.175\textwidth]{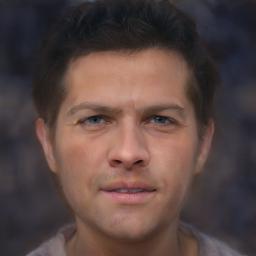}
        \tabularnewline
        \tabularnewline
        \includegraphics[width=0.175\textwidth]{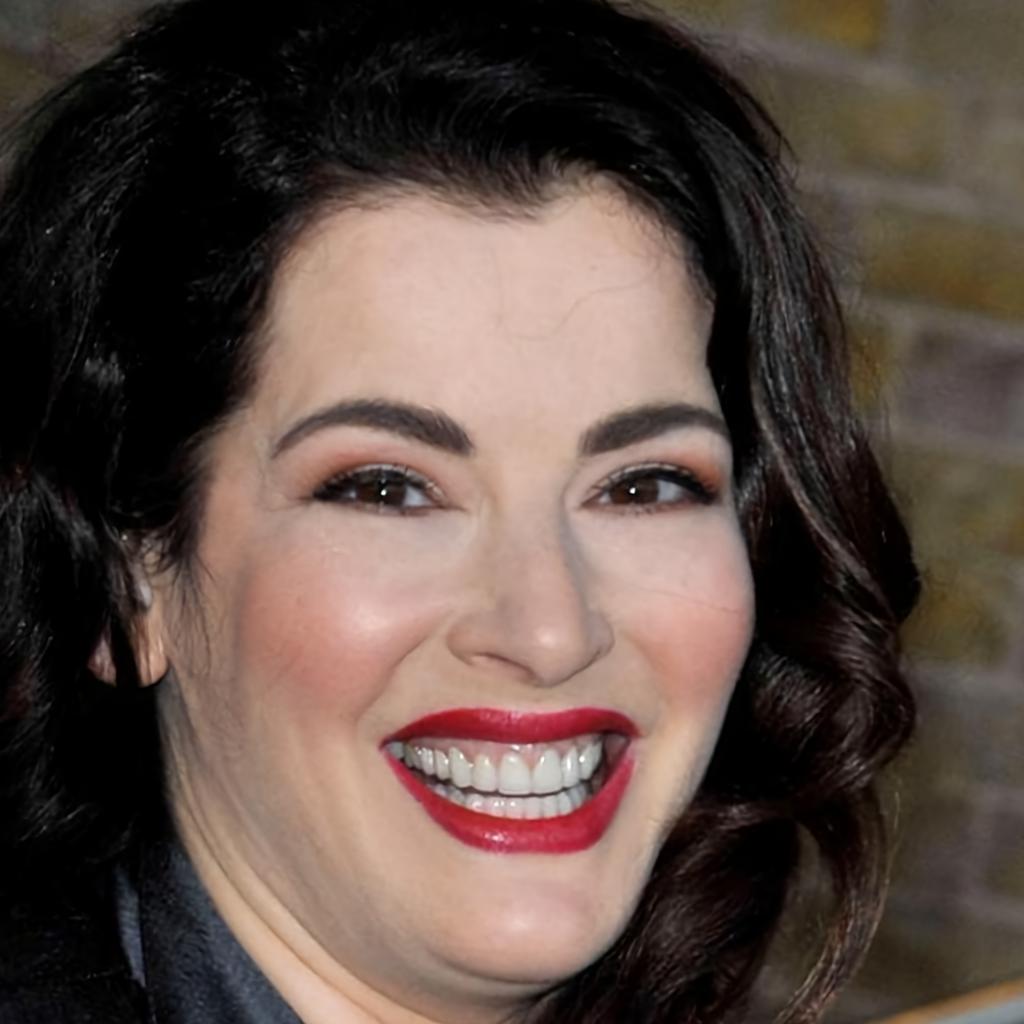}&
        \includegraphics[width=0.175\textwidth]{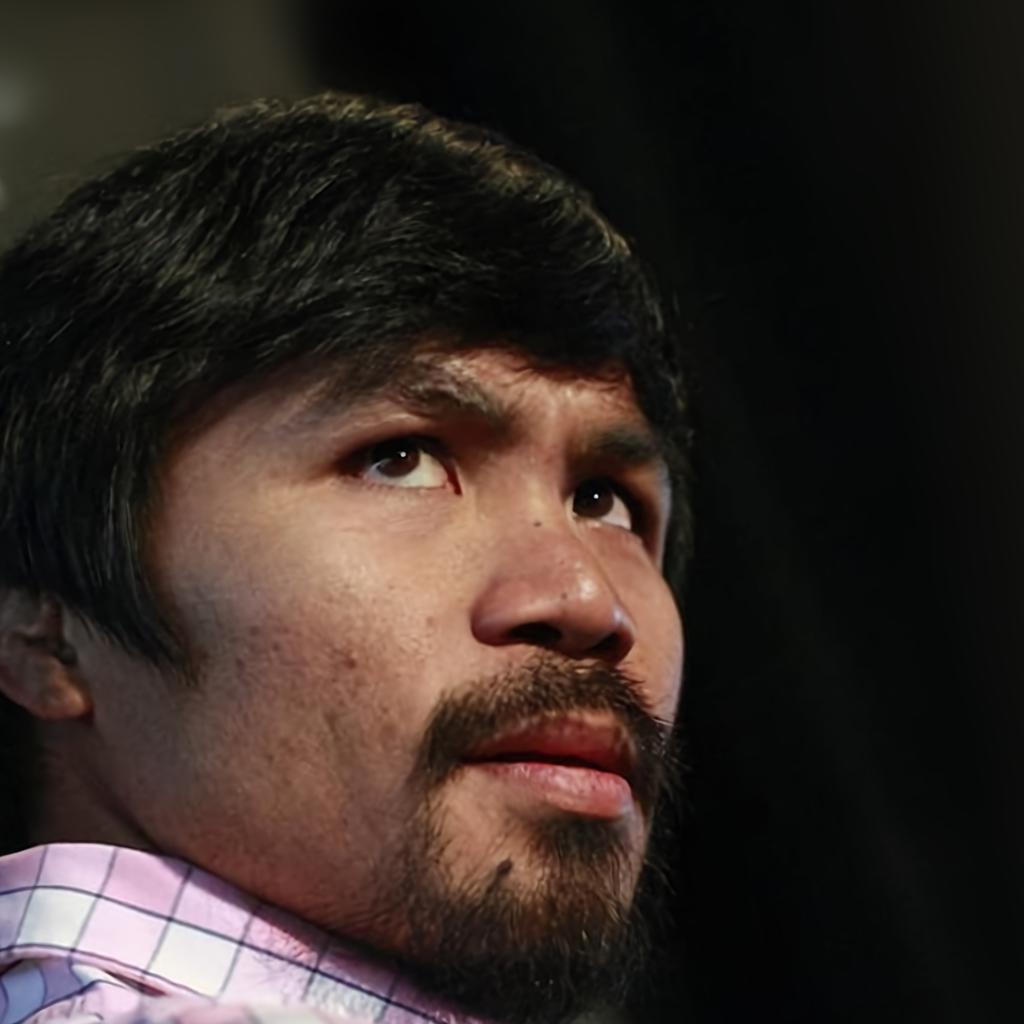}&
        \includegraphics[width=0.175\textwidth]{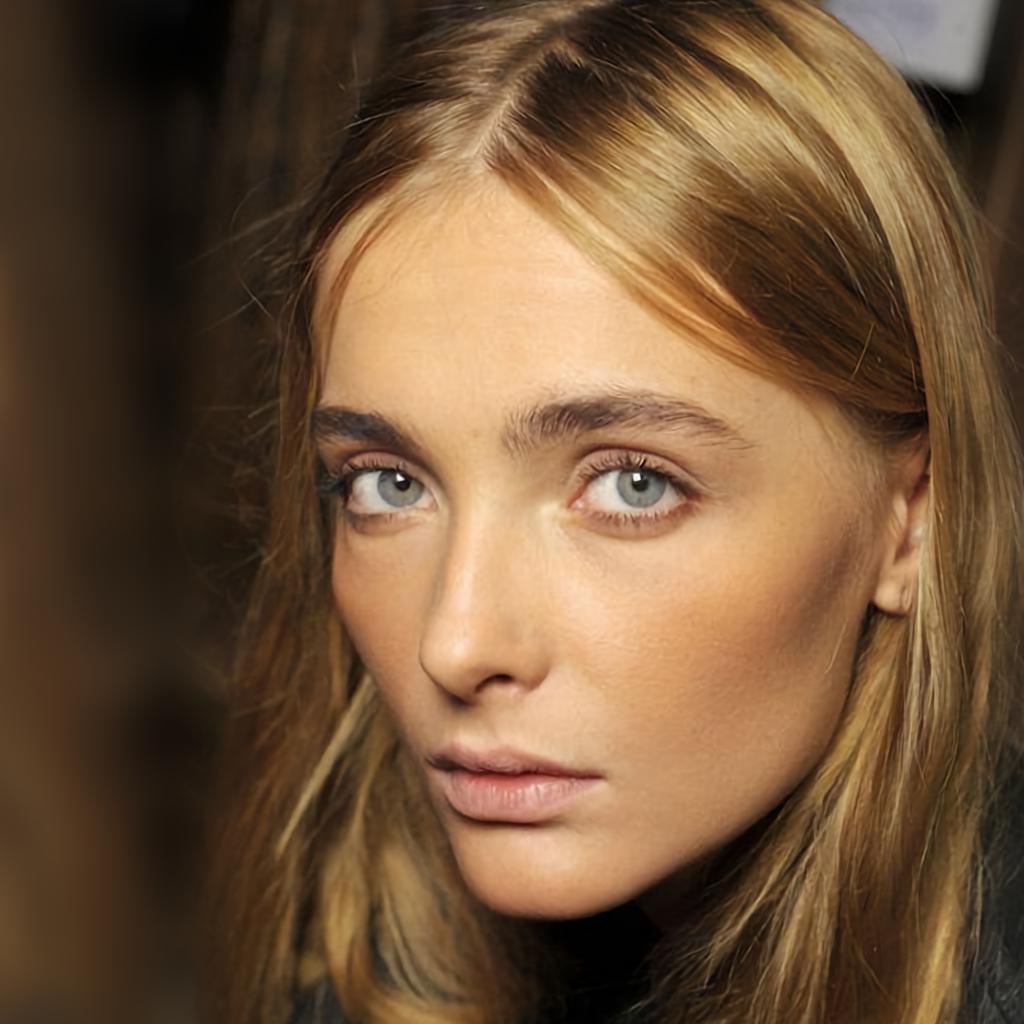}&
        \includegraphics[width=0.175\textwidth]{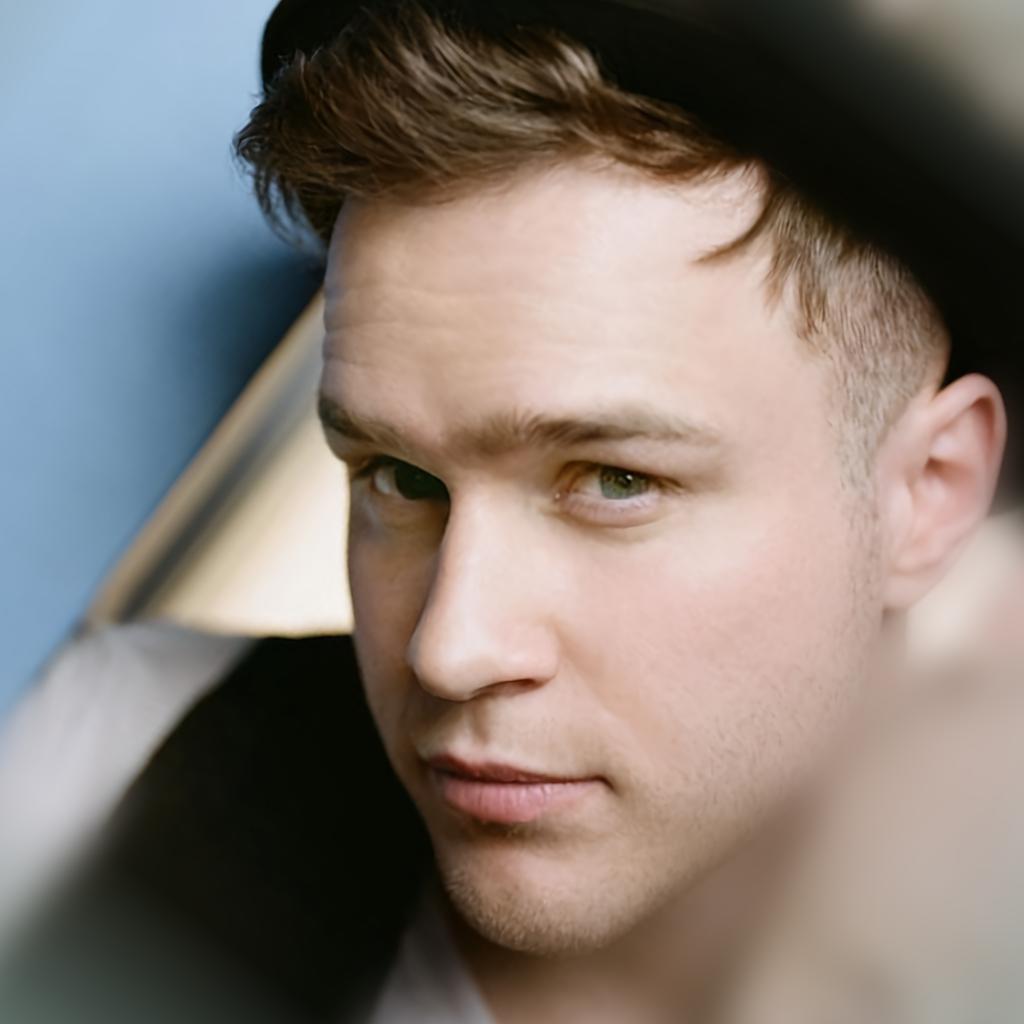}&
        \includegraphics[width=0.175\textwidth]{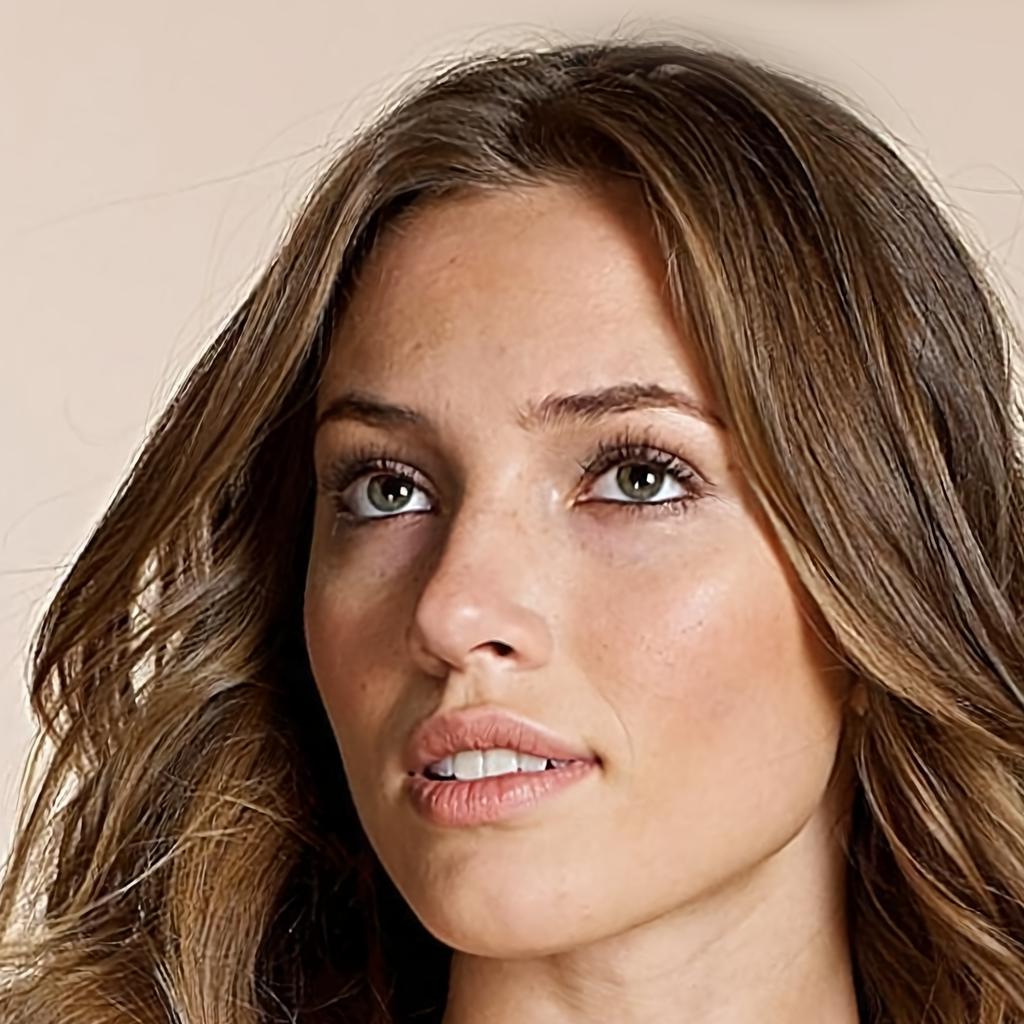}
        \tabularnewline
        \includegraphics[width=0.175\textwidth]{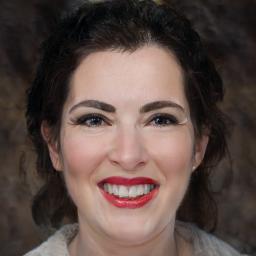}&
        \includegraphics[width=0.175\textwidth]{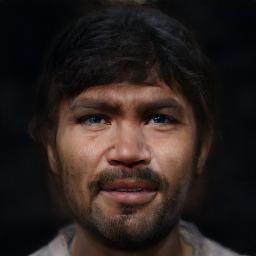}&
        \includegraphics[width=0.175\textwidth]{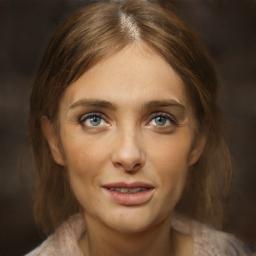}&
        \includegraphics[width=0.175\textwidth]{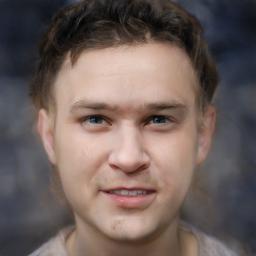}&
        \includegraphics[width=0.175\textwidth]{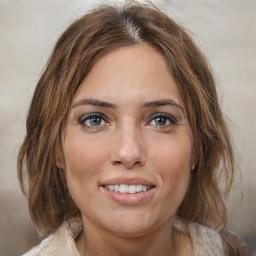}
        \tabularnewline
        \tabularnewline
        \includegraphics[width=0.175\textwidth]{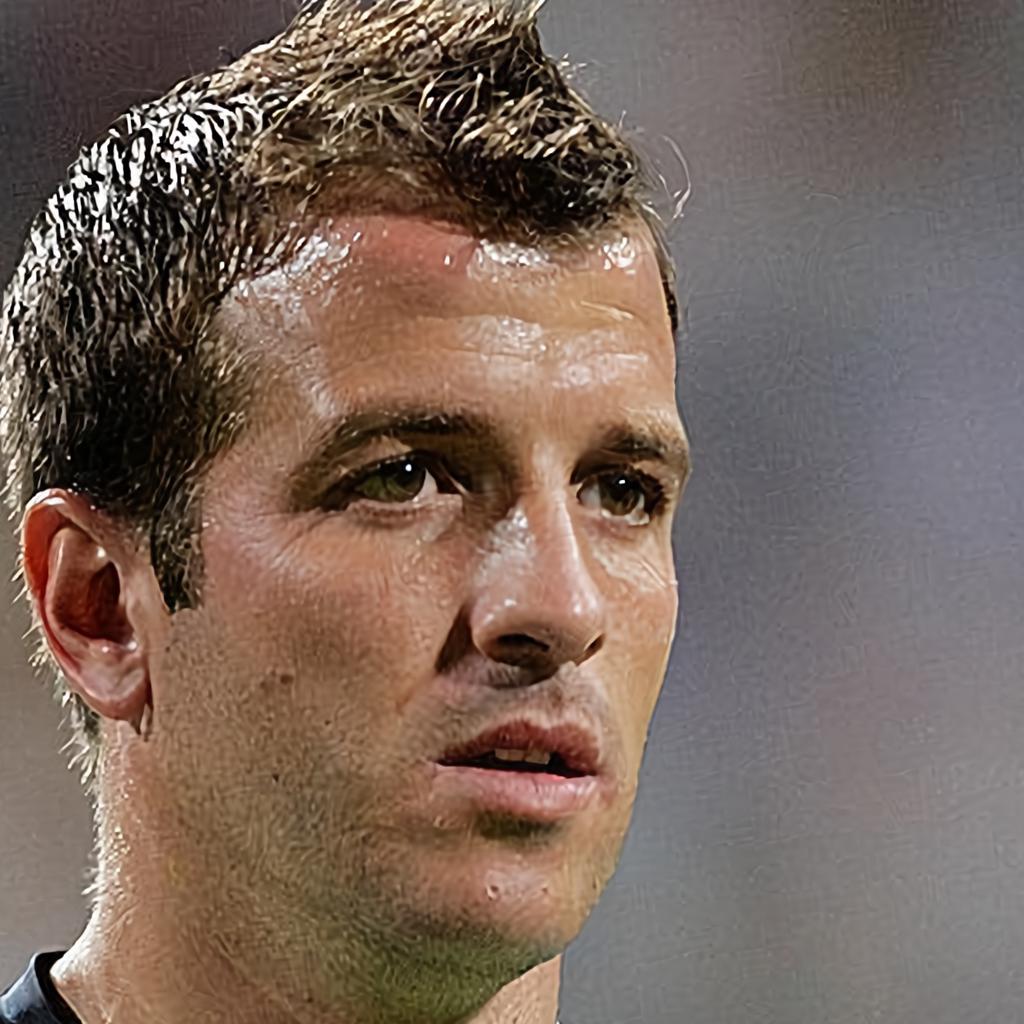}&
        \includegraphics[width=0.175\textwidth]{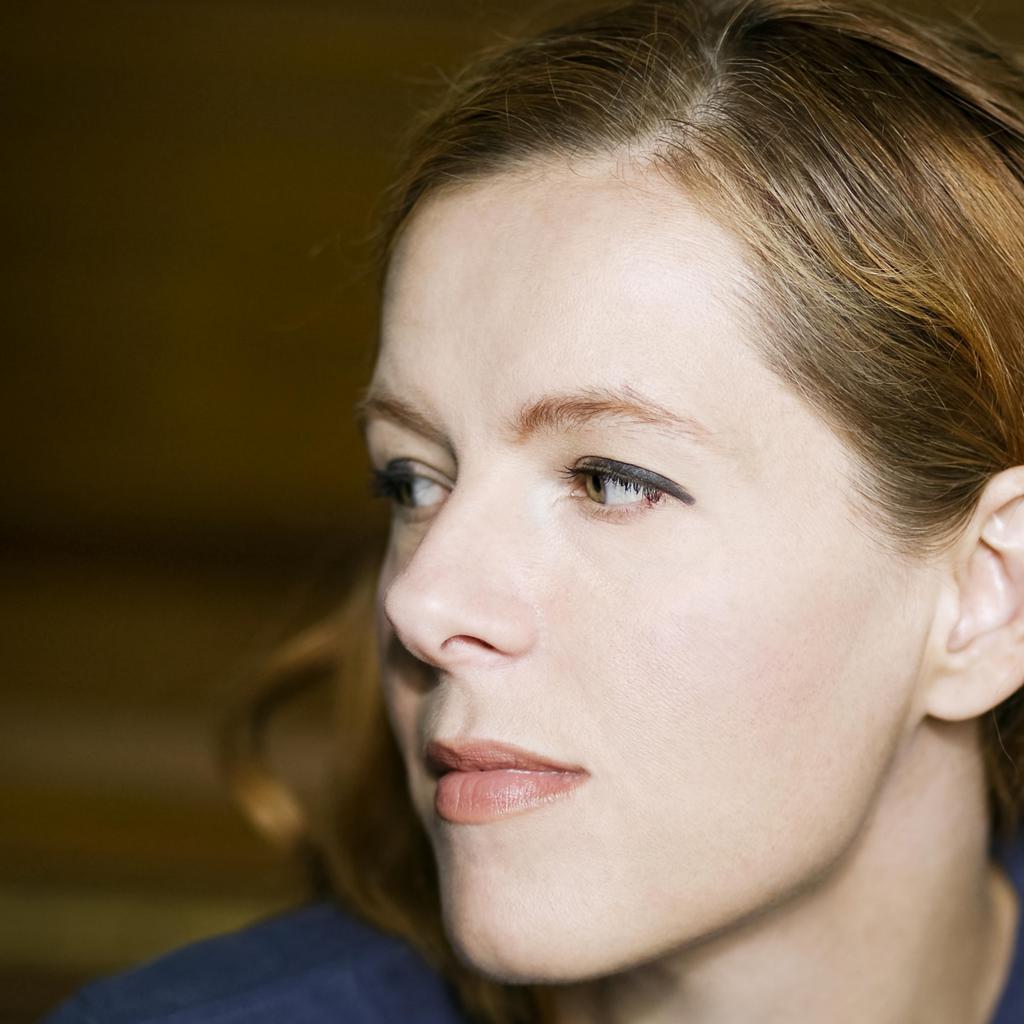}&
        \includegraphics[width=0.175\textwidth]{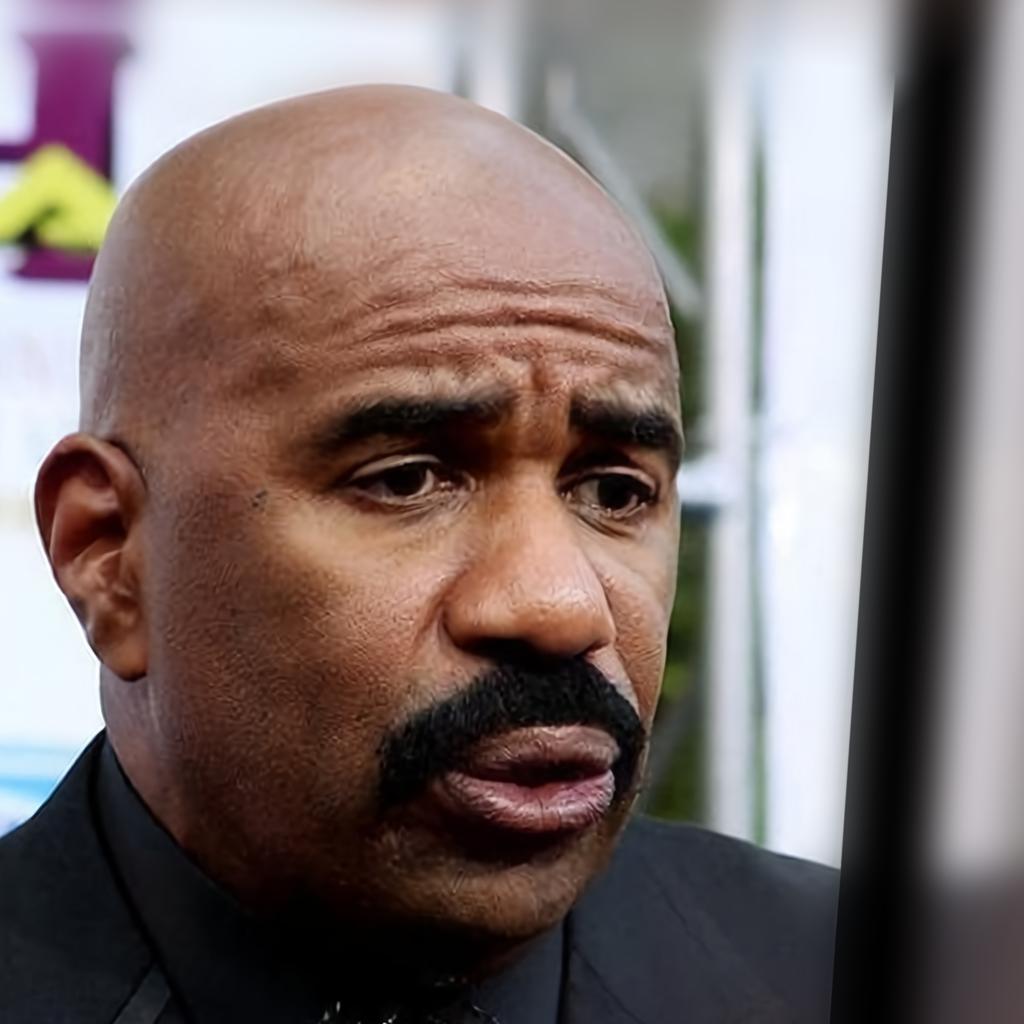}&
        \includegraphics[width=0.175\textwidth]{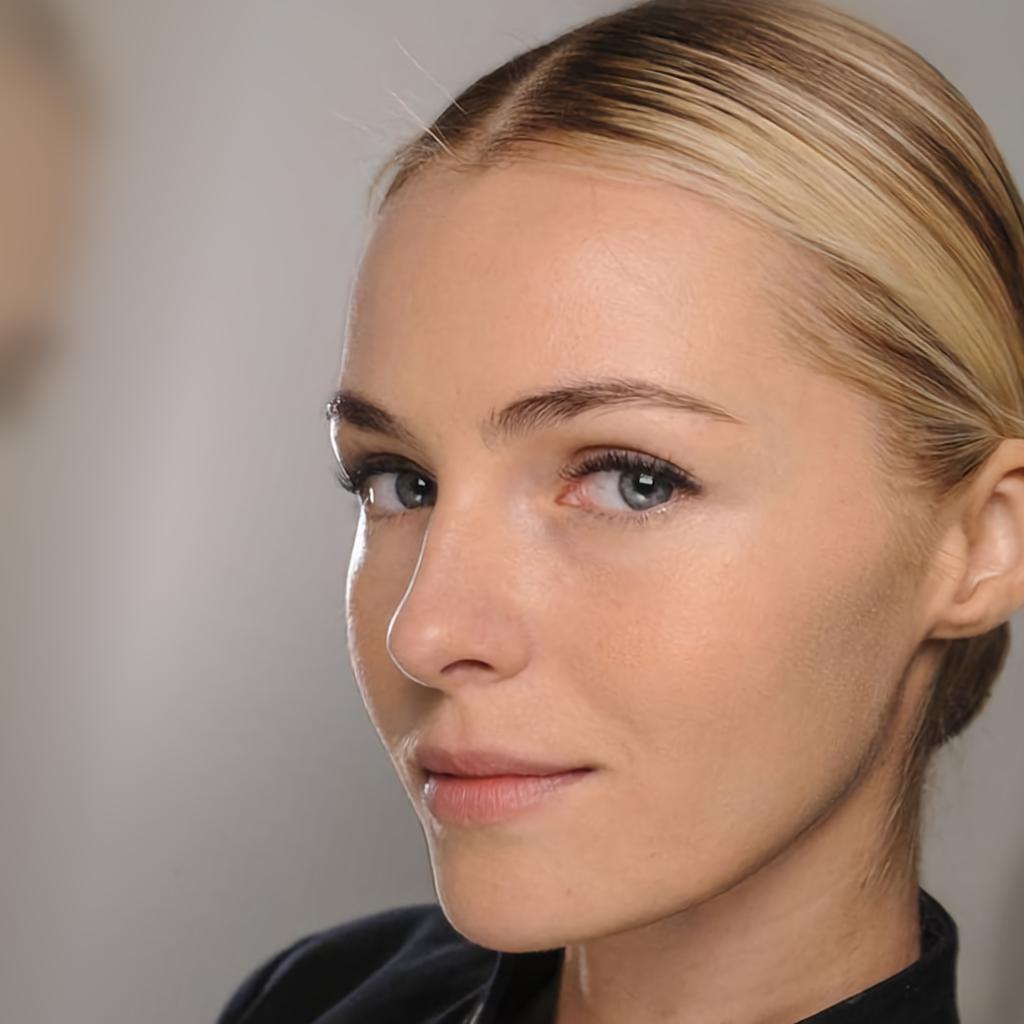}&
        \includegraphics[width=0.175\textwidth]{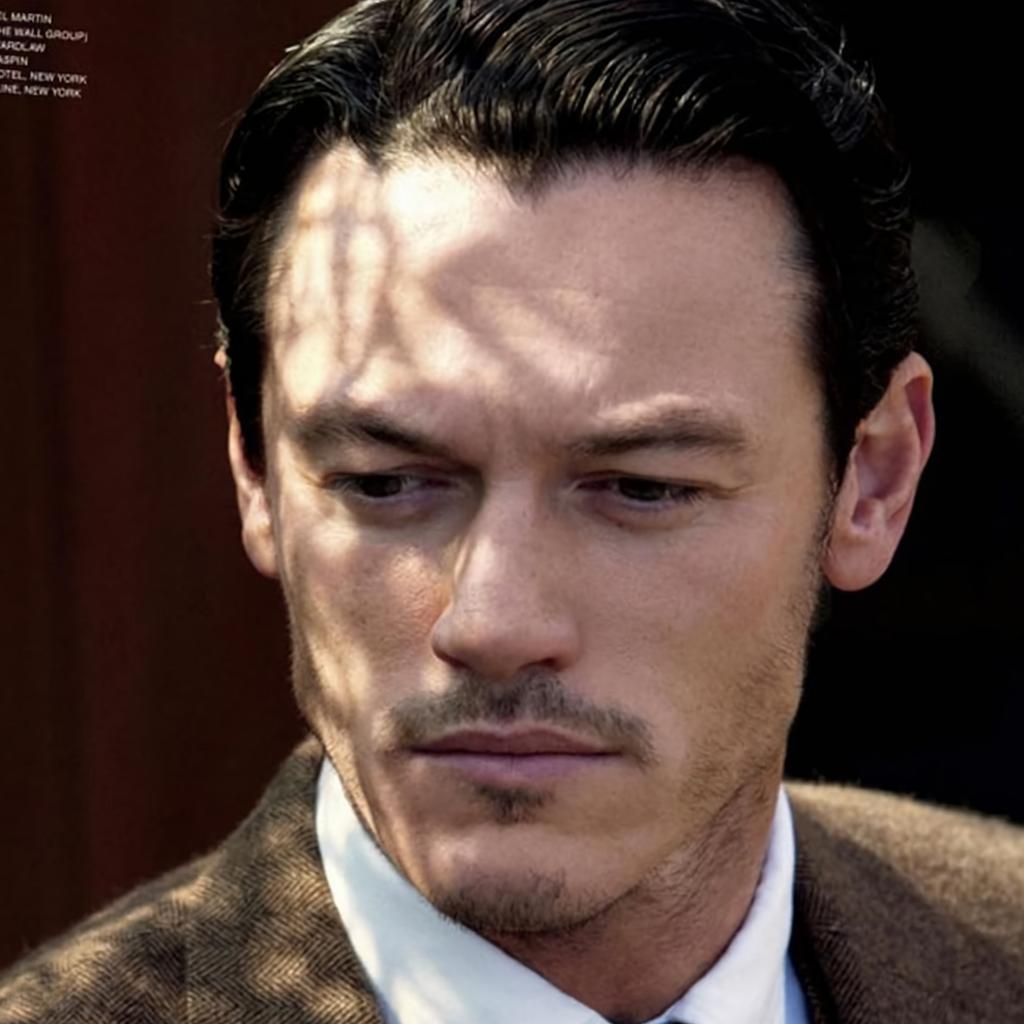}
        \tabularnewline
        \includegraphics[width=0.175\textwidth]{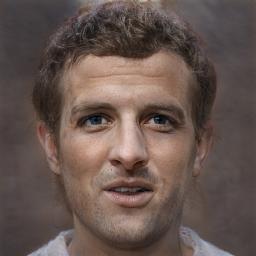}&
        \includegraphics[width=0.175\textwidth]{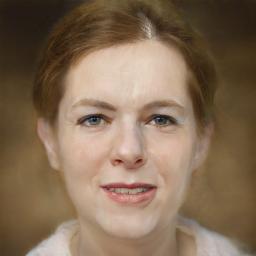}&
        \includegraphics[width=0.175\textwidth]{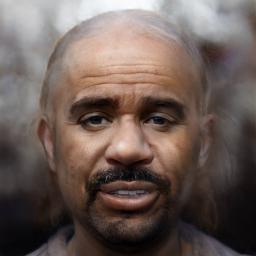}&
        \includegraphics[width=0.175\textwidth]{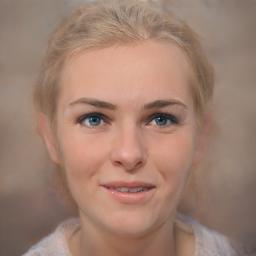}&
        \includegraphics[width=0.175\textwidth]{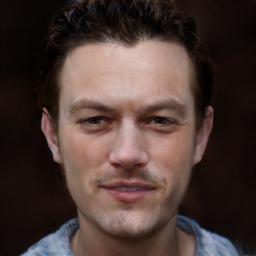}
    \end{tabular}
    \caption{Additional face frontalization results using pSp on the CelebA-HQ~\cite{karras2018progressive} test set.}
    \label{fig:additional_sketches}
\end{figure*}

\begin{figure*}
    \setlength{\tabcolsep}{1pt}
        \begin{tabular}{c c c c c c c}
            \raisebox{0.40in}{\rotatebox[origin=t]{90}{Input}}&
            \includegraphics[width=0.16\textwidth]{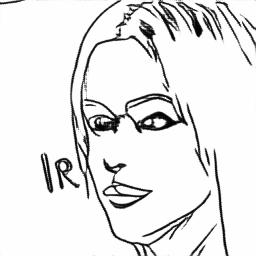}&   
            \includegraphics[width=0.16\textwidth]{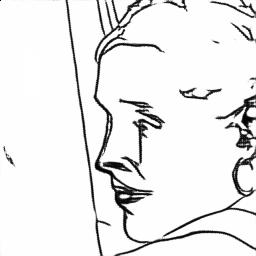}&
            \includegraphics[width=0.16\textwidth]{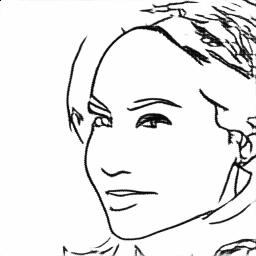}&
            \includegraphics[width=0.16\textwidth]{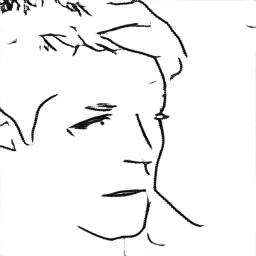}&
            \includegraphics[width=0.16\textwidth]{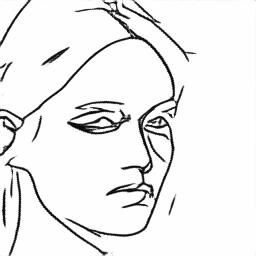}&  
            \includegraphics[width=0.16\textwidth]{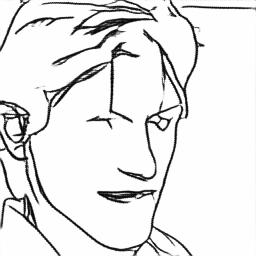}
            \tabularnewline
            \raisebox{0.4in}{\rotatebox[origin=t]{90}{pSp}}&
            \includegraphics[width=0.16\textwidth]{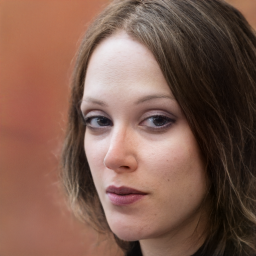}&    
            \includegraphics[width=0.16\textwidth]{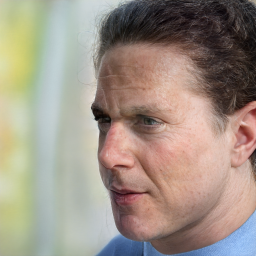}&
            \includegraphics[width=0.16\textwidth]{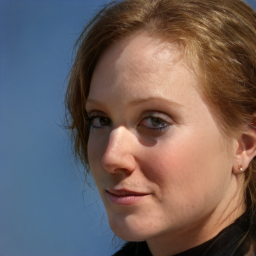}&
            \includegraphics[width=0.16\textwidth]{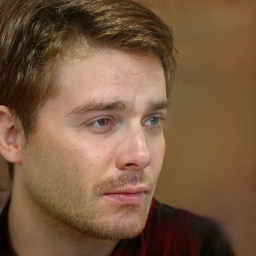}&
            \includegraphics[width=0.16\textwidth]{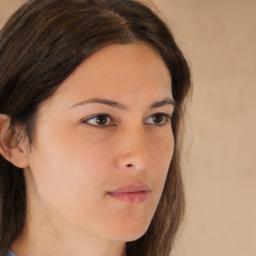}&
            \includegraphics[width=0.16\textwidth]{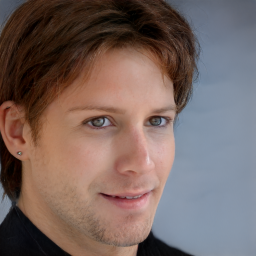}
            \tabularnewline
            \raisebox{0.45in}{\rotatebox[origin=t]{90}{pix2pixHD}}&
            \includegraphics[width=0.16\textwidth]{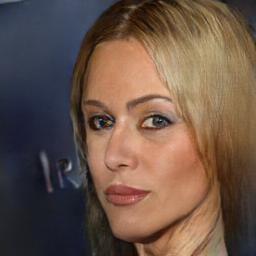}&
            \includegraphics[width=0.16\textwidth]{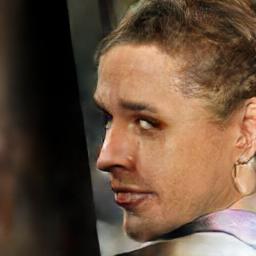}&
            \includegraphics[width=0.16\textwidth]{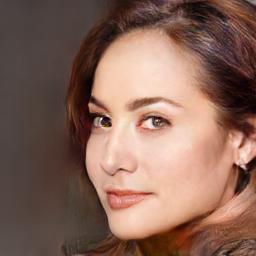}&
            \includegraphics[width=0.16\textwidth]{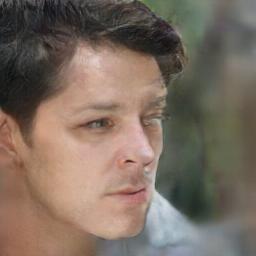}&
            \includegraphics[width=0.16\textwidth]{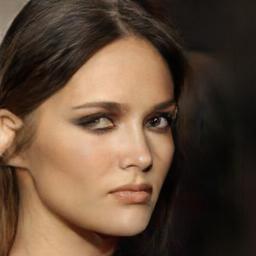}&   
            \includegraphics[width=0.16\textwidth]{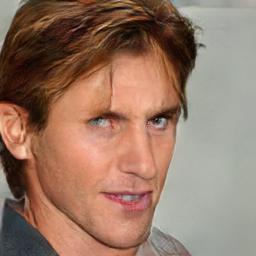}
            \end{tabular}
        \caption{Even for challenging, non-frontal face sketches, pSp is able to obtain high-quality, diverse outputs.}
        \label{fig:sketch_non_frontal}
    \end{figure*}

\begin{figure*}
\setlength{\tabcolsep}{1pt}
\centering
    \begin{tabular}{c c c c c}
        \includegraphics[width=0.20\textwidth]{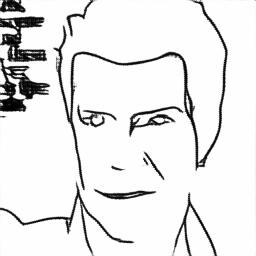}&
        \includegraphics[width=0.20\textwidth]{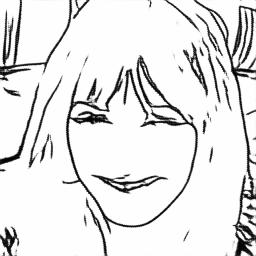}&
        \includegraphics[width=0.20\textwidth]{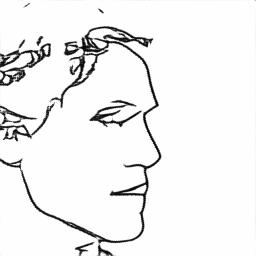}&
        \includegraphics[width=0.20\textwidth]{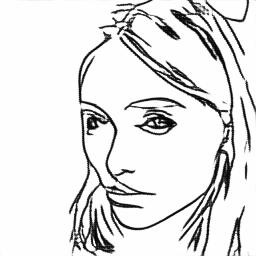}&
        \includegraphics[width=0.20\textwidth]{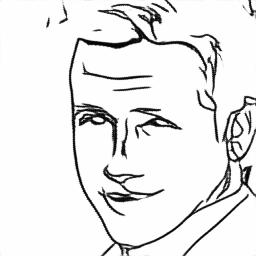}
        \tabularnewline
        \includegraphics[width=0.20\textwidth]{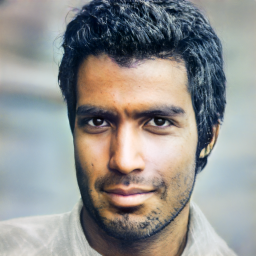}&
        \includegraphics[width=0.20\textwidth]{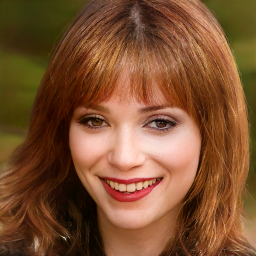}&
        \includegraphics[width=0.20\textwidth]{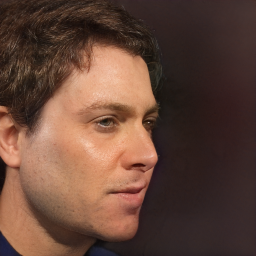}&
        \includegraphics[width=0.20\textwidth]{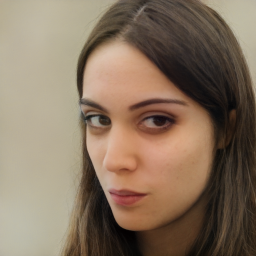}&
        \includegraphics[width=0.20\textwidth]{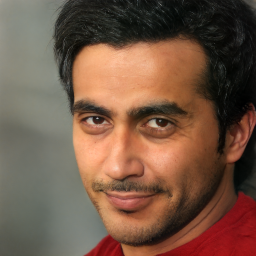}
        \tabularnewline
        \tabularnewline
        \includegraphics[width=0.20\textwidth]{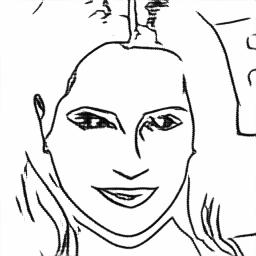}&
        \includegraphics[width=0.20\textwidth]{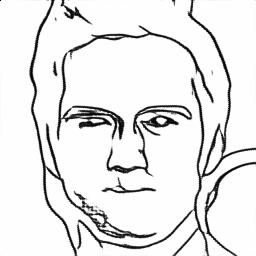}&
        \includegraphics[width=0.20\textwidth]{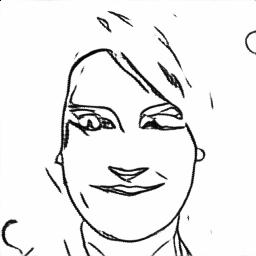}&
        \includegraphics[width=0.20\textwidth]{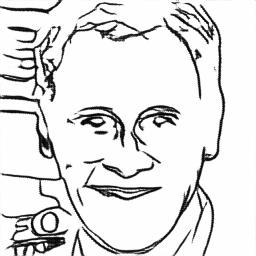}&
        \includegraphics[width=0.20\textwidth]{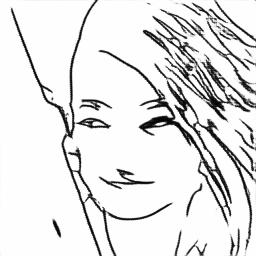}
        \tabularnewline
        \includegraphics[width=0.20\textwidth]{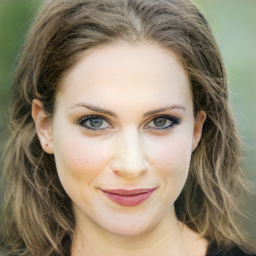}&
        \includegraphics[width=0.20\textwidth]{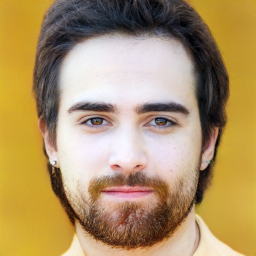}&
        \includegraphics[width=0.20\textwidth]{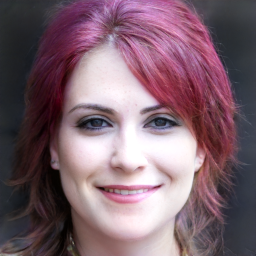}&
        \includegraphics[width=0.20\textwidth]{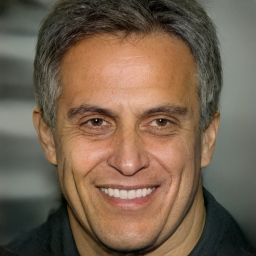}&
        \includegraphics[width=0.20\textwidth]{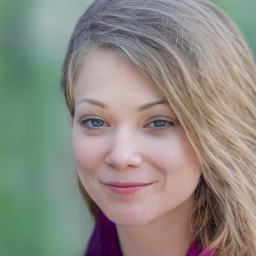}
    \end{tabular}
    \caption{Additional results using pSp for the generation of face images from sketches on the CelebA-HQ~\cite{karras2018progressive} test dataset.}
    \label{fig:additional_sketches}
\end{figure*}

\begin{figure*}
    \centering
    \includegraphics[width=\textwidth]{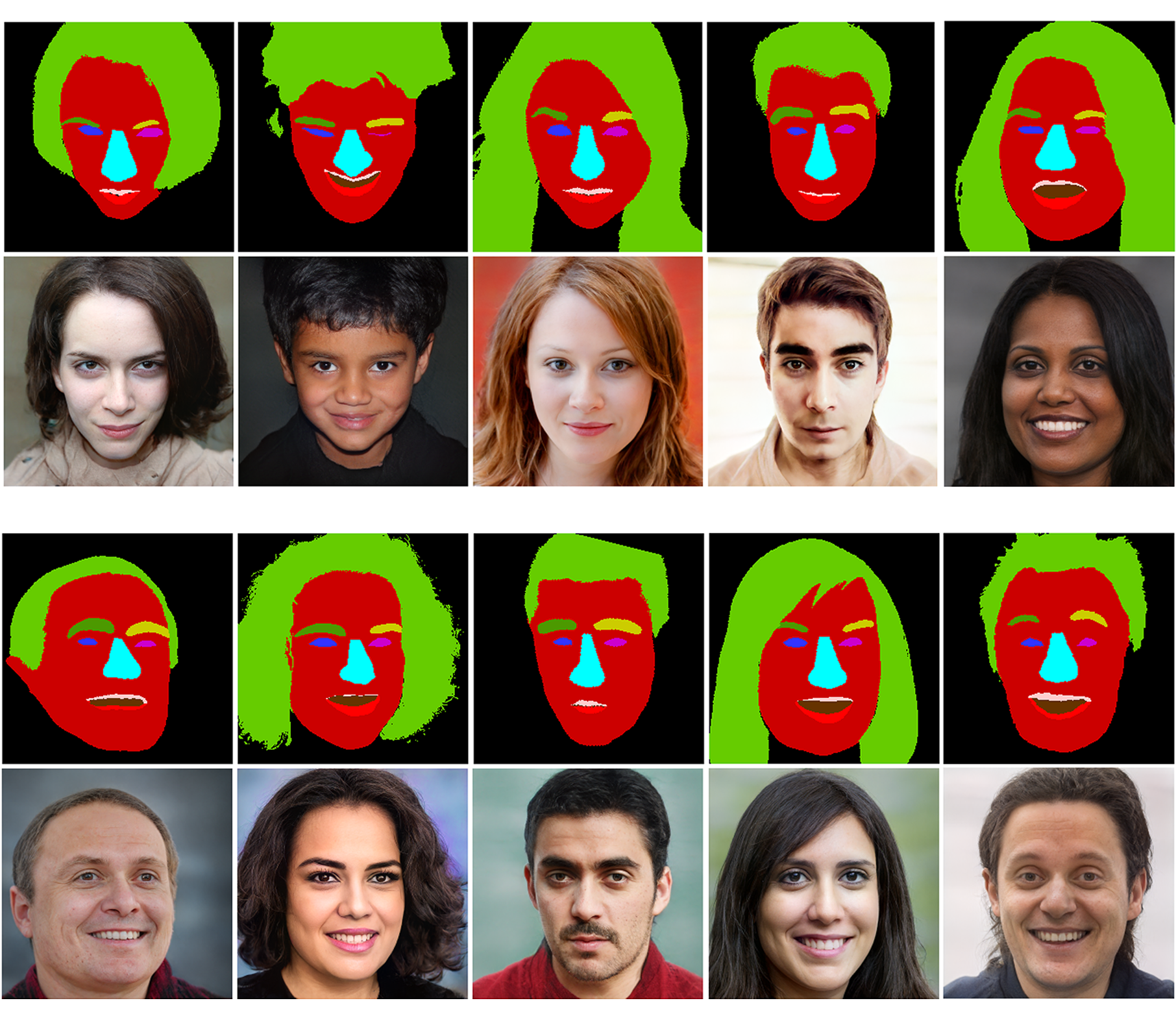}
    \captionof{figure}{Additional results on the Helen Faces \cite{HelenFaces} dataset using our proposed segmentation-to-image method.}
    \label{helen_segmentation_results}
\end{figure*}

\begin{figure*}
\setlength{\tabcolsep}{1pt}
\centering
    \begin{tabular}{c c c c c}
        \includegraphics[width=0.20\textwidth]{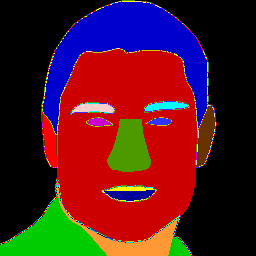}&
        \includegraphics[width=0.20\textwidth]{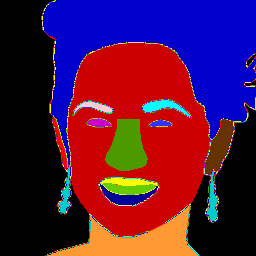}&
        \includegraphics[width=0.20\textwidth]{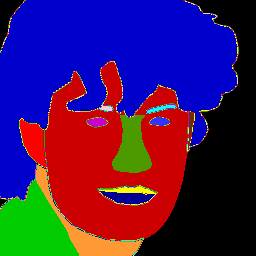}&
        \includegraphics[width=0.20\textwidth]{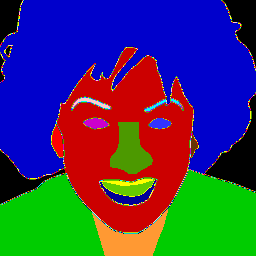}&
        \includegraphics[width=0.20\textwidth]{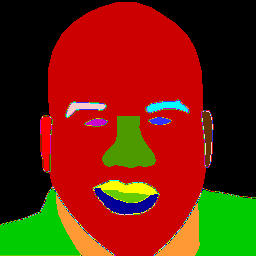}
        \tabularnewline
        \includegraphics[width=0.20\textwidth]{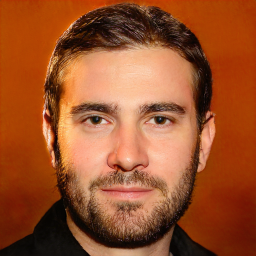}&
        \includegraphics[width=0.20\textwidth]{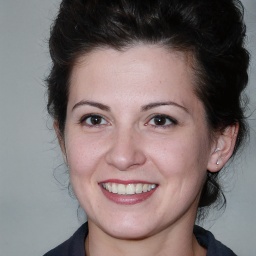}&
        \includegraphics[width=0.20\textwidth]{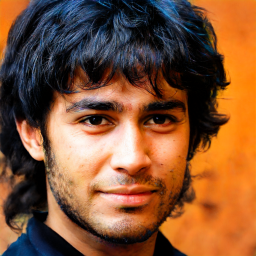}&
        \includegraphics[width=0.20\textwidth]{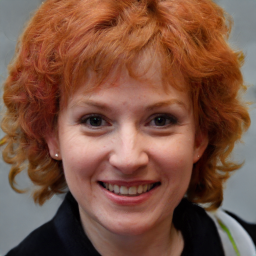}&
        \includegraphics[width=0.20\textwidth]{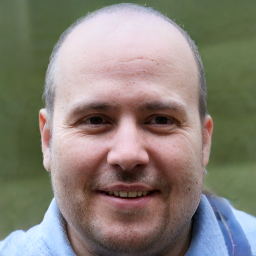}
        \tabularnewline
        \tabularnewline
        \includegraphics[width=0.20\textwidth]{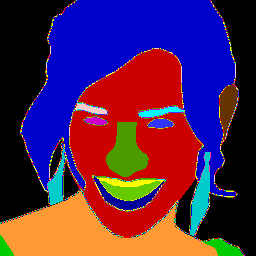}&
        \includegraphics[width=0.20\textwidth]{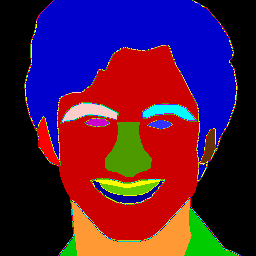}&
        \includegraphics[width=0.20\textwidth]{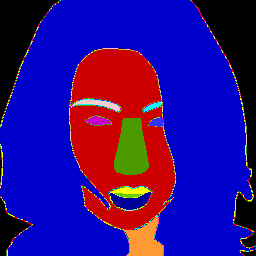}&
        \includegraphics[width=0.20\textwidth]{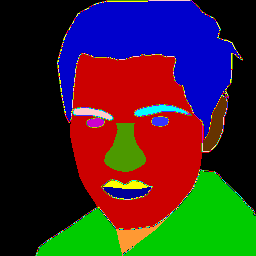}&
        \includegraphics[width=0.20\textwidth]{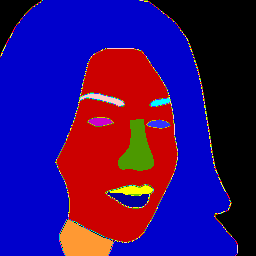}
        \tabularnewline
        \includegraphics[width=0.20\textwidth]{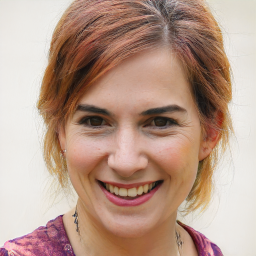}&
        \includegraphics[width=0.20\textwidth]{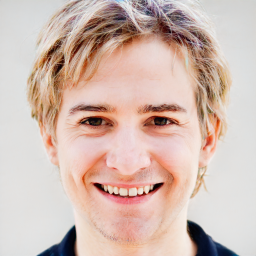}&
        \includegraphics[width=0.20\textwidth]{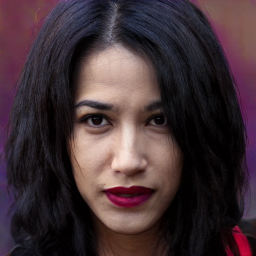}&
        \includegraphics[width=0.20\textwidth]{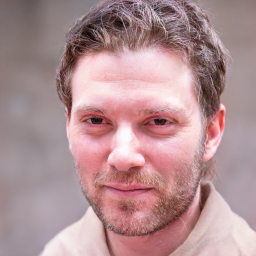}&
        \includegraphics[width=0.20\textwidth]{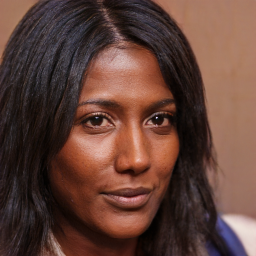}
    \end{tabular}
    \caption{Additional results on the CelebAMask-HQ~\cite{karras2018progressive} test set using our proposed segmentation-to-image method.}
    \label{fig:additional_segmentations}
\end{figure*}

\begin{figure*}
    \centering
        \includegraphics[width=0.9\textwidth]{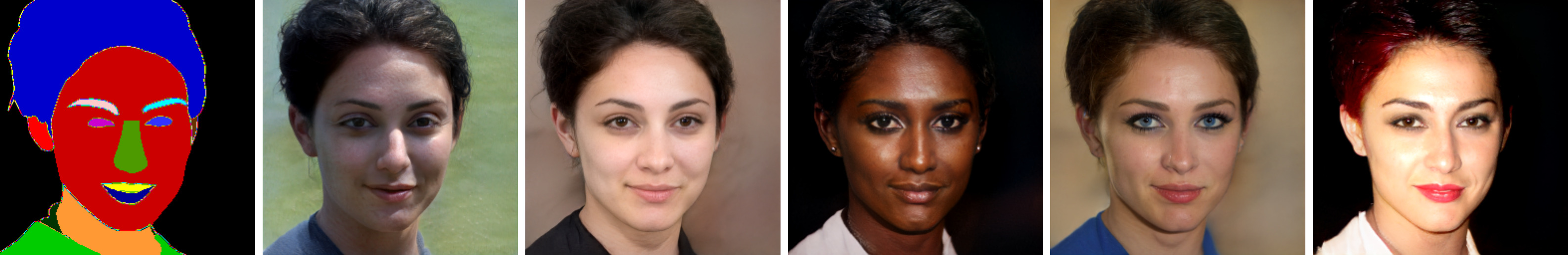}\\
        \includegraphics[width=0.9\textwidth]{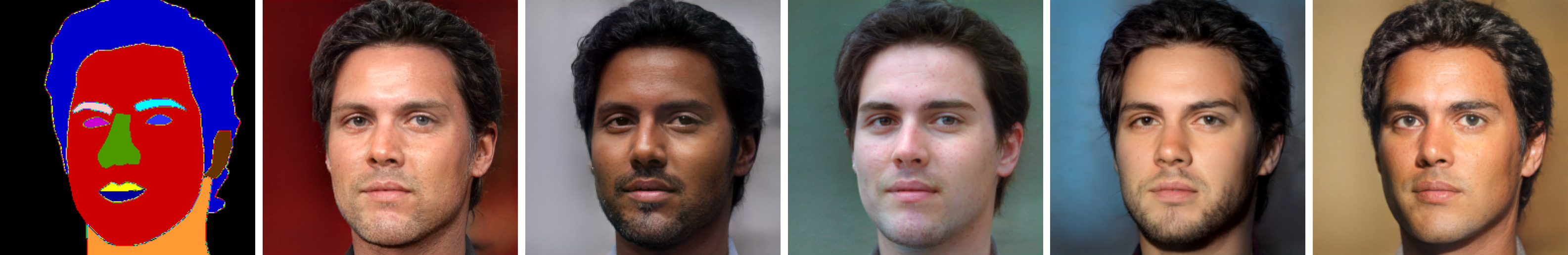}\\
        \includegraphics[width=0.9\textwidth]{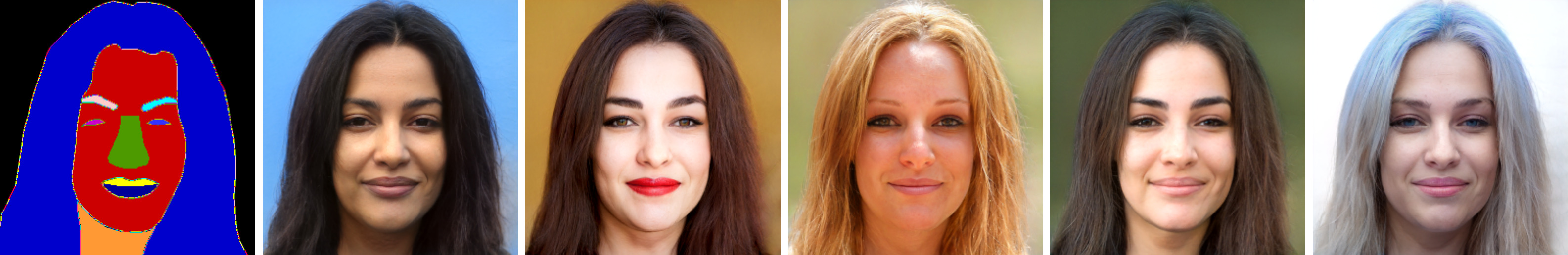}\\
        \includegraphics[width=0.9\textwidth]{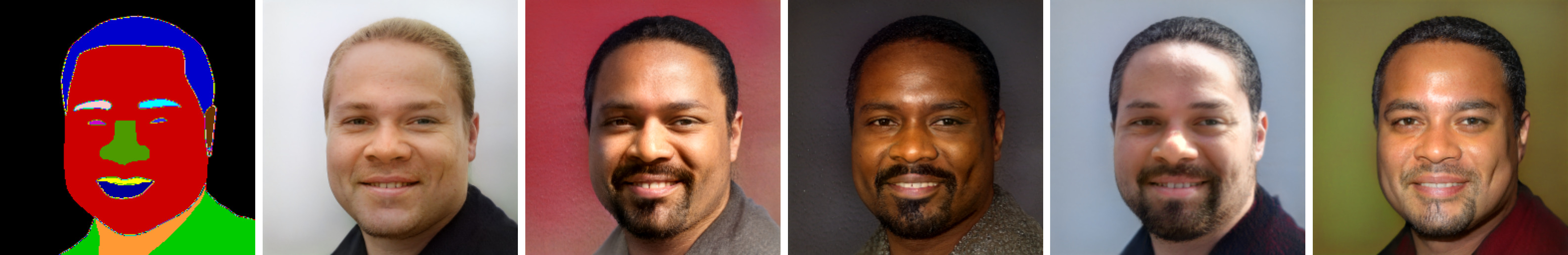}\\
        \vspace{0.5cm}
        \includegraphics[width=0.9\textwidth]{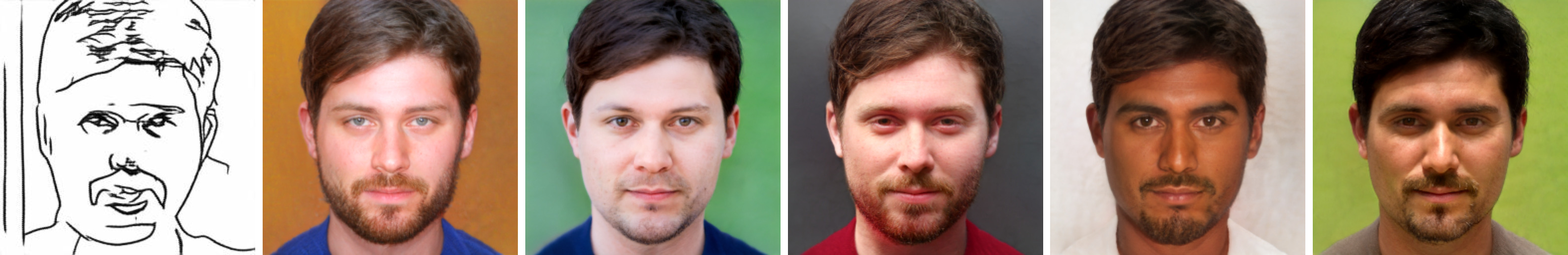}\\
        \includegraphics[width=0.9\textwidth]{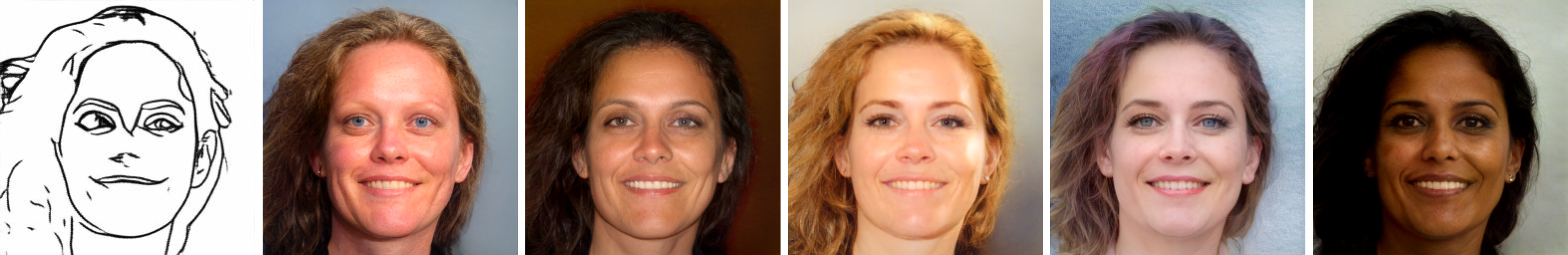}\\
        \includegraphics[width=0.9\textwidth]{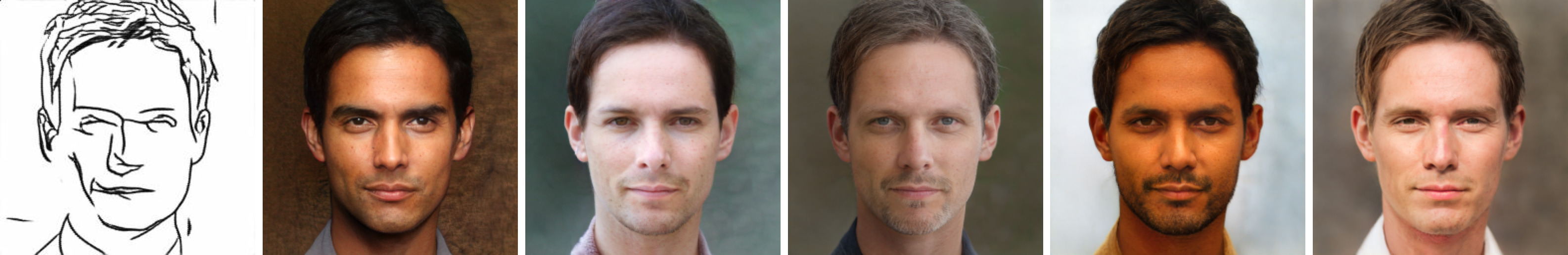}\\
        \includegraphics[width=0.9\textwidth]{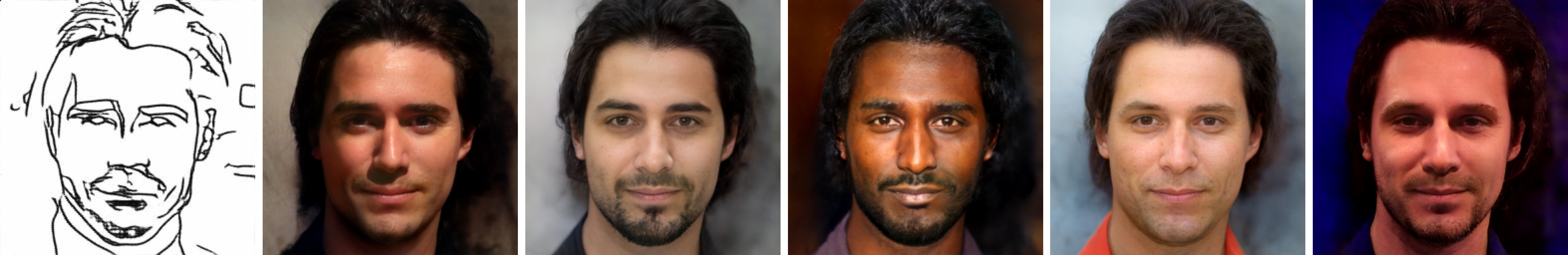}
    \captionof{figure}{Conditional image synthesis results from sketches and segmentation maps displaying the multi-modal property of our approach.}
    \label{fig:multi_modality_appendix}
\end{figure*}

\end{document}